\documentclass{article}

\usepackage{arxiv}

\usepackage[utf8]{inputenc} 
\usepackage[T1]{fontenc}    
\usepackage{hyperref}       
\usepackage{url}            
\usepackage{booktabs}       
\usepackage{amsfonts}       
\usepackage{nicefrac}       
\usepackage{microtype}      
\usepackage{lipsum}
\usepackage{graphicx}
\graphicspath{ {./images/} }

\usepackage{stmaryrd} 
\newcommand{\Ivs}[1]{\llbracket #1 \rrbracket} 
\usepackage{amsmath,amssymb}

\newcommand{\D}{{\cal D}}
\usepackage{adjustbox}
\usepackage{caption}
\usepackage{subcaption}
\usepackage{multirow}

\begin{document}

\title{Learning to Rank for Uplift Modeling}

\author{
Floris~Devriendt \\
Data Analytics Laboratory \\
Solvay Business School\\
Vrije Universiteit Brussel\\
Brussels, Belgium\\
\texttt{floris.devriendt@vub.be}\\
\And
Tias~Guns \\
Data Analytics Laboratory \\
Solvay Business School\\
Vrije Universiteit Brussel\\
Brussels, Belgium\\
\texttt{tias.guns@vub.be}
\And
Wouter~Verbeke \\
Data Analytics Laboratory \\
Solvay Business School\\
Vrije Universiteit Brussel\\
Brussels, Belgium\\
\texttt{wouter.verbeke@vub.be}
}
\maketitle
\begin{abstract}
Uplift modeling has effectively been used in fields such as marketing and customer retention, to target those customers that are most likely to respond due to the campaign or treatment. Uplift models produce uplift scores which are then used to essentially create a ranking. We instead investigate to learn to rank directly by looking into the potential of learning-to-rank techniques in the context of uplift modeling. We propose a unified formalisation of different global uplift modeling measures in use today and explore how these can be integrated into the learning-to-rank framework. Additionally, we introduce a new metric for learning-to-rank that focusses on optimizing the area under the uplift curve called the \textit{promoted cumulative gain} (PCG). We employ the learning-to-rank technique LambdaMART to optimize the ranking according to PCG and show improved results over standard learning-to-rank metrics and equal to improved results when compared with state-of-the-art uplift modeling. Finally, we show how learning-to-rank models can learn to optimize a certain targeting depth, however, these results do not generalize on the test set.
\end{abstract}

\section{Introduction}\label{sec:introduction}
Uplift modelling (UM) aims at establishing the net difference in customer behaviour when applying a certain \textit{treatment} to the customer versus not applying it, e.g. offering a discount for a certain product.

Uplift models produce for each customer an uplift score that indicates how susceptible (or \textit{persuadable}) they are by the treatment. This score is then used to rank unseen customers, and to target only a top fraction \cite{Devriendt:2018}. In order to make this ranking, uplift modeling aims to estimate a customers 'probability of persuasion' by the treatment. The main difficulty is that this is not directly measurable: we can observe the outcome after treating or not treating, but can not know what the outcome would have been for the opposite treatment choice, commonly known as \textit{The Fundamental Problem Of Causal Inference} \cite{Holland:1986}. Hence, uplift models have to be trained on data from an A/B test of treated and untreated customers and their respective outcome, and learn from that.

Because the output of an uplift model is used \textit{to create a ranking of customers}, we ask the question whether learning-to-rank would be beneficial as well. Learning-to-rank is a learning technique that stems from the information retrieval community \cite{Liu:2009}. It investigates techniques that optimize the quality of the predicted \textit{ranking} of instances directly, rather than the quality of the predicted value of the instances. 

Standard classification techniques can be seen as a 'pointwise' method to learning to rank. Hence, existing uplift modeling techniques can be seen as learning-to-rank already. 
A challenge in case of uplift modeling is that multiple evaluation measures have been proposed, often based on an interpretation of Uplift or Qini curves.
We will evaluate in how far existing uplift modeling techniques do or do not 'learn to rank', as well as the different ways in which it can be modelled as a learning-to-rank problem. We then consider how suited existing L2R measures are for uplift modeling, e.g. Discounted Cumulative Gain (DCG) and how to integrate uplift modeling measures directly into the LambdaMART learning-to-rank technique to optimize the area under the uplift curve.

Furthermore, in information retrieval, one is often interested in only the top-10 ranked results per query. Hence learning-to-rank has techniques and measures to optimize for the top-k specifically. Similarly, in many uplift modeling applications we are typically interested in a top fraction of the most susceptible customers. However, current uplift modeling techniques aim to optimize the entire ranking, whereas specifically targeting a top-k fraction has not been investigated before.

Our contributions are:
\begin{itemize} \itemsep0em
\item We investigate and compare the main uplift modeling measures in use today, and propose a unified formalization in which all measures can be unambiguously written.

\item We explore the different ways in which uplift modeling can be formulated as a learning-to-rank problem.

\item We show how Area Under the Uplift Curve can be directly translated as a learning-to-rank measure, with a specific discounting mechanism and relevance values.

\item We experimentally compare the different measures and their assumptions, as well as evaluating on real-life datasets the different learning-to-rank formulations.

\item We show how learning specifically for the top-k naturally fits the learning-to-rank setting, though the benefit of top-k learning is shown to be limited as the results do not generalize to the test set.

\item We compare different state-of-the-art uplift modelling techniques with our best performing learning-to-rank formulation and show learning-to-rank as a viable alternative to the existing uplift modeling methodology.
\end{itemize}

This paper is structured as followed: in Section~\ref{sec:RelatedWork} the existing evaluation measures from the uplift modeling literature are reviewed. Section~\ref{sec:L2R} explains learning-to-rank and its measures, and introduces a new metric for the learning-to-rank framework. In Section~\ref{Sec:Experiments} the experiments are described and the results reported. Section~\ref{Sec:Discussion} discusses the results and finally Section~\ref{sec:Conclusion} presents our conclusions.

\section{Uplift Modeling}\label{sec:RelatedWork}
In general, uplift modeling is about estimating the causal effect of an action or treatment on an outcome and therefore allows to determine which action to take or treatment to apply in order to optimize the outcome, i.e., the result or effect of the action or treatment. Uplift modeling is equivalent with individual causal modeling under a strong ignorability assumption \cite{Rubin:2005, Shalit:2017}. The field of uplift modeling was first introduced by \cite{Radcliffe:2007}, but has also been known as \textit{true-lift modeling}\cite{Lo:2002:TLM} or \textit{personalised treatment selection}\cite{Zhao:2017:MultipleTreatment}. 

We assume that a sample of customers is randomly divided into two groups, the treatment group and control group. A customer  is either in the treatment group, i.e., is exposed to the campaign, or is in the control group, i.e., is not exposed to the campaign. The uplift is then defined as the difference between response rates of the treatment and control group. As a formal definition, let $X$ be a vector of independent variables, also named features, $X = \{X_1, ..., X_n\}$ and let $y$ be the binary dependent or target variable, $y \in \{0,1\}$, that indicates response ($y=1$) or no response ($y=0$). Additionally, let the treatment variable $t$ denote whether a customer belongs to the treatment group, $t = 1$, or to the control group, $t = 0$. The expected value of a person responding in the treatment and control group is respectively $\mathbb{E}[y=1|X,do(t=1)]$ and $\mathbb{E}[y=1|X,do(t=0)]$. Under the assumption that the treatment was administered randomly and independent of $X$~\cite{Diemert:2018}, the causal effect of the treatment is then modelled as the uplift value:
\begin{equation}
U(X) := \mathbb{E}[y = 1 | X,t=1] - \mathbb{E}[y = 1 | X,t=0]
\label{Equation:FormulaUpliftExpVal}
\end{equation}

Uplift is here defined as the difference in probabilities, i.e., the probability of a customer to respond if treated minus the probability of the customer to respond when not treated.

When using a probability estimation tool $P$, such as a predictive model, for the two expectations above, we can estimate the uplift value of an individual instance as:
\begin{equation}
u(X_i) := P(y_i = 1 | X_i, t_i = 1) - P( y_i = 1 | X_i, t_i = 0)
\label{Equation:FormulaUplift}
\end{equation}

Uplift modeling has as goal to correctly estimate $u(X_i)$ for each instance in some way. The intended use is to then rank an unseen set of instances, e.g. customers, by their estimated $u(X_i)$ and to target a highly ranked fraction of the total population. For example in marketing or churn prediction, the size of the fraction is determined by the marketing budget, e.g. the top 1000 or 10000 customers. When given a limited budget, these are expected to be most likely to respond when treated by the campaign.

The main challenge in uplift modeling is that the uplift value $u(X_i)$ is unobservable. An instance can not be both treated and not-treated at the same time. This is the fundamental problem of causal inference, that we can observe only one of $P(y_i = 1 | X_i, t_i = 1)$ or $P( y_i = 1 | X_i, t_i = 0)$. 

\subsection{Related work}

Uplift modeling techniques can be grouped into data preprocessing and data processing approaches. In the preprocessing approaches, existing out-of-the-box learning methods are used, after pre- or post-processing of the data and outcomes. In the data processing approaches, new learning methods and methodologies are developed that aim to optimize expected uplift more directly. For an in depth discussion on uplift modeling techniques we refer to \cite{Devriendt:2018, Devriendt:2018b}.

A popular and generic \textit{data preprocessing} approach is the \textit{flipped label} approach, also called class transformation approach\cite{Lai:2006, Jaskowski:2012:UpliftClinical}. In this technique, a new target variable $Z \in {0,1}$ is created where $Z=1$ if either: treated with $Y=1$ or not treated with $Y=0$; and $Z=0$ otherwise.
Due to this class transformation the uplift modeling problem is converted into a binary classification problem with label $Z$, allowing to adopt any traditional classification technique \cite{Jaskowski:2012:UpliftClinical}.

Other data preprocessing approaches extend the set of predictor variables to allow for the estimation of uplift. An example is to group together the instances from both treatment and control group and then including a dummy variable that denotes whether an instance belongs to the treatment group (i.e. value 1) or control group (i.e. value 0). A model is then developed from: (1) the original predictor variables; (2) the added dummy variable; and (3) additional interaction variables between the predictor and dummy variables \cite{Lo:2002:TLM, Kane:2014}. The model can then be used to estimate two probabilities for a new instance, once as if the instance belongs to treatment group and once as if the instance belonged to the control group. In the latter case, the dummy variable is set to 0 and as a result the interaction variables are 0 as well. Subtracting the probabilities will return the uplift score.

\textit{Data processing} techniques include two categories: indirect and direct estimation approaches. Indirect estimation approaches include the two-model model approach. This approach builds two separate predictive models estimating churn, one for the treatment group and one for the control group. For a new instance, the aggregated uplift model computes the probability of positively responding with each model. Afterwards, the probabilities are subtracted to achieve an uplift score. An advantage of this approach is that it allows any standard predictive model to be employed.

Direct estimation approaches are typically adaptations from decision tree algorithms such as \textit{classification and regression trees} (CART) \cite{breiman1984} or \textit{chi-square automatic interaction detection} (CHAID) \cite{Kass:1980} methods. The adoptions include modified the splitting criteria and dedicated pruning techniques. Examples of tree-based uplift modeling approaches include the significance-based uplift trees proposed in \cite{Radcliffe:2011:RealWorldUplift}, decision trees making use of information theory-inspired splitting criteria presented in \cite{Rzepakowski:2012:SingleMultiple} and uplift random forest and causal conditional trees introduced in \cite{Guelman:2014}.

Another approach is the support vector machine approach of \cite{Jaroszewicz:2013:SVM, Zaniewicz:2017}. The model divides the space by constructing two hyperplanes and then predicts whether an instance will have a positive, neutral or negative reaction towards the action or treatment \cite{Jaroszewicz:2013:SVM, Zaniewicz:2017}. 

An alternative direct approach that is closer to learning to rank is the structured support vector machine approach of \cite{Kuusisto:2014}. It aims to directly optimizes the area under the uplift curve by using SVM-struct to maximizing with a single hyperplane, a weighted difference between that area under the roc curve on the treated subset and the area under the roc curve of the control subset, by adding pairwise constraints between the instances within the same group.

Our works differ from all the above in that we draw the link between uplift modelling and learning to rank, and we show how both standard learning-to-rank measures can be used, as well as how area under the uplift curve can naturally be formulated as a learning to rank measure; both when considering treated and control as separate subsets or as one joint one.

\subsection{Evaluation Measures}
\label{Sec:EvalMeasures}

Because $u(X_i)$ in Equation~\ref{Equation:FormulaUplift} is unobservable, we can not directly measure the quality of the estimated $u$ values, nor define a direct loss function. Instead, in uplift modeling the norm is to evaluate the quality of the \textit{ranking} resulting from ranking the instances by estimated uplift value. \cite{Radcliffe:2011:RealWorldUplift} propose to do this by computing and plotting the \textit{incremental} expected uplift value, for an incrementally larger subgroup of the ranked population. For example, for the top 10\%, 20\%, ..., 100\% of the instances. 

\begin{figure}[!t]
\centering
\includegraphics[width=10cm]{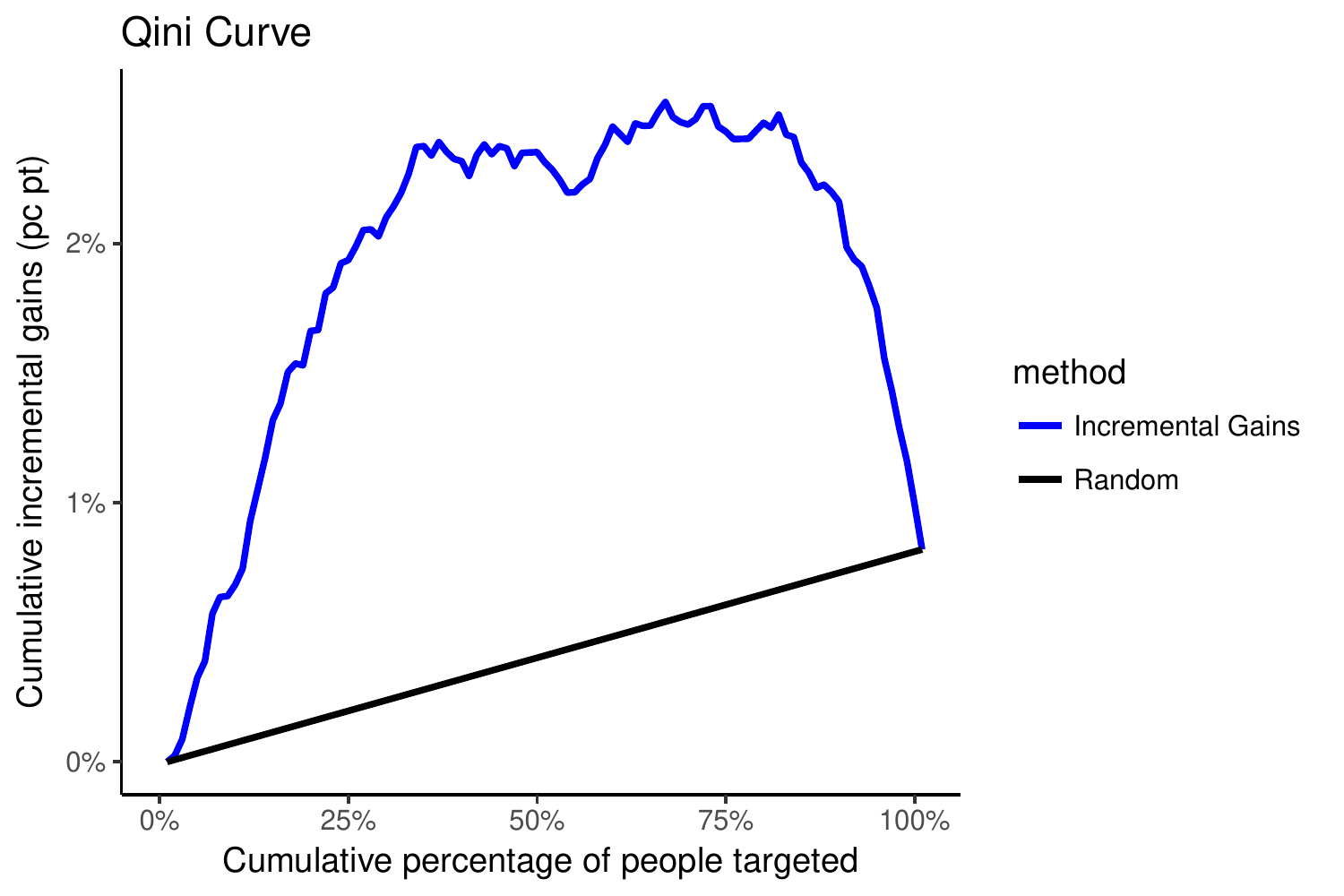}
\caption{An incremental gains curve (blue) and the expected gain from random treatments (black).}
\label{Fig:QiniCurve}
\end{figure}

Figure~\ref{Fig:QiniCurve} shows an example, where the blue line represents the cumulative incremental gains as a function of the selected fraction of the ranked population. The black line represents the expected value of a random subsample of that size called the random baseline.  A good uplift model will be able to rank individuals likely to respond when treated higher, leading to higher estimated uplift values in the early parts of the plot.

Interestingly, many different ways for computing the \textit{incremental} expected uplift have been proposed. Two variants exist with a different name, namely the Qini Curve and the Uplift Curve. We note that there is a further difference between the curves themselves, e.g. what values are plotted, and how a single measure such as area under the curve is derived (e.g. substracting the area under the random line or not). However, we here focus on the difference of how the curve values are computed.

Even for the two curves, there is no unique definition. We analyzed the literature, which frequently lacks a formal description of the curve computation, and identified two main differences. An overview is shown in Table~\ref{tab:evalMeasures}. The first difference relates to whether the ranking is computed for each group (i.e., treatment and control group) \textit{separately} or as one \textit{joint} group. If the ranking is computed separately, the 'top 10\%' instances are the 'top 10\%' instances of the treatment group and the 'top 10\%' instances of the control group. However, if the ranking is computed jointly, the 'top 10\%' instances will come from both groups. Hence, the proportion of instances represented in the ranking from each group can differ from the global proportion of the treatment and control group. The second difference is related to the potential imbalance of the two groups, and comes from whether the difference in effect is computed in (balanced) \textit{absolute} number of instances or in \textit{relative} incremental percentage, and if so, how this is done.

We first introduce the notation that will allow us to formalize the different variants, before discussing their differences in more detail. We assume a dataset ${\cal D} = \{ (X,y,t) \}$ of instances where $X$ is the feature vector, $y$ is a Boolean variable indicating whether the instance responded or not, and $t$ is a Boolean variable indicating whether the instance was treated or not. Note that it is generally accepted that some instances will respond without being treated (e.g. natural healing or subscribing to a product independent of advertising).

For evaluation, we assume the presence of a predictive model $\hat{u}$ by which a dataset ${\cal D}$ can be ranked. We write $\hat{u}$ because the score obtained is not necessarily obtained with the probabilities as in Equation~\ref{Equation:FormulaUplift}. We denote by $\pi$ the decreasing ordering of the dataset by $\hat{u}$, such that $\pi(\D, k)$ represents the first $k$ instances in $\D$ according to that ordering:
\begin{align*}
\pi({\cal D}, k) = \{ (X_i,y_i,t_i) \in \D \}_{i=1,\ldots,k}  ~\text{such that}~ \\
\hat{u}(X_i) \geq \hat{u}(X_l), X_i \in D, X_l \in D, \forall i \leq k, k < l
\end{align*}

We now denote the total \textit{number} of treated and control instances, among the top-$k$ ranked over the whole dataset as:
\begin{align}
N^T_\pi(\D, k) = \sum_{(X_i,y_i,1) \in \pi(\D, k)} \Ivs{t_i = 1} \\
N^C_\pi(\D, k) = \sum_{(X_i,y_i,0) \in \pi(\D, k)} \Ivs{t_i = 0}
\end{align}
where $\Ivs{\cdot}$ is the Iverson bracket to get the numeric value of the logical statement between brackets (1 for true, 0 for false). The number of treated and control \textit{responders} is, respectively:
\begin{align}
R^T_\pi(\D, k) = \sum_{(X_i,1,1) \in \pi(\D, k)} \Ivs{y_i = 1}\Ivs{t_i = 1} \\
R^C_\pi(\D, k) = \sum_{(X_i,1,0) \in \pi(\D, k)} \Ivs{y_i = 1}\Ivs{t_i = 0}
\end{align}

In this case, the entire data is ranked and the top-k is taken from that.

In literature, the lift in the treated and control group are often computed and compared separately. To formalize this, we use the subsets $T$ and $C$. With $T$ the subset of data that is treated, $T = \{ (X,y,t) \in {\cal D} | t = 1 \}$, and similarly $C$ the subset of data in the control group ($t = 0$), we can also consider the top $k$ ranked elements in the respective subgroups by $\pi(T, k)$ and $\pi(C, k)$.
We can now write (observe the $\pi(T, k)$ rather than $\pi(D, k)$ above):
\begin{align}
R_\pi(T, k) = \sum_{(X_i,y_i) \in \pi(T, k)} \Ivs{y_i = 1} \\
R_\pi(C, k) = \sum_{(X_i,y_i) \in \pi(C, k)} \Ivs{y_i = 1}
\end{align}
note that we can omit the indicator on $t_i$ as that condition is already satisfied by construction of $T$ and $C$.

\begin{table*}[hb]
\centering
\caption[Minimal Example]{Minimal example to demonstrate the notation style.}
\label{Table:ExNotation}
\begin{tabular}{|l|l|l|}
\hline
      & y & t \\ \hline
$i_1$ & 0 & 1 \\ \hline
$i_2$ & 1 & 1 \\ \hline
$i_3$ & 1 & 0 \\ \hline
$i_4$ & 1 & 1 \\ \hline
\end{tabular}
\end{table*}

To demonstrate the notation style and its difference, consider Table~\ref{Table:ExNotation} which shows a minimal example of a dataset. To obtain the amount of treatment instances amongst the top 1 and top 3 instances in the dataset, one would write $N^T(D,1) = 1$ and $N^T(D,3) = 2$ respectively. The difference in notation style becomes clear when comparing the notation for getting the amount of treatment responders in the joint data scenario with the separate data scenario. In the joint scenario, when counting the amount of treatment responders amongst the top 1 and top 3 one would write, $R^T(D,1) = 0$ and $R^T(D,3) = 1$ respectively, resulting in 0 treatment responders in the top 1 and 1 treatment responder in the top 3. However, in the separate scenario, when counting the treatment responder amongst the top 1 and top 3 one would write, $R(T,1) = 0$, and, $R(T,3) = 2$ respectively, resulting in 0 treatment responders in the top 1 and 2 treatment responders in the top 3. Notice the difference in outcome. With the joint scenario, control instances are included in the ranking as well, whereas in the separate scenario both the treatment and the control group have their own ranking.

\begin{table*}[t]
\centering
\caption[Evaluation Measures Uplift Modeling]{Evaluation measures for uplift modeling. Two main approaches are considered, the Qini Curve and the Uplift Curve, both over two dimensions: ranking the data separately per group or jointly over all data, and expressing the volumes in absolute or relative numbers.}
\begin{adjustbox}{max width=\textwidth}
\begin{tabular}{|c|c|l|l|}
\hline
\multicolumn{1}{|l|}{Rank} & \multicolumn{1}{l|}{Count} & Qini Curve & Uplift Curve \\ \hline
\multirow{4}{*}{Sep.}      & \multirow{2}{*}{Abs}       & $V(p) = R(T, p|T|) - R(C,p|C|)\frac{|T|}{|C|}$        & $V(p) = R(T, p|T|) - R(C,p|C|)$          \\
                           &                            & {\footnotesize \cite{Betlei:2018, Diemert:2018, Radcliffe:2007, Radcliffe:2011:RealWorldUplift}}       & {\footnotesize \cite{Kuusisto:2014}}         \\
                           & \multirow{2}{*}{Rel.}      &        & $V(p) = \frac{R(T, p|T|)}{|T|} - \frac{R(C, p|C|)}{|C|}$           \\
                           &                            &        & {\footnotesize \cite{Jaskowski:2012:UpliftClinical, Nassif:2013, Rzepakowski:2010, Rzepakowski:2012:SingleMultiple, Rzepakowski:2012:DirectMarketing, Soltys:2018, Soltys:2014, Zaniewicz:2017}}         \\ \hline
\multirow{4}{*}{Joint}     & \multirow{2}{*}{Abs.}      & $V(k) = R^T(\D, k) - R^C(\D, k)\frac{N^T(\D, k)}{N^C(\D, k)}$         & $V(k) = (\frac{R^T(\D, k)}{N^T(\D, k)} - \frac{R^C(\D, k)}{N^C(\D, k)}) * (N^T(\D, k) + N^C(\D, k))$           \\
                           &                            & {\footnotesize \cite{Gubela:2017, Betlei:2018}}       & {\footnotesize \cite{Gutierrez:2017}}         \\ \cline{3-4} 
                           & \multirow{2}{*}{Rel.}      & \multicolumn{2}{c|}{$V(k) = \frac{R^T(\D, k)}{|T|} - \frac{R^C(\D, k)}{|C|}$}   \\
                           &                            & \multicolumn{2}{c|}{{\footnotesize \cite{Guelman:2015, Diemert:2018}}} \\ \hline
\end{tabular}
\end{adjustbox}

\label{tab:evalMeasures}
\end{table*}

Table~\ref{tab:evalMeasures} shows the different types of evaluation measures proposed in the literature, where we omit the $\pi$ as it is clear from the context. We use $p$ in the 'separate' case to denote a percentage, e.g. 10\%, 20\%, where $p|T|$ is then the corresponding absolute number of instances, and $k$ in the 'joint' case to denote a number of instances. The \textbf{Value function $V()$} returns the incremental gains value of the first $p$ percentage of the population, or up to and including ranked instance $k$.

The first curve introduced in the literature was the Qini curve \cite{Radcliffe:2007}. The curve plots the absolute incremental responses of the treated group compared to as when there is no treatment. This is set in the \textit{separate} setting for which $p$ is used to denote the proportions of both the treatment and control groups. In case of uneven treatment and control groups, this offers a clear choice in value for $p$, e.g., if we have 100 treatment and 200 control observations, then $p|T| = \{10, 20, ..., 100\}$ for treatment and corresponding $p|C| = \{20, 40, ..., 200\}$ for control with $p=10\%$. The incremental responses of the control group are always adjusted in proportion to that of the treatment group for a balanced comparison. The values for the Qini Curve are then obtained by:
\begin{equation}
V(p) = R(T, p|T|) - R(C,p|C|)\frac{|T|}{|C|}
\end{equation} 

The Qini curve has since then been used and modified by several researchers to evaluate the performance of uplift models. 

Next to the Qini Curve, an alternative is the Uplift Curve. The uplift curve is described as the difference of two separate lift curves for treatment and control group using the same model. The resulting subtraction of the curves is called the uplift curve \cite{Rzepakowski:2012:SingleMultiple}. In \cite{Rzepakowski:2010} the authors measure the gain in profits by subtracting the profit obtained on $p$ percent of the highest scoring individuals of the control group from the profit obtained on $p$ percent of the highest total scoring individuals of the treatment group. A normalization factor, i.e., dividing by the respective total group sizes, is added to account for overall imbalance in treatment and control groups. The corresponding formula is then:
\begin{equation}
V(p) = \frac{R(T, p|T|)}{|T|} - \frac{R(C, p|C|)}{|C|}
\end{equation}

Kuusisto \cite{Kuusisto:2014} follows the same reasoning, however, the difference with the previous definition is that it measures the incremental gains in an absolute manner. The authors do not mention any normalization applied in calculating the points of the incremental gains curve.
\begin{equation}
V(p) = R(T, p|T|) - R(C,p|C|)
\end{equation}

The above definitions of the Qini and Uplift curve have all considered to evaluate the uplift models in a separate manner, i.e., the top $p$\% from the treatment group is compared with the top $p$\% from the control group. A different approach is to consider both the treatment and control group as one \textit{joint} group and target the top $k$ from that group, which is closer to how uplift models are intended to be used. An uplift model ranks observation by score and then selects the top $k$. 

In this setting, there is no clear distinction between the Qini curve and the Uplift curve, and it is modelled as follows \cite{Guelman:2015, Diemert:2018}:
\begin{equation}
V(k) = \frac{R^T(\D, k)}{|T|} - \frac{R^C(\D, k)}{|C|} \label{eq:joint_rel}
\end{equation}

The incremental gain can also be expressed in rebalanced absolute numbers as done in \cite{Gubela:2017}:
\begin{equation}
V(k) = R^T(\D, k) - R^C(\D, k)\frac{N^T(\D, k)}{N^C(\D, k)}
\end{equation}

Note that this measure rebalances the responder counts based on the proportions \textit{among the top-$k$ instances}, instead of the overall imbalance $\frac{|T|}{|C|}$.

An even different approach of obtaining rebalanced absolute numbers is proposed in \cite{Gutierrez:2017}, where the authors normalize before subtraction each of the responder counts by the proportion of the group under consideration, and then multiply it by the total number of instances in the group:
\begin{equation}
V(k) = (\frac{R^T(\D, k)}{N^T(\D, k)} - \frac{R^C(\D, k)}{N^C(\D, k)}) * (N^T(\D, k) + N^C(\D, k))
\end{equation}

\subsubsection*{From curves to a single measure}
The above definitions are used to construct the Qini or Uplift Curve. To measure and compare the performance among uplift models, we can calculate the Area Under the Uplift/Qini Curve (AUUC). This results into a numerical quantity which can be used to compare the performance among models. Sometimes this quantity is subtracted with the area under the curve of the baseline, e.g., for baseline \cite{Radcliffe:2007, Diemert:2018}. In this paper we will ignore this constant. Denoting any of the definitions from Table~\ref{tab:evalMeasures} as $V()$, the AUUC, in case of the joint setting, is defined as:

\begin{align}
AUUC = \int_0^1 V(x)dx \quad = \sum_{k=1}^n V(k)
\label{eq:AUUC}
\end{align}
and in case of the separate setting, we make the following approximation over 100 intervals:
\begin{align}
AUUC = \int_0^1 V(x)dx \quad \approx \sum_{p=1}^{100} V(p/100)
\label{eq:AUUC2}
\end{align}
In the above, for the separate setting, we arbitrarily chose to divide the area under the curve into 100 groups. In the literature, it is common that 10 group (or deciles) are used though this does not provide much granularity. When $|T|=|C|$ one can safely use $p=\frac{1}{|T|}, \frac{2}{|T|}, ..., \frac{|T|}{|T|}$, and similarly if the size of one is a multiple of the other. In other cases, the number of instances used per group, $p|T|$ or $p|C|$, may require rounding by which group sizes can slightly differ per group.

We will empirically compare the different measures in Section~\ref{sec:ComparingEvalMeasurers}.
However, our goal is not only to evaluate rankings but also to optimize the rankings for better performance. Therefore, we will look at learning-to-rank in the next section.

\section{Learning to Rank}
\label{sec:L2R}
Learning to rank finds its origin in the Information Retrieval (IR) domain. IR is defined as \textit{finding material (usually documents) of an unstructured nature (usually text) that satisfies an information need from within large collections (usually stored on computers)} \cite{Manning:2010}. Ranking is a core problem in IR, as many of IR problems are by nature ranking (e.g., document retrieval, collaborative filtering, product rating) \cite{Liu:2009}. As an example, we will use document retrieval in search to illustrate the working of learning-to-rank and to make the comparison with uplift modeling. 

Search engines are the most common examples of document retrieval in search. The web consists of an extremely large amount of documents (i.e., webpages), and finding relevant documents is a difficult task. A search engine has multiple components, but one of its most crucial ones is the ranker. The ranker is responsible for matching the request of the user, i.e., the query, with relevant indexed documents. The goal of a ranking algorithm is, given a query, to produce a ranked list of documents according to the relevance between the documents and the query \cite{Liu:2009}.

Learning-to-rank algorithms can be grouped into three approaches: pointwise approaches, pairwise approaches and listwise approaches. The pointwise approach predicts the relevance of each given document and uses these final scores to rank all the documents in the test set. Standard classification can be seen as a pointwise approach to ranking 'relevant' from 'non-relevant' documents, however, it also has its limitations: this approach does not consider the inter-dependency between documents. This means that the loss-function of a model does not consider the documents final place in the ranking \cite{Liu:2009}. The pairwise approach takes a pair of documents as input, and outputs which is more preferred between the two in terms of relevance. The limitation with a pairwise approach is that only the relative order of documents are considered, but from the output it is hard to derive the position of the documents in the final ranked list \cite{Liu:2009}. Finally, the listwise approach takes the entire set of documents for a query as input, and produces a ranked list as output. This approach is the closest to the learning-to-rank ideology as there is no mismatch between the learning stage of the model and the output of the model (as opposed to pointwise and pairwise approaches) \cite{Liu:2009}.

One of the most well-known and versatile techniques in learning-to-rank is LambdaMART, which is a listwise approach that combines techniques from two previous methods in the field, namely LamdaRank and Multiple Additive Regression Trees (MART). LambdaRank on itself is an extension of RankNet which uses a cost function to minimize the amount of inversions in a ranked list. A core idea to the LambdaRank algorithm is that it can directly optimize a ranking measure even if it is non-differentiable, by multiplying the gradients of a pairwise loss function with the cost difference of swapping the pair's positions in the ranking \cite{Burges:2010}. The Multiple Additive Regression Trees (MART) use gradient-boosted tree learning. Combining LambdaRank and MART gives us LambdaMART which combines the usage of gradients and loss functions with gradient boosted trees. For more information, we refer to \cite{Burges:2010}.

\subsection{Learning to Rank Evaluation Measures}
\label{Sec:L2RMeasures}
Several commonly used metrics in learning-to-rank exist \cite{Mcfee:2010, NIPS2009_3708, Mcfee:2010}: Precision (P) \cite{Mcfee:2010}, Mean Average Precision (MAP) \cite{Mcfee:2010}, Cumulative Gain (CG) \cite{Jarvelin:2002}, Discounted Cumulative Gain (DCG) \cite{Jarvelin:2002}. We briefly cover each one of them. Aferwards, we will compare them to uplift modeling measures in Section~\ref{sec:PointwiseVsListwise}.

Different metrics exist for when the relevance is binary and when its not binary, e.g., for when relevance values are graded into a five-star rating system.

\subsubsection{Binary relevance}
When working with binary relevance values, observations are either either 'relevant' or 'not relevant', i.e., $rel_i \in \{0,1\}$.

The Precision At K ($P(k)$) corresponds to the proportion of relevant instances among the first k instances, ranked according to the values produced by the learning system.

\begin{equation}
P(k) = \frac{\sum_{i=1}^k rel_i}{k}
\end{equation}

The Average Precision of a query $q$, consisting of $|q|$ documents (instances), sums over all documents in the query and computes the average of the $P(k)$ values at every rank $k$ where a relevant document is positioned. 

\begin{equation}
\textsl{AvgP}(q) = \frac{\sum^{|q|}_{i=1} P(i)}{\sum_{i=1}^{|q|} rel_i}
\end{equation}

The Mean Average Precision (MAP) is the mean of average precisions over a set $Q$ of queries.
\begin{equation}
MAP(Q) = \frac{\sum_{q\in Q} \textsl{AvgP}(q)}{|Q|} \label{eq:map}
\end{equation}

\subsubsection{Graded relevance}
Here, the relevance is assumed to be given as a graded score, e.g., $rel_i \in \{0,1,\ldots,5\}$, with higher being more relevant. Evaluation can be accordingly modified, as highly relevant documents are more valuable than marginally relevant documents, which in turn are more valuable than non-relevant documents. The Cumulative Gain (CG) of a ranked list is the sum of all relevance values of the ranked list of a single query, up to point $k$. The downside of CG is that it does not take into account the \textit{position} of a relevant document in the ranking \cite{Jarvelin:2002}.
\begin{equation}
CG(k) = \sum_{i=1}^k rel_i
\end{equation}

An alternative that does, is the Discounted Cumulative Gain (DCG) which will penalize highly relevant documents that appear lower in the ranking (as we want them to be ranked highly). The graded relevance value is discounted logarithmically proportional to the position in the ranking (the denominator in the equation) \cite{Jarvelin:2002}.
\begin{equation}
DCG^1(k) = \sum_{i=1}^k \frac{rel_i}{log_2(i+1)} \label{eq:dcg}
\end{equation}

An alternative formulation exists, which places stronger emphasis on the relevancy value of the observations \cite{Burges:2005}:
\begin{equation}
DCG^2(k) = \sum_{i=1}^k \frac{2^{rel_i}-1}{log_2(i+1)}
\end{equation}

Note that both formulations are equal when the relevance values are binary \cite{Croft:2010}. 

Different queries can have different amounts of documents. To fairly compare a rankers' performance among multiple queries of different size, one should normalize the DCG of each query to achieve scores in the range of $[0,1]$. The Normalized Discounted Gain (NDCG) measure proposes to normalize by dividing the achieved discounted cumulative gains of each query by its Ideal Discounted Cumulative Gain (IDCG). The IDCG is created by sorting all documents according to their relevance and so creating the maximum possible DCG. The NDCG is formulated as \cite{Jarvelin:2002}: 
\begin{align}
IDCG(k) &= \sum_{i=1}^k \frac{rel'_i}{log_2(i+1)} \\
NDCG(k) &= \frac{DCG(k)}{IDCG(k)}
\end{align}

where $rel'_i$ is the best possible ranking at position $i$. The total value of $DCG$ and $NDCG$ over a set of queries $Q$ is calculated by the mean of the value per query (similarly to Equation~\eqref{eq:map}).

\subsection{Uplift Modeling as Learning To Rank}

To apply uplift modeling in a learning-to-rank framework, we define two distinct steps: (1) we instantiate uplift modeling to learning-to-rank by identifying the correct definitions of queries, documents and relevance values and how they relate to Table~\ref{tab:evalMeasures}; and (2) we describe an alternative measure which allows to directly optimize the area under the uplift curve.

\subsubsection{Learning To Rank in context of Uplift Modeling}
\label{sec:UMasL2RMeasure}

Several concepts of learning-to-rank need to be translated in the context of uplift modeling: query, documents and relevance values. 

\paragraph*{Queries and documents}
In uplift modeling an individual can be in one of two groups: the treatment group (which was exposed to the treatment) or the control group (which was not exposed to the treatment). As shown in Table~\ref{tab:evalMeasures}, quality measures can be computed on the groups separately or as one joint group. Likewise in learning-to-rank, one or multiple queries can be considered.

In the separate setting, we consider both the treatment and control groups as separate queries, i.e., a learning-to-rank technique will be run with two queries in total, and the instances in each query are the documents. In the joint setting, we consider only one query, combining the individuals of the treatment and control group into a single set of documents.

\paragraph*{Relevance}
The equivalent of the relevance of a document in uplift modeling is the uplift-value we get from an instance, which is determined by which group it belongs to and what the response was. An instance can belong to one of the following four categories: Treatment Responder (TR), Treatment Non-Responder (TNR), Control Responder (CR) and Control Non-Responder (CNR). We now determine relevance values for each possible definition of uplift, shown in Table~\ref{tab:evalMeasures}.

\paragraph*{Separate Setting}
In the separate setting there are two queries, one for the treatment group and one for the control group. Each query optimizes their own rankings. For each of the aforementioned categories, the impact on the value function (as introduced in Table~\ref{tab:evalMeasures}) is checked. Table~\ref{tab:gainsSep} shows the relevant definitions of the value function and the corresponding relevance values. We can tell the impact of the categories TR and TNR when looking at the left component of the subtraction in the value function definitions. An instance of the category TR will increase the overall value obtained, whereas an instance of the TNR category does not impact the value-function at all (in both absolute and relative definitions). Similarly the right component shows the impact of the CR and CNR categories. An increase on the right component will lower the overall value obtained. Therefore, a negative relevance value is associated with an instance of the CR category. Depending on the value function definition, the relevance associated can change. On the other hand, an instance of the CNR category does not have an impact on the overall value.

\begin{table*}[t]
\begin{center}
\caption[Relevance Values - Separate Queries]{Relevance values for separate queries according to value function definition.\label{tab:gainsSep}}
\begin{adjustbox}{max width=\textwidth}
{\small
\begin{tabular}{l|l|l|l|l|}
\hline
\multicolumn{1}{|l|}{Value Function} & Treatment Responder & Treatment Non-Responder & Control Responder & Control Non-Responder \\
\multicolumn{1}{|l|}{}		& $rel_{TR}$ & $rel_{TNR}$ & $rel_{CR}$ & $rel_{CNR}$ \\ \hline 
\multicolumn{1}{|l|}{$V(p) = R(T, p|T|) - R(C,p|C|)$} & 1 & 0 & -1 & 0 \\
\multicolumn{1}{|l|}{$V(p) = R(T, p|T|) - R(C,p|C|)\frac{|T|}{|C|}$} & 1 & 0 & $-\frac{|T|}{|C|}$ & 0 \\
\multicolumn{1}{|l|}{$V(p) = \frac{R(T, p|T|)}{|T|} - \frac{R(C, p|C|)}{|C|}$} & $\frac{1}{|T|}$ & 0 & $-\frac{1}{|C|}$ & 0 \\ \hline
& \multicolumn{2}{c|}{Query 1} & \multicolumn{2}{c|}{Query 2} \\ \cline{2-5}
\end{tabular}
}
\end{adjustbox}
\end{center}
\end{table*}

\paragraph{Joint Setting}
In the joint setting the treatment and control group are combined into one query. When considering \textit{relative} values, the relevance values are as before with the separate queries (Table~\ref{tab:gainsJoint}). An instance from the TR category will raise the value, whereas an instance from the CR category will decrease it. An instance from the TNR or CNR category has no effect on the overall value function.

However, in the absolute settings \cite{Betlei:2018}, the counts are rebalanced using the number of treated/control instances observed so far $(\frac{N^T(\D, k)}{N^C(\D, k)})$. Hence, instances from the $TNR$ and $CNR$ no longer have $0$ effect as a TNR increases this ratio, while a CNR decreases this ratio. Hence, a TNR instance will increase the right (negative) component of the value function and will hence lead to a decrease of the overall value, while a CNR instance will decrease the negative component and hence increase the overall value.
Quantifying the exact increase and decrease is not trivial, and we instead observe that TR instances increase the value most, then CNR instances, then TNR instances lead to a small decrease and CR instances to a large decrease. We simply encode this relation by using a relevance of 3 for TR, 2 for CNR, 1 for TNR and 0 for CR (Table~\ref{tab:gainsJoint}).

\begin{table*}[t]
\begin{center}
\caption[Relevance Values - Joint Queries]{Relevance values for joint queries according to value function definition. \label{tab:gainsJoint}}
\begin{adjustbox}{max width=\textwidth}
{\small
\begin{tabular}{l|l|l|l|l|}
\hline
\multicolumn{1}{|l|}{Value Function} & Treatment Responder & Treatment Non-Responder & Control Responder & Control Non-Responder \\
\multicolumn{1}{|l|}{}		& $rel_{TR}$ & $rel_{TNR}$ & $rel_{CR}$ & $rel_{CNR}$ \\ \hline 
\multicolumn{1}{|l|}{$V(k) = \frac{R^T(\D, k)}{|T|} - \frac{R^C(\D, k)}{|C|}$} & $\frac{1}{|T|}$ & 0 & $-\frac{1}{|C|}$ & 0 \\
\multicolumn{1}{|l|}{$V(k) = R^T(\D, k) - R^C(\D, k)\frac{N^T(\D, k)}{N^C(\D, k)}$} & 3 & 1 & 0 & 2 \\ \hline
& \multicolumn{4}{c|}{Query 1} \\ \cline{2-5}
\end{tabular}
}
\end{adjustbox}
\end{center}

\end{table*}

Above relevance values can be used with any learning-to-rank metric that accepts graded relevance values, i.e., $DCG$ and $NDCG$. The MAP measure accepts only binary relevance values. In the next section we consider a new measure, capable of using above graded relevance values, and in line with uplift modeling by optimizing the AUUC directly.

\subsubsection{Uplift based measure AUUC as learning-to-rank measure}
\label{sec:ModLambdaMart}

The goal is to come up with a learning-to-rank measure that is most similar to the Area Under the Uplift Curve, and that can be optimized using existing learning-to-rank systems.
We will use the \textit{relative} definitions as the experiments in Section~\ref{Sec:Experiments} will show that these are most robust to differences in group sizes.

Equation~\ref{eq:AUUC} states that the AUUC is a summation of the value function over all possible $k$ values. If we insert the value definition for the \textit{Joint Relative} uplift curve of Equation~\ref{eq:joint_rel}, we get the following:
\begin{align*}
\begin{split}
AUUC &= \sum_{k=1}^n V(k) \\
     &= \sum_{k=1}^n \left(\frac{R^T(\D, k)}{|T|} - \frac{R^C(\D, k)}{|C|}\right)
\end{split}
\end{align*}

To ease the notation, we define $\D^\pi_i$ to be the $i$th element in the list when ordering the data $\D$ according to ordering $\pi$. Likewise, $(X^\pi_i,y^\pi_i,t^\pi_i) = \D^\pi_i$. Hence we have:
\begin{align*}
\begin{split}
R^T_\pi(\D, k) &= \sum_{(X_i,y_i,t_i) \in \pi(\D, k)} \Ivs{y_i = 1}\Ivs{t_i = 1} \\
               &= \sum_{i=1}^k \Ivs{y^\pi_i = 1}\Ivs{t^\pi_i = 1}
\end{split}
\end{align*}

and similarly for $R^C_\pi(\D, k)$. Plugging this into the above, we obtain:
\begin{align*}
\begin{split}
AUUC &= \sum_{k=1}^n \left(\frac{R^T(\D, k)}{|T|} - \frac{R^C(\D, k)}{|C|}\right) \\
     &= \sum_{k=1}^n \left(\frac{\sum_{i=1}^k \Ivs{y^\pi_i = 1}\Ivs{t^\pi_i = 1}}{|T|} - \frac{\sum_{i=1}^k \Ivs{y^\pi_i = 1}\Ivs{t^\pi_i = 0}}{|C|}\right) \\
     &= \sum_{k=1}^n \sum_{i=1}^k \left(\frac{\Ivs{y^\pi_i = 1}\Ivs{t^\pi_i = 1}}{|T|} - \frac{\Ivs{y^\pi_i = 1}\Ivs{t^\pi_i = 0}}{|C|}\right)
\end{split}
\end{align*}
We now introduce helper function $g(i)$ as:
\begin{flalign*} 
g(i) = \begin{cases} 
            0 & \text{if } y^\pi_i = 0 \\
            1/|T| & \text{if } y^\pi_i = 1 \text{ and } t^\pi_i = 1  \\
            -1/|C| & \text{if } y^\pi_i = 1 \text{ and } t^\pi_i = 0  \\
      \end{cases} \label{eq:gi}
\end{flalign*}
Observing that these three cases are mutually exclusive and cover all possible assignment combinations to an $y^\pi_i$ and $t^\pi_i$, we can substitute this into the previous equation to obtain:
\begin{align*}
\begin{split}
AUUC &= \sum_{k=1}^n \sum_{i=1}^k \left(\frac{\Ivs{y^\pi_i = 1}\Ivs{t^\pi_i = 1}}{|T|} - \frac{\Ivs{y^\pi_i = 1}\Ivs{t^\pi_i = 0}}{|C|}\right) \\
     &= \sum_{k=1}^n \sum_{i=1}^k g(i) = \sum_{i=1}^n \sum_{k=i}^n g(i) \\
     &= \sum_{i=1}^n  g(i)\sum_{k=i}^n 1 = \sum_{i=1}^n (n-i+1)g(i)
\end{split}
\end{align*}

This can be related to the DCG measure in Equation~\ref{eq:dcg}, where each element has a relevance $rel_i$, in our case determined by $g(i)$, and instead of discounting by $1/log_2(i+1)$, we promote each element by $(n-i+1)$. We call this measure the \textit{Promoting Cumulative Gain} PCG:
\begin{equation}
PCG(k) = \sum_{i=1}^k rel_i*(n-i+1) \label{eq:pcg}
\end{equation}

A similar derivation can be made for the the \textit{separate} setting where the treated and controls are divided and ranked in separate groups. The formula then becomes (see Appendix A for details):
\begin{align}
\begin{split}
\sum_{i=1}^{k_1} ({|T|}-i+1)g^T(i) + \sum_{i=1}^{k_2} ({|C|}-i+1)g^C(i) \\
 = PCG^T(k_1) + PCG^C(k_2)
\end{split}
\end{align}
with $k_1 = p|T|$ and $k_2 = p|C|$ for some percentage $p$. Hence it can be computed on both sets separately and summed up. In a learning-to-rank setup, that would mean creating two queries, one for treatment and one for control, and using the PCG measure on each, after which the scores are aggregated. Note that the $k$ is different for each query, which requires a small modification to the learning systems.

\section{Experiments}
\label{Sec:Experiments}
In order to test the relationship between uplift modeling and learning-to-rank, multiple research questions are defined. We first give information on the datasets and software in Section~\ref{sec:ExpSetup}. Section~\ref{sec:ComparingEvalMeasurers} covers our first research question in which we analyze the concrete difference between the value function definitions in Table~\ref{tab:evalMeasures}. Section~\ref{sec:PointwiseVsListwise} compares uplift modeling with learning-to-rank approaches. Section~\ref{sec:ComparisonGradedRelevanceLabels} looks how different relevance values affect the performance of the learning-to-rank techniques. Section~\ref{sec:RQ4} looks at the impact of the k-value in terms of AUUC value. Finally, Section~\ref{sec:RQ5} compares the performance of the best performing learning-to-rank technique with the state-of-the-art uplift modeling.

\subsection{Experimental Setup}
\label{sec:ExpSetup}

\subsubsection{Datasets}

The experiments are performed on three public real-world datasets in order to ensure reproducibility (Table~\ref{tab:evalMeasures}). The first dataset is part of the \textit{Information} R-package\footnote{https://cran.r-project.org/web/packages/Information/index.html}. The data relates to a marketing campaign in the insurance industry and the target variable indicates whether or not a purchase happened. The second dataset is published on the website MineThatData\footnote{https://blog.minethatdata.com/2008/03/minethatdata-e-mail-analytics-and-data.html} and contains data from an e-mail marketing campaign concerning clothing merchandise. The dataset includes three target variables: visit (yes / no), purchase (yes / no) and conversion (numerical; the amount of money spent). The dataset includes 64,000 observations with $1/3$ targeted with an email campaign concerning men's merchandise, $1/3$ targeted with an email campaign concerning women's merchandise and $1/3$ customers who where not targeted. For this dataset, the 'visit' target variable was selected as the target variable of interest and the selected treatment is the email campaign for women's merchandise, in line with \cite{Kane:2014} to facilitate comparison (reducing the dataset to 42,693 observations). The last dataset is obtained from the Criteo AI Lab\footnote{https://ailab.criteo.com/criteo-uplift-prediction-dataset/} \cite{Diemert:2018}, which has data resulting from several incrementality tests in advertising. The total size reaches up to 25 million observations, but due to computation issues, a randomly selected subsample of $0.001\%$ was taken (reducing the dataset to 25,310 observations).

\begin{table*}[t]
\begin{center}
\caption[Datasets Experiments]{Overview of datasets used in the experiments. \label{tab:Datasets}}
\begin{tabular}{l|lll|}
\cline{2-4}
                                                      & Information & Hillstrom (Womens Clothing) & Criteo        \\ \hline
\multicolumn{1}{|l|}{Description}                     & Insurance   & Online Merchandise          & Marketing     \\
\multicolumn{1}{|l|}{Channel}                         & E-Mail      & E-Mail                      & Advertisement \\
\multicolumn{1}{|l|}{Total Size}                      & 10 000      & 64 000                      & 25 309 483    \\
\multicolumn{1}{|l|}{\# Treatment Observations}       & 4972        & 21 387                      & 21 409        \\
\multicolumn{1}{|l|}{\# Control Observations}         & 5028        & 21 306                      & 3 901         \\
\multicolumn{1}{|l|}{\# Variables}                    & 68          & 10                          & 14            \\
\multicolumn{1}{|l|}{Response variable (binary)}      & Purchase    & Visit                       & Visit         \\
\multicolumn{1}{|l|}{Treatment-to-Control size ratio} & 0.99:1      & 1:1                         & 5.48:1        \\
\multicolumn{1}{|l|}{Treatment Positive Rate}         & 20.37 \%    & 15.14 \%                    & 4.41 \%       \\
\multicolumn{1}{|l|}{Control Positive Rate}           & 19.55 \%    & 10.62 \%                    & 2.61 \%       \\
\multicolumn{1}{|l|}{Uplift Initial Campaign}         & 0.82 \%     & 4.52\%                      & 1.80 \%       \\ \hline
\end{tabular}
\end{center}
\end{table*}

\subsubsection{Experiments}

All learning-to-rank techniques are from the open-source RankLib package\footnote{https://sourceforge.net/p/lemur/wiki/RankLib/} which is implemented in Java. The selected learning-to-rank technique is LamdaMART, which is implemented as a gradient boosted tree. For fair comparison all uplift modeling techniques make use of gradient boosted trees, which are implemented in R with the aid of the \textit{xgboost}-package \cite{R:xgboost}. 

Parameter tuning was done upfront and the hyperparameters are kept identical for all experiments in the paper, namely 500 trees and a learning rate of 0.01. 
The experiments are each repeated 10 times and the results are averaged, with plots visualising the minimum/maximum range as a shaded area.

\subsection{Question 1: Comparing evaluation measure differences through simulation}
\label{sec:ComparingEvalMeasurers}

To optimize learning-to-rank techniques with an uplift modeling loss-function we need to consider which definition of the evaluation approaches introduced in Table~\ref{tab:evalMeasures} to use. Looking at the table, one can see that the normalization that accounts for possible differences in treatment and control group size is a differentiating factor. To better understand the differences, we simulate data into three different scenarios all related to the sizes of the treatment and control group.

To evaluate a ranking-based evaluation metric we only need a different set of rankings to compare with. Simulating a theoretical ranking allows us to produce rankings for any scenario. The simulation consists of two populations: the treatment group with a response rate of 7\% and the control group with a response rate of 5\%. Each of the individuals is uniformly sampled an uplift score between 0 and 1 depending on the category the observation is in (i.e., TR, TNR, CR or CNR). To simulate uplift, the categories TR and CNR will be sampled from a higher uplift score range between $[0.2, 1.0]$, and the CR and TNR will be sampled from a lower uplift score range between $[0.0, 0.8]$. These ranges ensure that TR and CNR observations will appear higher in the overall ranking compared to CR and TNR observations. In a next step, we sample from both treatment and control group to create three scenario's: (1) a balanced setting with equal observations in both treatment and control groups ($|T|=|C|$), (2) an imbalanced setting with a larger treatment group ($|T| = 9|C|$) and (3) an imbalanced setting with a larger control group ($|C|=9|T|$). Because each observation has an uplift score, we can create for each scenario a ranking which is then to be tested on ranking-based evaluaton metrics.

We visualise the curves obtained when using the different measures in Table~\ref{tab:evalMeasures} to study the effect. The goal is to find out if there is an uplift measure that is stable under the three different scenarios. We split up the analysis in absolute count and relative count due to them having different units of measurement.

\begin{figure*}[t]
\centering
\begin{subfigure}{.45\textwidth}
  \centering
  \includegraphics[width=1\linewidth]{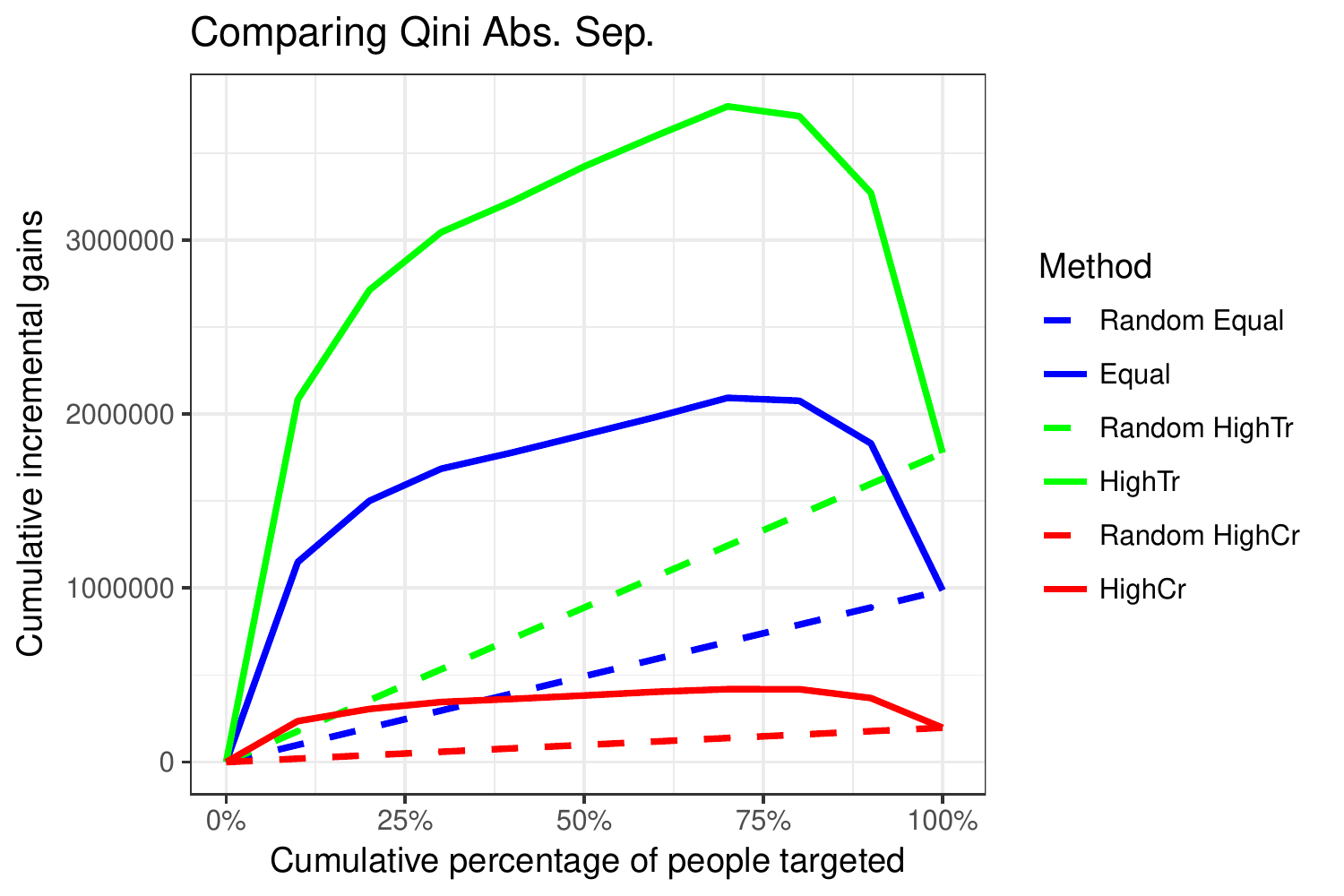}
  \caption{Qini Curve - Absolute Separate}
  \label{fig:SCQiniCurveAbsSep}
\end{subfigure}%
\begin{subfigure}{.45\textwidth}
  \centering
  \includegraphics[width=1\linewidth]{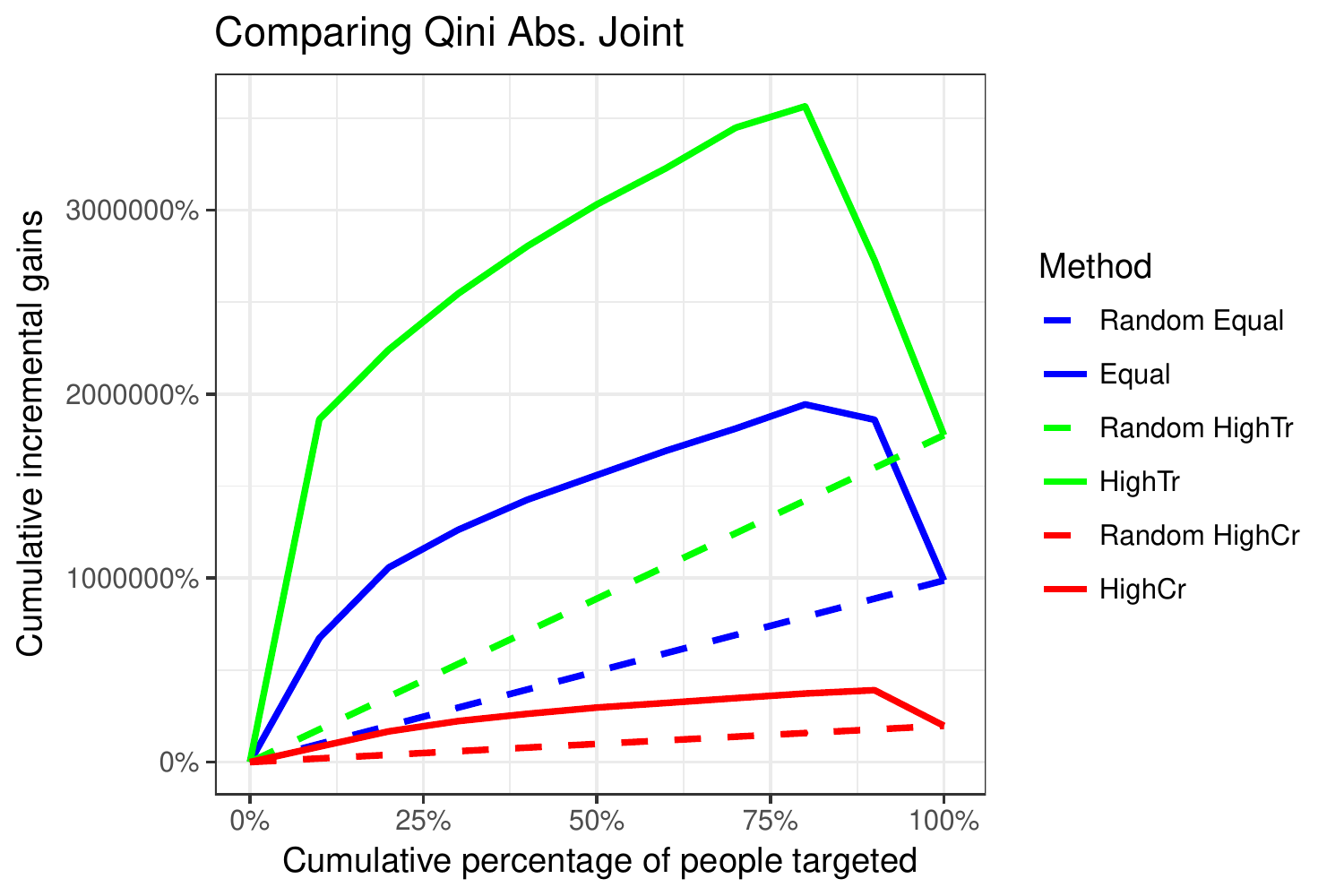}
  \caption{Qini Curve - Absolute Joint}
    \label{fig:SCQiniCurveAbsJoint}
\end{subfigure}
\begin{subfigure}{.45\textwidth}
  \centering
  \includegraphics[width=1\linewidth]{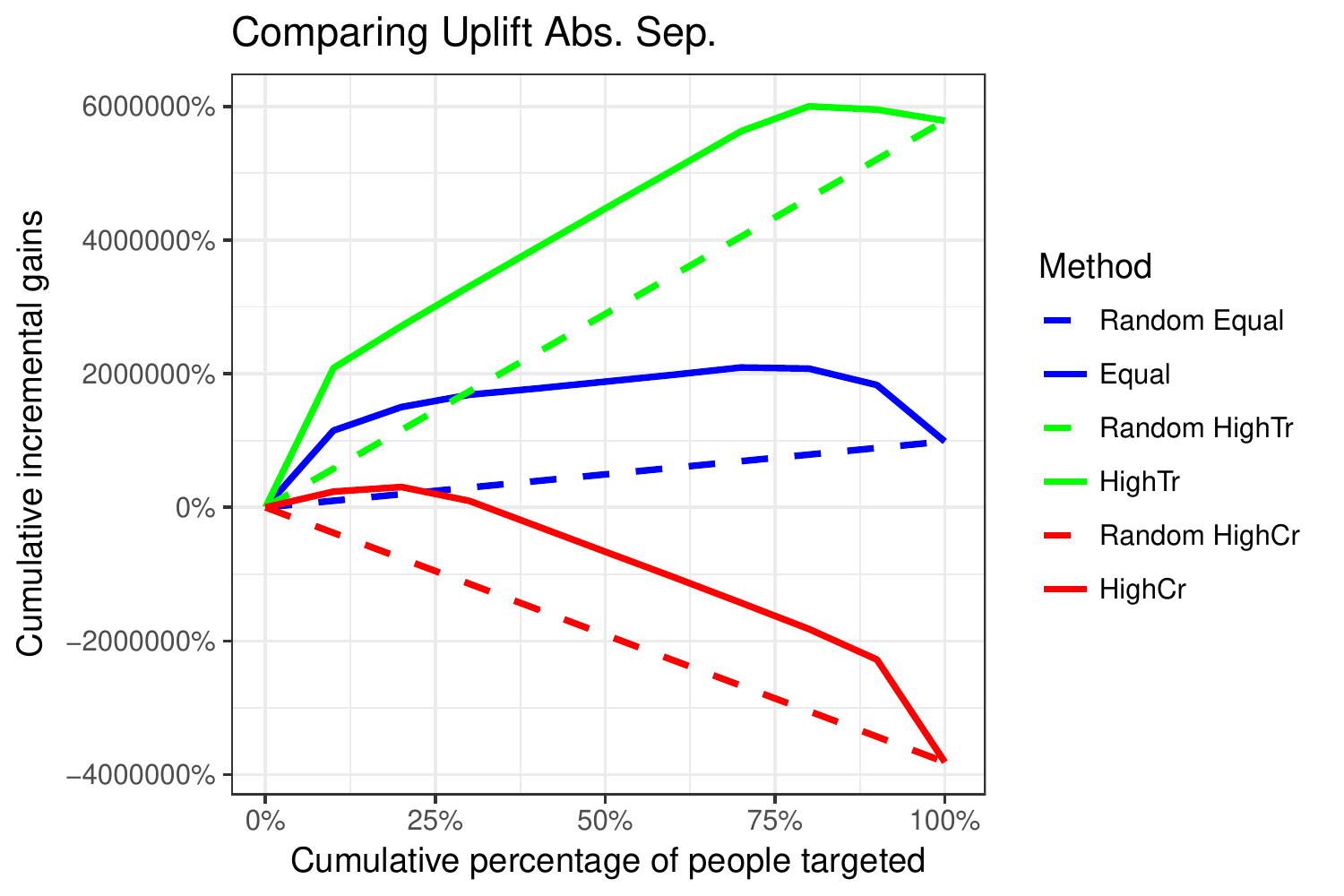}
  \caption{Uplift Curve - Absolute Separate}
      \label{fig:SCUpliftCurveAbsSep}
\end{subfigure}%
\begin{subfigure}{.45\textwidth}
  \centering
  \includegraphics[width=1\linewidth]{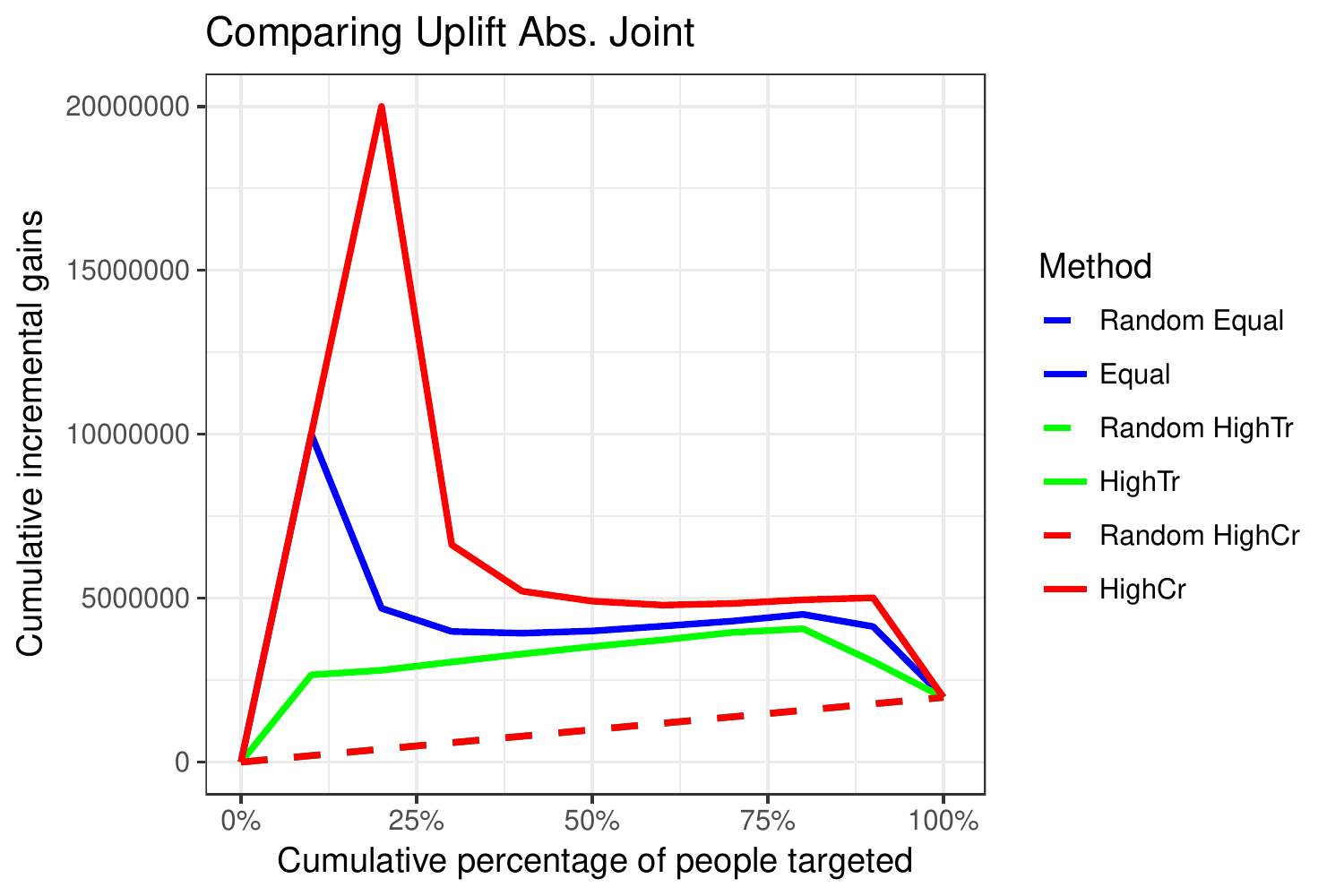}
  \caption{Uplift Curve - Absolute Joint}
      \label{fig:SCUpliftCurveAbsJoint}
\end{subfigure}
\caption[Experiment 1 - Simulation Results Absolute Curves]{Simulation of the Absolute Curves in both Separate and Joint setting. In each figure the baseline of each respective scenario is represented with the dashed line. In Figure~\ref{fig:SCUpliftCurveAbsJoint} the baselines have the same values in each scenario.}
\label{fig:SimulationCurvesAbs}
\end{figure*}

\paragraph{\bf Absolute Count}

To analyze the behavior of the absolute count we take all four absolute definitions from Table~\ref{tab:evalMeasures} and plot their Qini curve, with corresponding baseline, based on the ranking of the three scenario's defined above (Figure~\ref{fig:SimulationCurvesAbs}). The qini curve in separate and joint setting behave quite similar (Figure~\ref{fig:SCQiniCurveAbsSep} and \ref{fig:SCQiniCurveAbsJoint}). All AUUC values are positive, but the values differ significantly depending on the sizes of the group. The uplift curve in the absolute setting shows similar behavior, but when the control group has significantly more observations we see that the AUUC values turn negative. This is due to the fact that there is no normalization between the two groups. In contrast, both qini definitions and the uplift definition in joint setting have this. Finally, the uplift curve in the joint setting with absolute count shows highly different values in the first 30\% of the population among the different scenarios. This definition seems to favor when more uplift is captured in the top fractions of the population, however, it is not consistent over the three scenarios set up. The scenario with the high amount of treatment observations does not seem to follow this pattern.

\paragraph{\bf Relative Count}

\begin{figure*}[t]
\centering
\begin{subfigure}{.32\textwidth}
  \centering
  \includegraphics[width=1\linewidth]{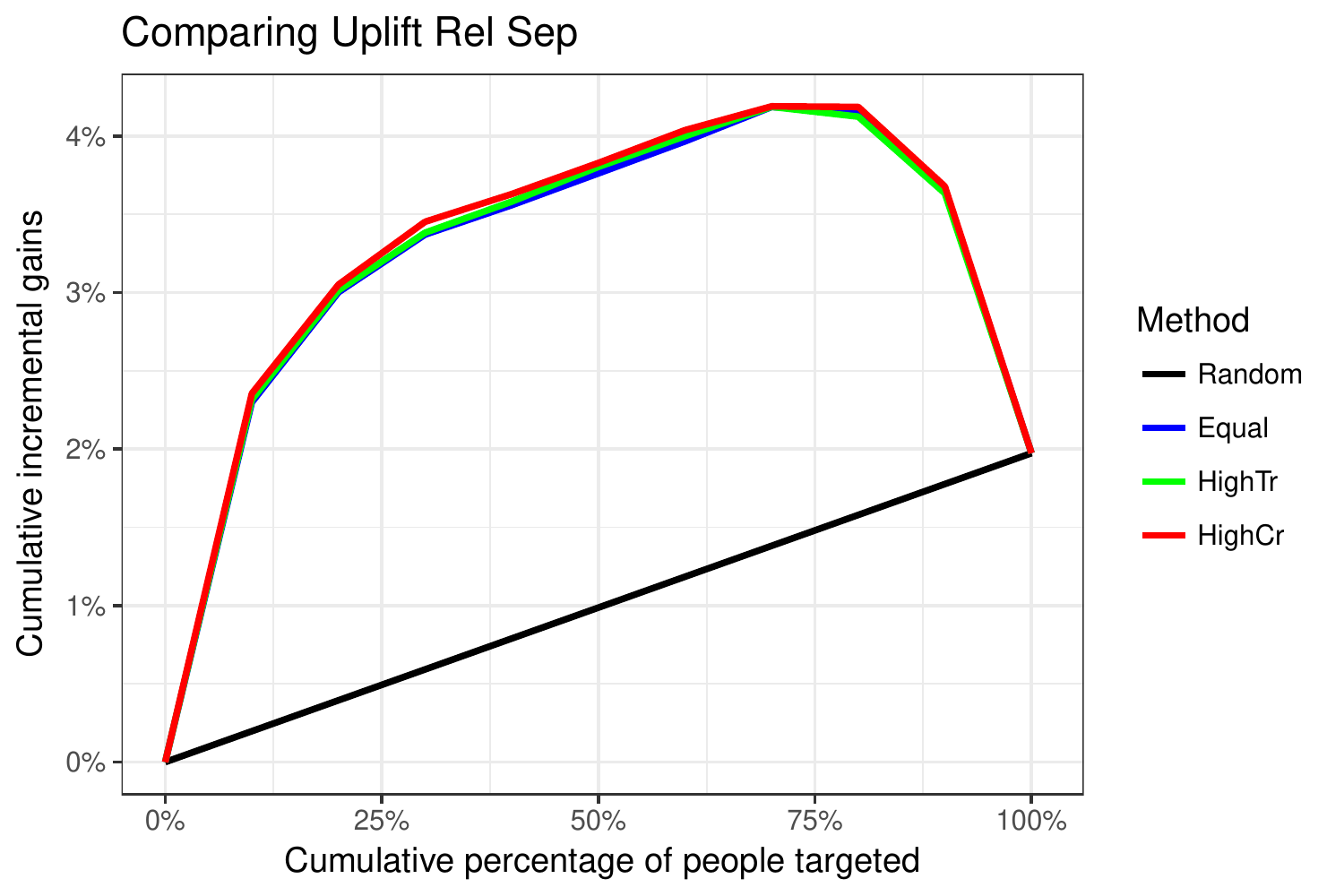}
  \caption{Uplift Curve - Relative Separate}
  \label{fig:SCUpliftCurveRelSep}
\end{subfigure}%
\begin{subfigure}{.32\textwidth}
  \centering
  \includegraphics[width=1\linewidth]{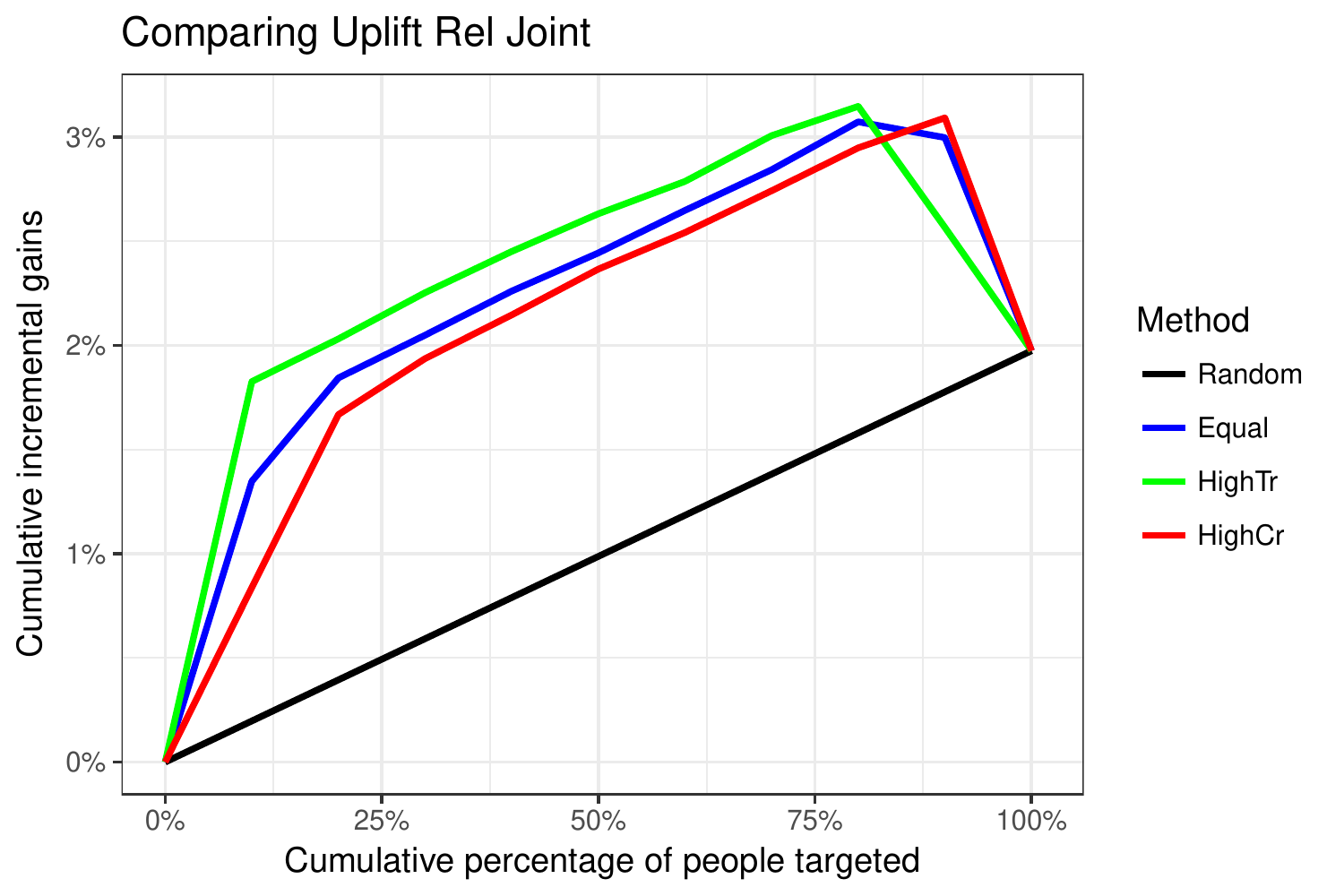}
  \caption{Uplift Curve - Relative Joint}
    \label{fig:SCUpliftCurveRelJoint}
\end{subfigure}%
\begin{subfigure}{.32\textwidth}
  \centering
  \includegraphics[width=1\linewidth]{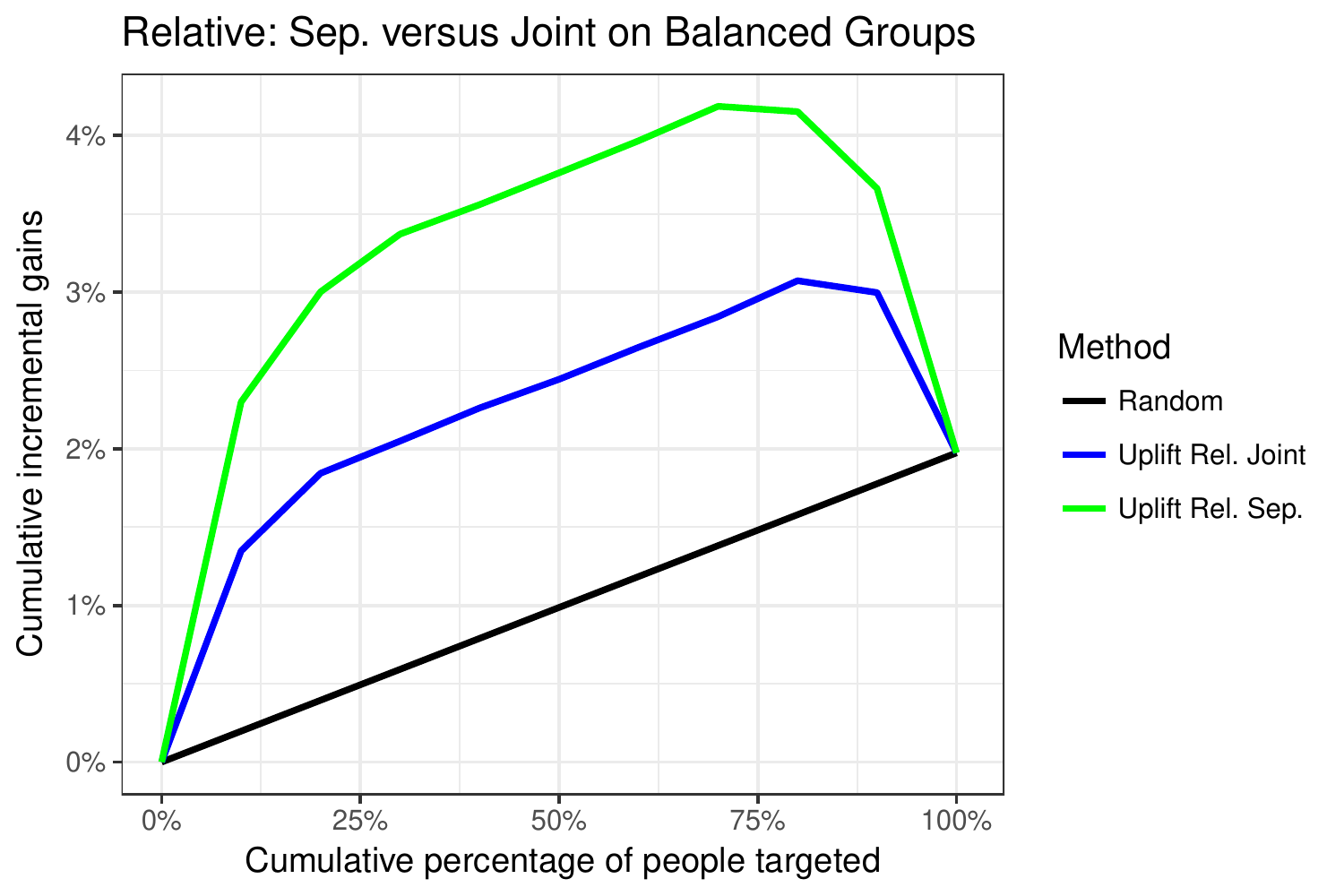}
  \caption{Uplift Curve - Relative Separate versus Relative Joint}
\end{subfigure}

\caption[Experiment 1 - Simulation Results Relative Curves]{Simulation of the Relative Curves in both Separate and Joint setting.}
\label{fig:SimulationCurvesRel}
\end{figure*}

Second, we analyze the behavior of the relative count definitions from Table~\ref{tab:evalMeasures} which has only two definitions: the uplift curve with relative count in separate and joint setting. As a reminder, the qini curve with relative count in the joint setting is identical to its uplift counterpart. Figure~\ref{fig:SCUpliftCurveRelSep} shows the uplift curve in the separate setting. We see no difference in shape of the curve, nor the corresponding AUUC values. The separate setting and the uniform sampling, causes the response rates from both treatment and control to be similar at all targeted proportions of the ranking. When looking at the joint setting in Figure~\ref{fig:SCUpliftCurveRelJoint}, we see a different story. The different sizes of the treatment and control group cause a shift in the corresponding uplift curves. A possible explanation is that one group is overrepresented in the first selected fraction, which causes the uplift to be one-sided. The corresponding AUUC values however do remain close.  

More interestingly, when comparing the separate setting versus the joint setting on the balanced scenario, we see that we get a higher incremental gains curve for the separate setting, and thus a higher AUUC. This is also true for the other scenarios. The reason the uplift is so high, is because in the separate setting the amount TR's is much higher than in the joint setting. In the joint setting we see a lot of CNR's among the first, which heavily influences the possible uplift achieved at the early stages. 

Given the above comparisons on the simulated rankings, we consider the relative setting to be more robust to differences in treatment and control group sizes. The qini curves in the absolute separate setting are also promising and close to the original definition of the qini curve. However, this definition is closely related to the uplift curve in the separate setting. Dividing the value function by $|T|$ results in:
\begin{equation}
V(p) = \frac{R(T, p|T|)}{|T|} - \frac{R(C, p|C|)}{|C|} = \frac{R(T, p|T|) - R(C,p|C|)\frac{|T|}{|C|}}{|T|}
\end{equation}

The two formulations are equivalent up to a constant factor, hence when optimizing the same results will be found. Therefore we select two measures to use in the learning-to-rank setting. The first definition is the most used definition in the literature (Table~\ref{tab:evalMeasures}) and the second definition is desirable due to its joint setting which is closer to the intended use of uplift modeling's approach to rank new instances. 

\subsection{Question 2: Pointwise vs Listwise}
\label{sec:PointwiseVsListwise}

\textit{How well does the pointwise single model approach with the transformed outcome variable compare to a listwise learning-to-rank approach?} The flipped label approach is in effect a pointwise learning-to-rank method in which a single model has to rank treated responders higher than non-responders and control non-responders higher than control responders. Hence, both treatment responders and control non-responders are considered the positive class, whereas the rest is considered the negative class. In a pointwise method, there is no need to separate instances in different query groups. 

We compare the baseline uplift model with standard learning-to-rank techniques. The technique used for the baseline uplift model is extreme gradient boosted trees (from \textit{xgboost}-package in R \cite{R:xgboost}). This technique was chosen due to it being similar to the chosen learning-to-rank technique LambdaMART, which also uses gradient boosted trees.

The flipped label approach will group TR's and CNR's together as the positive class, but no distinction is made between these two groups. Likewise TNR's and CR's are grouped up in the negative class without any distinction between the two. This is very similar to the separate setting in learning-to-rank as the relevant documents of each query have no relation to each other. Therefore, the flipped label approach will be compared to a \textit{separate} listwise learning-to-rank technique, i.e., LambdaMART. For relevance values we choose binary values as this also corresponds with the flipped label approach.

The treatment and control groups will be seen as separate queries, each preferring one type of responder above another (Table~\ref{tab:gainsSep}). Typically, LambdaMART optimizes up to the first $k$ observations (e.g., in search engines people typically only focus on the first 10 results), however, in order to compare with the baseline we use the LambdaMART to optimize over the entire population. So, for this experiment, $k$ will equal the amount of training observations of the largest group (between treatment and control). We use LambdaMART to optimize on four different metrics: $MAP$, $DCG$, $NDCG$ and $PCG$.

\begin{figure*}[t]
\centering
\begin{subfigure}{.32\textwidth}
  \centering
  \includegraphics[width=1\linewidth]{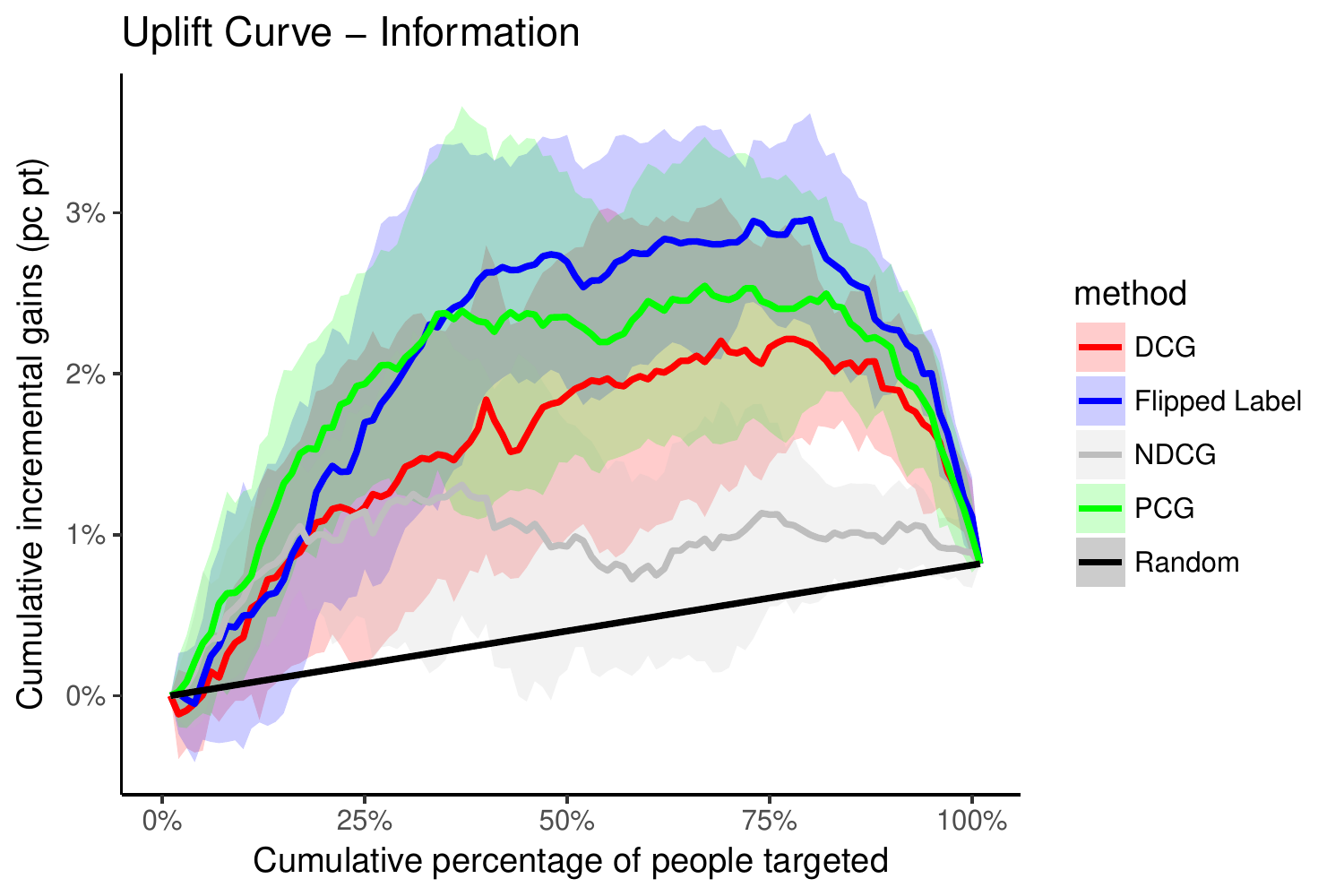}
  \caption{Dataset 1 - Insurance}
  \label{fig:Exp1Insurance}
\end{subfigure}%
\begin{subfigure}{.32\textwidth}
  \centering
  \includegraphics[width=1\linewidth]{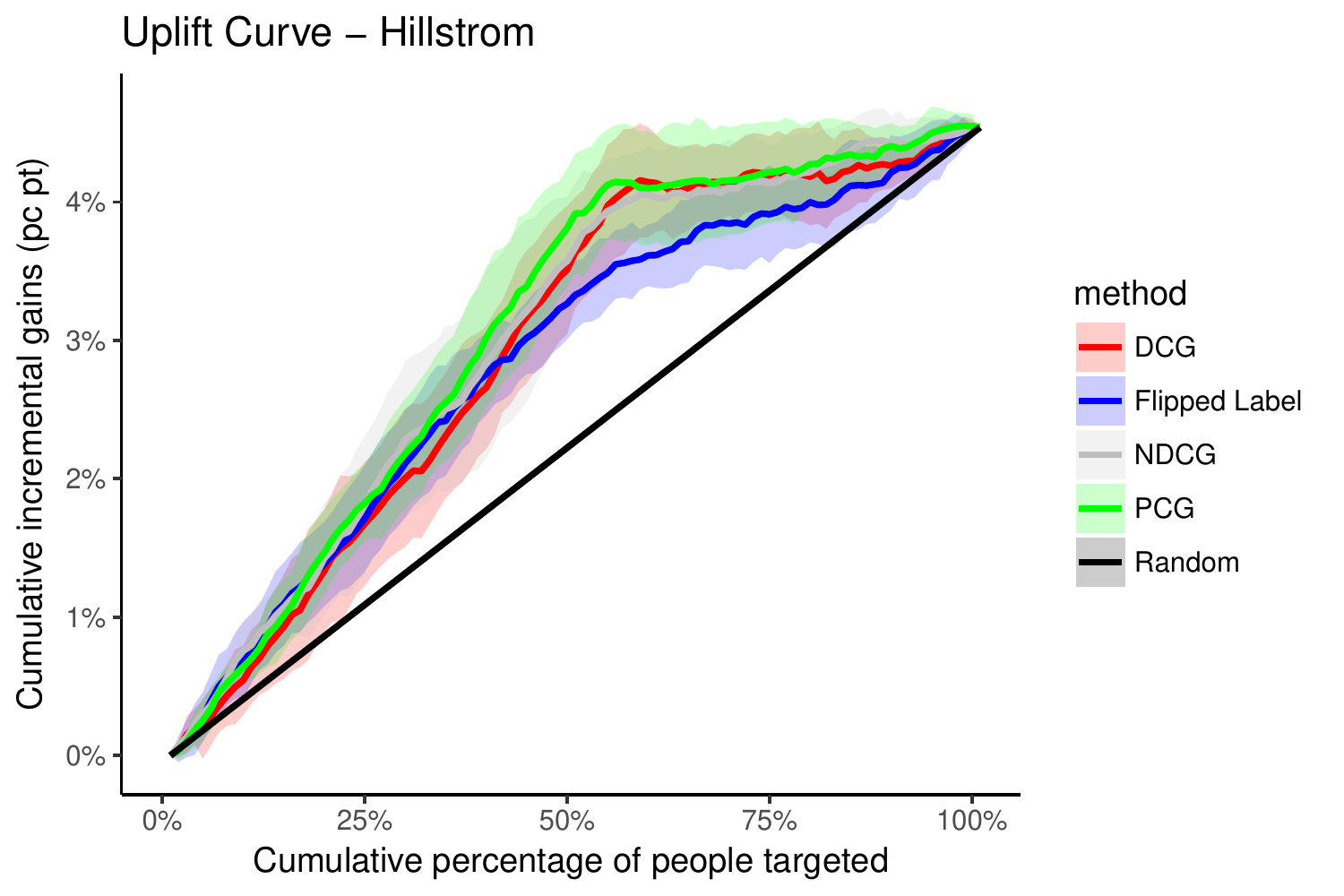}
  \caption{Dataset 2 - Clothing}
    \label{fig:Exp1Clothing}
\end{subfigure}%
\begin{subfigure}{.32\textwidth}
  \centering
  \includegraphics[width=1\linewidth]{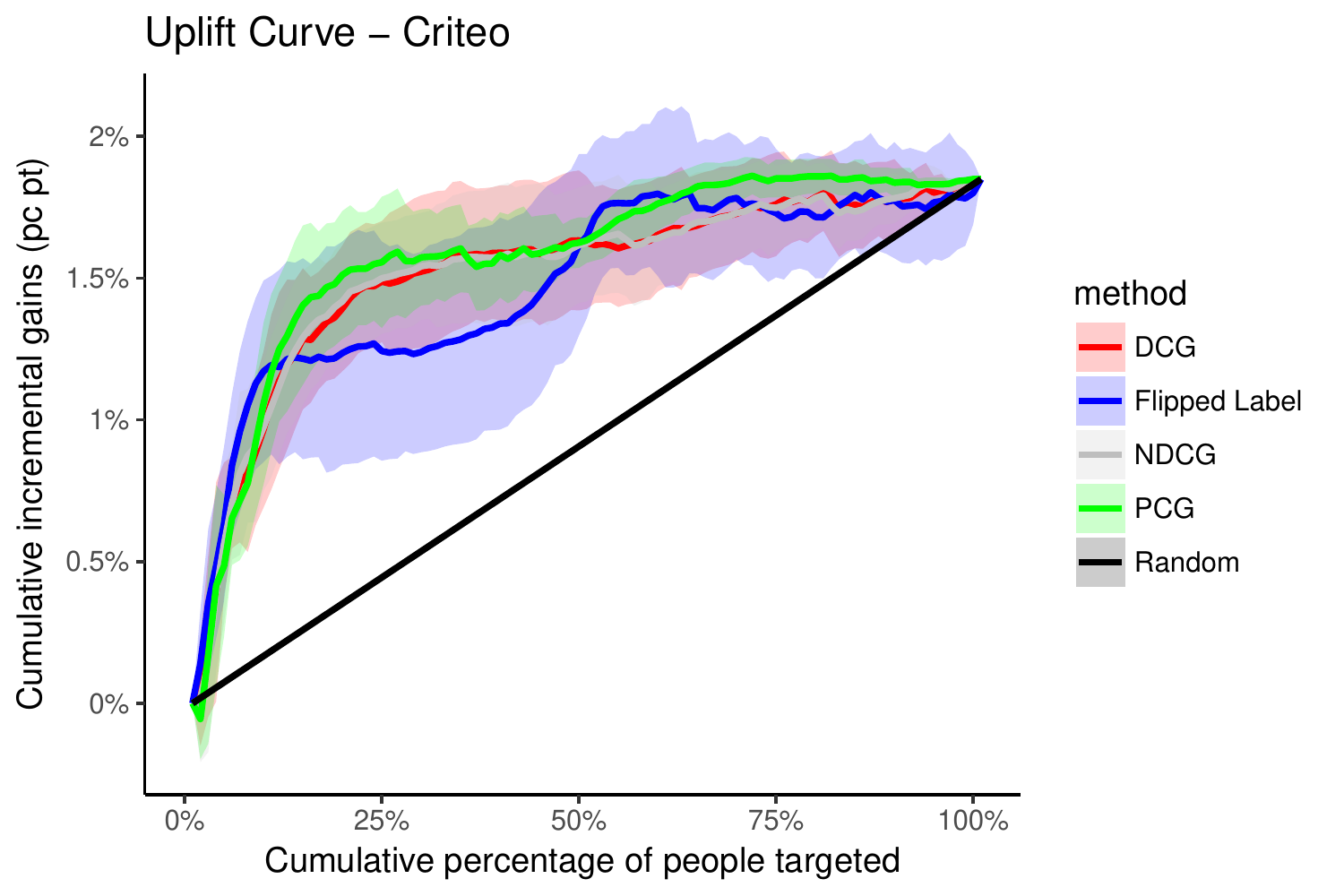}
  \caption{Dataset 3 - Marketing}
    \label{fig:Exp1Marketing}
\end{subfigure}
\caption[Experiment 2 - Results]{Results Experiment 2. The x-axis shows the percentage of the population targeted. The y-axis shows the incremental gains achieved. Black represents random targeting, blue represents the uplift model baseline and the other colors represent the LamdaMART techniques.}
\label{fig:Exp1}
\end{figure*}

We can assess the performance of the models by plotting the Uplift curves and by calculating the AUUC of each technique. Figure~\ref{fig:Exp1} shows the incremental gains curves, using the value function of Table~\ref{tab:evalMeasures} for the separate relative setting of the uplift curve. For dataset 1 (Figure~\ref{fig:Exp1Insurance}), the pointwise approach performs significantly better when compared with the standard LambdaMART approaches (DCG, NDCG and MAP). However, our modified metric PCG shows higher incremental gains at the earlier stages when compared to the pointwise approach. On dataset 2 (Figure~\ref{fig:Exp1Clothing}), the listwise LambdaMART approach with MAP, DCG, NDCG and PCG perform equally well compared to the pointwise approach, and perform significantly better from 40\% of the targeted population onwards. Finally, on dataset 3 (Figure~\ref{fig:Exp1Marketing}), all techniques achieve high incremental gains in the early portions of the targeted population (first 10\%). However, the pointwise approach drops in performance between 10\% and 40\% of the targeted population when compared to the listwise approaches. 

We further analyze the results by analyzing the AUUC values (Table~\ref{tab:Exp1AUUCAbs}). First, we take a look at the AUUC values from the relative separate uplift curve. The pointwise approach performs significantly better than the listwise approaches on the information dataset with only the listwise approach with PCG coming close. The listwise approaches marginally perform better on both the hillstrom and criteo datasets. Second, we take a look at the AUUC values from the relative joint uplift curve. On the first dataset the pointwise approach performs better than the listwise approaches. On the second and third dataset both pointwise and listwise approaches achieve similar results. We do conclude that among all listwise measures, the PCG always leads to the best results. 

Overall listwise approaches on the first dataset achieve better at low percentage values, whereas for higher percentage values slightly worse than the pointwise approach. On the second and third dataset listwise approaches show the same trend, but also achieve equal-to-better results at higher percentages. These results indicate that listwise approaches are a viable alternative to pointwise approaches. 

\begin{table*}[tb]
\begin{center}
\caption[Experiment 2 - Results]{AUUC values when run in a separate setting with absolute relevance values. Both AUUC values of the separate relative uplift curve and the joint relative uplift curve are presented. (*): significantly different from the pointwise approach. Bold: best value on that dataset. \label{tab:Exp1AUUCAbs}}
\begin{adjustbox}{max width=\textwidth}
{\small
\begin{tabular}{l|l|l|l|l|l|l|}
\cline{2-7}
                                               & \multicolumn{3}{l|}{AUUC Uplift Relative Separate}     & \multicolumn{3}{l|}{AUUC Uplift Relative Joint} \\ \hline
\multicolumn{1}{|l|}{Technique}                & Information        & Hillstrom         & Criteo        & Information      & Hillstrom      & Criteo      \\ \hline
\multicolumn{1}{|l|}{Pointwise Flipped-label Approach}  & \textbf{0.02052}  & 0.02858           & 0.01479          & \textbf{0.02050}          & 0.02865               &  0.01418              \\ \hline
\multicolumn{1}{|l|}{Listwise LambdaMART MAP}   & 0.01237(*) & 0.03038 & 0.01556 & 0.01185(*) & 0.03039 & 0.0161(*) \\ \hline
\multicolumn{1}{|l|}{Listwise LambdaMART DCG}   & 0.01520 (*)        & 0.02960          & 0.01522          & 0.01612         & 0.02955              & 0.01580 \\ \hline
\multicolumn{1}{|l|}{Listwise LambdaMART NDCG}  & 0.00935 (*)        & 0.03032          & 0.01523          & 0.00844 (*)          & 0.03051              & 0.01578               \\ \hline
\multicolumn{1}{|l|}{Listwise LambdaMART PCG}   & 0.01938          & \textbf{0.03077} & \textbf{0.01578}          & 0.01894                & \textbf{0.03067}               &  \textbf{0.01662(*)}               \\ \hline
\end{tabular}
}
\end{adjustbox}
\end{center}
\end{table*}

\subsection{Question 3: Comparison of graded relevance labels}
\label{sec:ComparisonGradedRelevanceLabels}

In Section~\ref{sec:UMasL2RMeasure} we identified possible relevance values to use in the learning-to-rank setup. Four possible scenarios can be created depending on whether the relevance values are absolute or relative, and depending on whether the groups are considered separate or as one joint group. Table~\ref{tab:gainValues} shows the relevance values tested in this experiment. We consider both the separate and joint approaches. The learning-to-rank techniques with $DCG$, $NDCG$ and $PCG$ as optimization metric are run on all scenarios from Table~\ref{tab:gainValues} to test their effect in terms of AUUC values. These metrics are chosen because they can handle graded relevances. 

\begin{table*}[t]
\begin{center}
\caption[Experiment 3 - Relevance Values]{Possible relevance values depending on the approach taken. \label{tab:gainValues}}
\begin{adjustbox}{max width=\textwidth}
{\small
\begin{tabular}{|l|l|l|l|l|}
\hline
Setting              & Treatement Responder & Treatment Non-Responder & Control Responder & Control Non-Responder \\ \hline
Absolute Relevance 1 & 1                     & 0                        & 0                  & 1                       \\ \hline
Absolute Relevance 2 & 1                     & 0                        & -1                  & 0                      \\ \hline
Absolute Relevance 3 & 3                    & 1                        & 0                  & 2                      \\ \hline
Relative Relevance   & $\frac{1}{|T|} $                    & 0                        & $-\frac{1}{|C|}$   & 0          \\ \hline
\end{tabular}
}
\end{adjustbox}
\end{center}
\end{table*}

\subsubsection{Separate Queries}

When using separate queries, we notice that in all possible settings, PCG consistently performs best when compared to other metrics. On the first dataset the NDCG performs significantly worse compared to the other approaches, with the absolute third setting being the exception. On the second and third dataset the performance fairly equal to that of the DCG metric. Despite that the PCG with relative labels is exactly the same as optimizing AUUC, using absolute labels performs marginally better on the first two datasets.

\begin{table*}[t]
\begin{center}
\caption[Experiment 3 - Separate Queries]{AUUC values of the \textit{separate} relative uplift curve. Bold: best value in that column, Underline: best value on that dataset.\label{tab:ExpGainSepQuery}}
\begin{adjustbox}{max width=\textwidth}
{\small
\begin{tabular}{l|l|l|l||l|l|l||l|l|l||}
\cline{2-10}
                            & \multicolumn{3}{c||}{Information} & \multicolumn{3}{c||}{Hillstrom} & \multicolumn{3}{c||}{Criteo} \\ \cline{2-10} 
                            & DCG       & NDCG       & PCG      & DCG      & NDCG      & PCG      & DCG     & NDCG     & PCG     \\ \hline
\multicolumn{1}{|l||}{Abs 1} & \textbf{0.01520} & 0.00935 & \textbf{\underline{0.01938}} & 0.02960 & \textbf{0.03032} & \textbf{\underline{0.03077}} & 0.01522 & 0.01523 & 0.01578 \\ 
\multicolumn{1}{|l||}{Abs 2} & \textbf{0.01520} & 0.00678 & \textbf{\underline{0.01938}} & 0.02960 & 0.02893 & \textbf{\underline{0.03077}} & 0.01522 & \textbf{0.01555} & 0.01578 \\ 
\multicolumn{1}{|l||}{Abs 3} & \textbf{0.01520} & \textbf{0.01524} & \textbf{\underline{0.01938}} & 0.0296 & 0.02953 & \textbf{\underline{0.03077}} & 0.01522 & 0.01538 & 0.01578 \\ 
\multicolumn{1}{|l||}{Rel}   & 0.01382 & 0.00677 & 0.01829 & \textbf{0.02961} & 0.02893 & 0.03055 & \textbf{0.01549} & \textbf{0.01555} & \textbf{\underline{0.01601}} \\ \hline
\end{tabular}
}
\end{adjustbox}
\end{center}
\end{table*}

\subsubsection{Joint Queries}
When using joint queries, we notice that PCG consistently outperforms other metrics again. Furthermore, the absolute relevance values now perform best over all datasets and techniques. The third absolute setting perform best on the first two datasets, whereas the second setting performs best on the last. Finally, the NDCG shows improved and nearly equal results to the DCG on all datasets including the first. When the NDCG has only one query, the results should indeed be equal to that of DCG metric. 

\begin{table*}[t]
\begin{center}
\caption[Experiment 3 - Joint Queries]{AUUC values of the \textit{joint} relative uplift curve. Bold: best value in that column, Underline: best value on that dataset.}
\label{tab:ExpGainJointQuery}
\begin{adjustbox}{max width=\textwidth}
{\small
\begin{tabular}{l|l|l|l||l|l|l||l|l|l||}
\cline{2-10}
                            & \multicolumn{3}{c||}{Information} & \multicolumn{3}{c||}{Hillstrom} & \multicolumn{3}{c||}{Criteo} \\ \cline{2-10} 
                            & DCG       & NDCG       & PCG      & DCG      & NDCG      & PCG      & DCG     & NDCG     & PCG     \\ \hline
\multicolumn{1}{|l||}{Abs 1} & 0.01396 & 0.01452 & 0.01940 & 0.02957 & 0.02957 & 0.03002 & 0.01497 & 0.01494 & 0.01469 \\ 
\multicolumn{1}{|l||}{Abs 2} & 0.01101 & 0.01116 & 0.01536 & 0.02935 & 0.02935 & 0.03051 &\textbf{0.01607} &\textbf{ 0.01607} & \textbf{\underline{0.01669}}\\ 
\multicolumn{1}{|l||}{Abs 3} & \textbf{0.01563} & \textbf{0.01473} & \textbf{\underline{0.02300}} & \textbf{0.02968} & \textbf{0.02968} & \textbf{\underline{0.03063}} & 0.01568 & 0.01568 & 0.01536\\ 
\multicolumn{1}{|l||}{Rel}   & 0.01052 & 0.01031 & 0.01573 & 0.02954 & 0.02954 & 0.03027 & 0.01541 & 0.01543 & 0.01554\\  \hline
\end{tabular}
}
\end{adjustbox}
\end{center}
\end{table*}

These results are somewhat surprising, while they do confirm PCG to be better suited, the use of less theoretically motivated relevance values is shown to be able to give good results too. Especially the use of the third absolute setting is interesting, as it is the only setting that values 'control non-responders' over 'treated non-responders'. 

\subsection{Question 4: Optimisation through k-value}
\label{sec:RQ4}

One of the properties of LambdaMART and other learning-to-rank systems is their ability to optimize on a specific targeting depth or $k$ value. The $k$ value indicates on how many documents/observations the ranking needs to be optimized. In our experiments this has always been the size of the largest group (treatment or control) from our datasets. This is the equivalent of optimizing our ranker for the entire dataset, that is, scoring the ranking of all instances. 

In uplift modeling, as in information retrieval, one is often interested in only a fraction of the top ranked instances, as dictated for example, by the campaign budget that limits the amount of customers that can be treated in a marketing setting \cite{Devriendt:2018b}. In this experiment, we check whether using a specific $k$ value leads to increased performance in the top proportions of our data. We optimize our learning-to-rank techniques on $k$-values equal to: 10\%, 30\% and 50\%. For each of these settings, we then check the uplift curves and the AUUC values at all corresponding cut-offs. Additionally we compare the results with our previous experiments in which we optimize for 100\% of the data. We present the results of the PCG metric with relative gains, as it is closest to the theoretical goal of optimizing AUUC. The results for DCG and NDCG are presented in Appendix C. The same experiments for the third absolute relevance setting is given in Appendix D.

\begin{figure*}[t]
\centering
\begin{subfigure}{.32\textwidth}
  \centering
  \includegraphics[width=1\linewidth]{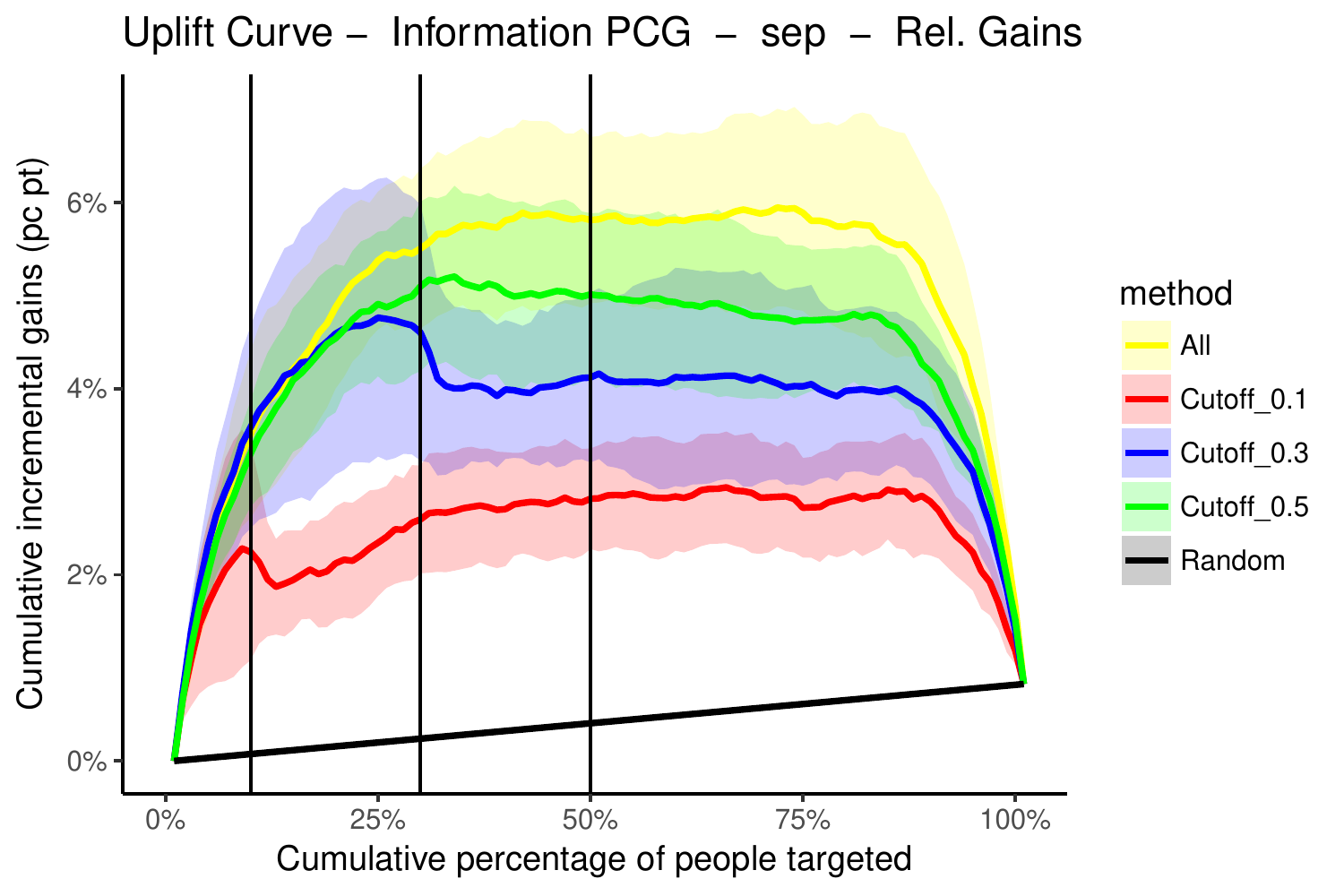}
  \caption{Information - optimization}
\end{subfigure} %
\begin{subfigure}{.32\textwidth}
  \centering
  \includegraphics[width=1\linewidth]{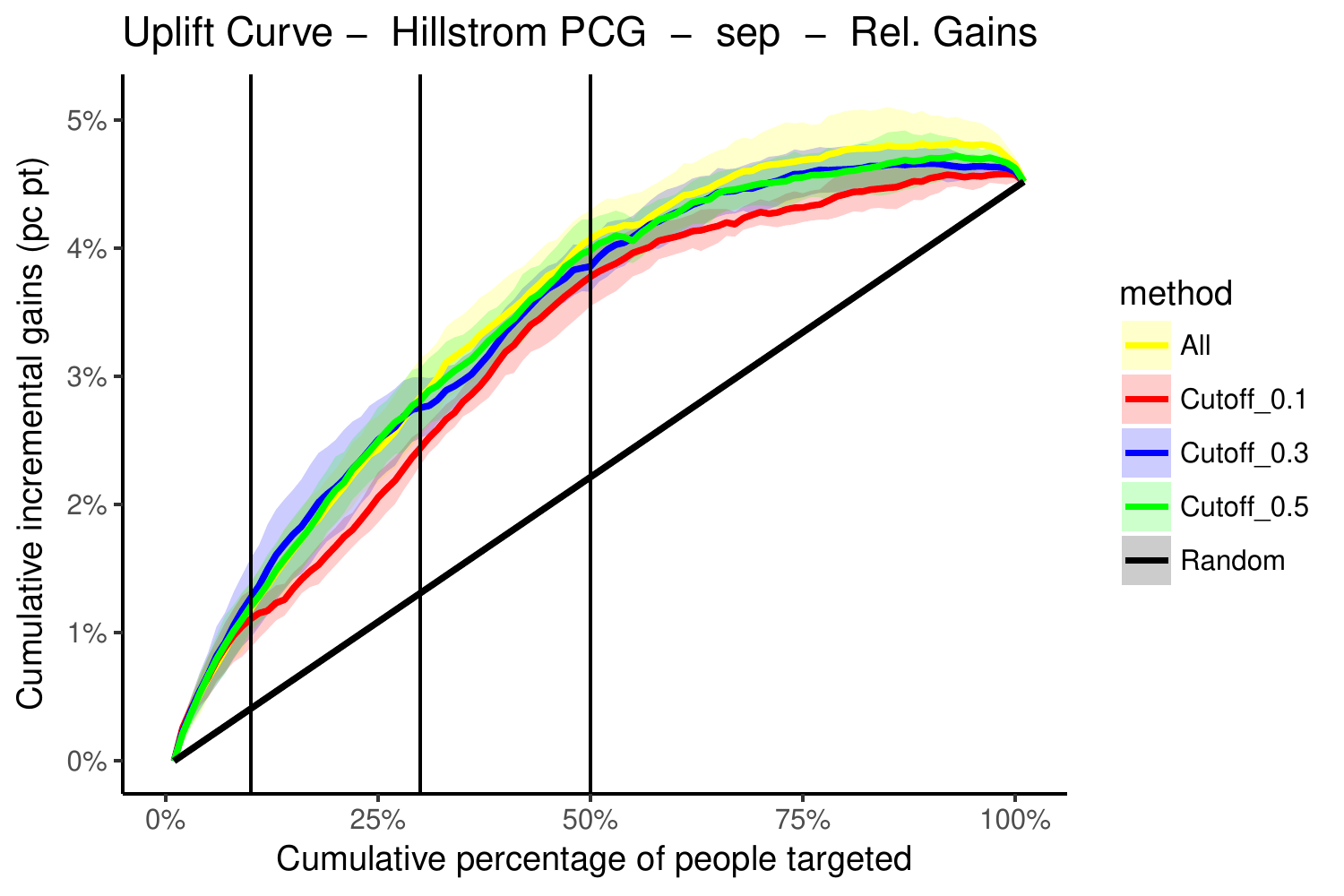}
  \caption{Hillstrom - optimization}
\end{subfigure} %
\begin{subfigure}{.32\textwidth}
  \centering
  \includegraphics[width=1\linewidth]{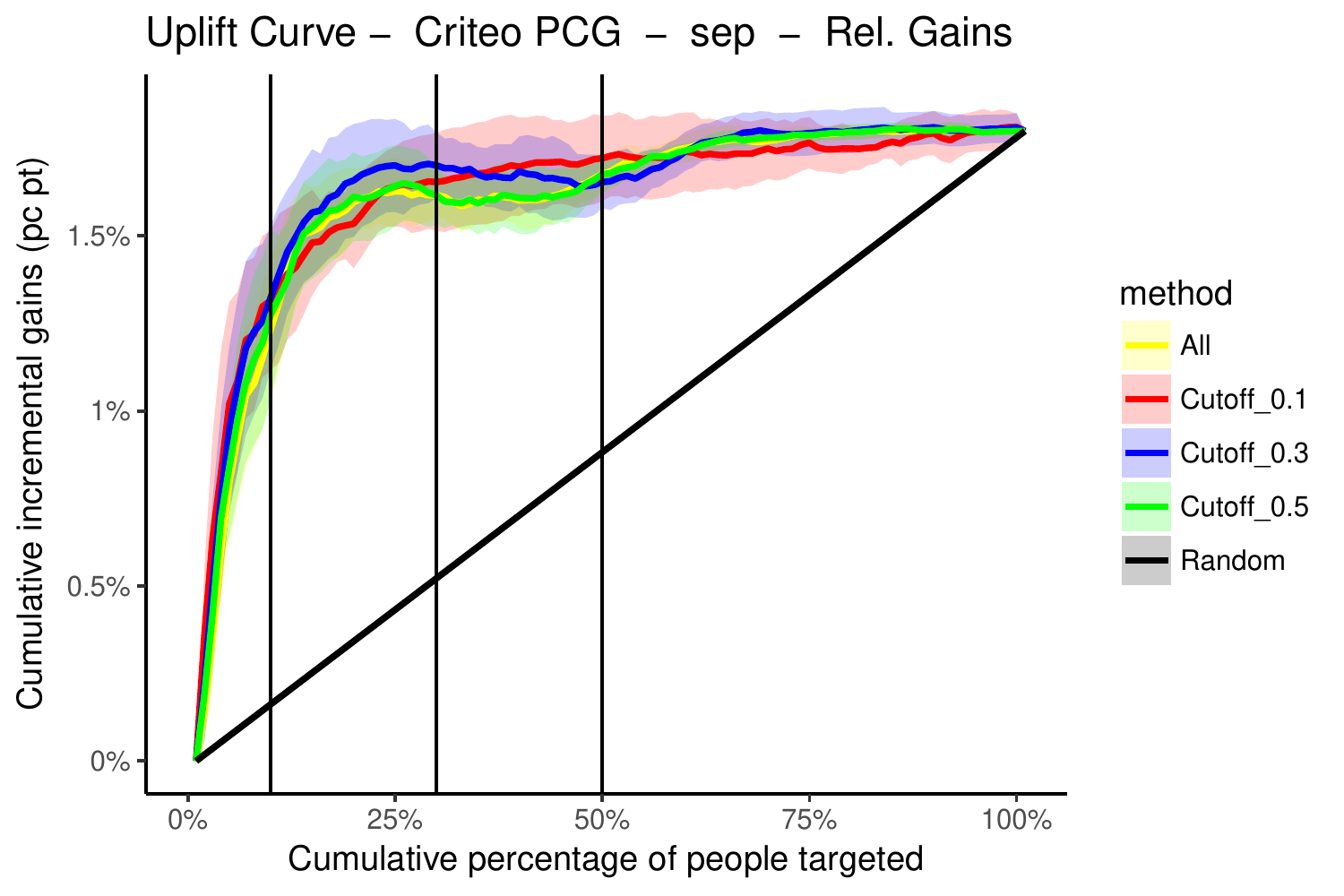}
  \caption{Criteo - optimization}
\end{subfigure}

\begin{subfigure}{.32\textwidth}
  \includegraphics[width=1\linewidth]{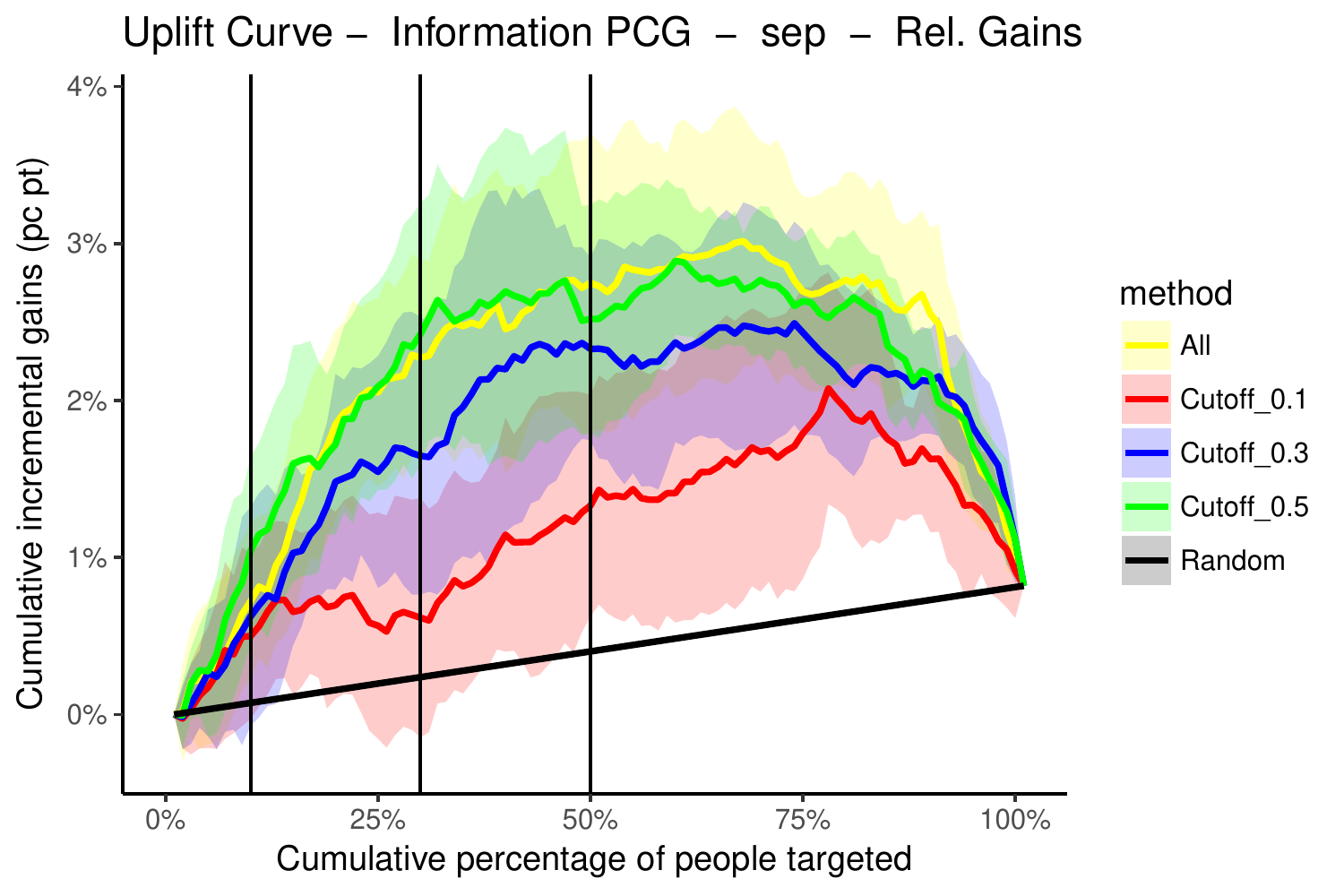}
  \caption{Information - generalization}
\end{subfigure} %
\centering
\begin{subfigure}{.32\textwidth}
  \centering
  \includegraphics[width=1\linewidth]{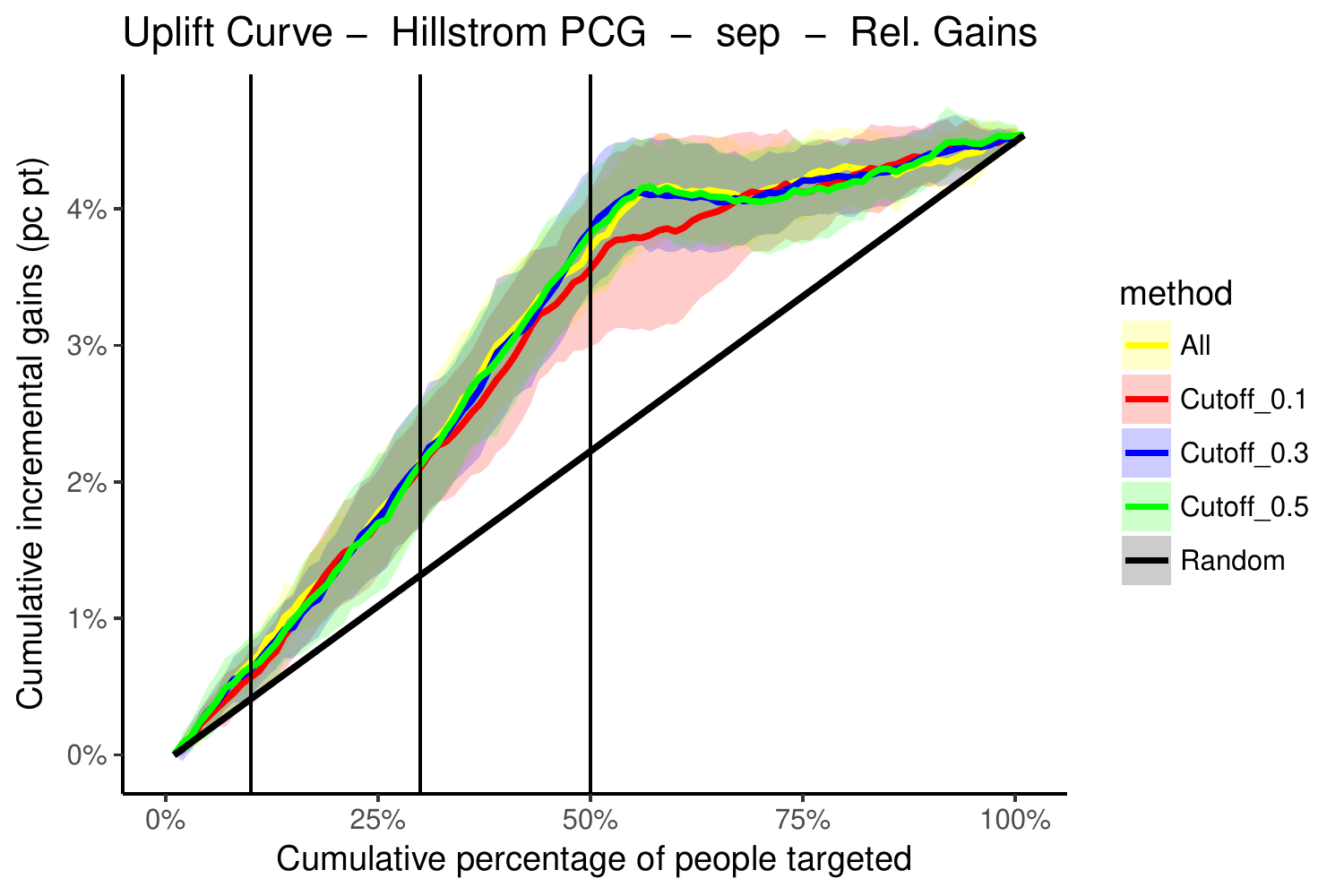}
  \caption{Hillstrom - generalization}
\end{subfigure} %
\begin{subfigure}{.32\textwidth}
  \centering
  \includegraphics[width=1\linewidth]{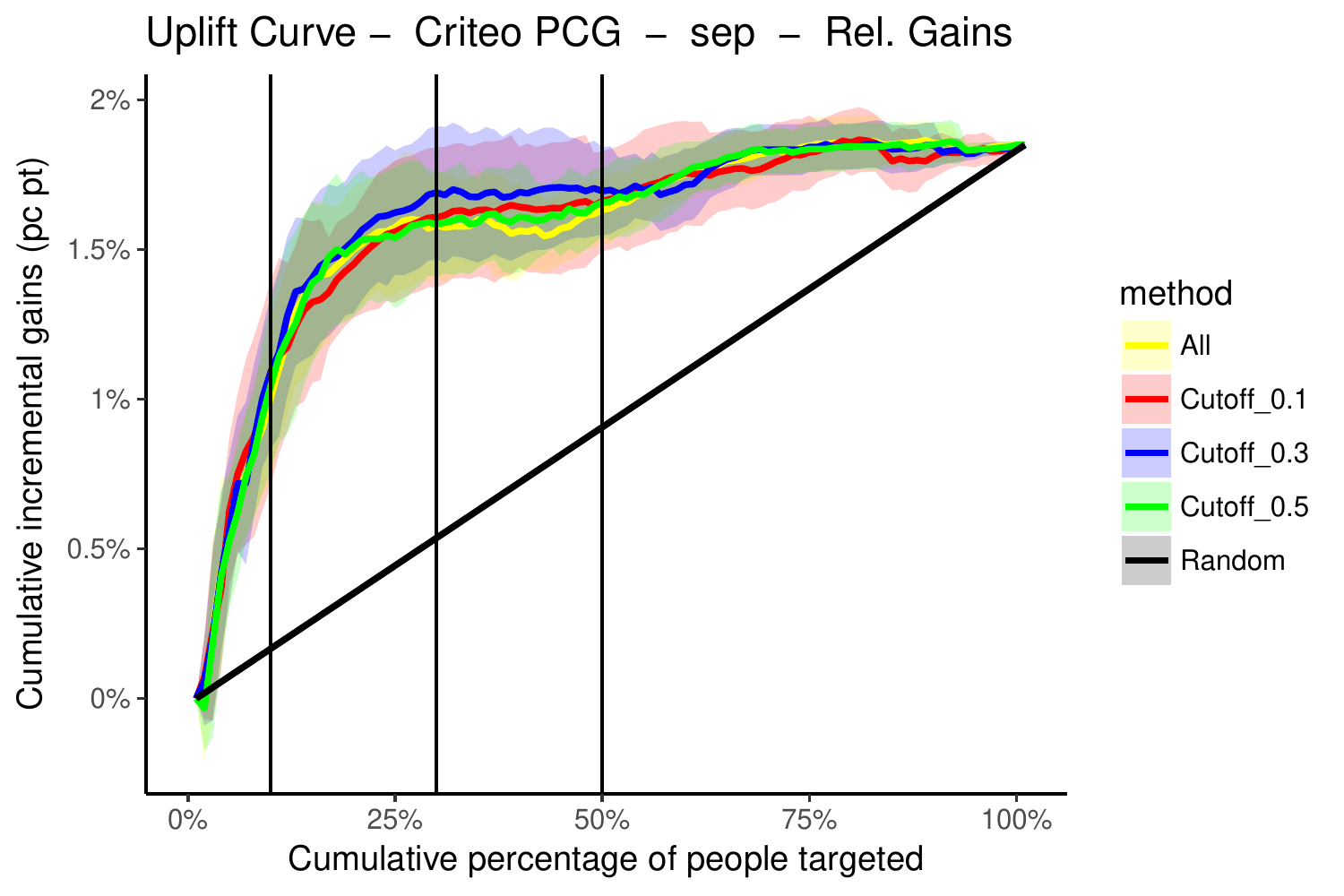}
  \caption{Criteo - generalization}
\end{subfigure}
\caption[Experiment 4 - Results - Separate Setting]{Separate setting at multiple cutoffs with relative gains. The first row is tested on the training set. The second row is tested on the test set. \label{fig:Exp4SepSetting}} 
\end{figure*}

\begin{figure*}[t]
\centering
\begin{subfigure}{.32\textwidth}
  \centering
  \includegraphics[width=1\linewidth]{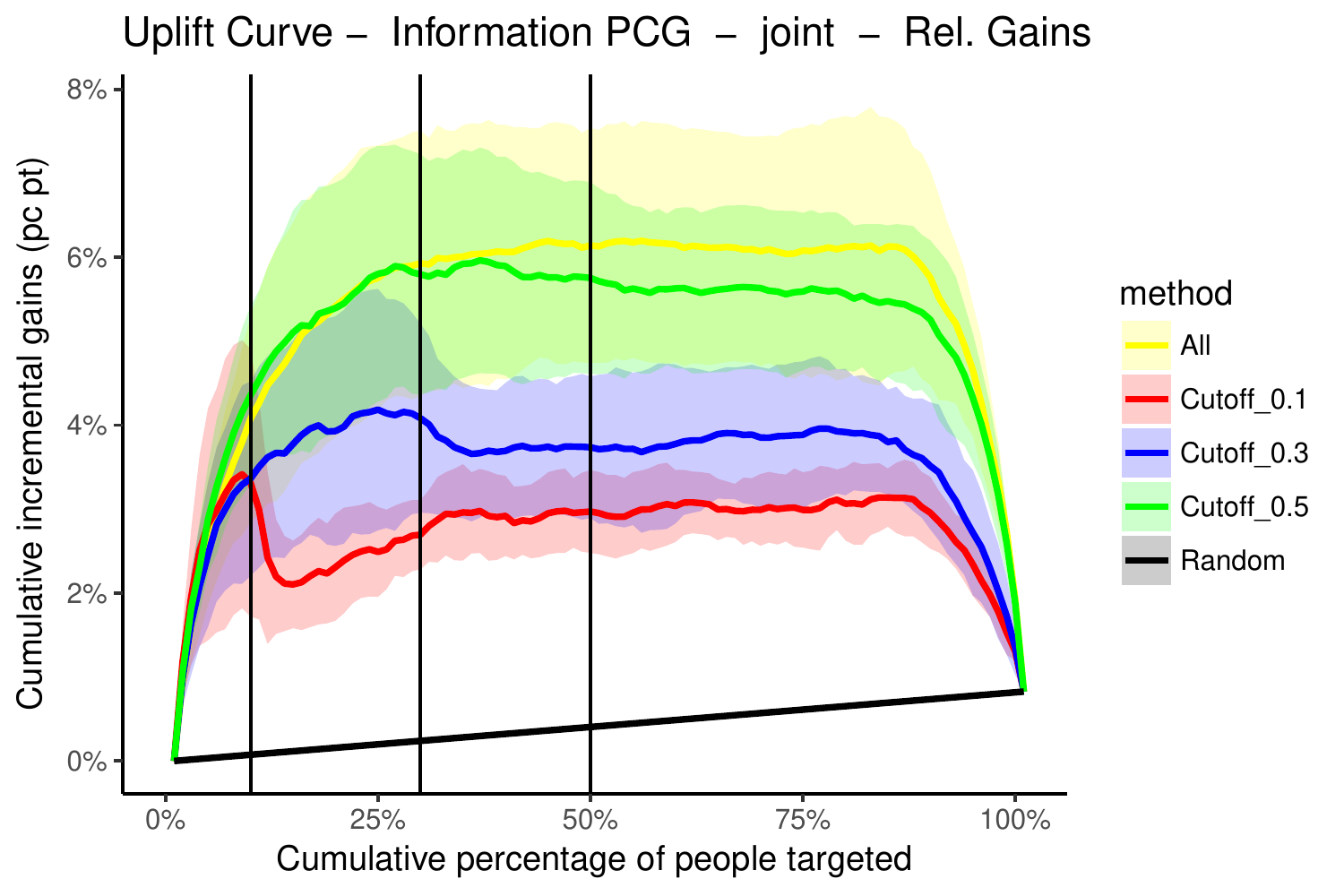}
  \caption{Information - optimization}
\end{subfigure} %
\begin{subfigure}{.32\textwidth}
  \centering
  \includegraphics[width=1\linewidth]{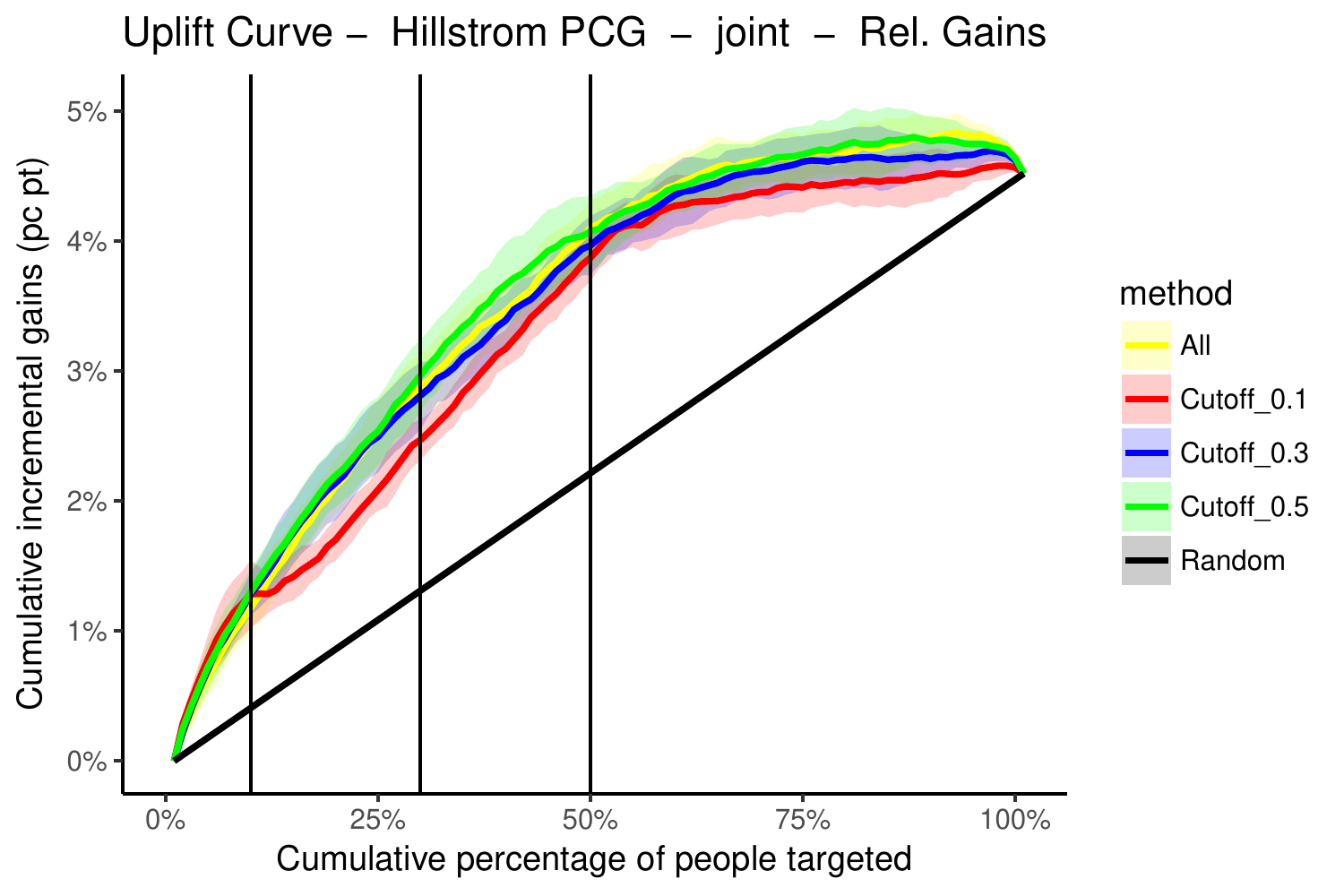}
  \caption{Hillstrom - optimization}
\end{subfigure} %
\begin{subfigure}{.32\textwidth}
  \centering
  \includegraphics[width=1\linewidth]{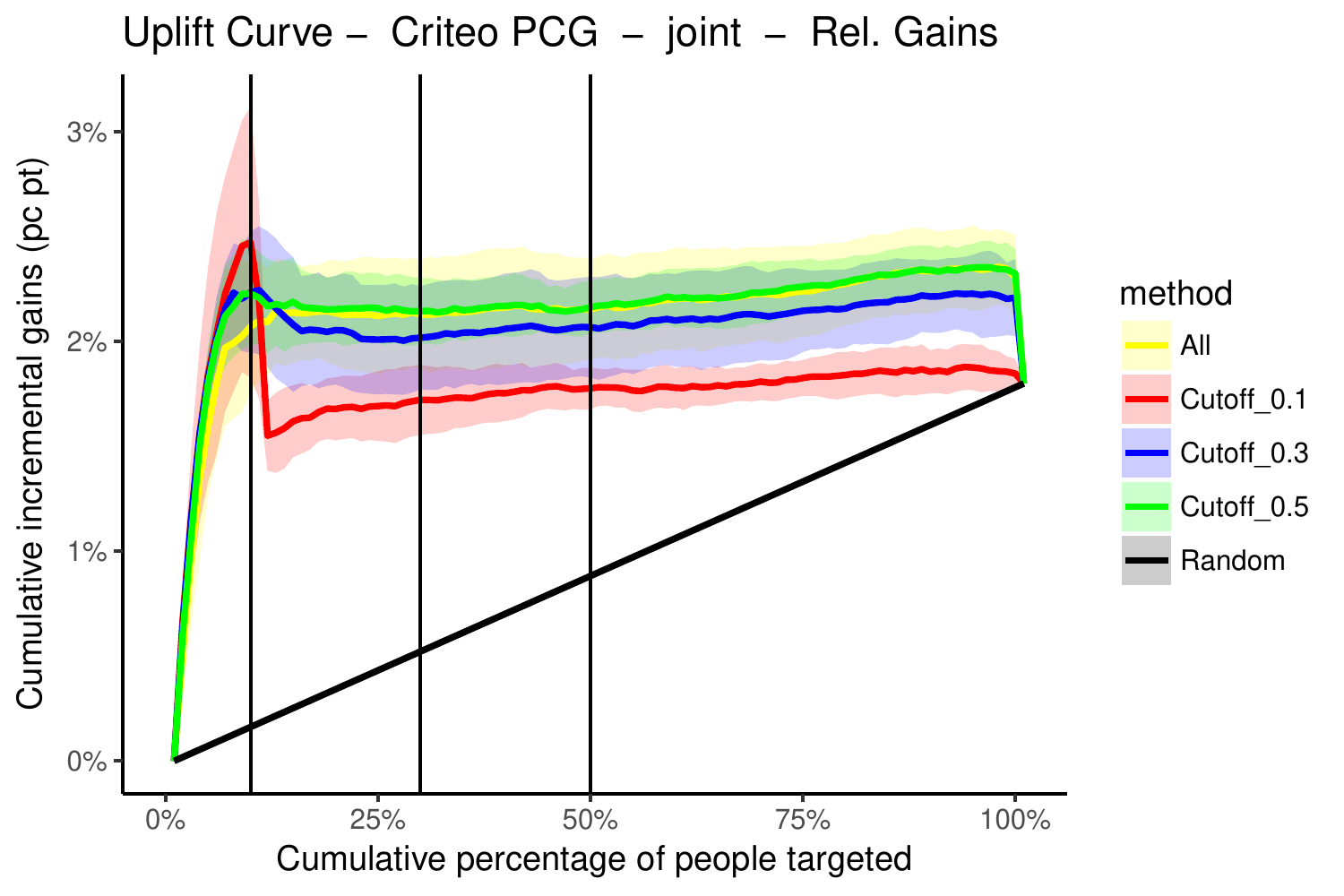}
  \caption{Criteo - optimization}
\end{subfigure}

\begin{subfigure}{.32\textwidth}
  \includegraphics[width=1\linewidth]{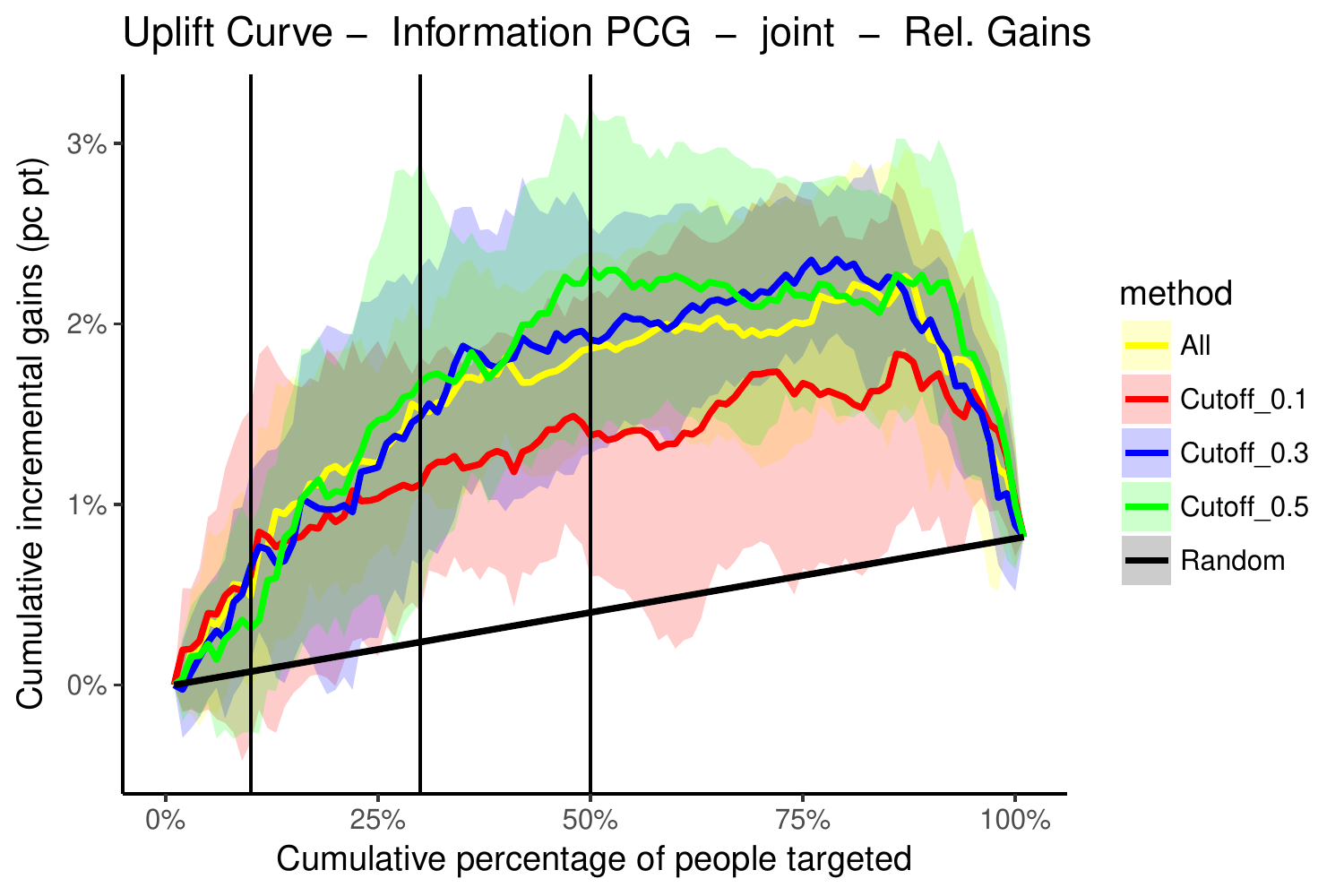}
  \caption{Information - generalization}
\end{subfigure} %
\centering
\begin{subfigure}{.32\textwidth}
  \centering
  \includegraphics[width=1\linewidth]{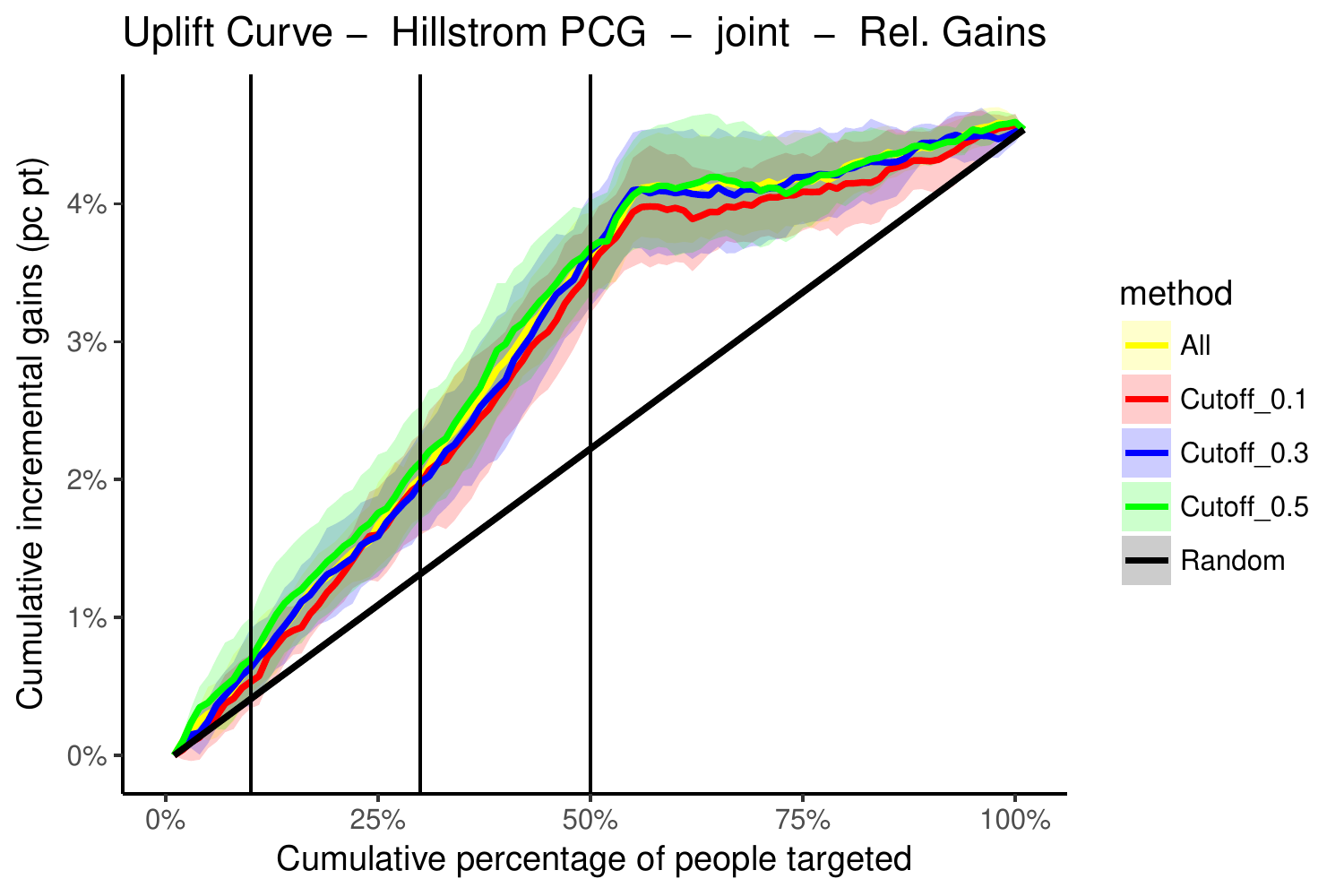}
  \caption{Hillstrom - generalization}
\end{subfigure} %
\begin{subfigure}{.32\textwidth}
  \centering
  \includegraphics[width=1\linewidth]{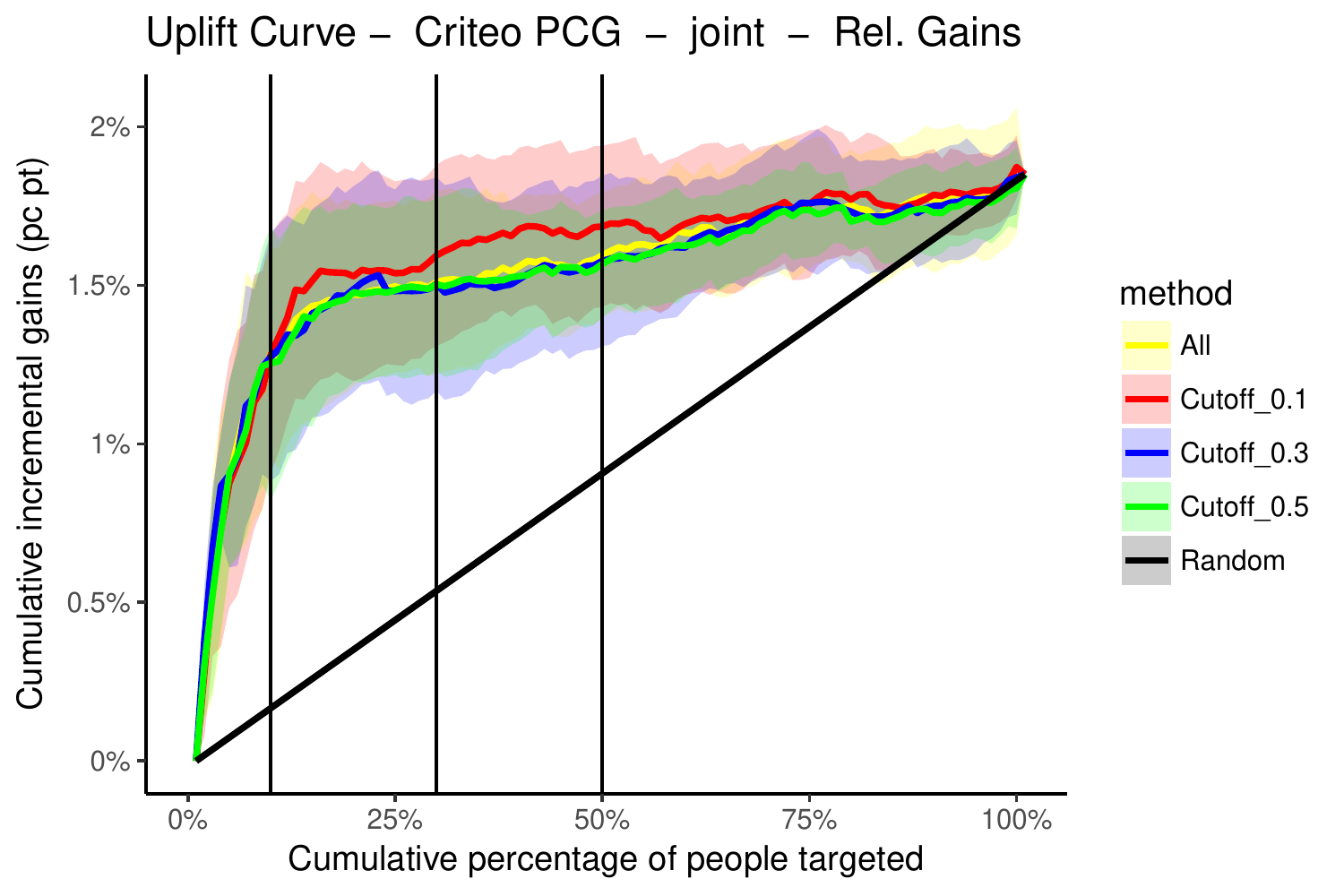}
  \caption{Criteo - generalization}
\end{subfigure}
\caption[Experiment 4 - Results - Joint Setting]{Experiment 4 - PCG for Joint Setting at multiple cutoffs with relative gains. The first row is tested on the training set. The second row is tested on the test set. \label{fig:Exp4JointSetting}}
\end{figure*}

When taking a look at the actual uplift curves we present two perspectives. The optimization perspective, which shows the results of the models when evaluated on the training set (which is only indicative of the effect of training), and the generalization perspective, which shows the results of the model on the test set (the proper way to evaluate the model).  The results are shown in Figure~\ref{fig:Exp4SepSetting}, for the separate setting, and~\ref{fig:Exp4JointSetting}, for the joint setting.

The optimization perspective shows us in both the separate and joint setting that the learning-to-rank techniques can optimize for a specific threshold. We can see that these curves have change-points, after which their behavior changes. This is especially pronounced for the top-10\% trained models in the joint setting, which show a peak followed by a decline around the 10\% mark. Overall, the effect of optimization is most clearly visible in the joint setting.

When investigating the effect on the test set, e.g., how well the results generalize to new data, the figures provide a less clear answer. Table~\ref{tab:MultipleCutoff} provides a closer look by showing the results on each dataset from the generalization perspective at the different cut-offs. The bold values indicate whether a technique with a specific $k$-value has a better AUUC-value than when using all data (i.e., 100\%). We first take a look at the AUUC values of the relative separate uplift curve. In some cases, optimizing for a lower 'k' value shows better performance than when targeting the entire dataset. This is particularly the case on the criteo dataset. However, all of those results are not significantly different when performing a Student's t-test with p-value 0.05. Only PCG at 10\% is significantly worse than PCG at 100\% for the information dataset and cutoffs $30\%, 50\%, 100\%$. In general, we see that there is no direct relation between learning for a specific cut-off percentage and the results it obtains for that percentage on the test set, nor for other precentages. This is hence a negative result: while learning to optimize AUUC at a specific cut-off is possible, on these three datasets there is no significant benefit compared to learning on the total AUUC.

\begin{table*}[t]
\begin{center}
\caption[Experiment 4 - Results]{AUUC-values at multiple cut-offs for all datasets, left half is in the separate setting and right half in the joint setting. Bold values indicate performance of PCG@k is higher than PCG@100\% in that column. (-) indicates statistically significantly worse results with a students t-test with p-value 0.05, all other results are not statistically significantly better or worse than PCG@100\%.}
\label{tab:MultipleCutoff}
\begin{adjustbox}{max width=\textwidth}
{\small
\begin{tabular}{ll||l|l|l|l||l|l|l|l|}
\cline{3-10}
                                                   &              & \multicolumn{4}{c||}{\textit{Separate} Relative Uplift AUUC at cutoff} & \multicolumn{4}{c|}{\textit{Joint} Relative Uplift AUUC at cutoff} \\ 
                                                   &              & 10\%          & 30\%          & 50\%         & 100\%         & 10\%         & 30\%         & 50\%         & 100\%        \\ \hline \hline

\multicolumn{1}{|l|}{\multirow{4}{*}{Information}} & PCG @ 100\% & 0.00037 & 0.00384 & 0.0090 & \underline{0.02178} & 0.00034 & \underline{0.00271} & 0.00616 & 0.01573 \\ \cline{2-10} 
\multicolumn{1}{|l|}{}                             & PCG @ 10\% & 0.00026 & 0.00159(-) & 0.00368(-) & 0.0115(-) & \textbf{\underline{0.00040}} & 0.00232 & 0.00495 & 0.01264   \\ 
\multicolumn{1}{|l|}{}                             & PCG @ 30\% & 0.00030 & 0.00299 & 0.00731 & 0.01819 & 0.00030 & 0.00246 & 0.00611 & \textbf{0.01608} \\ 
\multicolumn{1}{|l|}{}                             & PCG @ 50\% & \textbf{\underline{0.00049}} & \textbf{\underline{0.00419}} & \textbf{\underline{0.00942}} & 0.02143 & 0.00021 & 0.00257 & \textbf{\underline{0.00646}} & \textbf{\underline{0.0169}} \\ \hline \hline

\multicolumn{1}{|l|}{\multirow{4}{*}{Hillstrom}}   & PCG @ 100\% & 0.00037 & \underline{0.00332} & \underline{0.00947} & \underline{0.0306} & 0.00037 & 0.00319 & 0.00905 & 0.03027 \\ \cline{2-10} 
\multicolumn{1}{|l|}{}                             & PCG @ 10\% & 0.00032 & 0.00318 & 0.00901 & 0.0298 & 0.00029 & 0.00296 & 0.00853 & 0.0292 \\ 
\multicolumn{1}{|l|}{}                             & PCG @ 30\% & 0.00037 & 0.00323 & 0.00934 & 0.03045 & 0.00035 & 0.00313 & 0.00887 & 0.03001 \\ 
\multicolumn{1}{|l|}{}                             & PCG @ 50\% & \textbf{\underline{0.00038}} & 0.00321 & 0.00935 & 0.03043 & \textbf{\underline{0.00043}} & \textbf{\underline{0.00347}} & \textbf{\underline{0.0095}} & \textbf{\underline{0.03073}} \\ \hline \hline

\multicolumn{1}{|l|}{\multirow{4}{*}{Criteo}}      & PCG @ 100\% & 0.00059 & 0.00354 & 0.00671 & 0.01575 & 0.00088 & 0.0038 & 0.00691 & 0.01554   \\ \cline{2-10}
\multicolumn{1}{|l|}{}                             & PCG @ 10\% & \textbf{\underline{0.00063}} & 0.00353 & \textbf{0.00681} & \textbf{0.01577} & 0.00087 & \textbf{\underline{0.00392}} & \textbf{\underline{0.00725}} & \textbf{\underline{0.01599}}  \\ 
\multicolumn{1}{|l|}{}                             & PCG @ 30\% & \textbf{0.00062} & \textbf{\underline{0.00368}} & \textbf{\underline{0.00706}} & \textbf{\underline{0.01608}} & \textbf{\underline{0.00092}} & \textbf{0.00383} & 0.00689 & 0.01545  \\ 
\multicolumn{1}{|l|}{}                             & PCG @ 50\% & 0.00058 & 0.00353 & \textbf{0.00676} & \textbf{0.0158} & 0.00088 & 0.00377 & 0.00684 & 0.01533   \\ \hline
\end{tabular}
}
\end{adjustbox}
\end{center}
\end{table*}

\subsection{Question 5: LambdaMART PCG vs state-of-the-art uplift modeling}
\label{sec:RQ5}

In a final experiment we test the listwise LambdaMART technique with the PCG metric against state-of-the-art uplift modeling techniques. Experiment 2 (Section~\ref{sec:PointwiseVsListwise}) demonstrated the learning-to-rank techniques capabilities against Lai's approach, but this time we include other uplift modeling techniques. The additional techniques are the dummy treatment approach \cite{Lo:2002:TLM}, the two model approach \cite{Hansotia:2002} and the uplift random forest \cite{Guelman:2014}. From these techniques the uplift random forest has the most consistent performances according to a previous benchmark \cite{Devriendt:2018}. For the learning-to-rank technique we choose the LambdaMART PCG with relative relevance values in the separate setting. We have also done the experiment with absolute relevance values, however, the results are not presented as the conclusions remain the same.

\begin{figure*}[t]
\centering
\begin{subfigure}{.32\textwidth}
  \centering
  \includegraphics[width=1\linewidth]{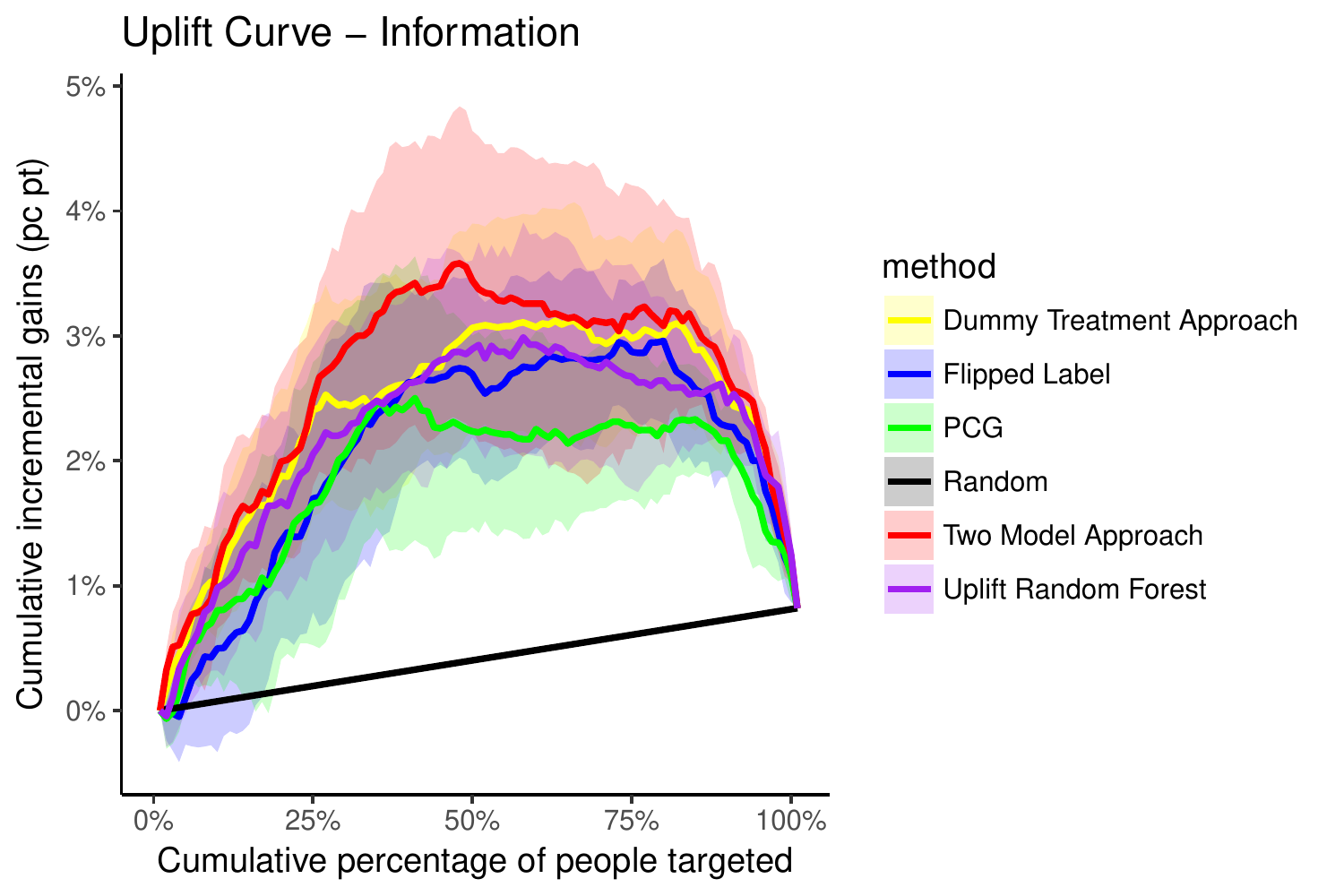}
  \caption{Dataset 1 - Insurance}
  \label{fig:Exp5Insurance}
\end{subfigure}%
\begin{subfigure}{.32\textwidth}
  \centering
  \includegraphics[width=1\linewidth]{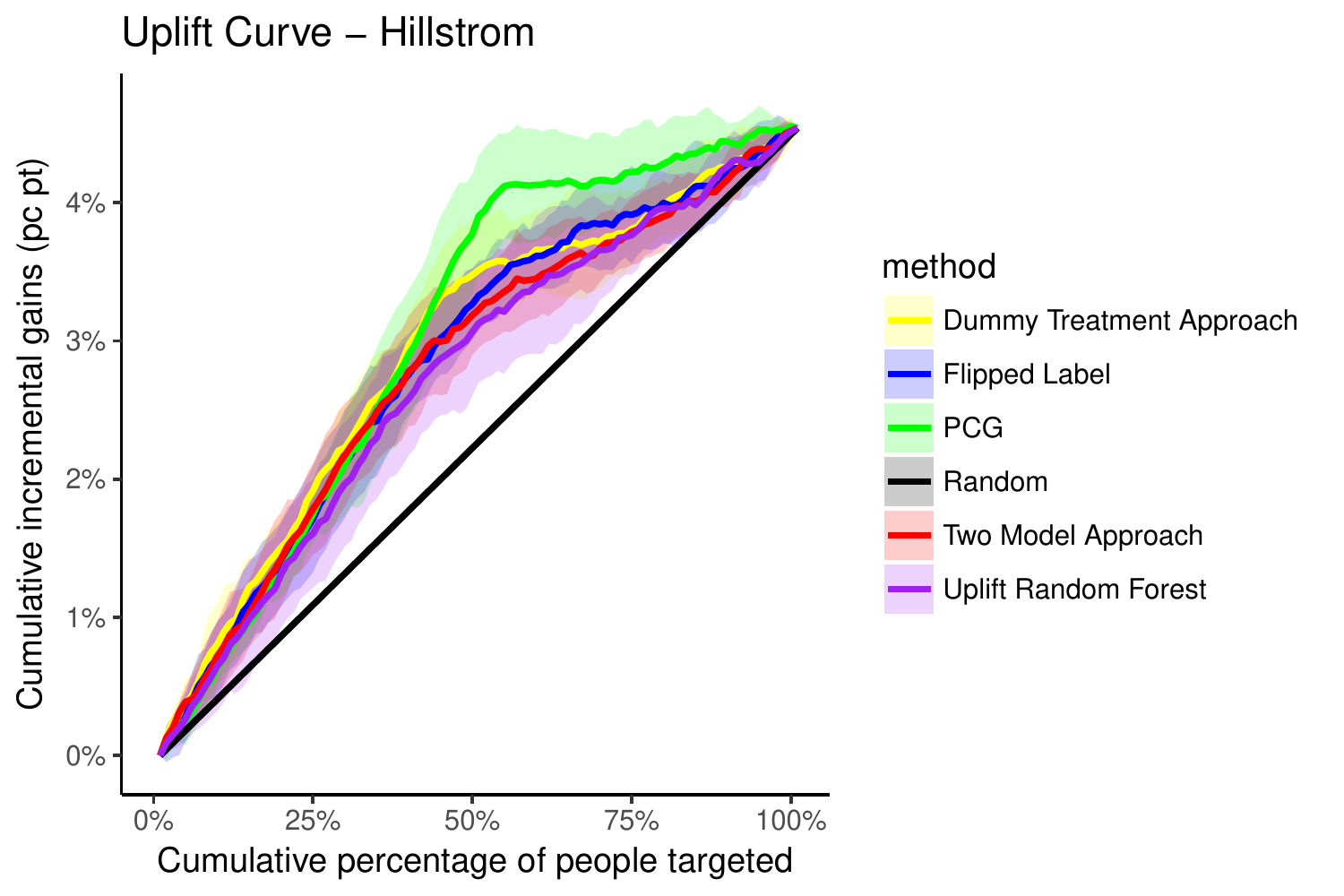}
  \caption{Dataset 2 - Clothing}
    \label{fig:Exp5Clothing}
\end{subfigure}%
\begin{subfigure}{.32\textwidth}
  \centering
  \includegraphics[width=1\linewidth]{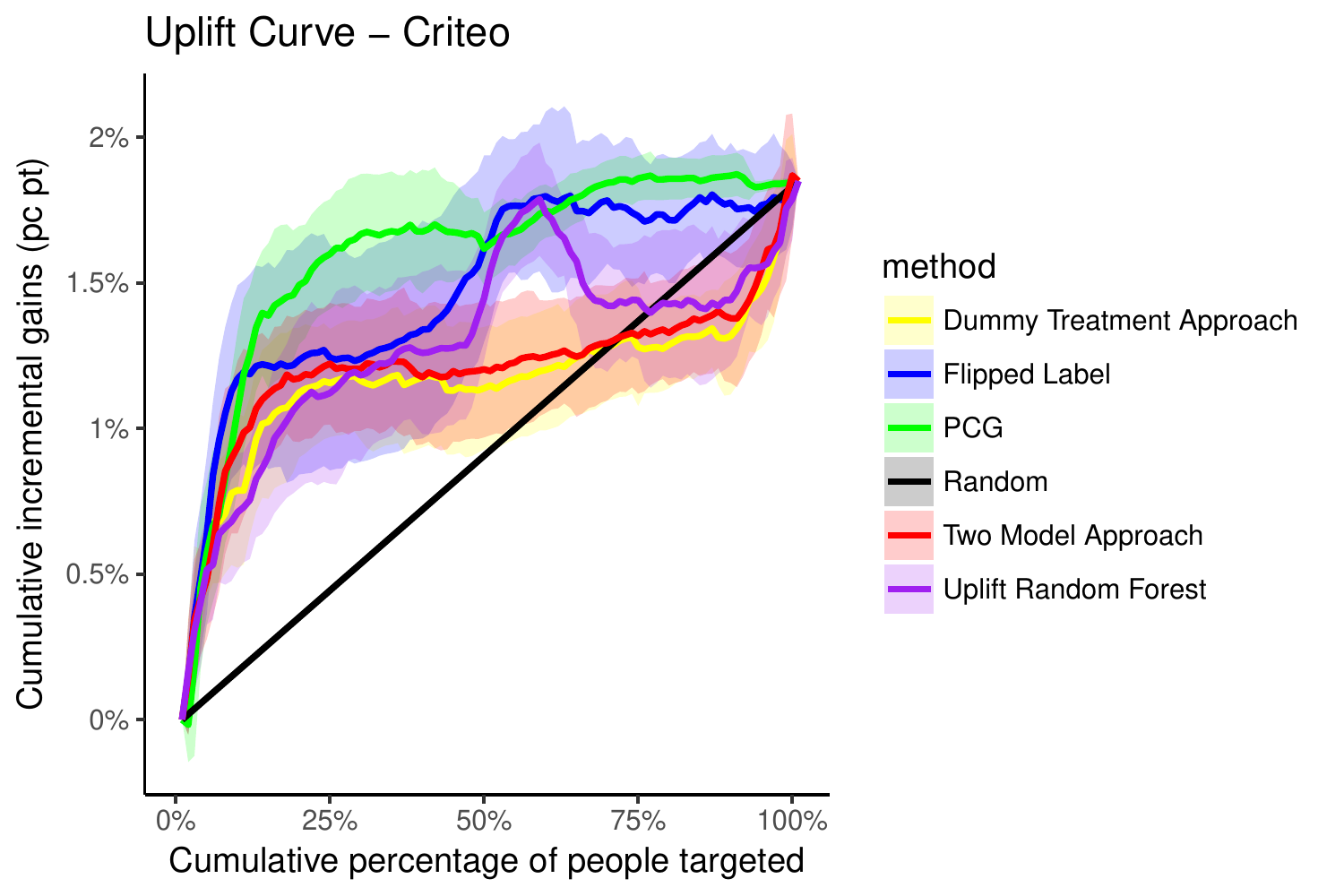}
  \caption{Dataset 3 - Marketing}
    \label{fig:Exp5Marketing}
\end{subfigure}
\caption[Experiment 5 - Results]{Results Experiment 5. Comparison of LambdaMART PCG with state-of-the-art uplift modeling techniques.}
\label{fig:Exp5}
\end{figure*}

Figure~\ref{fig:Exp5} shows the uplift curves on all three datasets. Most uplift modeling techniques perform better on the first dataset. With the two model approach being the best performing uplift modeling technique. Lai's approach, which we compared earlier with the learning-to-rank techniques, is relatively similar in performance to the dummy treatment approach and the uplift random forest. However, on the second and third dataset the LamdaMART PCG shows improved performance over the uplfit modeling techniques. On the second dataset the learning-to-rank technique achieves 4\% incremental gains as the only technique at 50\% of the targeted population, whereas the other techniques only achieve this when targeting close to everyone. The uplift modeling techniques show little difference in performance amongst themselves. On the final dataset we see Lai's approach do well in the first 10\%, however after that the PCG clearly dominates between 10 to 25\%, whereas only Lai's approach and the uplift random forest achieve similar performance when targeting near 50\% percent of the population. Some of the uplift modeling techniques even perform worse than baseline near the end. 

Next to the uplift curve plots we analyze the AUUC values in Table~\ref{tab:Exp5AUUC}. On the first dataset, the two model approach and the dummy treatment approach appear significantly better in performance than the LambdaMART PCG. There is no significant differences between the learning-to-rank approach, Lai's approach and the uplift random forest. The LambdaMART PCG is significantly better than the two model-approach and the uplift random forest on both the second and third dataset, and significantly better than the dummy treatment approach on the third dataset. We have additionally presented the results when evaluated with the relative joint setting, however, the main storyline does not change. These results do show that the learning-to-rank techniques can compete and even do better in some scenarios than existing state-of-the-art uplift modeling techniques. 

\begin{table*}[t]
\begin{center}
\caption[Experiment 5 - Results]{AUUC values when run in a separate setting with absolute relevance values. Both AUUC values of the separate relative uplift curve and the joint relative uplift curve are presented. Results that are significantly better (+) or worse (-) from LambdaMART PCG are marked as such.  \label{tab:Exp5AUUC}}
\begin{adjustbox}{max width=\textwidth}
{\small
\begin{tabular}{l|l|l|l|l|l|l|}
\cline{2-7}
                                               & \multicolumn{3}{c|}{\textit{Separate} relative AUUC}     & \multicolumn{3}{c|}{\textit{Joint} relative AUUC} \\ 
\multicolumn{1}{r|}{Dataset:}                & Information        & Hillstrom         & Criteo        & Information      & Hillstrom      & Criteo      \\ \hline
\multicolumn{1}{|l|}{Listwise LambdaMART PCG}   & 0.01829 & \textbf{0.03055} & \textbf{0.01601} & 0.01791 & \textbf{0.03057} & \textbf{0.01677} \\ \hline
\multicolumn{1}{|l|}{Dummy Treatment Approach}  & 0.02392(+) & 0.02935 & 0.01165(-) & 0.02283 & 0.0295 & 0.01181(-)  \\ 
\multicolumn{1}{|l|}{Two Model Approach}   & \textbf{0.02610}(+) & 0.02820(-) & 0.01213(-) & \textbf{0.02578}(+) & 0.02840 & 0.01224(-) \\ 
\multicolumn{1}{|l|}{Lai's Approach}  & 0.02052 & 0.02858 & 0.01479 & 0.02050 & 0.02865 & 0.01418(-)               \\ 
\multicolumn{1}{|l|}{Uplift Random Forest}  & 0.0221 & 0.02744(-) & 0.01287(-) & 0.02163 & 0.02746(-) & 0.01174(-)   \\ \hline
\end{tabular}
}
\end{adjustbox}
\end{center}
\end{table*}

\section{Discussion}
\label{Sec:Discussion}

Our first research question compared multiple definitions of the qini and uplift curve by analyzing the performance results on a simulated ranking. The results showed that counting the  incremental gains in absolute units would depend too much on the sizes of the treatment and control groups. Counting incremental gains with relative units gives a more robust evaluation (i.e., not affected by the treatment and control group sizes). Given those results we continued our experiments by evaluating the area under the uplift curve with relative units. This, in both separate or joint setting, depending on what learning method used as the learning-to-rank techniques can also be run in a separate or joint setting. Either we consider the treatment and control as one joint group, or we consider each group to be a separate setting. Note, we only consider the binary treatment case, where either a person is subject to the treatment, or is not. However, this could be extended to multiple treatments, by creating a query for each treatment available as this is very straightforward in learning-to-rank.

In the second experiment standard learning-to-rank techniques in the separate setting are compared to Flipped label approach from uplift modeling. We test these learning-to-rank techniques with both standard metrics such as the DCG and the NDCG, and with our own AUUC-centric metric, PCG. The results show that learning-to-rank techniques perform equally-to-better at the early percentages and slightly worse at later stages. This result already shows that learning-to-rank techniques is a viable alternative and can be added to the uplift modeling toolbox of techniques. 

We further investigated learning-to-rank approaches with research question 3 and 4. First, we investigated whether different sets of relevance values would significantly impact the performance of these techniques. In all settings the PCG clearly outperforms standard learning-to-rank metrics. When running PCG with the relative relevance values we directly optimize the AUUC. However, the results indicate that using non-symmetrical absolute relevance values show marginal improvements over relative relevance values, especially so for the NDCG when using separate queries. Our proposed set of relevance values seem equivalent, with no clear set outperforming the others significantly. This might be due to the preferences given over each category. In the separate setting we typically gave higher values to treatment responders, whereas in the control setting higher values are given to control non-responders over responders. Even when the values themselves changes, the preference order does not. In the joint setting the relation between treatment non-responders and control non-responders does not necessarily have no impact, hence, the non-symmetrical absolute relevance values were used. Further research could investigate different set of relevance values.

Research question 4 was about optimizing the AUUC among top fractions as many uplift modeling applications are typically about targeting a top fraction. Learning-to-rank techniques achieve this by using lower k-values. From an optimization perspective, we often did observe a clear change in behavior. Each model, optimized at a certain k-value, shows a clear peak in performance at the selected k-value. This indicates that we can optimize on lower k-values. Although these results were observed in both the separate and joint setting, it is more distinguishable in the joint setting. This indicates that the joint setting is more suitable for optimization.

From a generalization perspective, i.e., when testing on the test set, these results did not generalize well. Dataset 2 shows very little difference in performance, no matter which k-value was optimized. On dataset 1 and 3 the models could not replicate the peak-performance at each optimized k-value on all runs. Checking the standard deviation, shown by the error regions on the plots, shows that some runs do achieve some peak performance on their optimized k-value when compared to models optimized on the entire ranking, however, these results did not generalize over all runs. These experiments indicate optimizing for earlier fractions of the population is possible, but the actual benefit from it is marginal and not significant. 

Finally, we have tested LambdaMART PCG against existing state-of-the-art uplift modeling techniques. The uplift modeling techniques show varying performances on all datasets. On the first dataset, the uplift modeling techniques do seem to perform better than the LambdaMART PCG. However, on the second and third dataset the results show the LambdaMART PCG comes out on top in terms of AUUC. On the third dataset, the uplift modeling techniques show peculiar behavior. Although in the early stages all techniques manage to get positive incremental gains, most of them have near worse performance than baseline near the 70\% mark. Even the uplift random forest, which performed consistently good on previous benchmarks \cite{Devriendt:2018b}, shows odd behavior by both achieving high incremental gains to then performing worse than baseline. The LambdaMART PCG shows that it clearly identifies the most impact full instances at lower percentages targeted from the population when compared to the uplift modeling techniques. 

\section{Conclusion}
\label{sec:Conclusion}

Uplift modeling typically aims at identifying those instances most likely to respond as an effect of being treated.

Uplift models produce an uplift score which is then used to build a ranking from which a fraction is selected for targeting. Learning-to-rank techniques specifically focus on building quality rankings rather than quality predictions. A typical example being document retrieval for search engines. This paper explores the possibility of using learning-to-rank techniques in an uplift modeling context.

Before uplift modeling was brought into the learning-to-rank framework, an analysis of the current evaluation metrics of uplift modeling was done. This analysis shows conflicting definitions in the literature. The main differences among definitions were (1) whether the treatment and control group were considered as separate groups or jointly as one group and (2) whether the value of the incremental gains were in absolute units or in relative percentages. 

When comparing with the two selected definitions, the relative separate uplift curve and the relative joint uplift curve, we see that standard learning-to-rank techniques have shown to be capable models in terms of area under the uplift curve (AUUC) captured. With the promoted cumulative gain (PCG) we created a new metric which instead of discounting values further in the ranking, promotes values earlier in the ranking. The PCG is exactly the AUUC metric from uplift modeling and can be used in readily existing learning-to-rank techniques. Techniques optimizing the PCG show better results in terms of AUUC than standard learning-to-rank metrics and achieve equal or better results when compared with the baseline uplift model. 

This paper also tested the potential of learning-to-rank techniques optimization on lower k-values. Techniques have been trained to optimize their rankings at 10\%, 30\%, 50\% and 100\%. When analyzing from an optimization perspective one can see positive results in the sense that peaks are measured at the corresponding thresholds. However, these promising results do not generalize. Lower k-values usually means a lower overall performance, and not an improved performance at the specified threshold. 

Finally, with the last experiment we have shown that the LambdaMART PCG can compete with existing state-of-the-art uplift modeling techniques and even offer improved performances in terms of AUUC. Overall, these results confirm learning-to-rank as a viable alternative to existing uplift modeling methodology by focussing on modeling the ranking directly instead of the prediction value. This research opens up new research questions for future research:
\begin{itemize}
\item Future research could look into other listwise learning-to-rank techniques to compare performances.
\item Further investigation should also look into the effect of using different relevance values for CNR and TNR.
\item One of the open questions in uplift modeling is the use case of having multiple treatments \cite{Rzepakowski:2012:SingleMultiple}. The learning-to-rank framework readily allows one to plug in mulitple treatments by considering each treatment as a separate query.
\item Finally, as illustrated with the PCG metric, the learning-to-rank framework allows to optimize for custom metrics. This creates the opportunity to include profit-centric metrics into the modeling phase. The Maximum Profit Uplift (MPU), specifically created for uplift modeling \cite{Devriendt:2018b}, can be implemented into learning-to-rank with the aim of achieving more profitable models. 
\end{itemize}   
\bibliographystyle{plain}
\bibliography{Bibliography}

\begin{thebibliography}{10}

\bibitem{Betlei:2018}
Artem Betlei, Eustache Diemert, and Massih-Reza Amini.
\newblock Uplift prediction with dependent feature representation in imbalanced
  treatment and control conditions.
\newblock In {\em International Conference on Neural Information Processing},
  pages 47--57. Springer, 2018.

\bibitem{breiman1984}
L.~Breiman, J.~Friedman, R.~Olshen, and C.~Stone.
\newblock {\em Classification and Regression Trees}.
\newblock Chapman and Hall, New York, NY, U.S.A., 1984.

\bibitem{Burges:2005}
Chris Burges, Tal Shaked, Erin Renshaw, Ari Lazier, Matt Deeds, Nicole
  Hamilton, and Greg Hullender.
\newblock Learning to rank using gradient descent.
\newblock In {\em Proceedings of the 22Nd International Conference on Machine
  Learning}, ICML '05, pages 89--96, New York, NY, USA, 2005. ACM.

\bibitem{Burges:2010}
Chris~J.C. Burges.
\newblock From ranknet to lambdarank to lambdamart: An overview.
\newblock Technical Report MSR-TR-2010-82, June 2010.

\bibitem{R:xgboost}
Tianqi Chen, Tong He, Michael Benesty, Vadim Khotilovich, Yuan Tang, Hyunsu
  Cho, Kailong Chen, Rory Mitchell, Ignacio Cano, Tianyi Zhou, Mu~Li, Junyuan
  Xie, Min Lin, Yifeng Geng, and Yutian Li.
\newblock {\em xgboost: Extreme Gradient Boosting}, 2018.
\newblock R package version 0.81.0.1.

\bibitem{NIPS2009_3708}
Wei Chen, Tie yan Liu, Yanyan Lan, Zhi ming Ma, and Hang Li.
\newblock Ranking measures and loss functions in learning to rank.
\newblock In Y.~Bengio, D.~Schuurmans, J.~D. Lafferty, C.~K.~I. Williams, and
  A.~Culotta, editors, {\em Advances in Neural Information Processing Systems
  22}, pages 315--323. Curran Associates, Inc., 2009.

\bibitem{Croft:2010}
W~Bruce Croft, Donald Metzler, and Trevor Strohman.
\newblock {\em Search engines: Information retrieval in practice}, volume 520.
\newblock Addison-Wesley Reading, 2010.

\bibitem{Devriendt:2018}
Floris Devriendt, Darie Moldovan, and Wouter Verbeke.
\newblock A literature survey and experimental evaluation of the
  state-of-the-art in uplift modeling: A stepping stone toward the development
  of prescriptive analytics.
\newblock {\em Big data}, 6(1):13--41, 2018.

\bibitem{Devriendt:2018b}
Floris Devriendt and Wouter Verbeke.
\newblock The case for prescriptive analytics: a novel maximum profit measure
  for evaluating and comparing customer churn prediction and uplift models.
\newblock WorkingPaper~12, Vrije Universiteit Brussel, Faculteit Economische en
  Sociale Wetenschappen \& Solvay Business School, Belgium, 4 2018.

\bibitem{Diemert:2018}
{Diemert Eustache, Betlei Artem}, Christophe Renaudin, and Amini Massih-Reza.
\newblock A large scale benchmark for uplift modeling.
\newblock In {\em Proceedings of the AdKDD and TargetAd Workshop, KDD,
  London,United Kingdom, August, 20, 2018}. ACM, 2018.

\bibitem{Gubela:2017}
Robin~Marco Gubela, Stefan Lessmann, Johannes Haupt, Annika Baumann, Tillmann
  Radmer, and Fabian Gebert.
\newblock Revenue uplift modeling.
\newblock 2017.

\bibitem{Guelman:2015}
Leo Guelman.
\newblock Optimal personalized treatment learning models with insurance
  applications.
\newblock 2015.

\bibitem{Guelman:2014}
Leo Guelman, Montserrat Guillen, and Ana~M. P{\'e}rez-Mar{\'i}n.
\newblock Optimal personalized treatment rules for marketing interventions: A
  review of methods, a new proposal, and an insurance case study.
\newblock Working Papers 2014-06, Universitat de Barcelona, UB Riskcenter,
  2014.

\bibitem{Gutierrez:2017}
Pierre Gutierrez and Jean-Yves Gerardy.
\newblock Causal inference and uplift modelling: A review of the literature.
\newblock In Claire Hardgrove, Louis Dorard, Keiran Thompson, and Florian
  Douetteau, editors, {\em Proceedings of The 3rd International Conference on
  Predictive Applications and APIs}, volume~67 of {\em Proceedings of Machine
  Learning Research}, pages 1--13, Microsoft NERD, Boston, USA, 11--12 Oct
  2017. PMLR.

\bibitem{Hansotia:2002}
Behram Hansotia and Brad Rukstales.
\newblock Incremental value modeling.
\newblock {\em J INTERACT MARK}, 16(3):35--46, 2002.

\bibitem{Holland:1986}
Paul~W Holland.
\newblock Statistics and causal inference.
\newblock {\em Journal of the American statistical Association},
  81(396):945--960, 1986.

\bibitem{Jarvelin:2002}
Kalervo J\"{a}rvelin and Jaana Kek\"{a}l\"{a}inen.
\newblock Cumulated gain-based evaluation of ir techniques.
\newblock {\em ACM Trans. Inf. Syst.}, 20(4):422--446, October 2002.

\bibitem{Jaskowski:2012:UpliftClinical}
Maciej Jaskowski and Szymon Jaroszewicz.
\newblock Uplift modeling for clinical trial data.
\newblock In {\em ICML Workshop on Clinical Data Analysis}, 2012.

\bibitem{Kane:2014}
Kathleen Kane, Victor~S.Y. Lo, and Jane Zheng.
\newblock Mining for the truly responsive customers and prospects using
  true-lift modeling: Comparison of new and existing methods.
\newblock {\em Journal of Marketing Analytics}, 2(4):218--238, Dec 2014.

\bibitem{Kass:1980}
Gordon~V Kass.
\newblock An exploratory technique for investigating large quantities of
  categorical data.
\newblock {\em Appl Stat}, pages 119--127, 1980.

\bibitem{Kuusisto:2014}
Finn Kuusisto, Vitor~Santos Costa, Houssam Nassif, Elizabeth Burnside, David
  Page, and Jude Shavlik.
\newblock Support vector machines for differential prediction.
\newblock In {\em Joint European Conference on Machine Learning and Knowledge
  Discovery in Databases}, pages 50--65. Springer, 2014.

\bibitem{Lai:2006}
Lily Yi-Ting Lai.
\newblock {\em Influential marketing: a new direct marketing strategy
  addressing the existence of voluntary buyers}.
\newblock PhD thesis, School of Computing Science-Simon Fraser University,
  2006.

\bibitem{Liu:2009}
Tie-Yan Liu et~al.
\newblock Learning to rank for information retrieval.
\newblock {\em Foundations and Trends{\textregistered} in Information
  Retrieval}, 3(3):225--331, 2009.

\bibitem{Lo:2002:TLM}
Victor S.~Y. Lo.
\newblock The true lift model: A novel data mining approach to response
  modeling in database marketing.
\newblock {\em SIGKDD Explor. Newsl.}, 4(2):78--86, December 2002.

\bibitem{Manning:2010}
Christopher Manning, Prabhakar Raghavan, and Hinrich Sch{\"u}tze.
\newblock Introduction to information retrieval.
\newblock {\em Natural Language Engineering}, 16(1):100--103, 2010.

\bibitem{Mcfee:2010}
Brian McFee and Gert~R Lanckriet.
\newblock Metric learning to rank.
\newblock In {\em Proceedings of the 27th International Conference on Machine
  Learning (ICML-10)}, pages 775--782, 2010.

\bibitem{Nassif:2013}
Houssam Nassif, Finn Kuusisto, Elizabeth~S Burnside, and Jude~W Shavlik.
\newblock Uplift modeling with roc: An srl case study.
\newblock In {\em ILP (Late Breaking Papers)}, pages 40--45, 2013.

\bibitem{Radcliffe:2007}
Nicholas~J Radcliffe.
\newblock Using control groups to target on predicted lift: Building and
  assessing uplift models.
\newblock {\em Direct Market J Direct Market Assoc Anal Council}, 1:14--21,
  2007.

\bibitem{Radcliffe:2011:RealWorldUplift}
Nicholas~J Radcliffe and Patrick~D Surry.
\newblock Real-world uplift modelling with significance-based uplift trees.
\newblock {\em White Paper TR-2011-1, Stochastic Solutions}, 2011.

\bibitem{Rubin:2005}
Donald~B Rubin.
\newblock Causal inference using potential outcomes: Design, modeling,
  decisions.
\newblock {\em Journal of the American Statistical Association},
  100(469):322--331, 2005.

\bibitem{Rzepakowski:2010}
Piotr Rzepakowski and Szymon Jaroszewicz.
\newblock Decision trees for uplift modeling.
\newblock In {\em Proceedings of the 2010 IEEE International Conference on Data
  Mining}, ICDM '10, pages 441--450, Washington, DC, USA, 2010. IEEE Computer
  Society.

\bibitem{Rzepakowski:2012:SingleMultiple}
Piotr Rzepakowski and Szymon Jaroszewicz.
\newblock Decision trees for uplift modeling with single and multiple
  treatments.
\newblock {\em Knowl Inf Syst}, 32(2):303--327, 2012.

\bibitem{Rzepakowski:2012:DirectMarketing}
Piotr Rzepakowski and Szymon Jaroszewicz.
\newblock Uplift modeling in direct marketing.
\newblock {\em Journal of Telecommunications and Information Technology}, pages
  43--50, 2012.

\bibitem{Shalit:2017}
Uri Shalit, Fredrik~D Johansson, and David Sontag.
\newblock Estimating individual treatment effect: generalization bounds and
  algorithms.
\newblock In {\em Proceedings of the 34th International Conference on Machine
  Learning-Volume 70}, pages 3076--3085. JMLR. org, 2017.

\bibitem{Soltys:2018}
Micha{\l} So{\l}tys and Szymon Jaroszewicz.
\newblock Boosting algorithms for uplift modeling.
\newblock {\em arXiv preprint arXiv:1807.07909}, 2018.

\bibitem{Soltys:2014}
Michal Soltys, Szymon Jaroszewicz, and Piotr Rzepakowski.
\newblock Ensemble methods for uplift modeling.
\newblock {\em Data Min Knowl Discov}, pages 1--29, 2014.

\bibitem{Jaroszewicz:2013:SVM}
L.~Zaniewicz and S.~Jaroszewicz.
\newblock Support vector machines for uplift modeling.
\newblock In {\em Data Mining Workshops (ICDMW), 2013 IEEE 13th International
  Conference on}, pages 131--138, 2013.

\bibitem{Zaniewicz:2017}
{\L}ukasz Zaniewicz and Szymon Jaroszewicz.
\newblock Lp-support vector machines for uplift modeling.
\newblock {\em Knowledge and Information Systems}, 53(1):269--296, 2017.

\bibitem{Zhao:2017:MultipleTreatment}
Yan Zhao, Xiao Fang, and David Simchi-Levi.
\newblock Uplift modeling with multiple treatments and general response types.
\newblock In {\em Proceedings of the 2017 SIAM International Conference on Data
  Mining}, pages 588--596. SIAM, 2017.

\end{thebibliography}

\newpage
\section*{Appendix}
\subsection*{Appendix A: AUUC derivation for the separate relative setting}
Substituting Separate Relative into it, and accounting for $p$ being a percentage as $|T|$ and $|C|$ can be different in size:
\begin{align}
AUUC &= \sum_{p=1}^{100} V(p/100) \\
     &= \sum_{p=1}^{100} \left(\frac{R(T, p|T|/100)}{|T|} - \frac{R(C, p|C|/100)}{|C|}\right) \\
     &\approx \sum_{k=1}^{|T|} \frac{R(T, k)}{|T|} - \sum_{k=1}^{|C|} \frac{R(C, k)}{|C|}
\end{align}

To ease the derivation, we define $\pi^T$ to be the ordering of \textit{only} the instances in $T$, and similarly for $\pi^C$, where $y^{\pi_T}_i$ denotes the $y$ value of the $i$th ranked instance of $T$ according to $\pi$. Hence, we have:
\begin{align}
R_\pi(T, k) &= \sum_{(X_i,y_i) \in \pi(T, k)} \Ivs{y_i = 1} \\
            &= \sum_{i=1}^k \Ivs{y^{\pi_T}_i = 1}
\end{align}
and similarly for $R_\pi(C, k)$. Plugging this into the above, we obtain:
\begin{align}
AUUC &= \sum_{k=1}^{|T|} \frac{R(T, k)}{|T|} - \sum_{k=1}^{|C|} \frac{R(C, k)}{|C|} \\
     &= \sum_{k=1}^{|T|} \frac{\sum_{i=1}^{k} \Ivs{y^{\pi_T}_i = 1}}{|T|} - \sum_{k=1}^{|C|} \frac{\sum_{i=1}^{k} \Ivs{y^{\pi_C}_i = 1}}{|C|}
\end{align}
We now introduce helper function $g(i)$ as:
\begin{flalign} 
g^T(i) = \begin{cases} 
            0 & \text{if } y^\pi_i = 0 \text{ and } t^\pi_i = 1 \\
            1/|T| & \text{if } y^\pi_i = 1 \text{ and } t^\pi_i = 1
      \end{cases} \label{eq:gti}
\end{flalign}
and
\begin{flalign} 
g^C(i) = \begin{cases} 
            0 & \text{if } y^\pi_i = 0 \text{ and } t^\pi_i = 0 \\
            -1/|C| & \text{if } y^\pi_i = 1 \text{ and } t^\pi_i = 0
      \end{cases} \label{eq:gci}
\end{flalign}
After substitution, we obtain:
\begin{align}
AUUC &= \sum_{k=1}^{|T|} \frac{\sum_{i=1}^{k} \Ivs{y^{\pi_T}_i = 1}}{|T|} - \sum_{k=1}^{|C|} \frac{\sum_{i=1}^{k} \Ivs{y^{\pi_C}_i = 1}}{|C|} \\
     &= \sum_{k=1}^{|T|} \sum_{i=1}^k g^T(i) + \sum_{k=1}^{|C|} \sum_{i=1}^k g^C(i) \\
     &= \sum_{i=1}^{|T|} \sum_{k=i}^n g^T(i) + = \sum_{i=1}^n  g(i)\sum_{k=i}^n 1 \\
     &= \sum_{i=1}^{|T|} ({|T|}-i+1)g^T(i) + \sum_{i=1}^{|C|} ({|C|}-i+1)g^C(i)
\end{align}

\newpage
\subsection*{Appendix B: Experiment 3 - Results Graded Relevance Comparison}

\begin{table*}[ht!]
\begin{center}
\caption{AUUC values of the relative separate uplift curve. Evaluated with both the separate uplift and joint uplift setting.}
\begin{adjustbox}{max width=\textwidth}
{\small
\begin{tabular}{ll|l|l|l|l|l|l|l|l|l|}
\cline{3-11}
                                                                &       & \multicolumn{3}{l|}{Information} & \multicolumn{3}{l|}{Hillstrom} & \multicolumn{3}{l|}{Criteo} \\ \cline{3-11} 
                                                                &       & DCG       & NCG       & PCG      & DCG      & NCG      & PCG      & DCG     & NCG     & PCG     \\ \hline
\multicolumn{1}{|l|}{\multirow{4}{*}{Separate AUUC Evaluation}} & Abs 1 &0.0152 & 0.00935 & 0.01938 & 0.0296 & 0.03032 & 0.03077 & 0.01522 & 0.01523 & 0.01578        \\ \cline{2-11} 
\multicolumn{1}{|l|}{}                                          & Abs 2 &0.0152 & 0.00678 & 0.01938 & 0.0296 & 0.02893 & 0.03077 & 0.01522 & 0.01555 & 0.01578          \\ \cline{2-11} 
\multicolumn{1}{|l|}{}                                          & Abs 3 &0.0152 & 0.01524 & 0.01938 & 0.0296 & 0.02953 & 0.03077 & 0.01522 & 0.01538 & 0.01578         \\ \cline{2-11} 
\multicolumn{1}{|l|}{}                                          & Rel   &0.01382 & 0.00677 & 0.01829 & 0.02961 & 0.02893 & 0.03055 & 0.01549 & 0.01555 & 0.01601         \\ \hline
\multicolumn{1}{|l|}{\multirow{4}{*}{Joint AUUC Evaluation}}    & Abs 1 & 0.01612 & 0.00844 & 0.01894 & 0.02955 & 0.03051 & 0.03067 & 0.0158 & 0.01578 & 0.01662         \\ \cline{2-11} 
\multicolumn{1}{|l|}{}                                          & Abs 2 &0.01612 & 0.00649 & 0.01894 & 0.02955 & 0.02887 & 0.03067 & 0.0158 & 0.01629 & 0.01662        \\ \cline{2-11} 
\multicolumn{1}{|l|}{}                                          & Abs 3 &0.01612 & 0.01661 & 0.01894 & 0.02955 & 0.02948 & 0.03067 & 0.0158 & 0.01597 & 0.01662         \\ \cline{2-11} 
\multicolumn{1}{|l|}{}                                          & Rel   &0.01459 & 0.00651 & 0.01791 & 0.0296 & 0.02887 & 0.03057 & 0.01601 & 0.01629 & 0.01677        \\ \hline
\end{tabular}
}
\end{adjustbox}
\label{tab:ExpGainSepQueryApp}
\end{center}
\end{table*}

\begin{table*}[ht!]
\begin{center}
\caption{AUUC values of the relative joint uplift curve. Evaluated with both the separate uplift and joint uplift setting.}
\begin{adjustbox}{max width=\textwidth}
{\small
\begin{tabular}{ll|l|l|l|l|l|l|l|l|l|}
\cline{3-11}
                                                                &       & \multicolumn{3}{l|}{Information} & \multicolumn{3}{l|}{Hillstrom} & \multicolumn{3}{l|}{Criteo} \\ \cline{3-11} 
                                                                &       & DCG       & NCG       & PCG      & DCG      & NCG      & PCG      & DCG     & NCG     & PCG     \\ \hline
\multicolumn{1}{|l|}{\multirow{4}{*}{Separate AUUC Evaluation}} & Abs 1 &0.01396 & 0.01452 & 0.0194 & 0.02957 & 0.02957 & 0.03002 & 0.01497 & 0.01494 & 0.01469\\  \cline{2-11} 
\multicolumn{1}{|l|}{}                                          & Abs 2 &0.01101 & 0.01116 & 0.01536 & 0.02935 & 0.02935 & 0.03051 & 0.01607 & 0.01607 & 0.01669\\  \cline{2-11} 
\multicolumn{1}{|l|}{}                                          & Abs 3 &0.01563 & 0.01473 & 0.023 & 0.02968 & 0.02968 & 0.03063 & 0.01568 & 0.01568 & 0.01536\\  \cline{2-11} 
\multicolumn{1}{|l|}{}                                          & Rel   &0.01052 & 0.01031 & 0.01573 & 0.02954 & 0.02954 & 0.03027 & 0.01541 & 0.01543 & 0.01554\\  \hline

\multicolumn{1}{|l|}{\multirow{4}{*}{Joint AUUC Evaluation}}    & Abs 1 & 0.01396 & 0.01452 & 0.0194 & 0.02957 & 0.02957 & 0.03002 & 0.01497 & 0.01494 & 0.01469\\  \cline{2-11} 
\multicolumn{1}{|l|}{}                                          & Abs 2 &0.01101 & 0.01116 & 0.01536 & 0.02935 & 0.02935 & 0.03051 & 0.01607 & 0.01607 & 0.01669\\  \cline{2-11} 
\multicolumn{1}{|l|}{}                                          & Abs 3 &0.01563 & 0.01473 & 0.023 & 0.02968 & 0.02968 & 0.03063 & 0.01568 & 0.01568 & 0.01536\\  \cline{2-11} 
\multicolumn{1}{|l|}{}                                          & Rel   &0.01052 & 0.01031 & 0.01573 & 0.02954 & 0.02954 & 0.03027 & 0.01541 & 0.01543 & 0.01554\\  \hline
\end{tabular}
}
\end{adjustbox}
\label{tab:ExpGainJointQueryApp}
\end{center}
\end{table*}


\clearpage

\subsection*{Appendix C: Experiment 4 - Plots - Relative Relevance}

\begin{figure}[ht!]
\centering
\begin{subfigure}{.32\textwidth}
  \centering
  \includegraphics[width=1\linewidth]{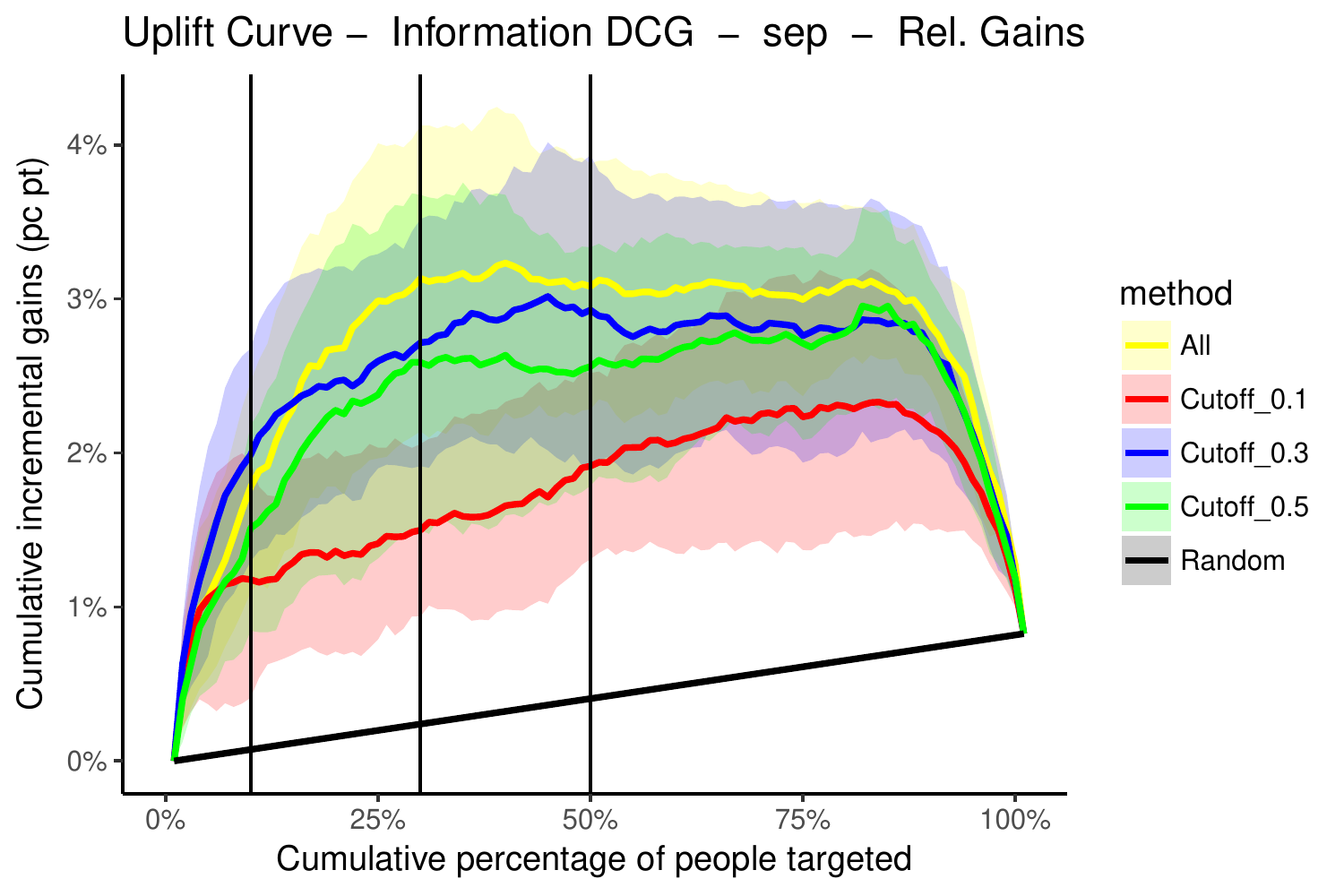}
  \caption{Uplift Curve - Information}
\end{subfigure} %
\begin{subfigure}{.32\textwidth}
  \centering
  \includegraphics[width=1\linewidth]{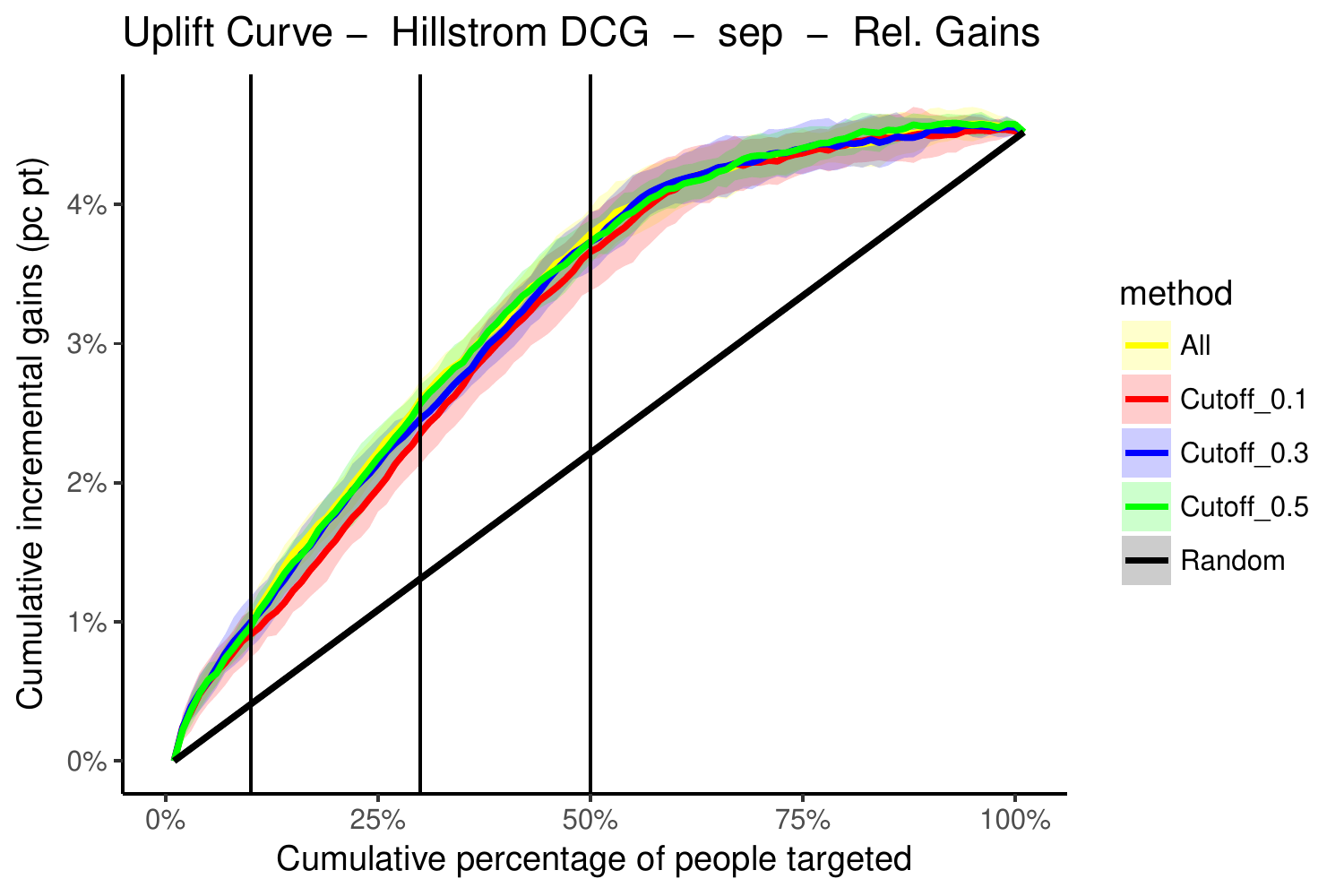}
  \caption{Uplift Curve - Hillstrom}
\end{subfigure} %
\begin{subfigure}{.32\textwidth}
  \centering
  \includegraphics[width=1\linewidth]{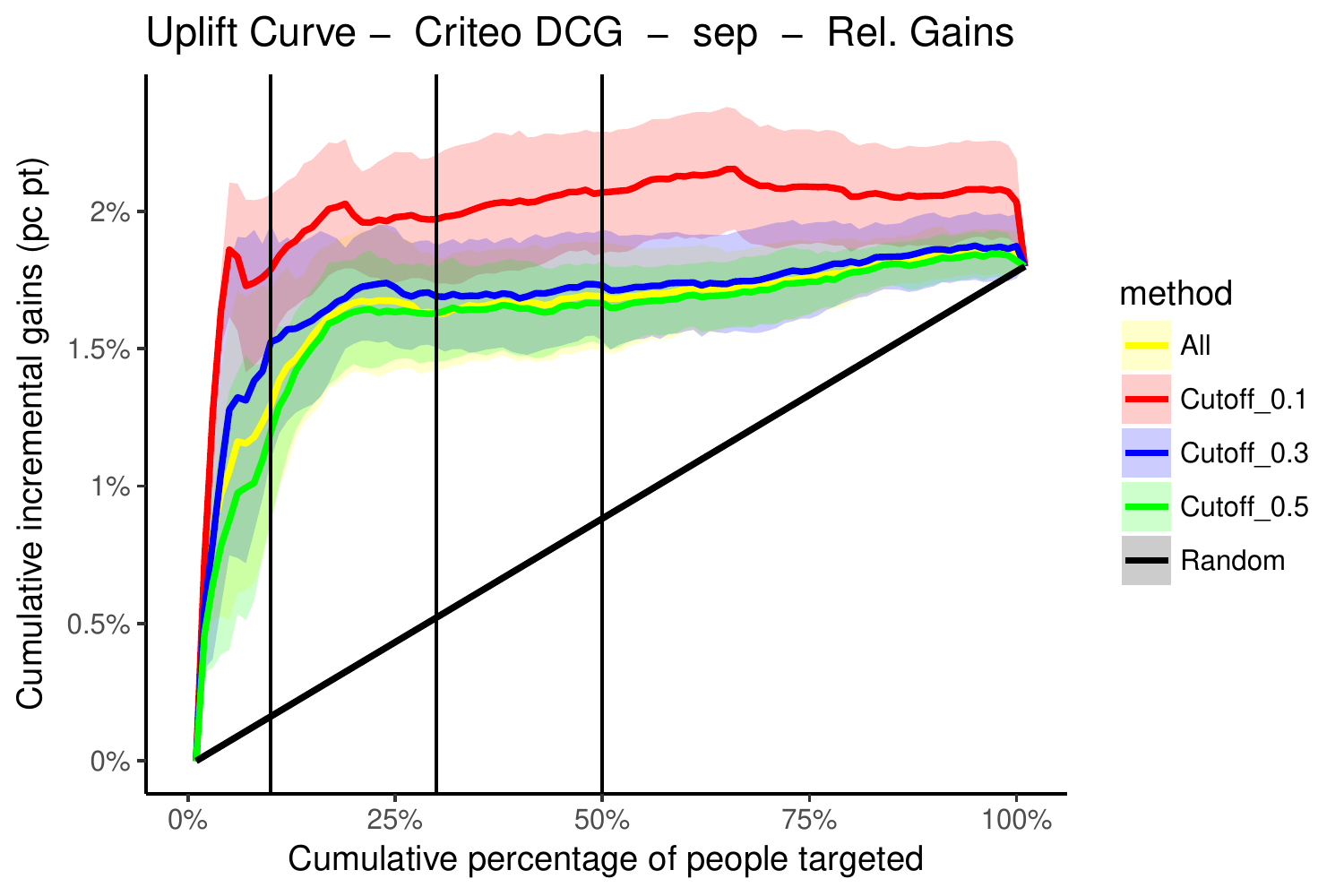}
  \caption{Uplift Curve - Criteo}
\end{subfigure}

\begin{subfigure}{.32\textwidth}
  \centering
  \includegraphics[width=1\linewidth]{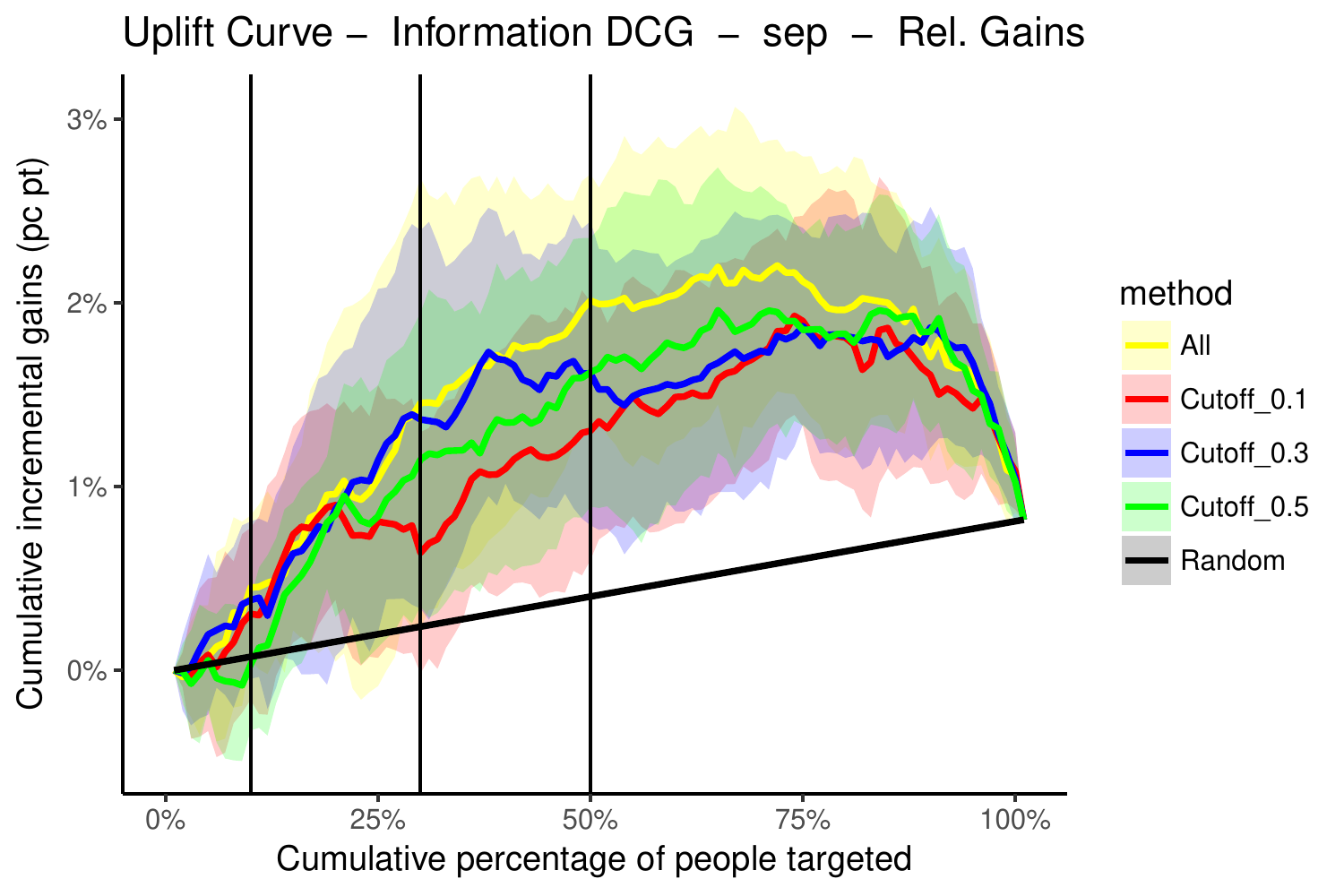}
  \caption{Uplift Curve - Information}
\end{subfigure} %
\begin{subfigure}{.32\textwidth}
  \centering
  \includegraphics[width=1\linewidth]{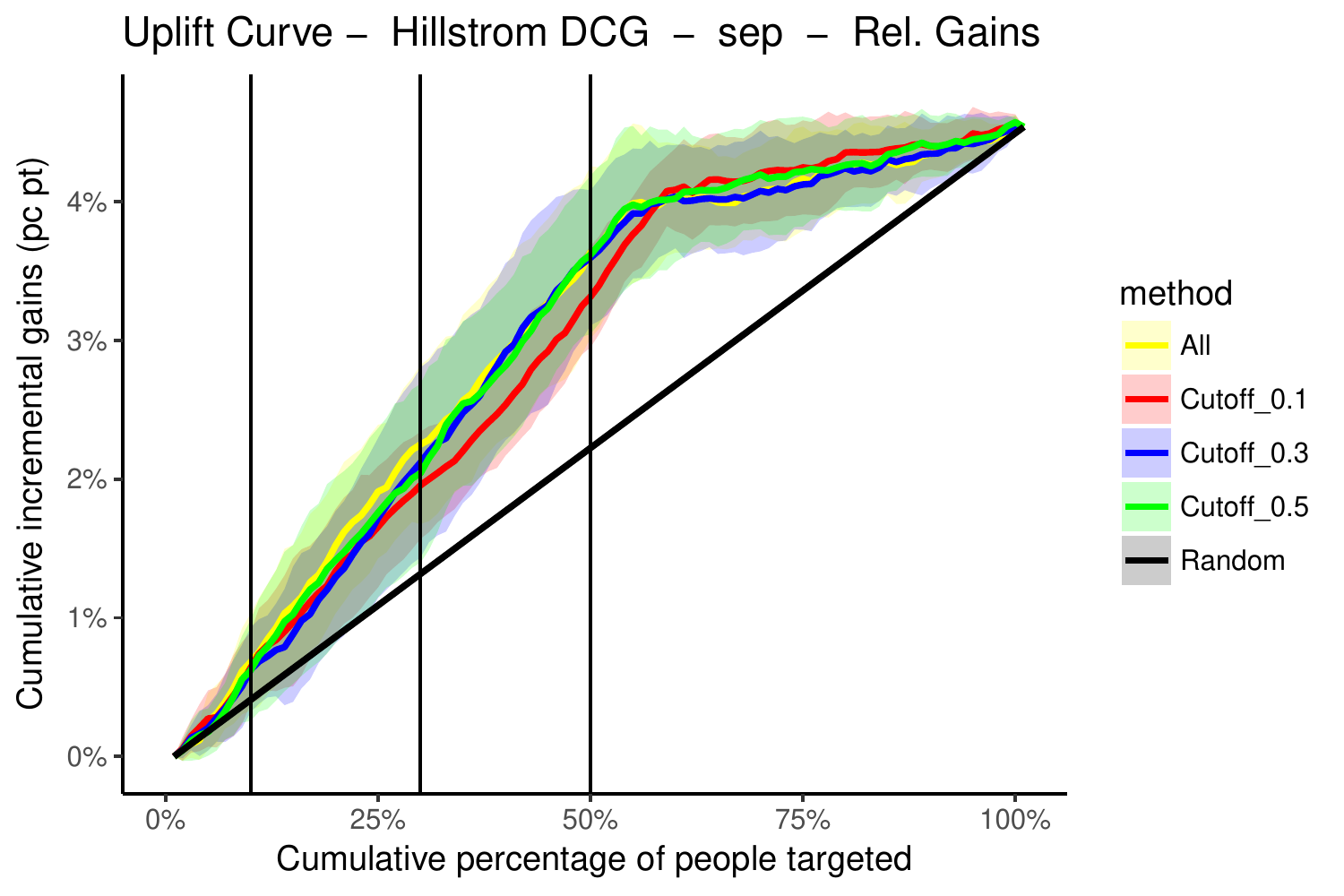}
  \caption{Uplift Curve - Hillstrom}
\end{subfigure} %
\begin{subfigure}{.32\textwidth}
  \centering
  \includegraphics[width=1\linewidth]{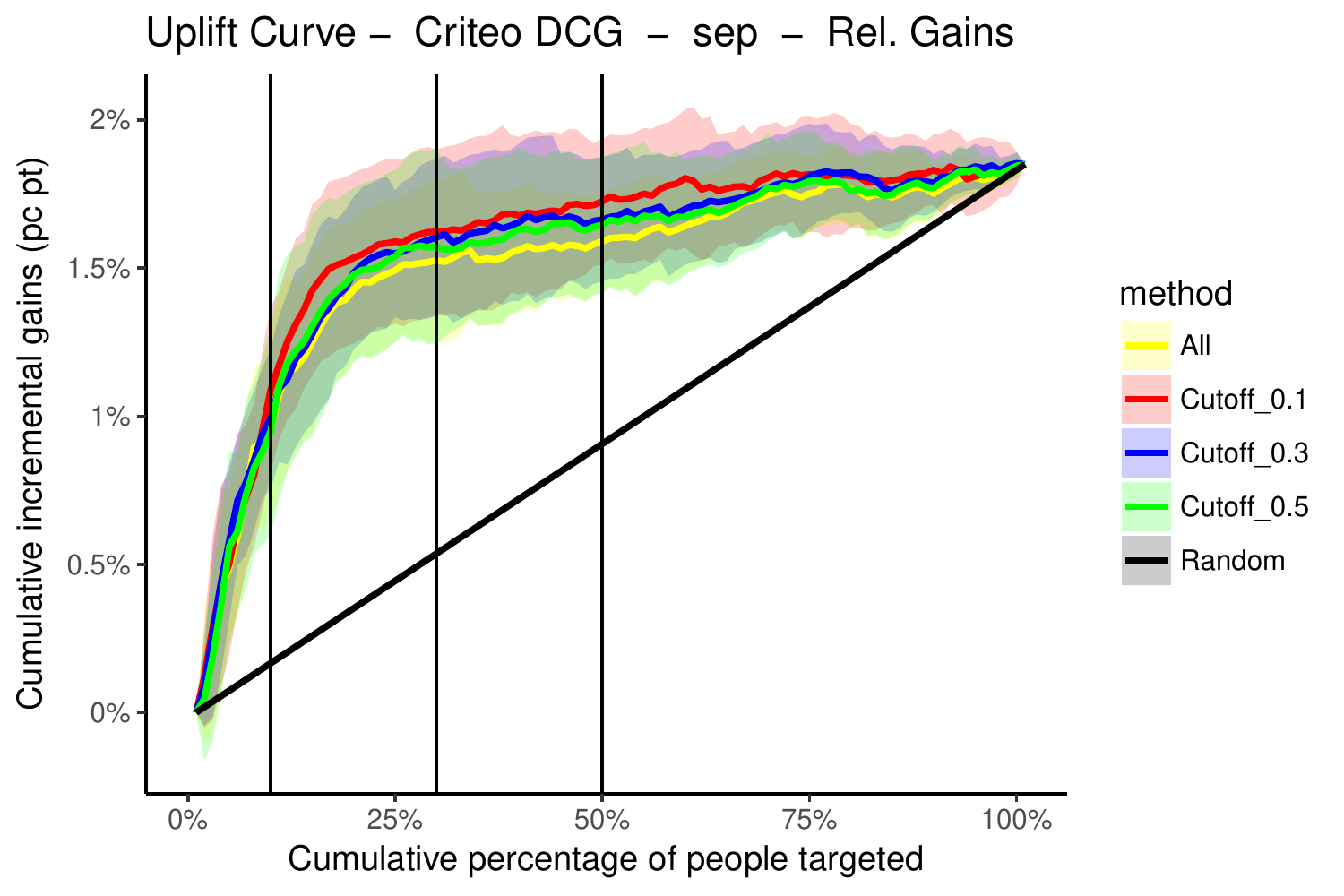}
  \caption{Uplift Curve - Criteo}
\end{subfigure}
\caption{Experiment 2 - DCG for Separate Setting at multiple cutoffs with relative gains.}
\end{figure}

\begin{figure}[ht!]
\centering
\begin{subfigure}{.32\textwidth}
  \centering
  \includegraphics[width=1\linewidth]{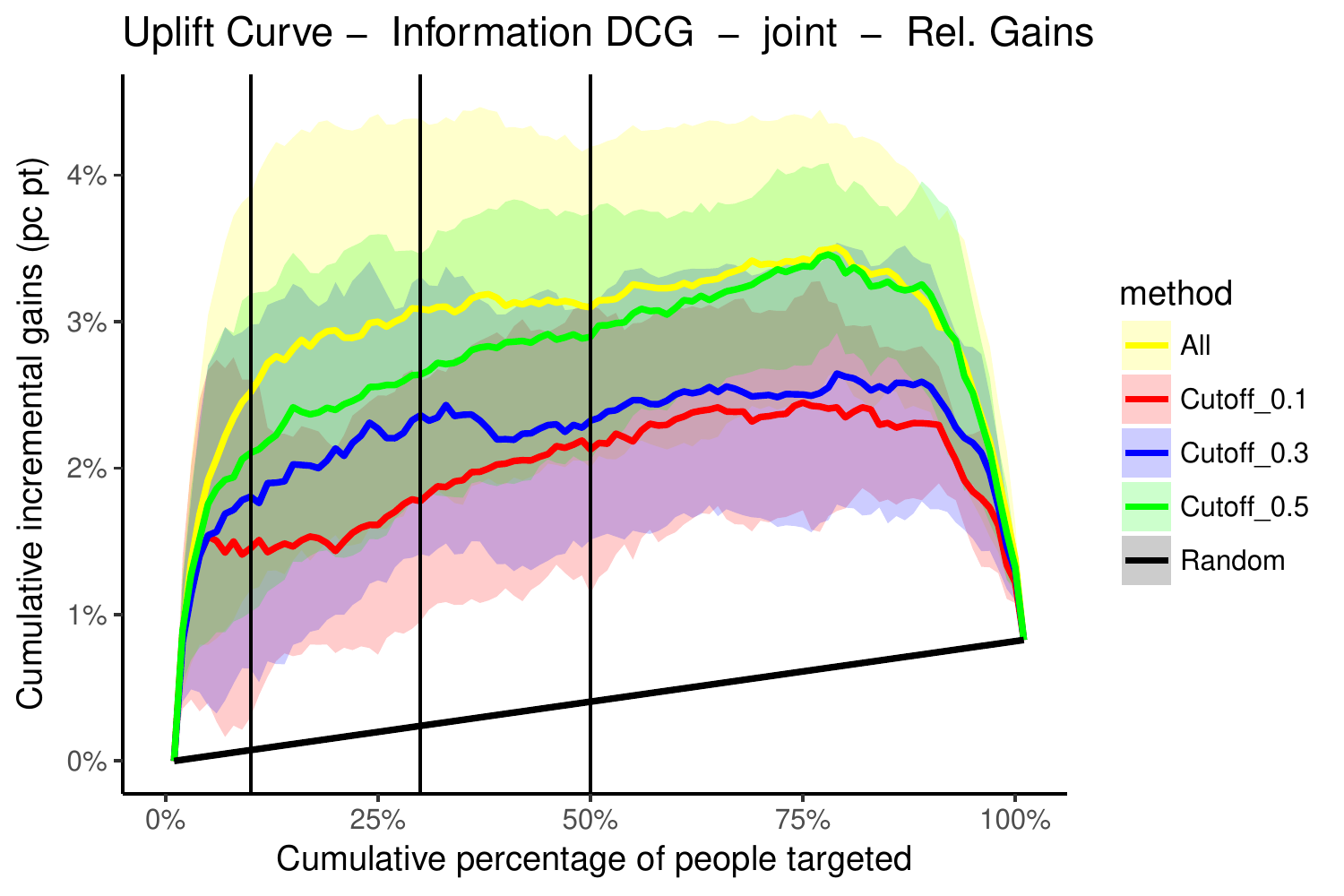}
  \caption{Uplift Curve - Information}
\end{subfigure} %
\begin{subfigure}{.32\textwidth}
  \centering
  \includegraphics[width=1\linewidth]{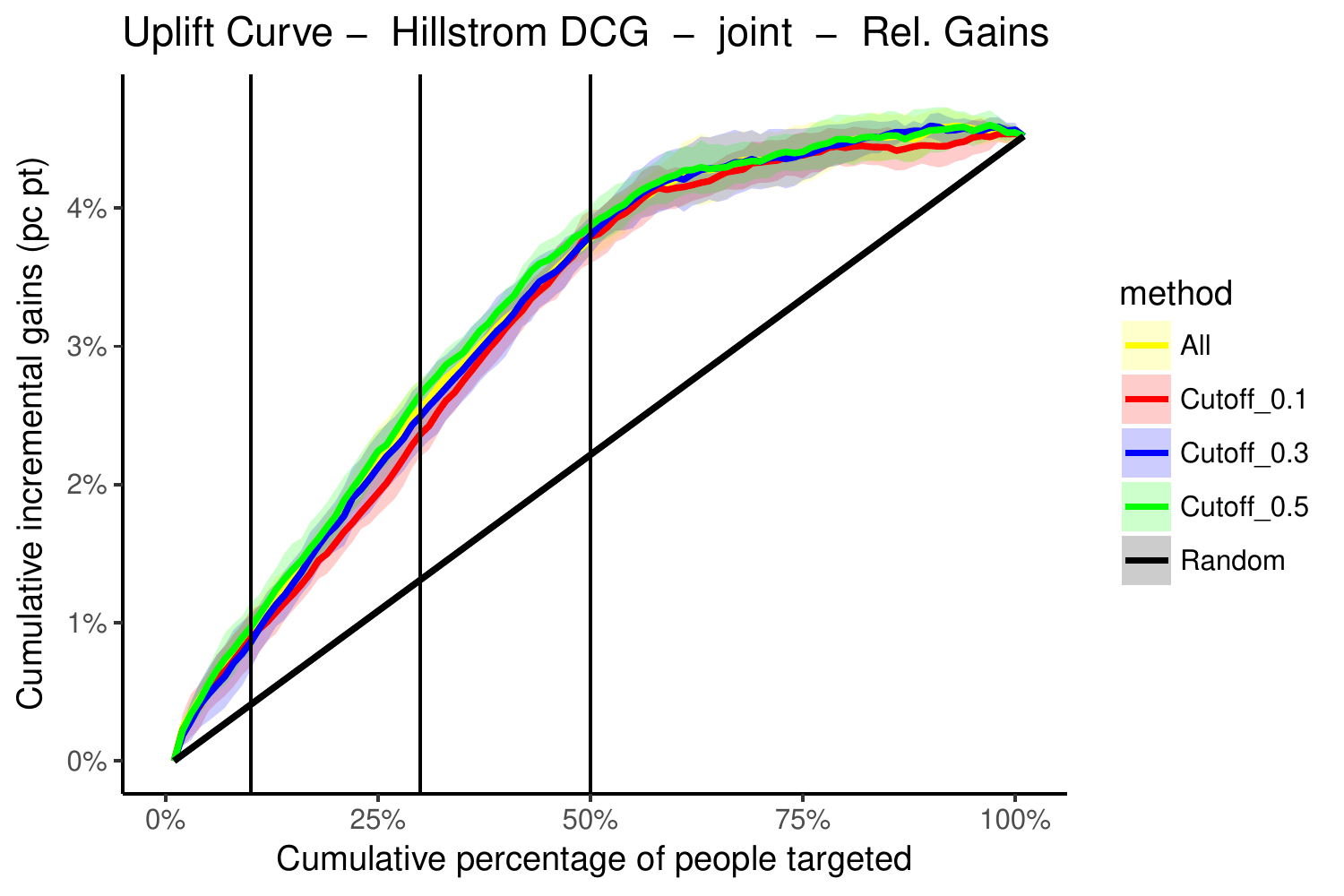}
  \caption{Uplift Curve - Hillstrom}
\end{subfigure} %
\begin{subfigure}{.32\textwidth}
  \centering
  \includegraphics[width=1\linewidth]{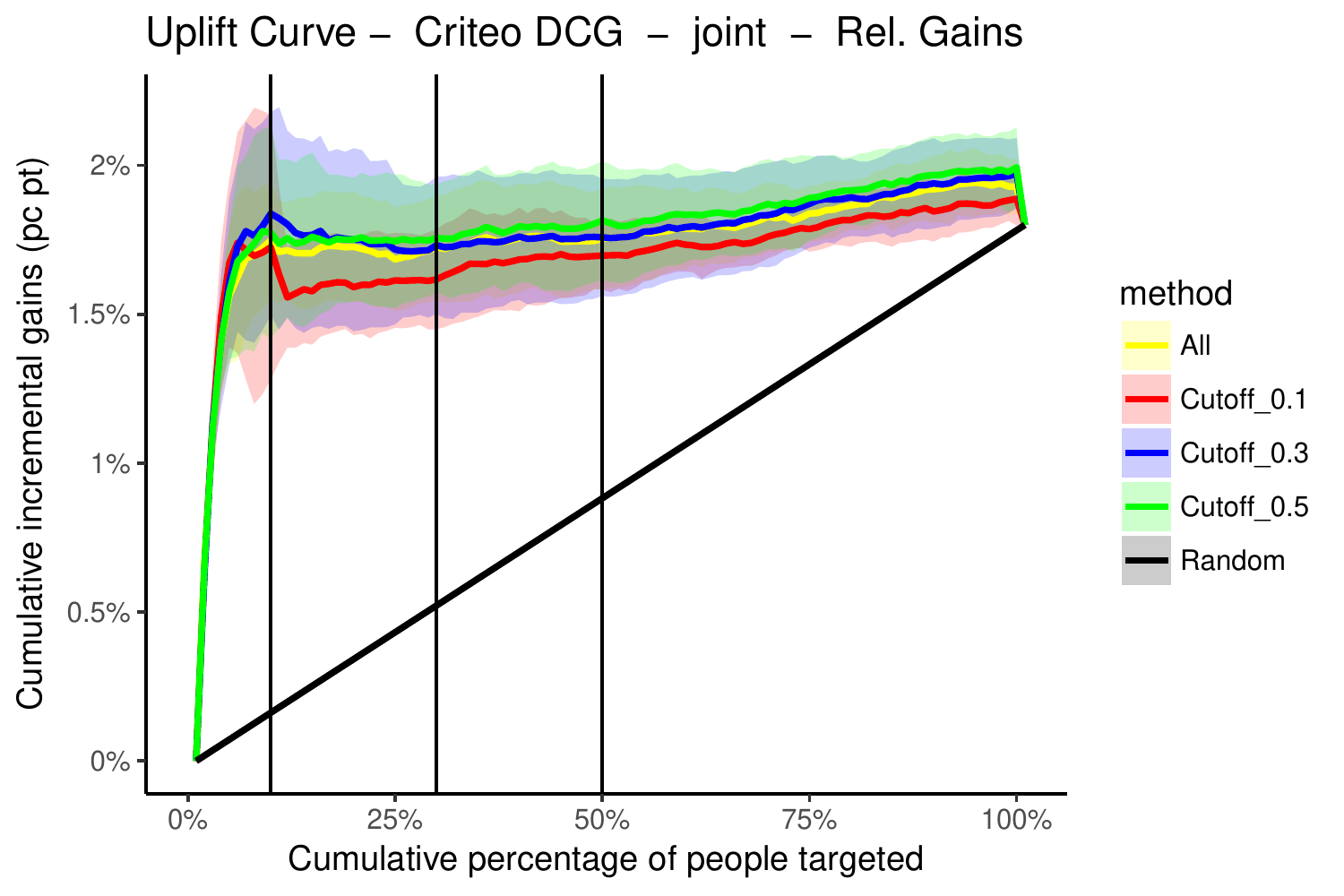}
  \caption{Uplift Curve - Criteo}
\end{subfigure}

\begin{subfigure}{.32\textwidth}
  \centering
  \includegraphics[width=1\linewidth]{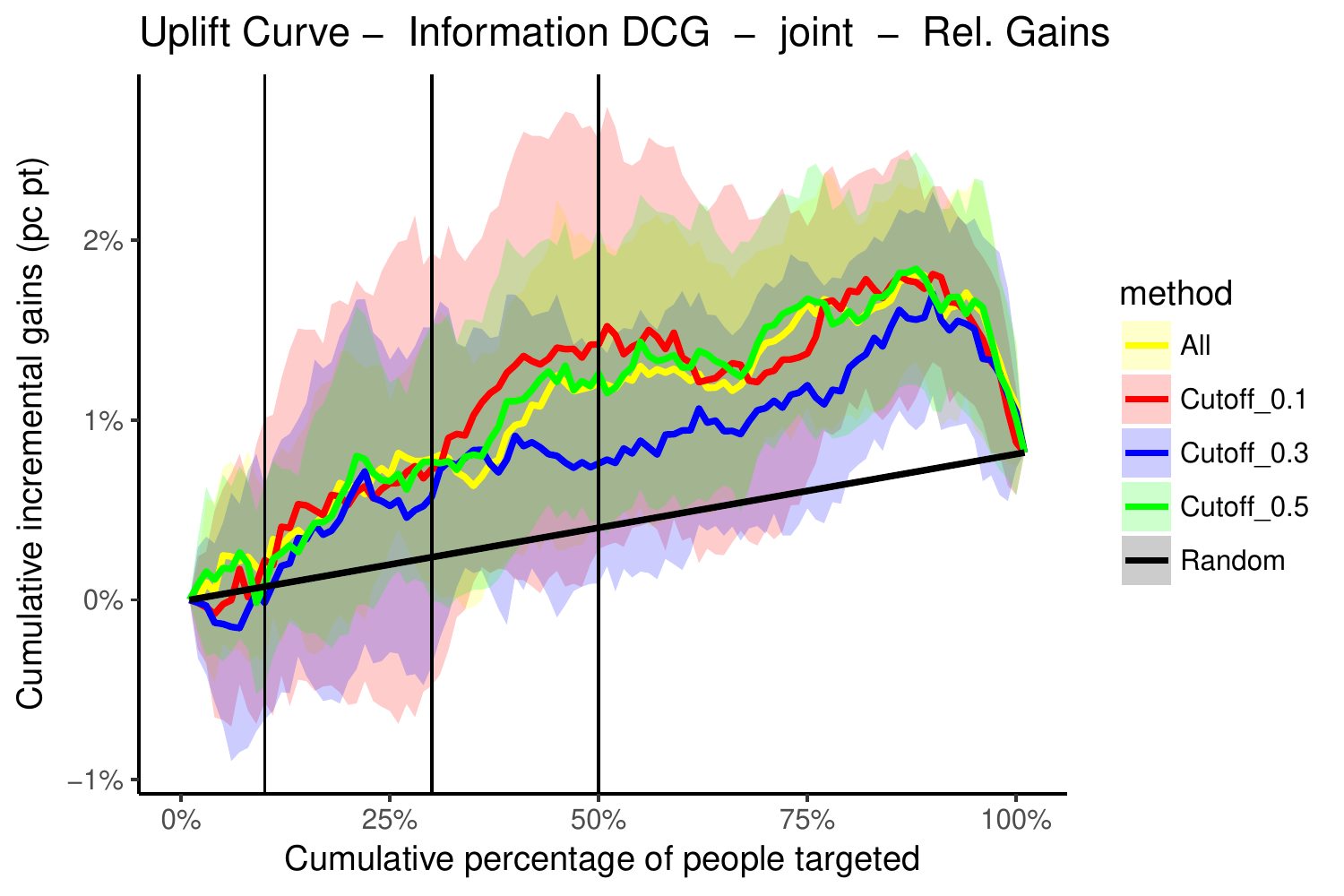}
  \caption{Uplift Curve - Information}
\end{subfigure} %
\begin{subfigure}{.32\textwidth}
  \centering
  \includegraphics[width=1\linewidth]{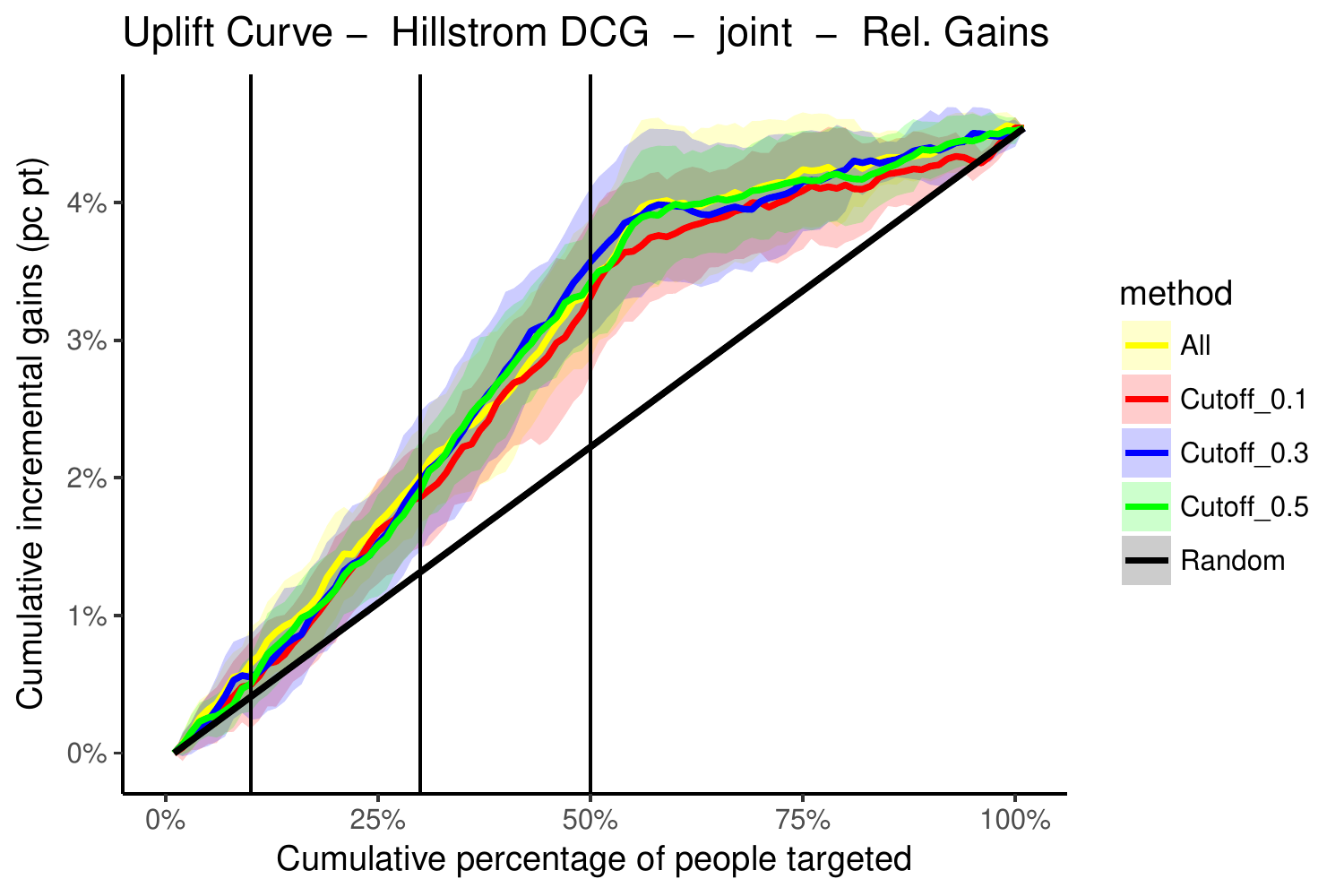}
  \caption{Uplift Curve - Hillstrom}
\end{subfigure} %
\begin{subfigure}{.32\textwidth}
  \centering
  \includegraphics[width=1\linewidth]{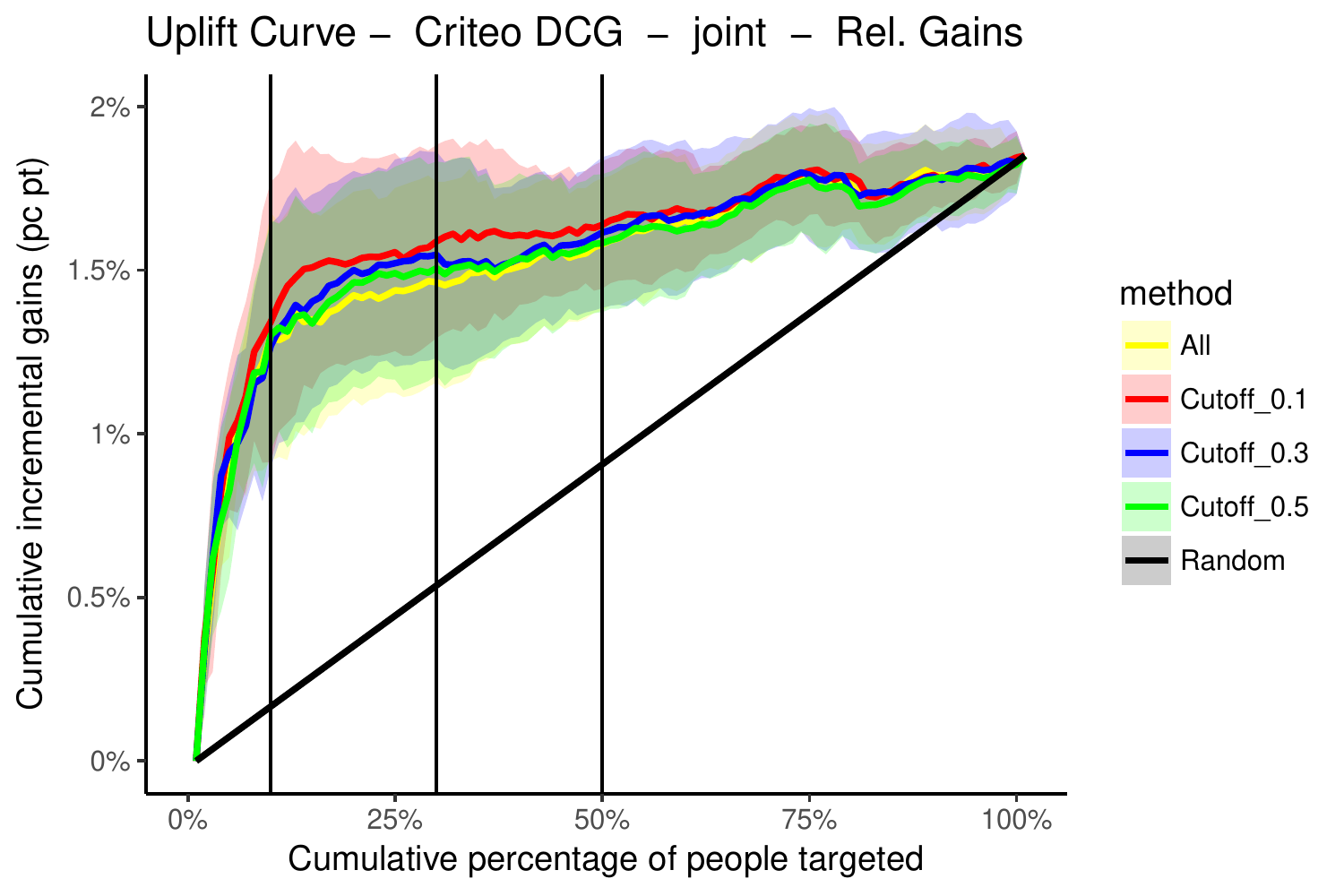}
  \caption{Uplift Curve - Criteo}
\end{subfigure}

\caption{Experiment 2 - DCG for Joint Setting at multiple cutoffs with relative gains.}
\end{figure}


\begin{figure}[ht!]
\centering
\begin{subfigure}{.32\textwidth}
  \centering
  \includegraphics[width=1\linewidth]{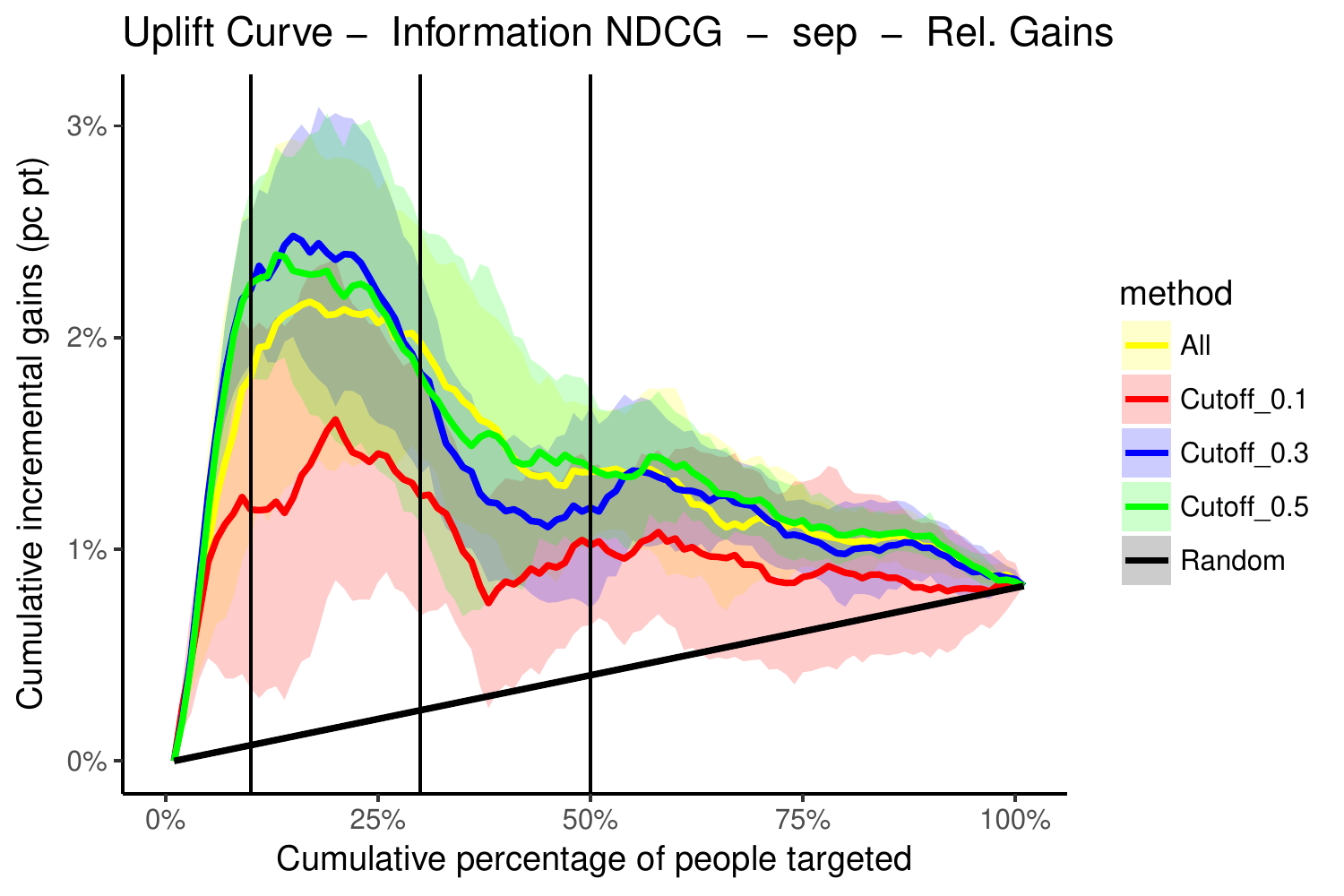}
  \caption{Uplift Curve - Information}
\end{subfigure} %
\begin{subfigure}{.32\textwidth}
  \centering
  \includegraphics[width=1\linewidth]{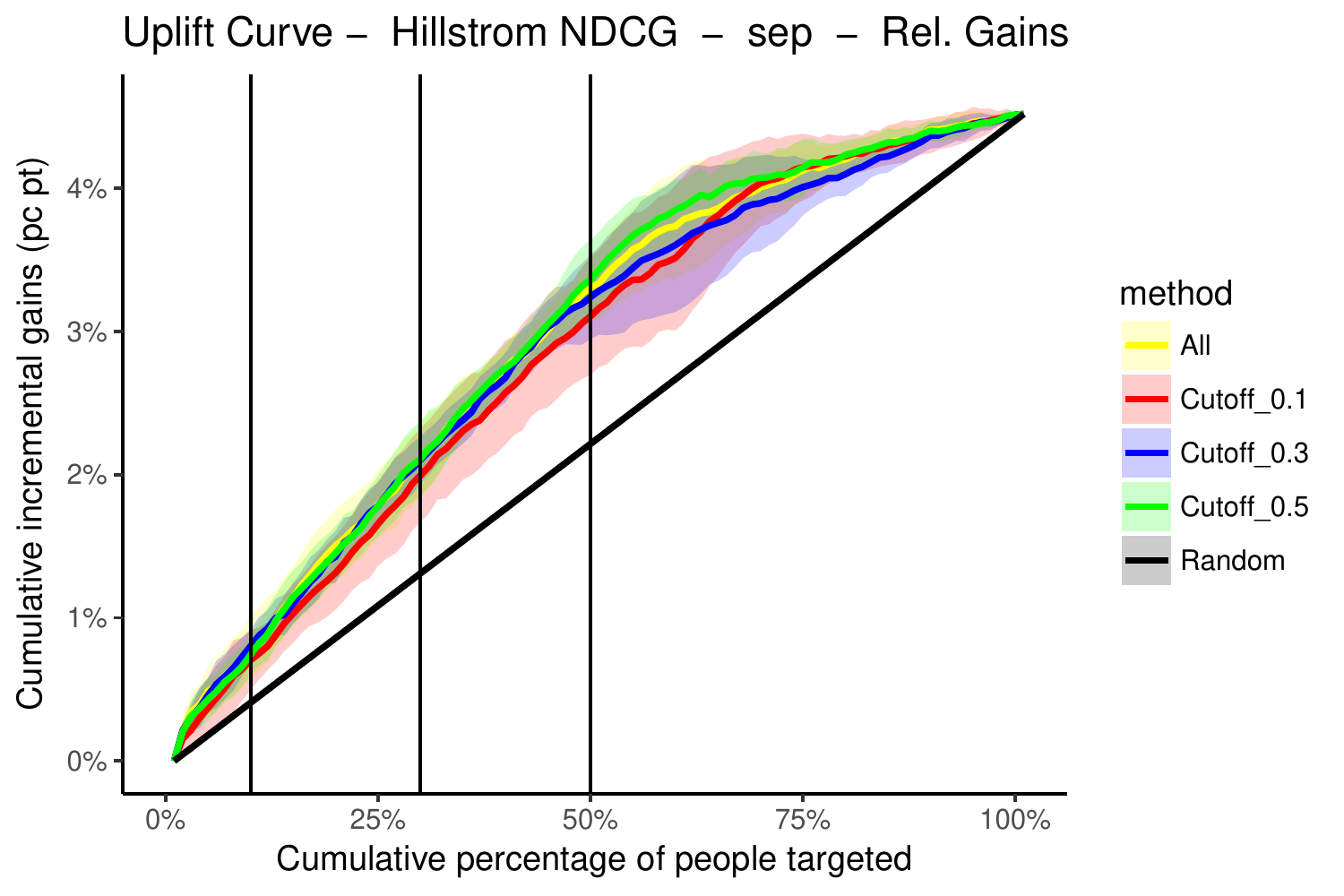}
  \caption{Uplift Curve - Hillstrom}
\end{subfigure} %
\begin{subfigure}{.32\textwidth}
  \centering
  \includegraphics[width=1\linewidth]{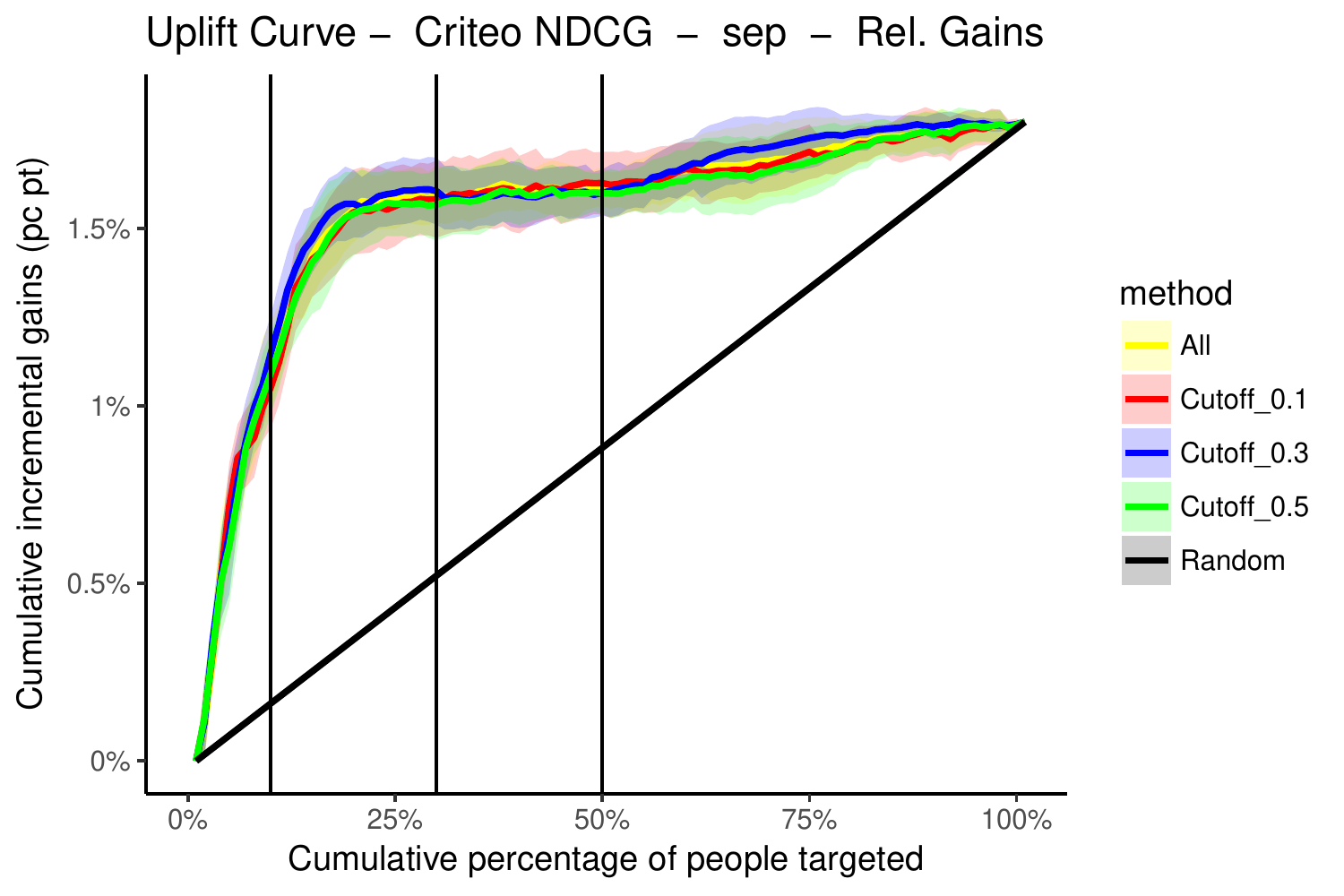}
  \caption{Uplift Curve - Criteo}
\end{subfigure}

\begin{subfigure}{.32\textwidth}
  \centering
  \includegraphics[width=1\linewidth]{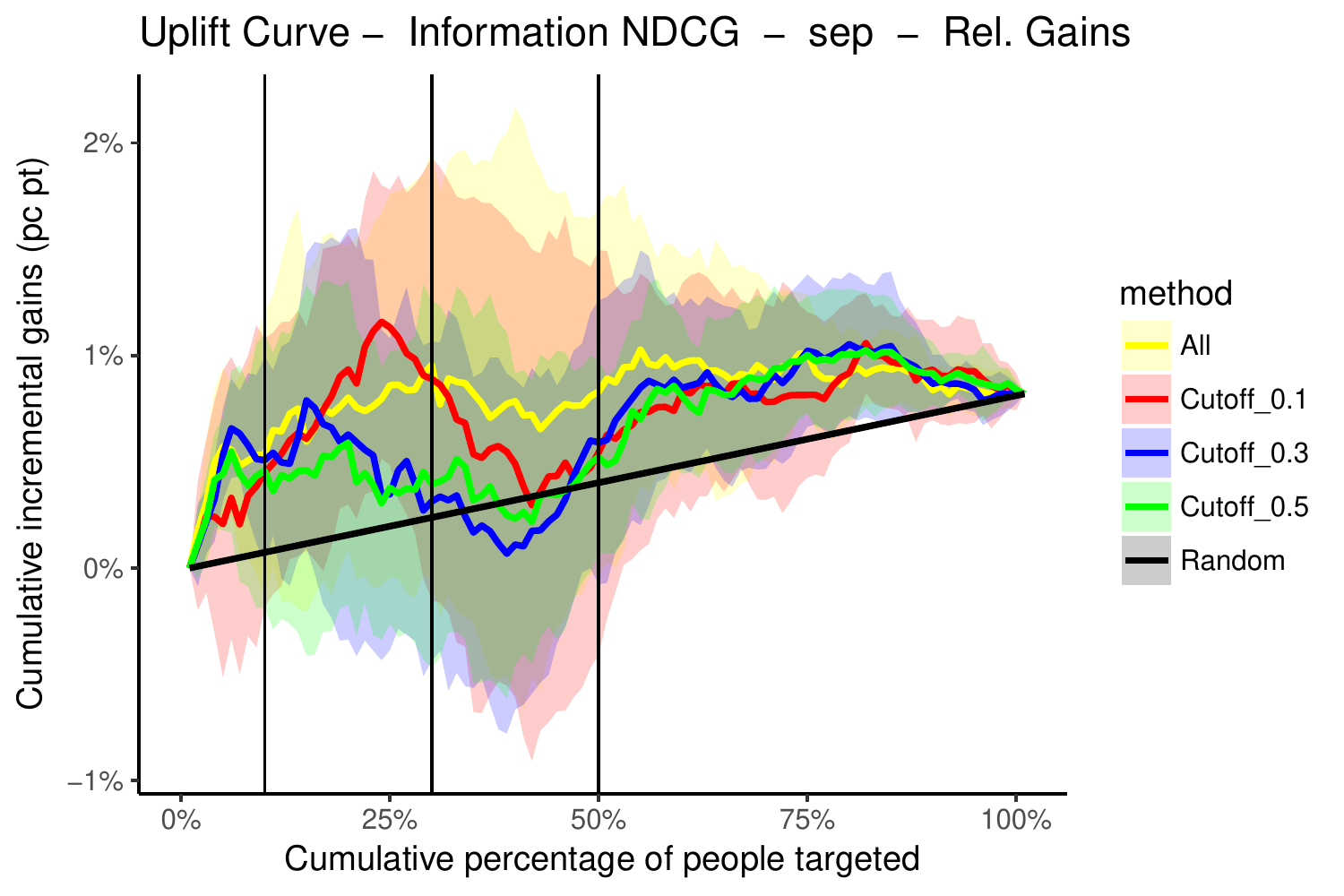}
  \caption{Uplift Curve - Information}
\end{subfigure} %
\begin{subfigure}{.32\textwidth}
  \centering
  \includegraphics[width=1\linewidth]{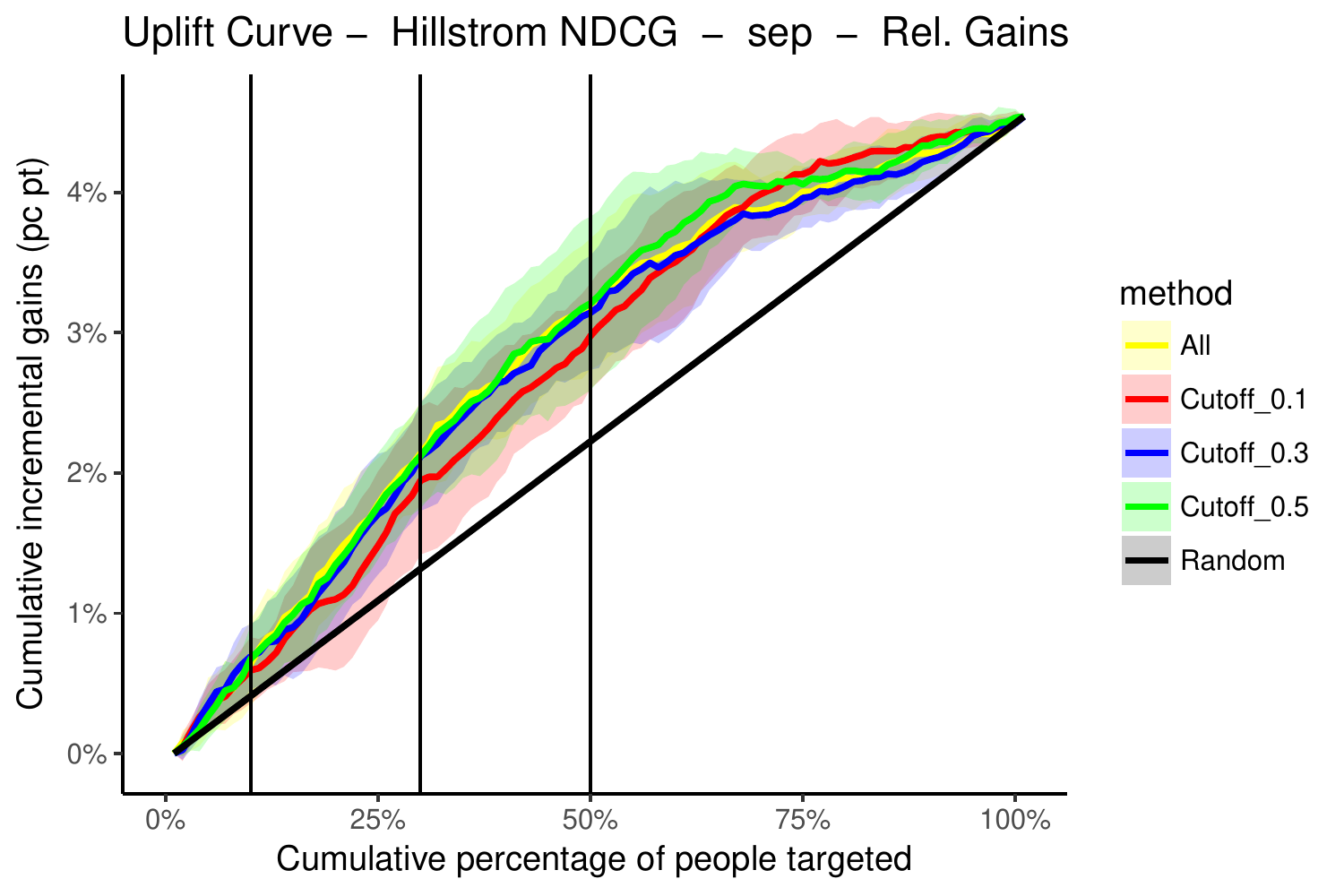}
  \caption{Uplift Curve - Hillstrom}
\end{subfigure} %
\begin{subfigure}{.32\textwidth}
  \centering
  \includegraphics[width=1\linewidth]{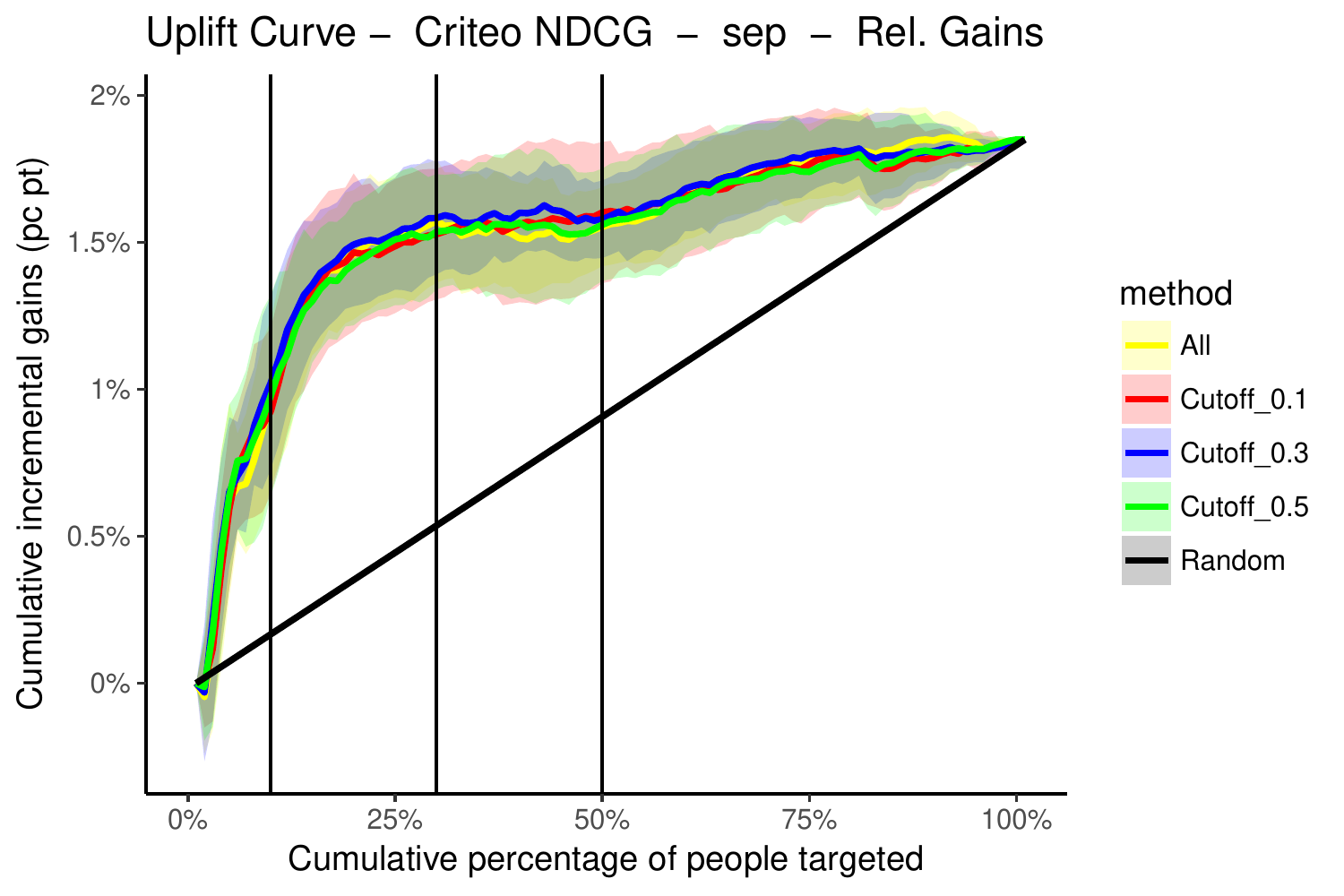}
  \caption{Uplift Curve - Criteo}
\end{subfigure}
\caption{Experiment 2 - NDCG for Separate Setting at multiple cutoffs with relative gains.}
\end{figure}

\begin{figure}[ht!]
\centering
\begin{subfigure}{.32\textwidth}
  \centering
  \includegraphics[width=1\linewidth]{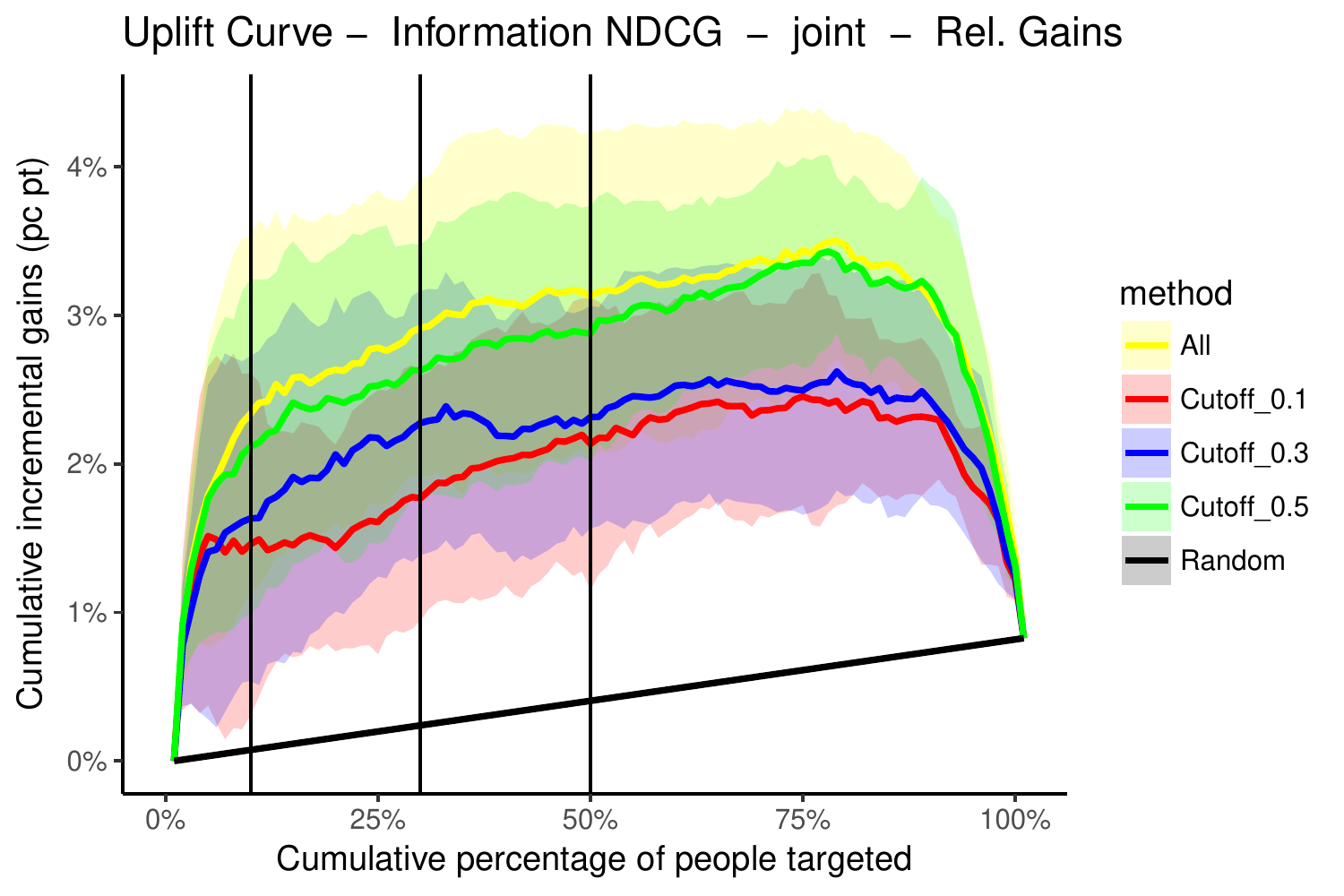}
  \caption{Uplift Curve - Information}
\end{subfigure} %
\begin{subfigure}{.32\textwidth}
  \centering
  \includegraphics[width=1\linewidth]{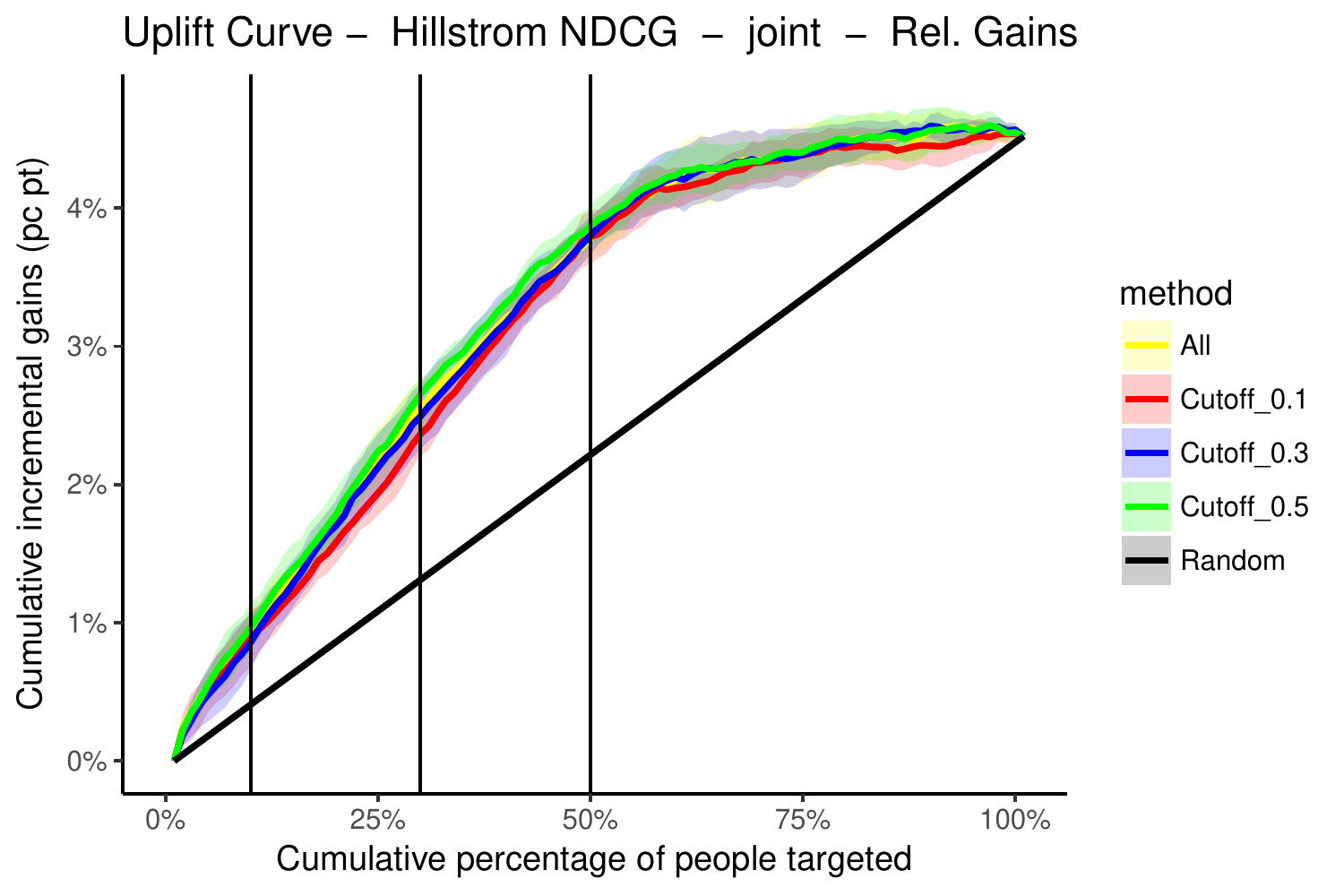}
  \caption{Uplift Curve - Hillstrom}
\end{subfigure} %
\begin{subfigure}{.32\textwidth}
  \centering
  \includegraphics[width=1\linewidth]{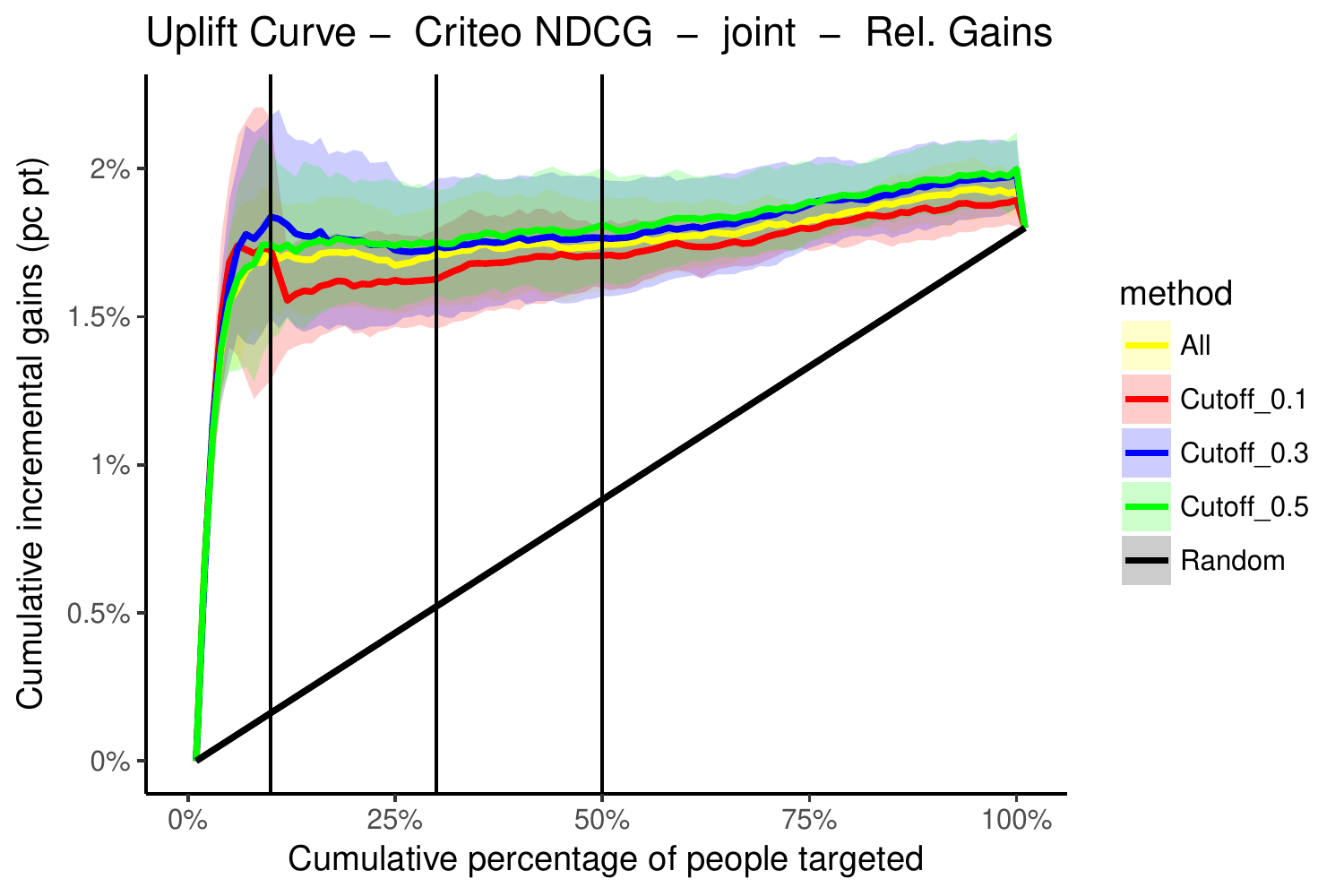}
  \caption{Uplift Curve - Criteo}
\end{subfigure}

\begin{subfigure}{.32\textwidth}
  \centering
  \includegraphics[width=1\linewidth]{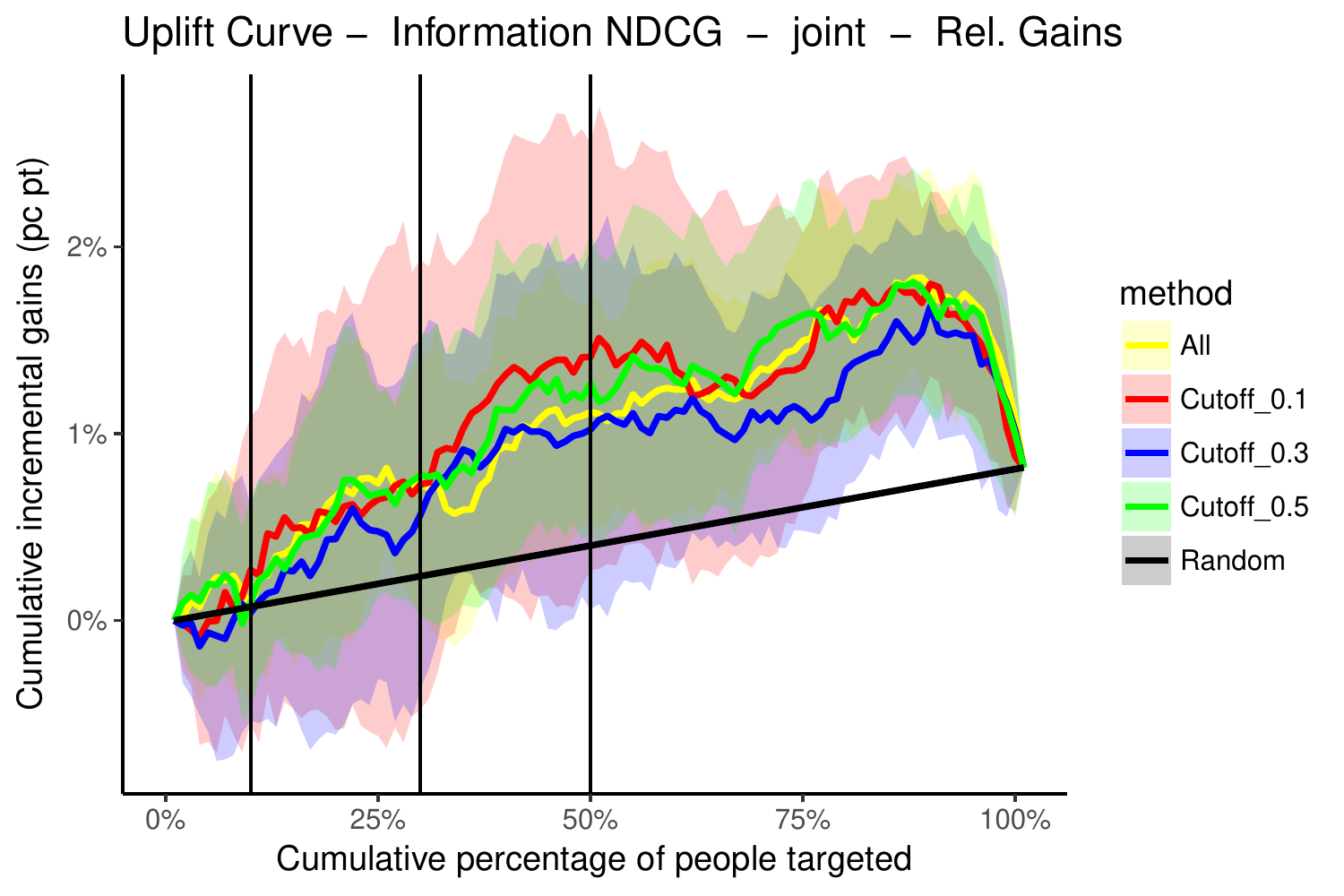}
  \caption{Uplift Curve - Information}
\end{subfigure} %
\begin{subfigure}{.32\textwidth}
  \centering
  \includegraphics[width=1\linewidth]{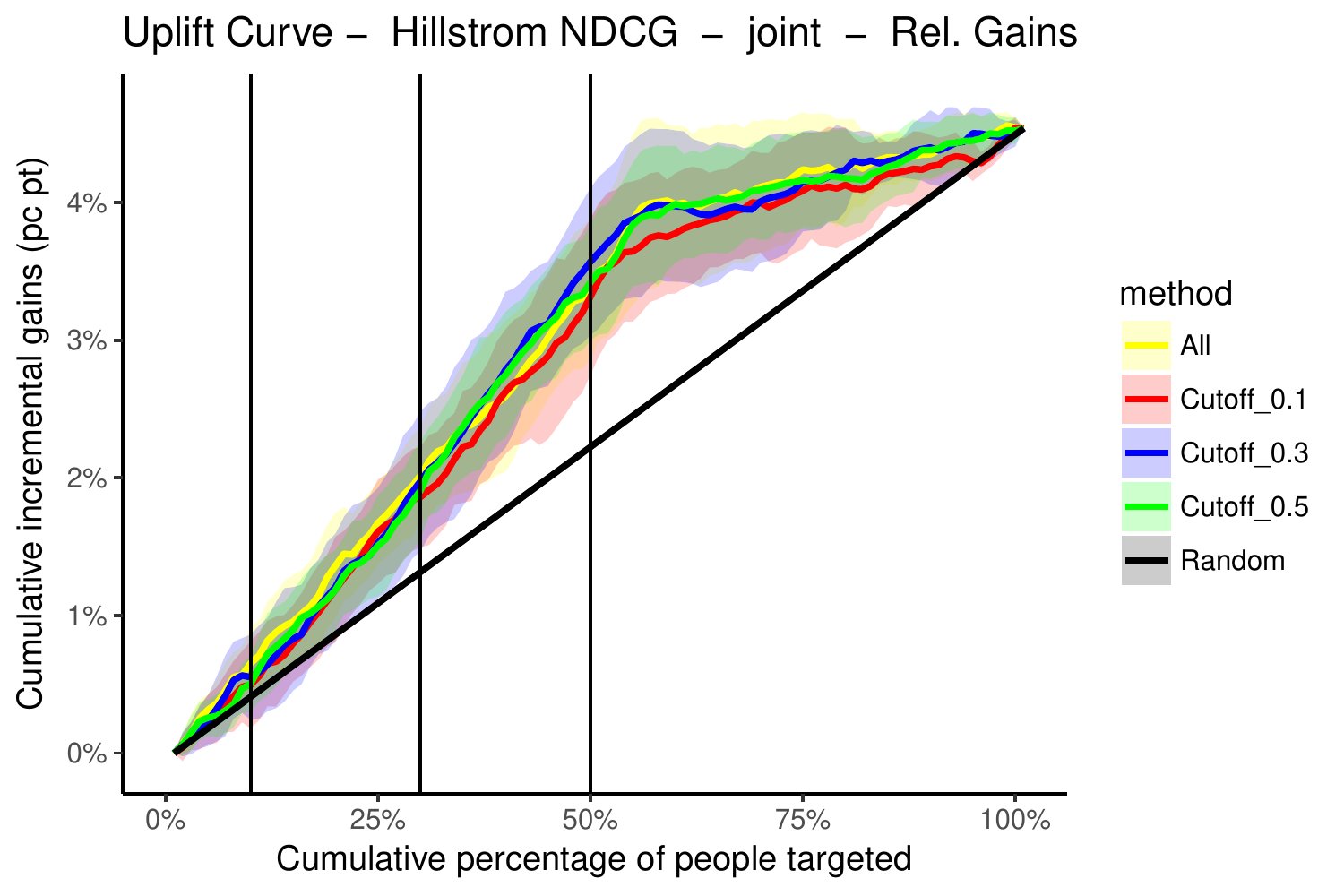}
  \caption{Uplift Curve - Hillstrom}
\end{subfigure} %
\begin{subfigure}{.32\textwidth}
  \centering
  \includegraphics[width=1\linewidth]{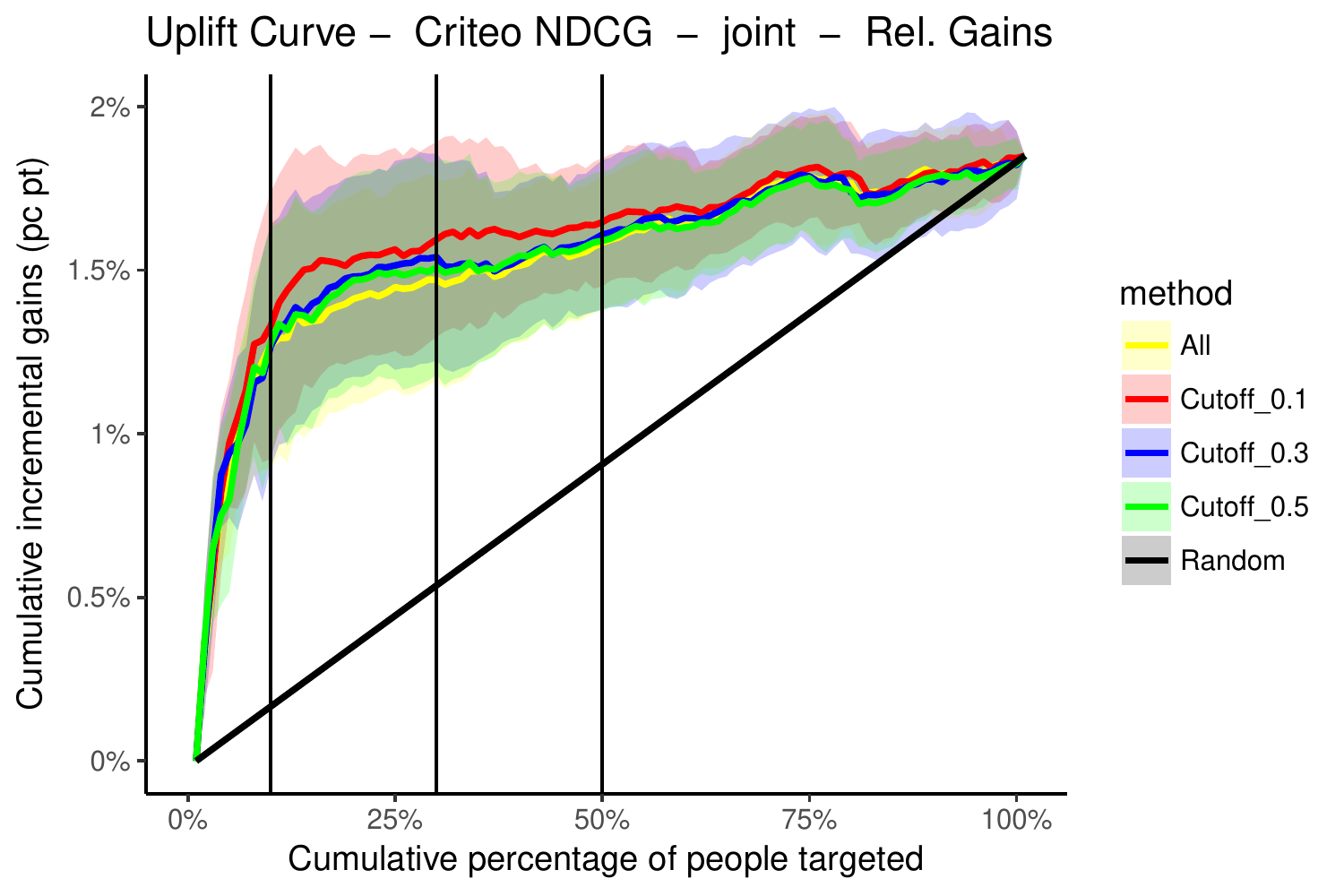}
  \caption{Uplift Curve - Criteo}
\end{subfigure}

\caption{Experiment 2 - NDCG for Joint Setting at multiple cutoffs with relative gains.}
\end{figure}

\clearpage
\subsection*{Appendix D: Experiment 4 - Plots - Absolute Relevance}

\begin{figure}[ht!]
\centering
\begin{subfigure}{.32\textwidth}
  \centering
  \includegraphics[width=1\linewidth]{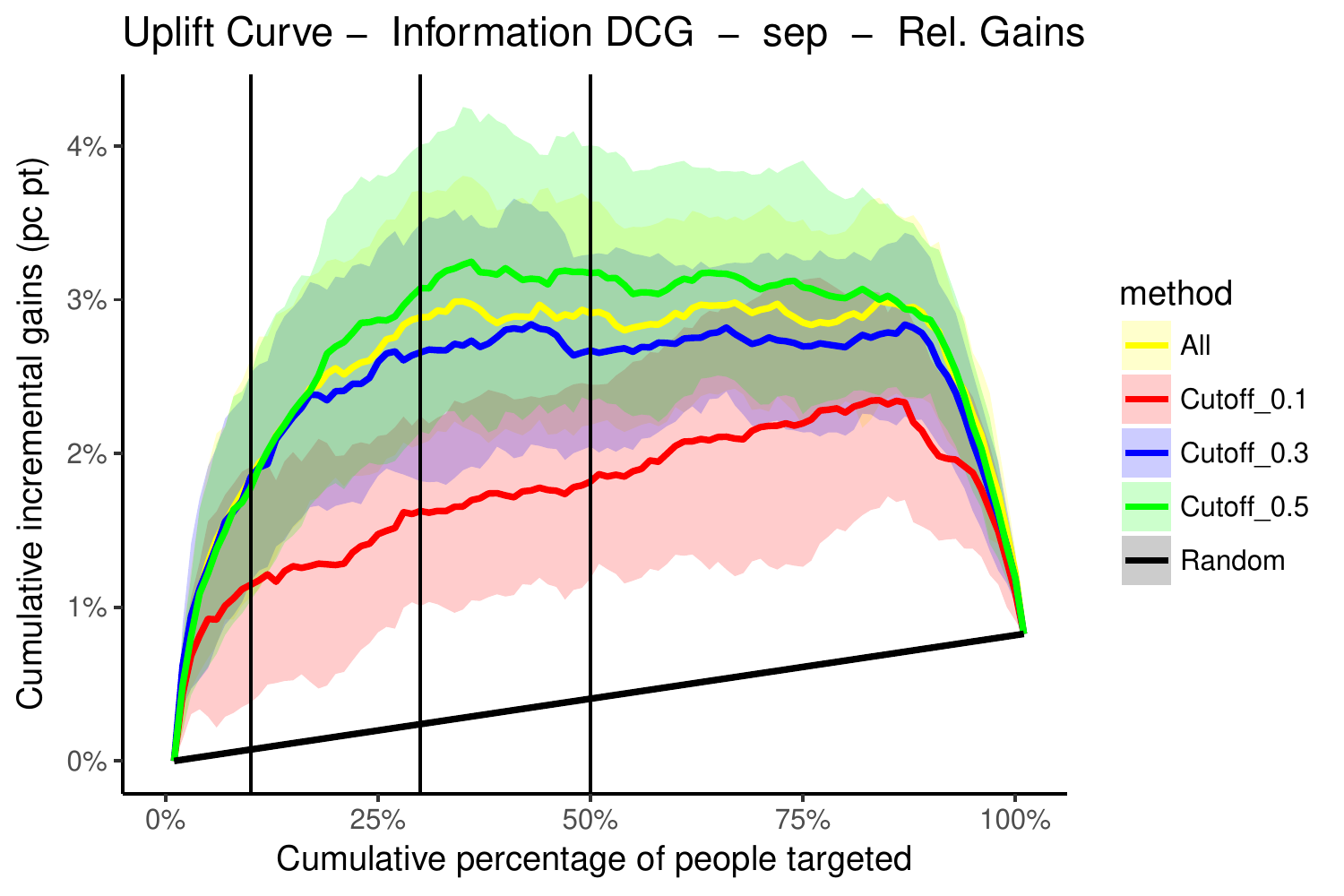}
  \caption{Uplift Curve - Information}
\end{subfigure} %
\begin{subfigure}{.32\textwidth}
  \centering
  \includegraphics[width=1\linewidth]{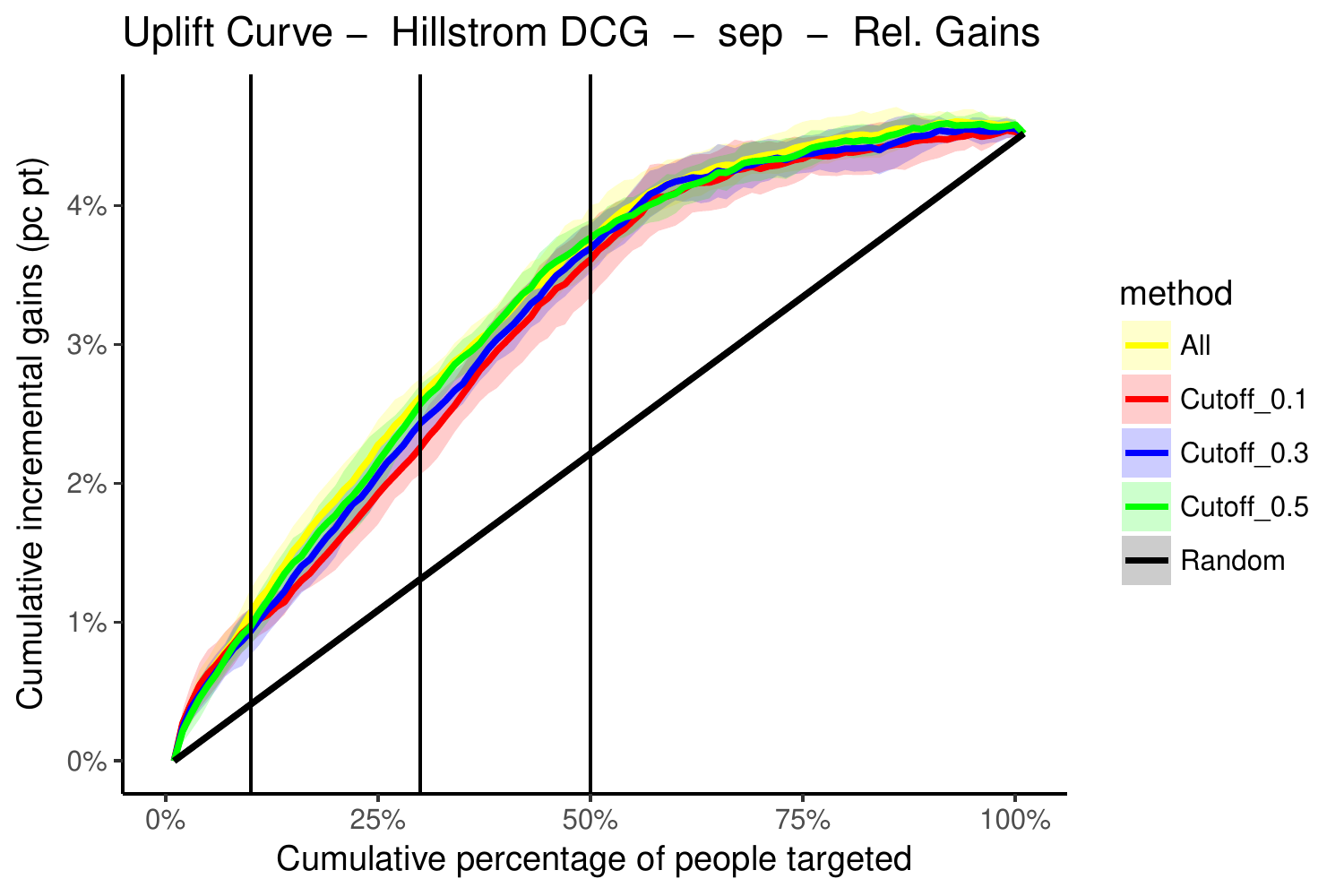}
  \caption{Uplift Curve - Hillstrom}
\end{subfigure} %
\begin{subfigure}{.32\textwidth}
  \centering
  \includegraphics[width=1\linewidth]{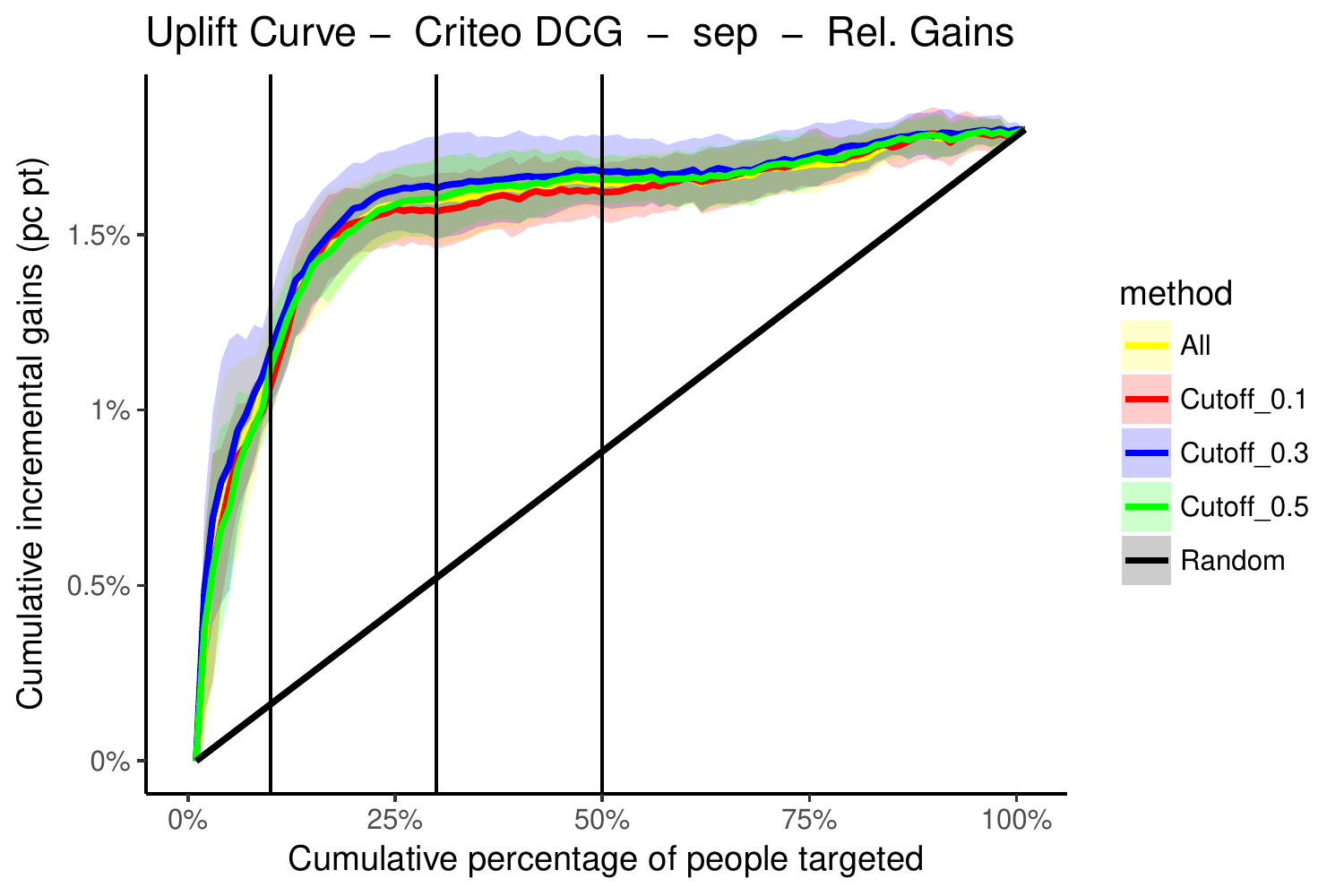}
  \caption{Uplift Curve - Criteo}
\end{subfigure}

\begin{subfigure}{.32\textwidth}
  \centering
  \includegraphics[width=1\linewidth]{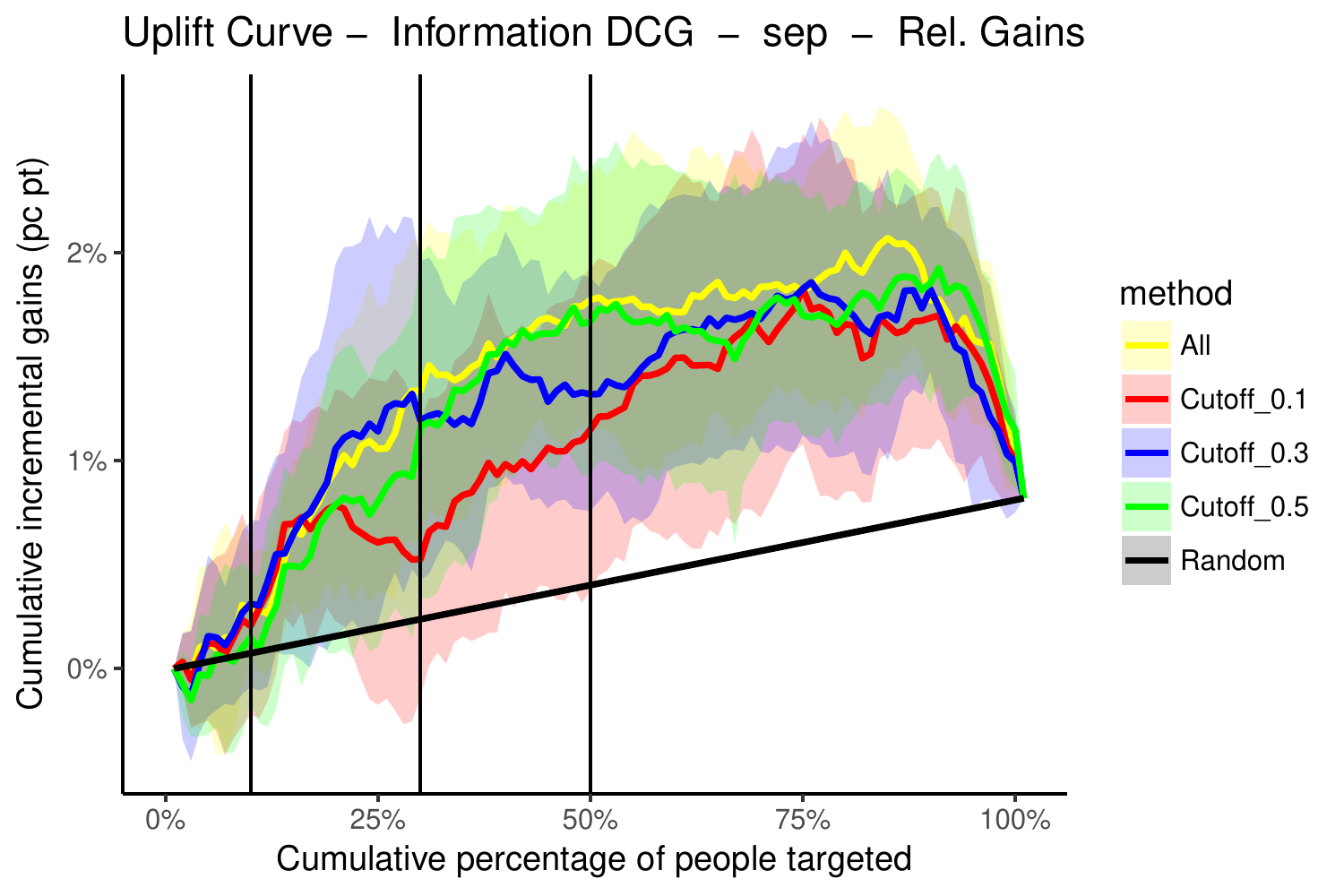}
  \caption{Uplift Curve - Information}
\end{subfigure} %
\begin{subfigure}{.32\textwidth}
  \centering
  \includegraphics[width=1\linewidth]{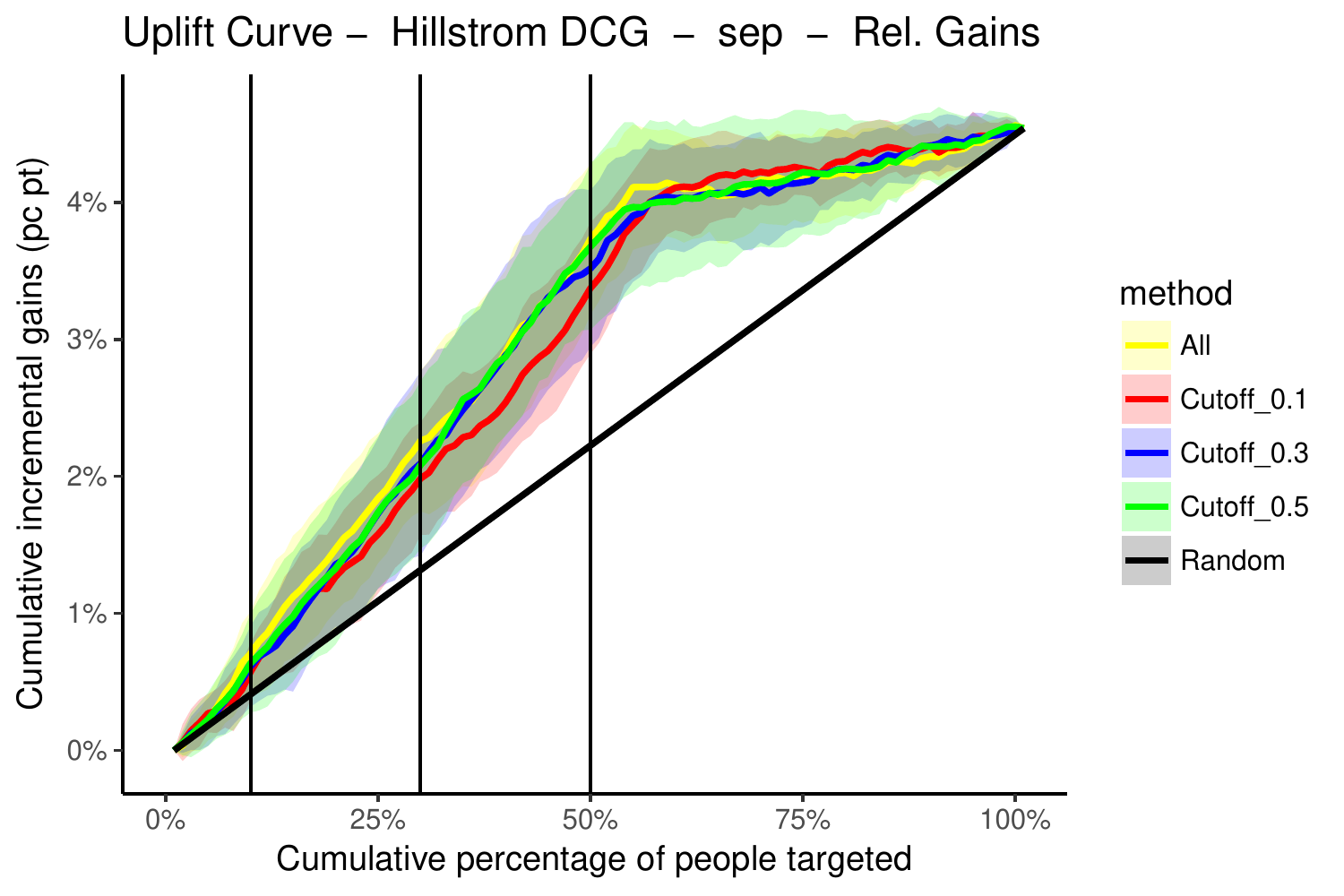}
  \caption{Uplift Curve - Hillstrom}
\end{subfigure} %
\begin{subfigure}{.32\textwidth}
  \centering
  \includegraphics[width=1\linewidth]{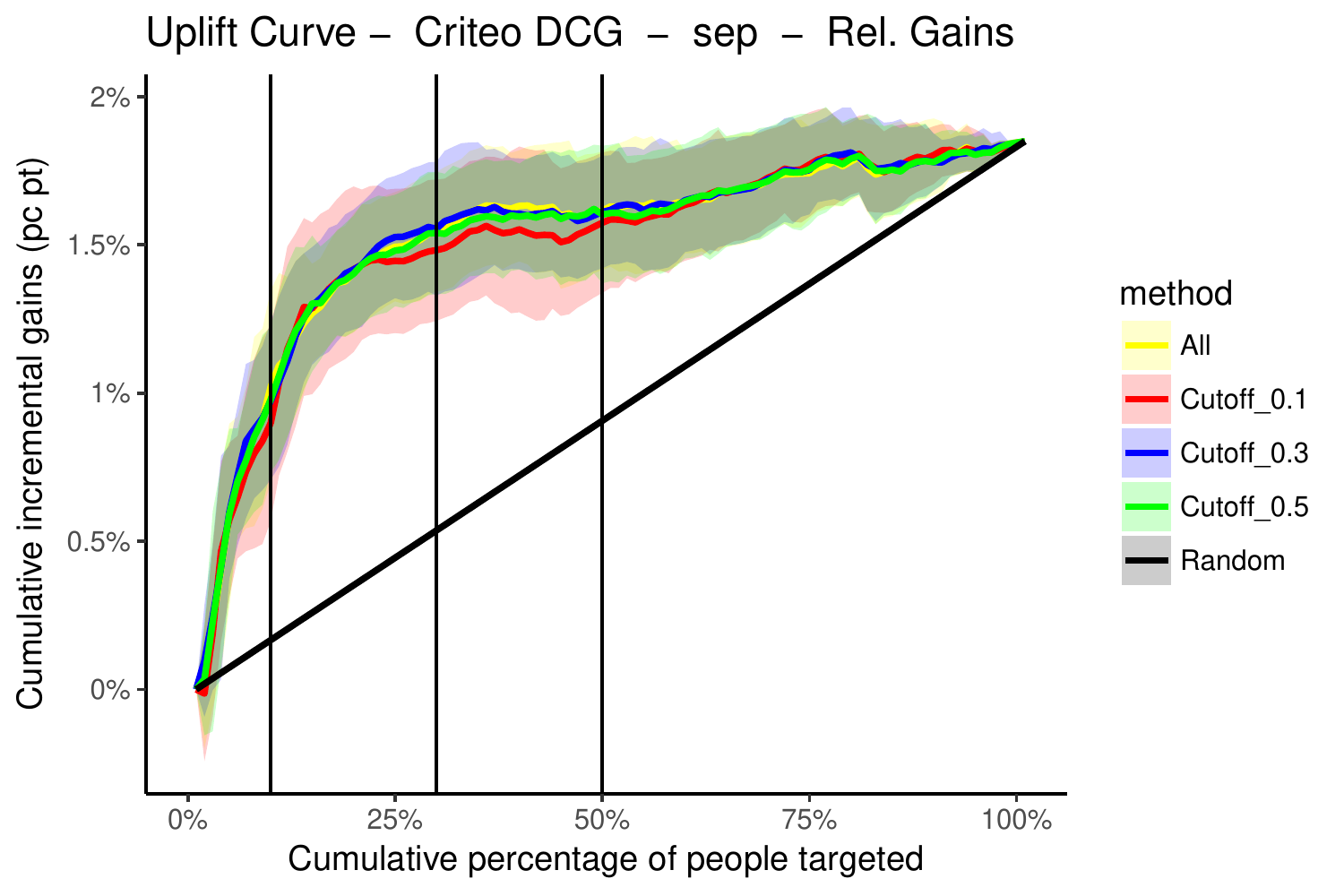}
  \caption{Uplift Curve - Criteo}
\end{subfigure}
\caption{Experiment 2 - DCG for Separate Setting at multiple cutoffs with relative gains.}
\end{figure}

\begin{figure}[ht!]
\centering
\begin{subfigure}{.32\textwidth}
  \centering
  \includegraphics[width=1\linewidth]{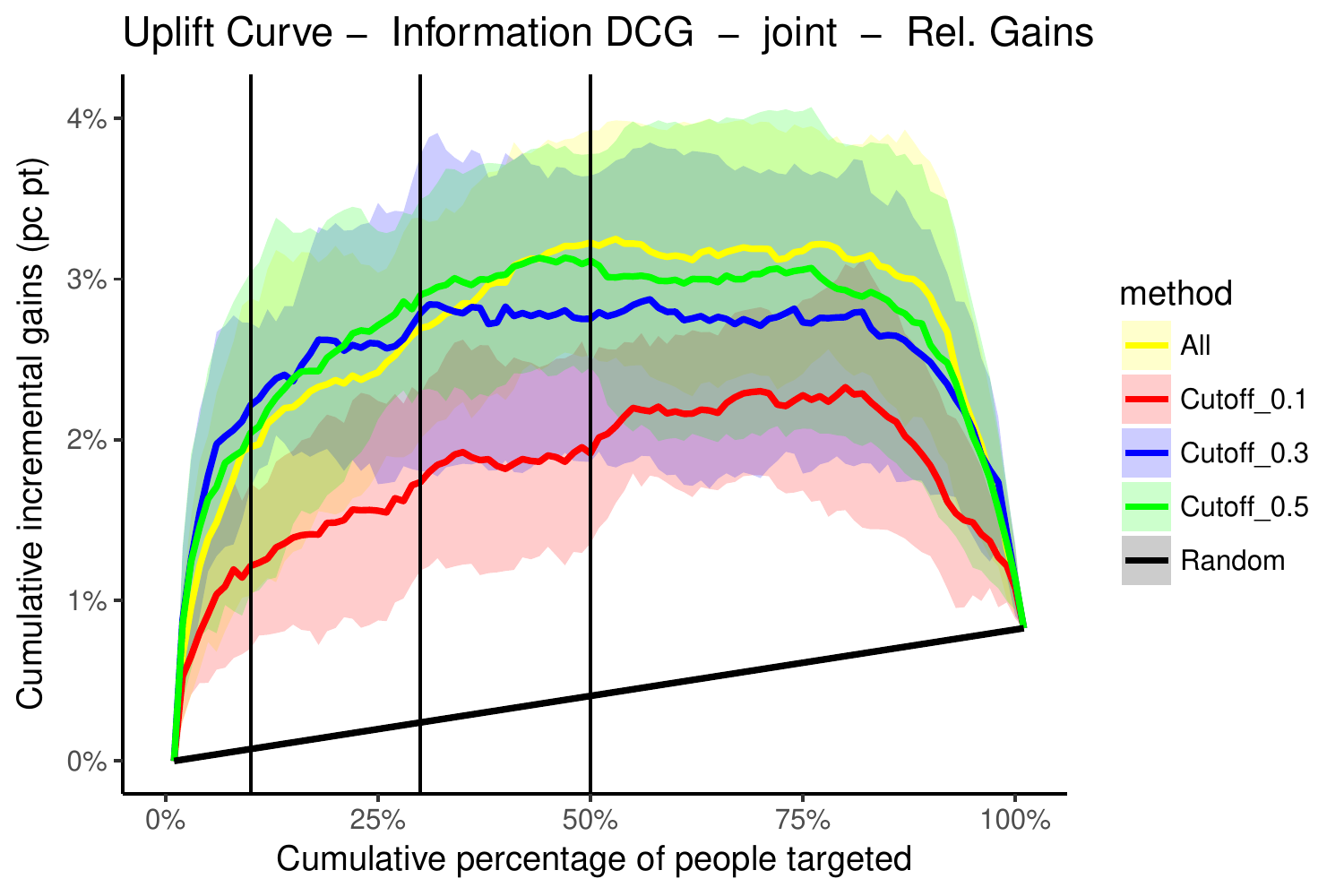}
  \caption{Uplift Curve - Information}
\end{subfigure} %
\begin{subfigure}{.32\textwidth}
  \centering
  \includegraphics[width=1\linewidth]{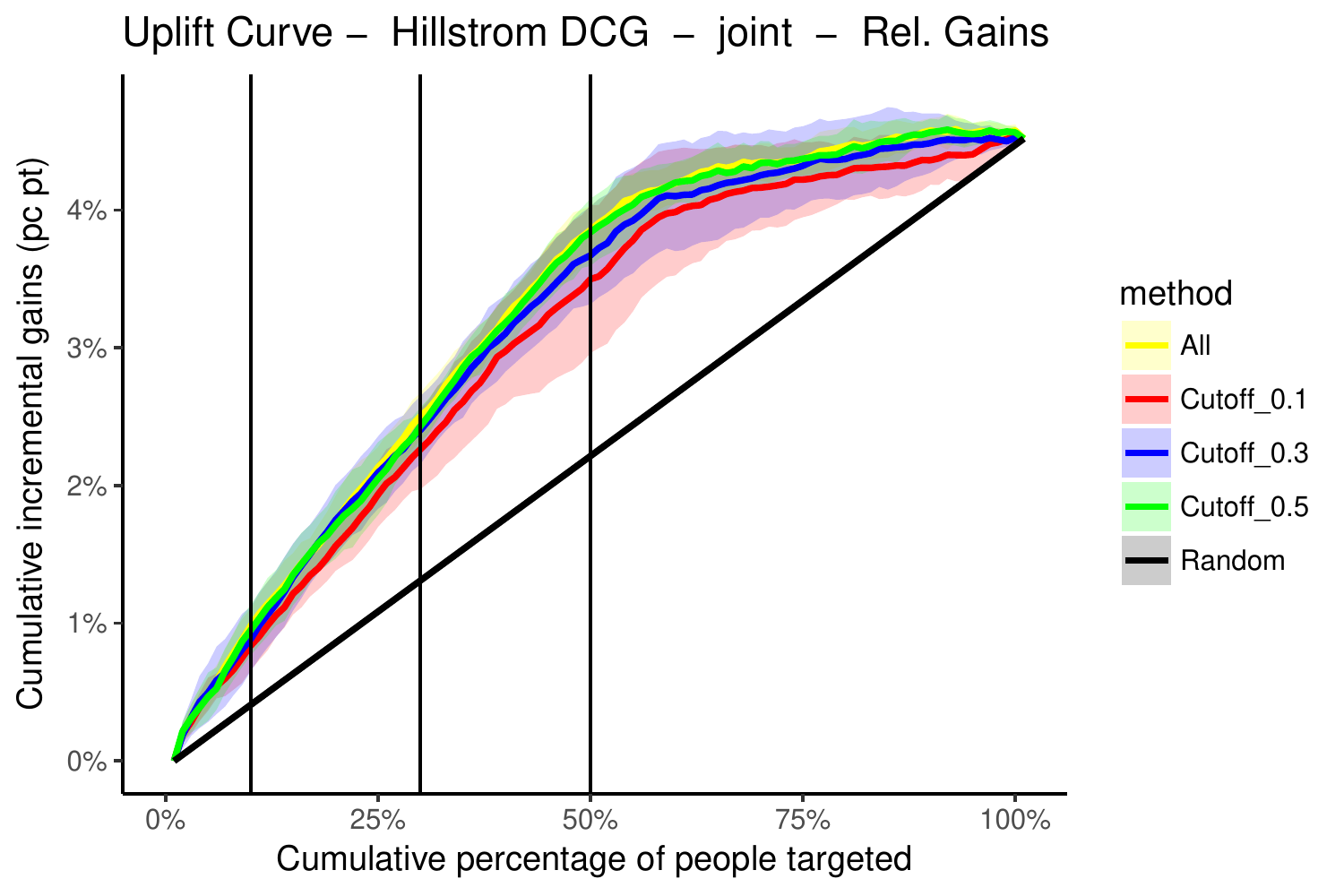}
  \caption{Uplift Curve - Hillstrom}
\end{subfigure} %
\begin{subfigure}{.32\textwidth}
  \centering
  \includegraphics[width=1\linewidth]{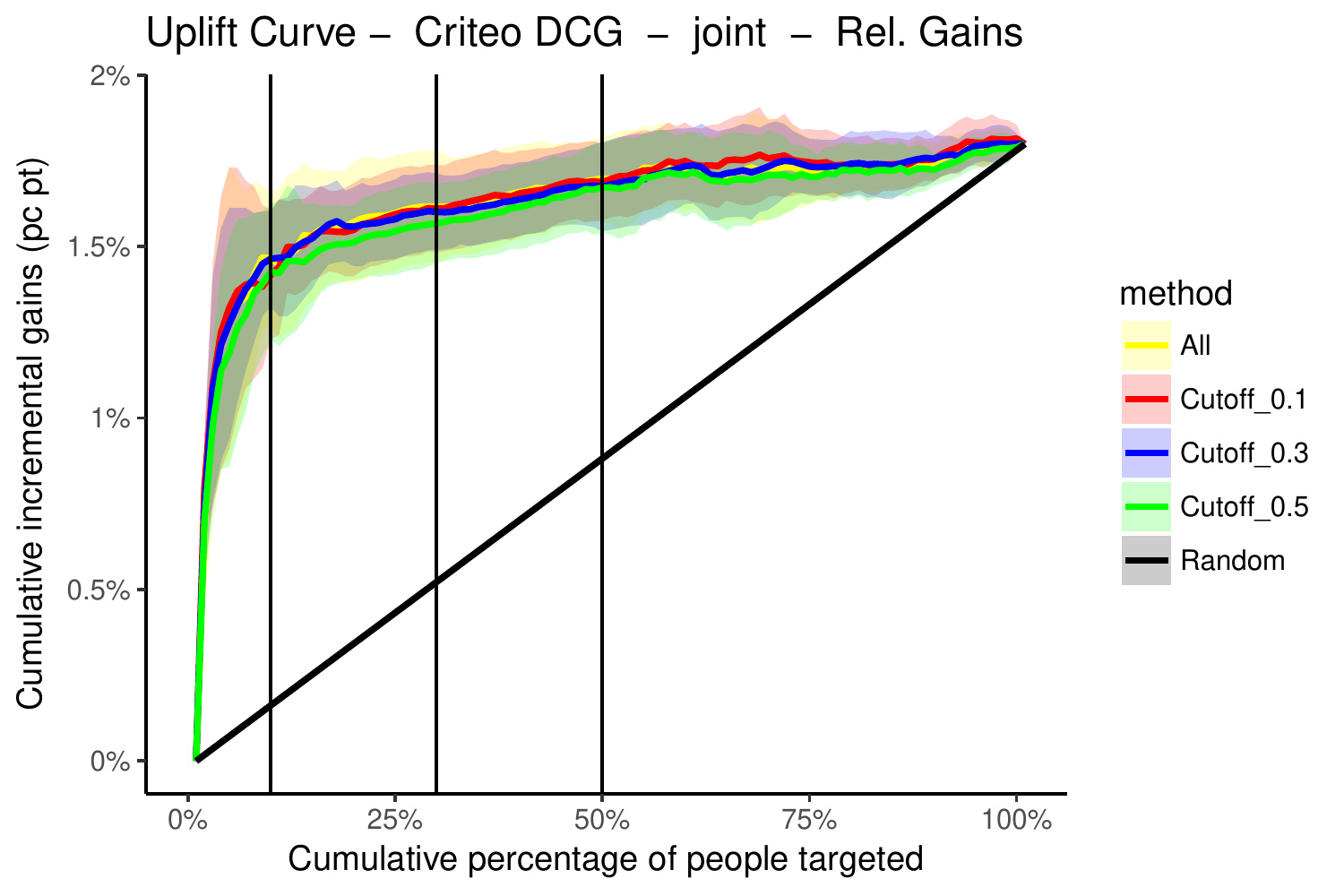}
  \caption{Uplift Curve - Criteo}
\end{subfigure}

\begin{subfigure}{.32\textwidth}
  \centering
  \includegraphics[width=1\linewidth]{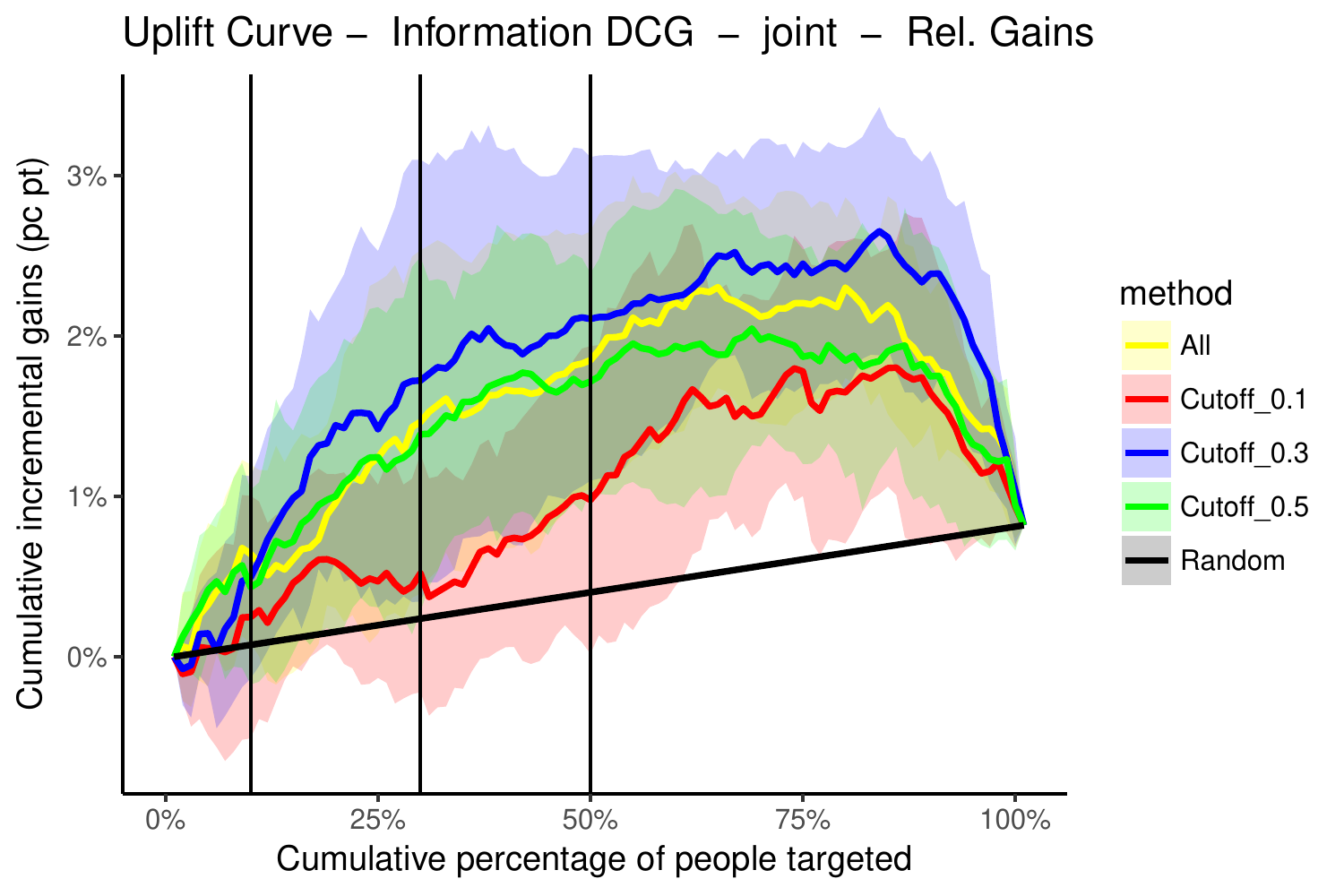}
  \caption{Uplift Curve - Information}
\end{subfigure} %
\begin{subfigure}{.32\textwidth}
  \centering
  \includegraphics[width=1\linewidth]{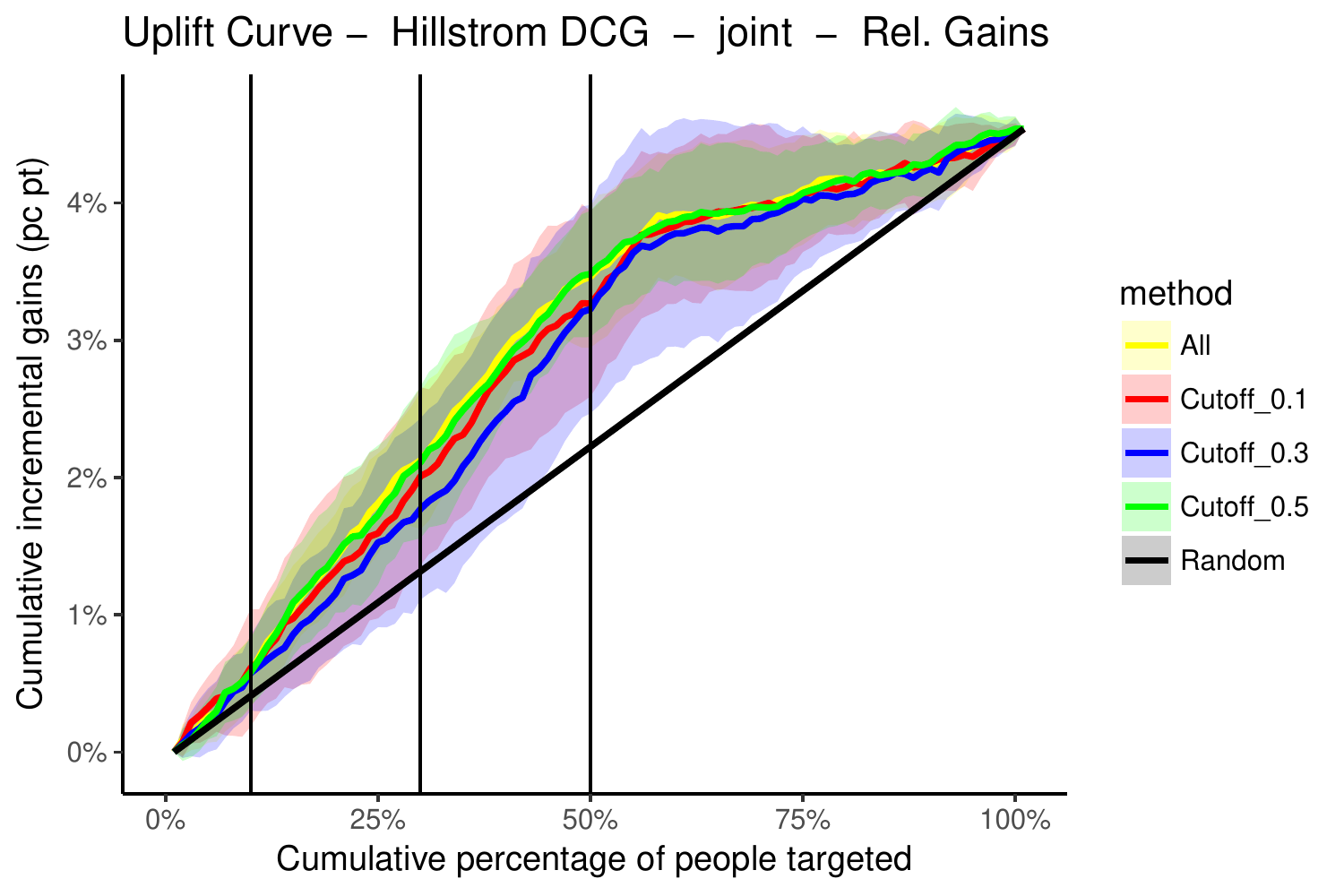}
  \caption{Uplift Curve - Hillstrom}
\end{subfigure} %
\begin{subfigure}{.32\textwidth}
  \centering
  \includegraphics[width=1\linewidth]{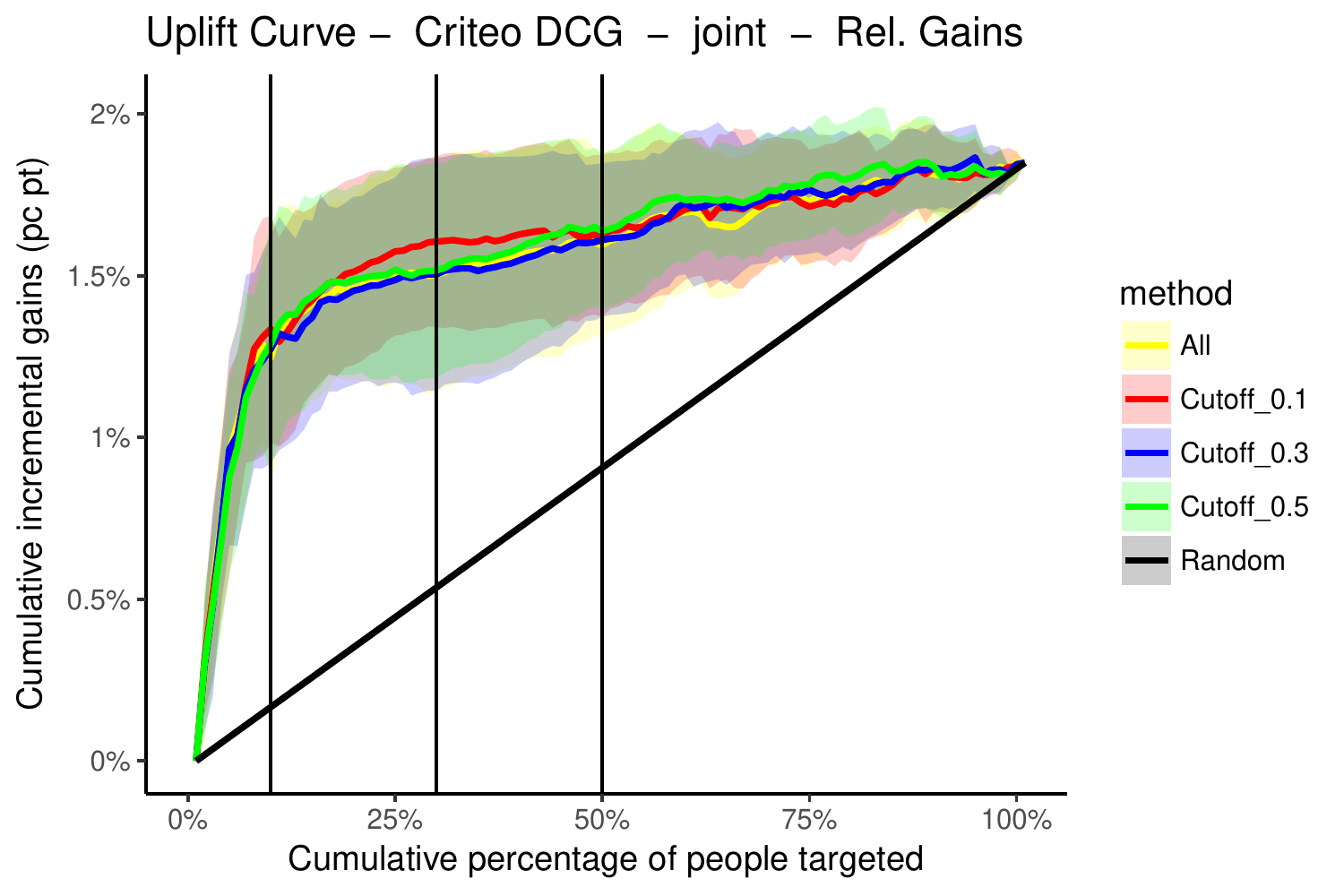}
  \caption{Uplift Curve - Criteo}
\end{subfigure}

\caption{Experiment 2 - DCG for Joint Setting at multiple cutoffs with relative gains.}
\end{figure}

\begin{figure}[ht!]
\centering
\begin{subfigure}{.32\textwidth}
  \centering
  \includegraphics[width=1\linewidth]{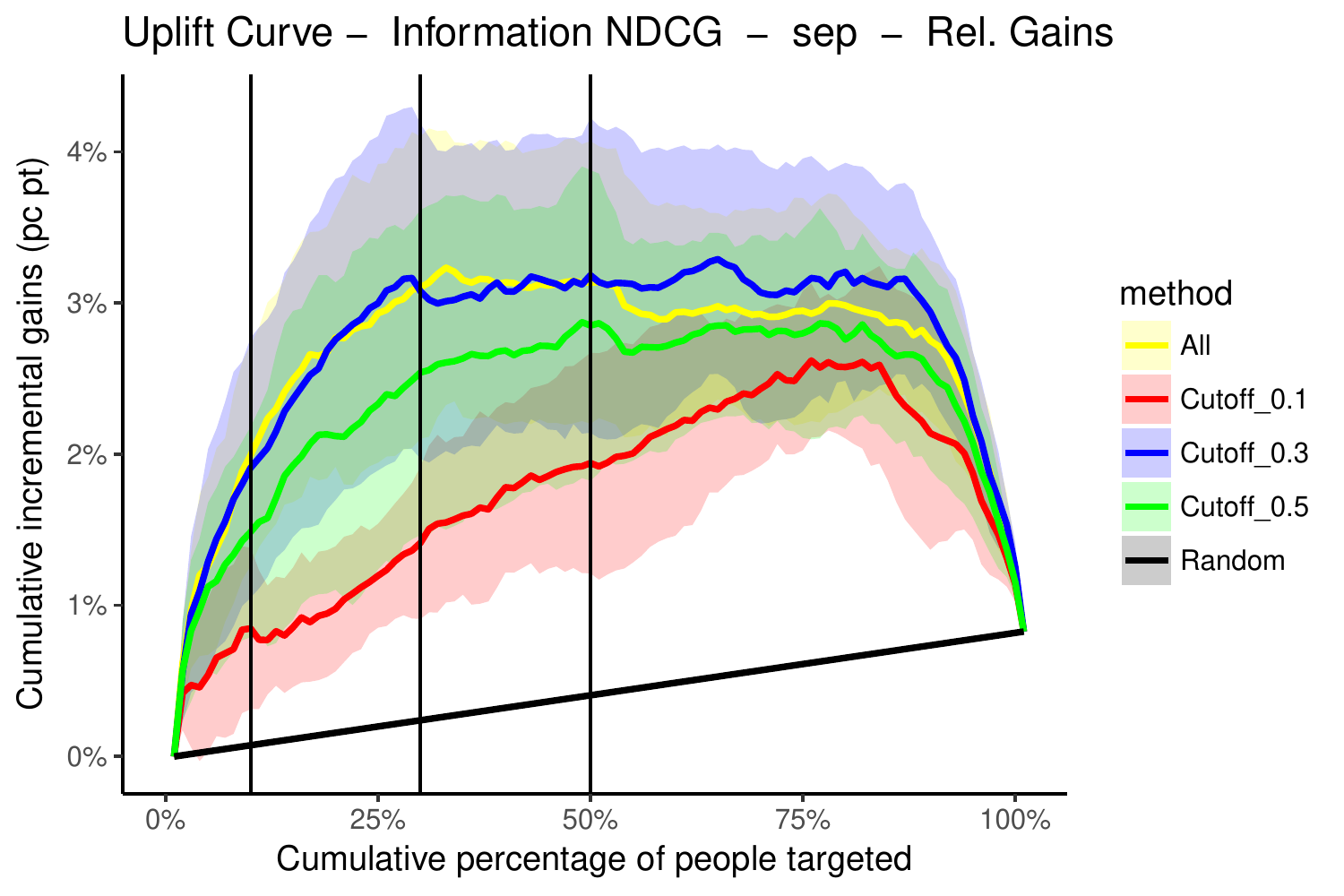}
  \caption{Uplift Curve - Information}
\end{subfigure} %
\begin{subfigure}{.32\textwidth}
  \centering
  \includegraphics[width=1\linewidth]{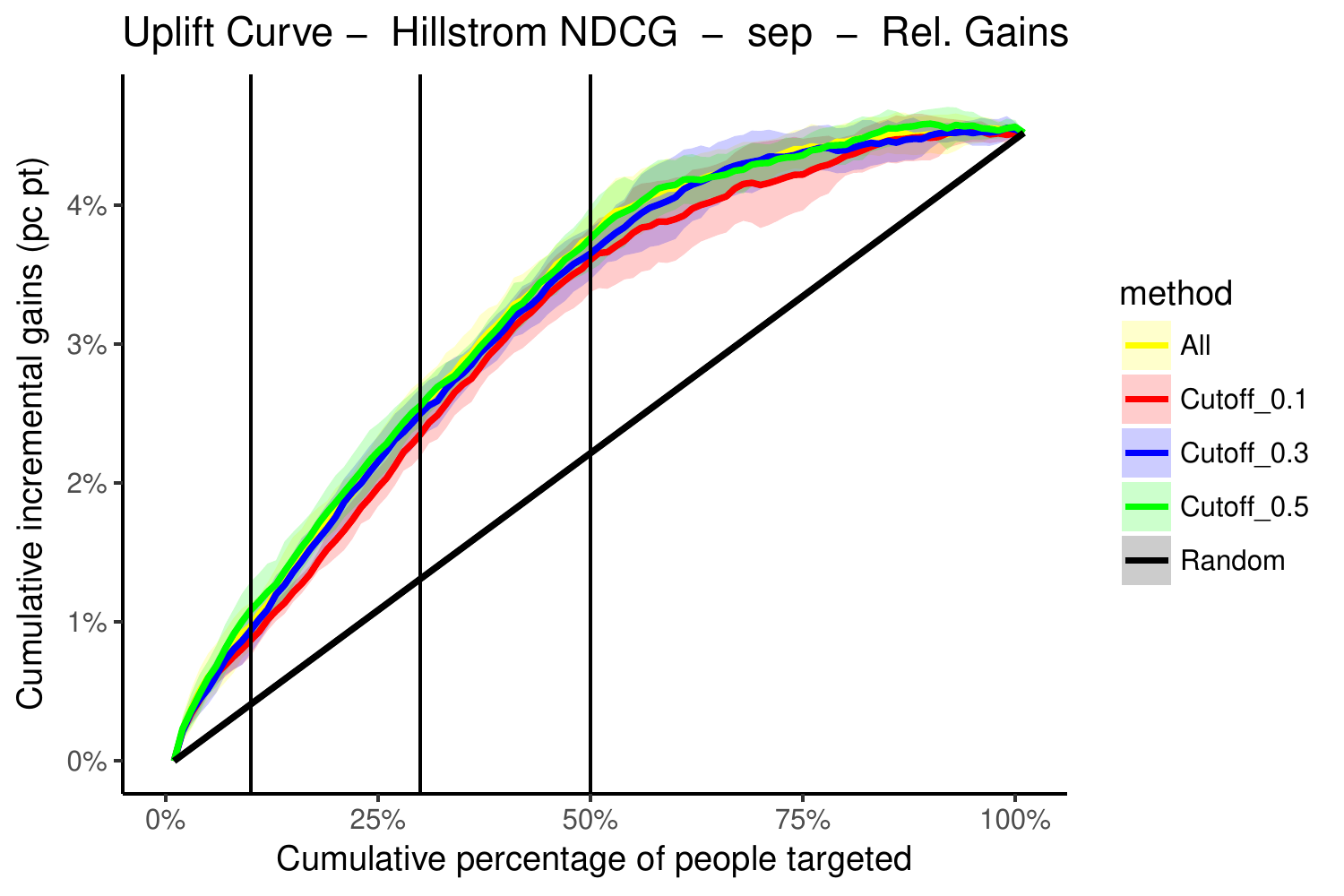}
  \caption{Uplift Curve - Hillstrom}
\end{subfigure} %
\begin{subfigure}{.32\textwidth}
  \centering
  \includegraphics[width=1\linewidth]{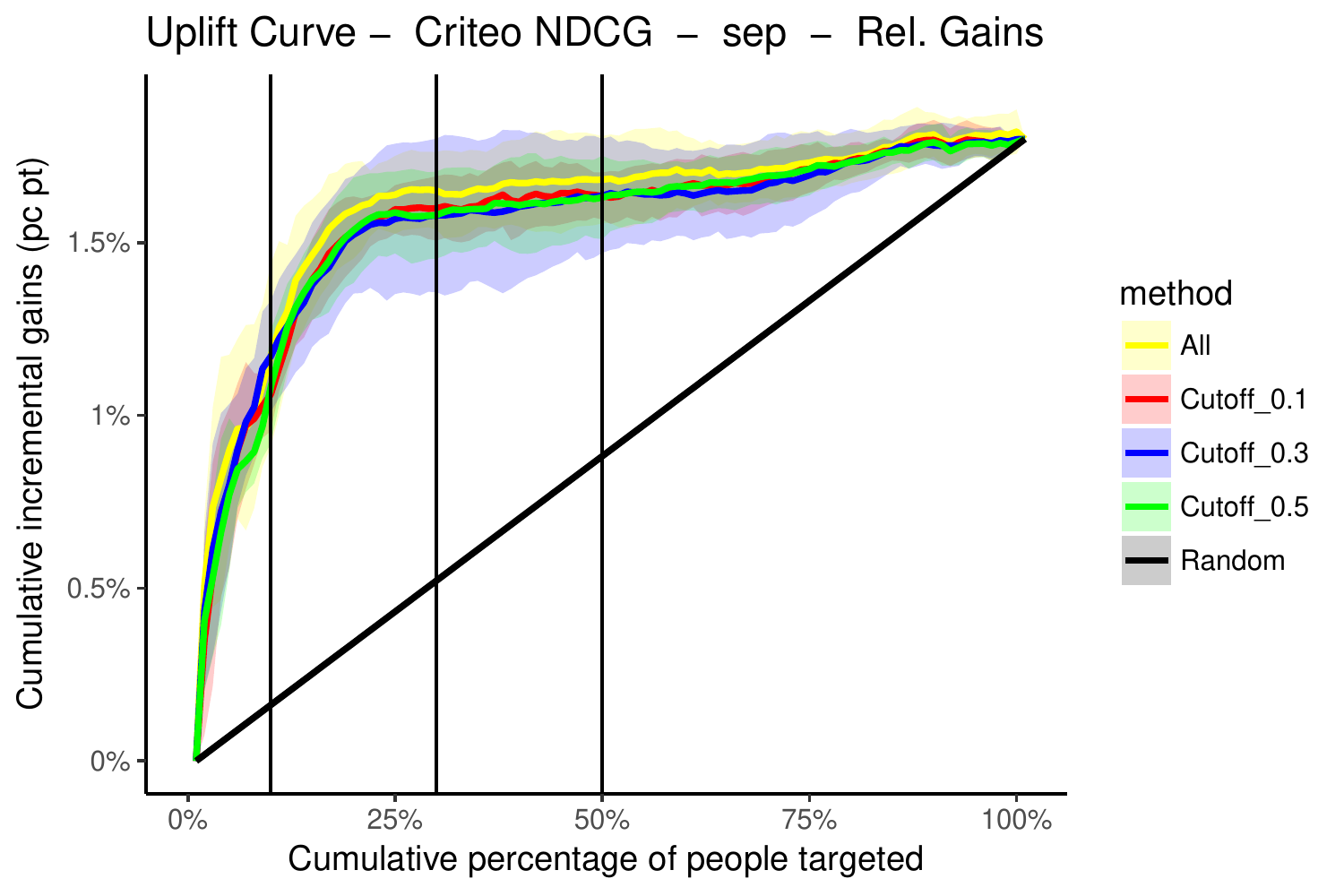}
  \caption{Uplift Curve - Criteo}
\end{subfigure}

\begin{subfigure}{.32\textwidth}
  \centering
  \includegraphics[width=1\linewidth]{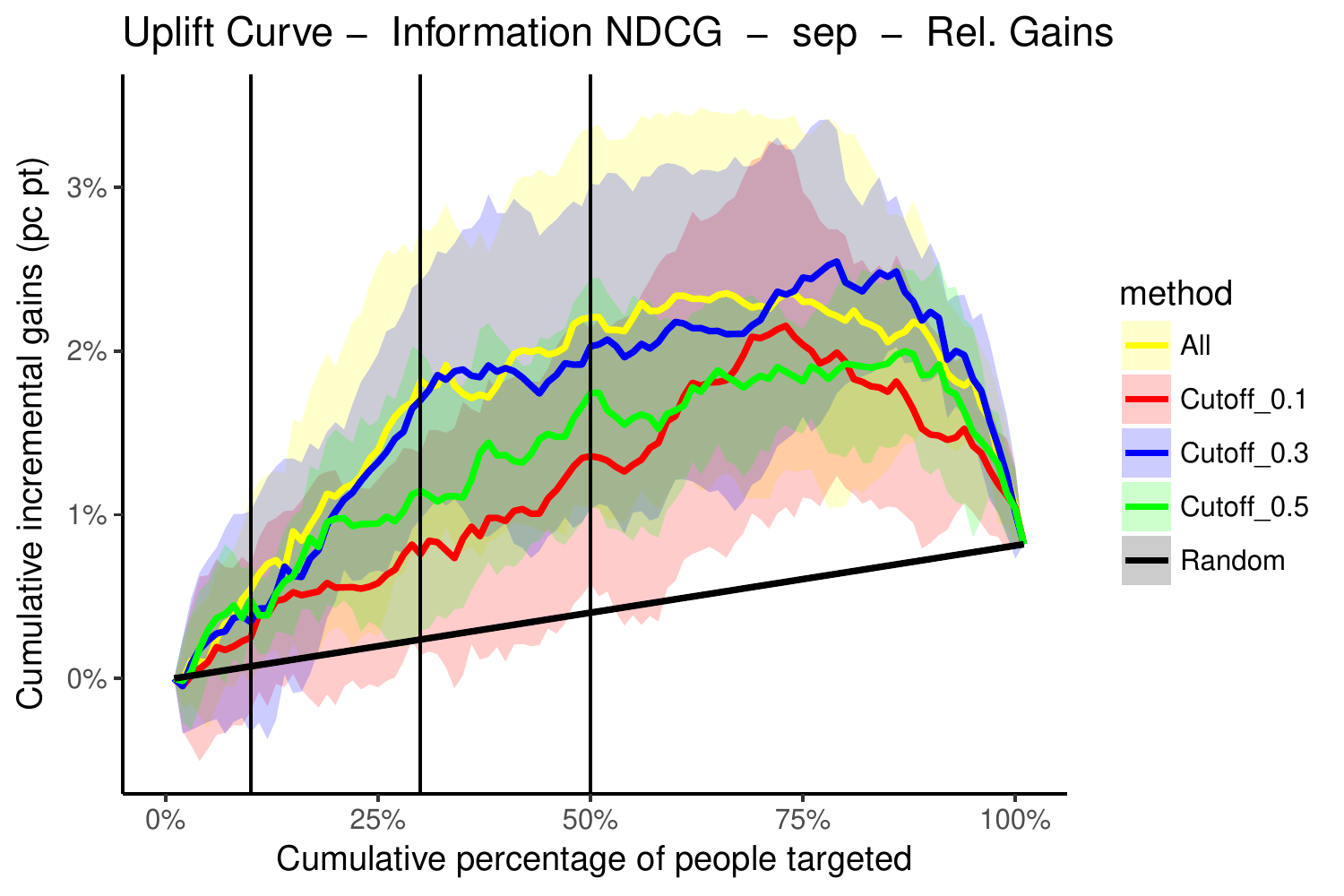}
  \caption{Uplift Curve - Information}
\end{subfigure} %
\begin{subfigure}{.32\textwidth}
  \centering
  \includegraphics[width=1\linewidth]{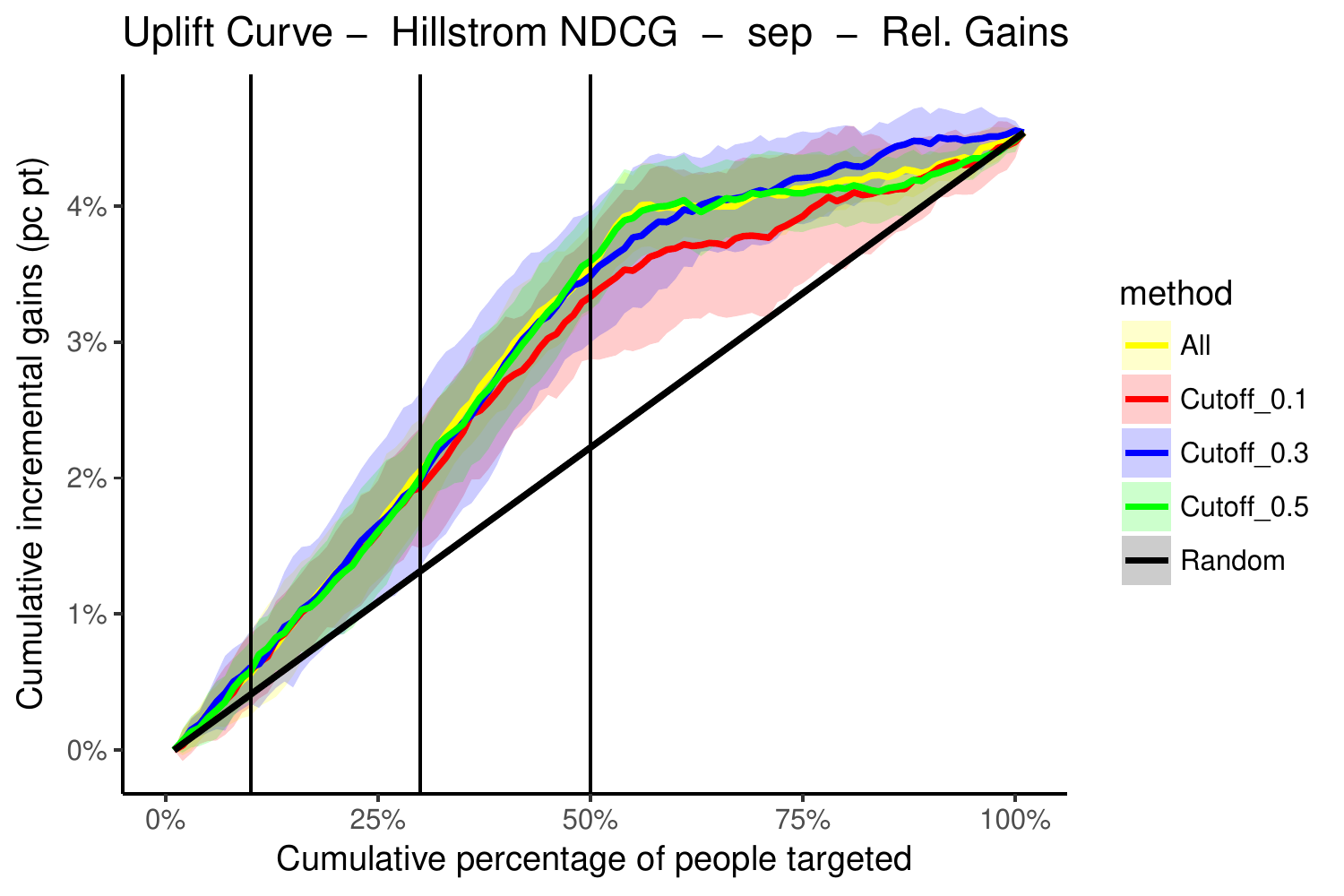}
  \caption{Uplift Curve - Hillstrom}
\end{subfigure} %
\begin{subfigure}{.32\textwidth}
  \centering
  \includegraphics[width=1\linewidth]{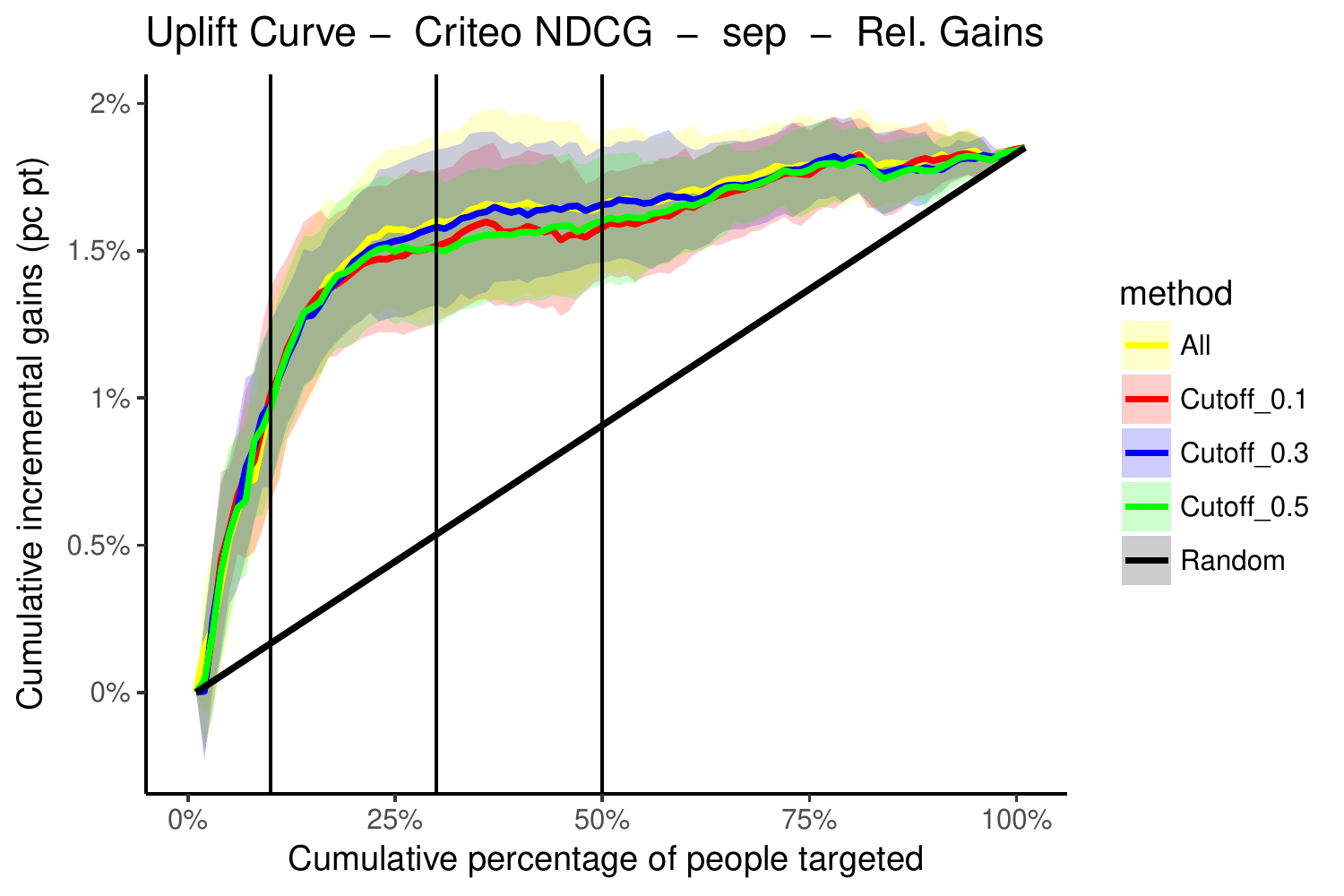}
  \caption{Uplift Curve - Criteo}
\end{subfigure}
\caption{Experiment 2 - NDCG for Separate Setting at multiple cutoffs with relative gains.}
\end{figure}

\begin{figure}[ht!]
\centering
\begin{subfigure}{.32\textwidth}
  \centering
  \includegraphics[width=1\linewidth]{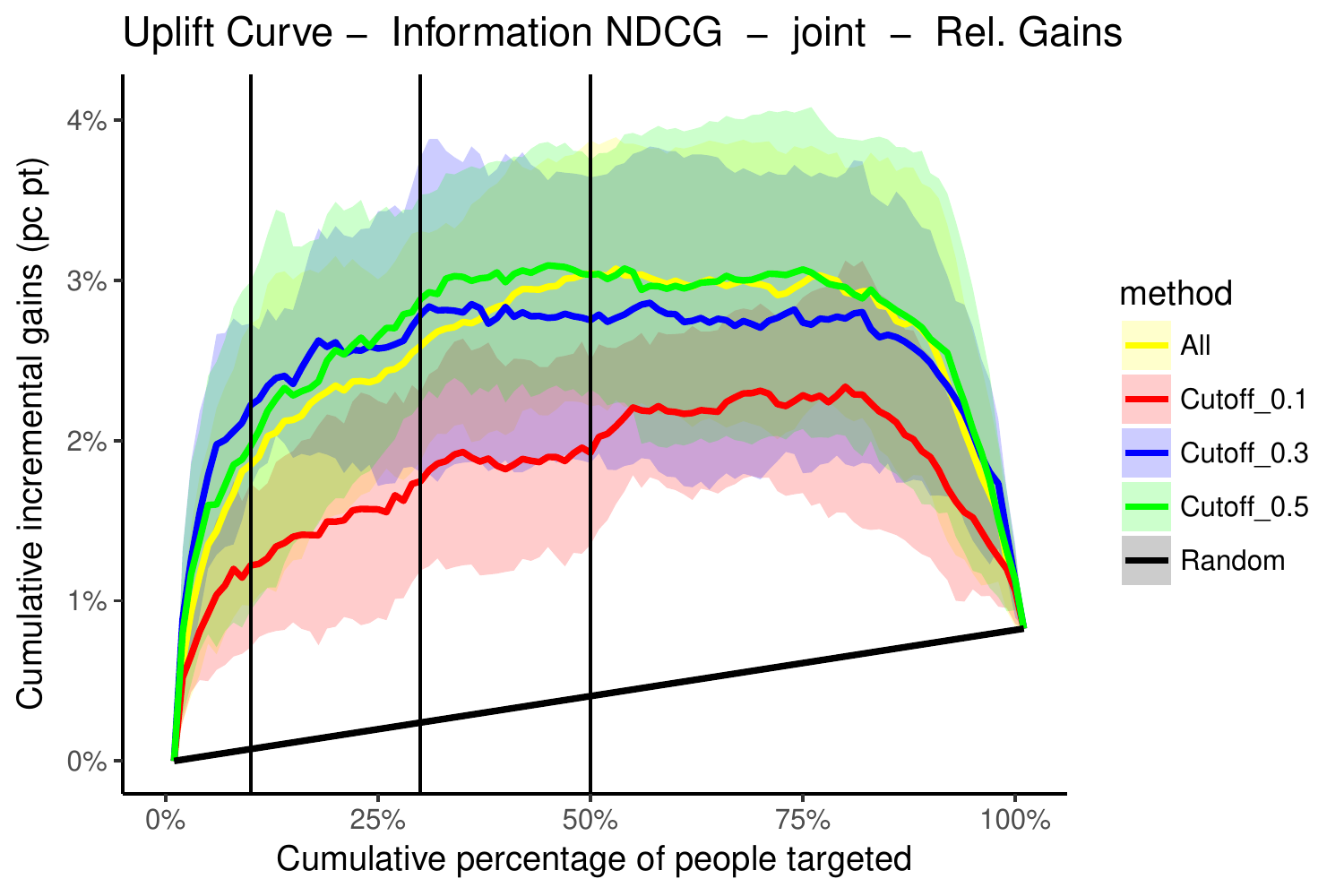}
  \caption{Uplift Curve - Information}
\end{subfigure} %
\begin{subfigure}{.32\textwidth}
  \centering
  \includegraphics[width=1\linewidth]{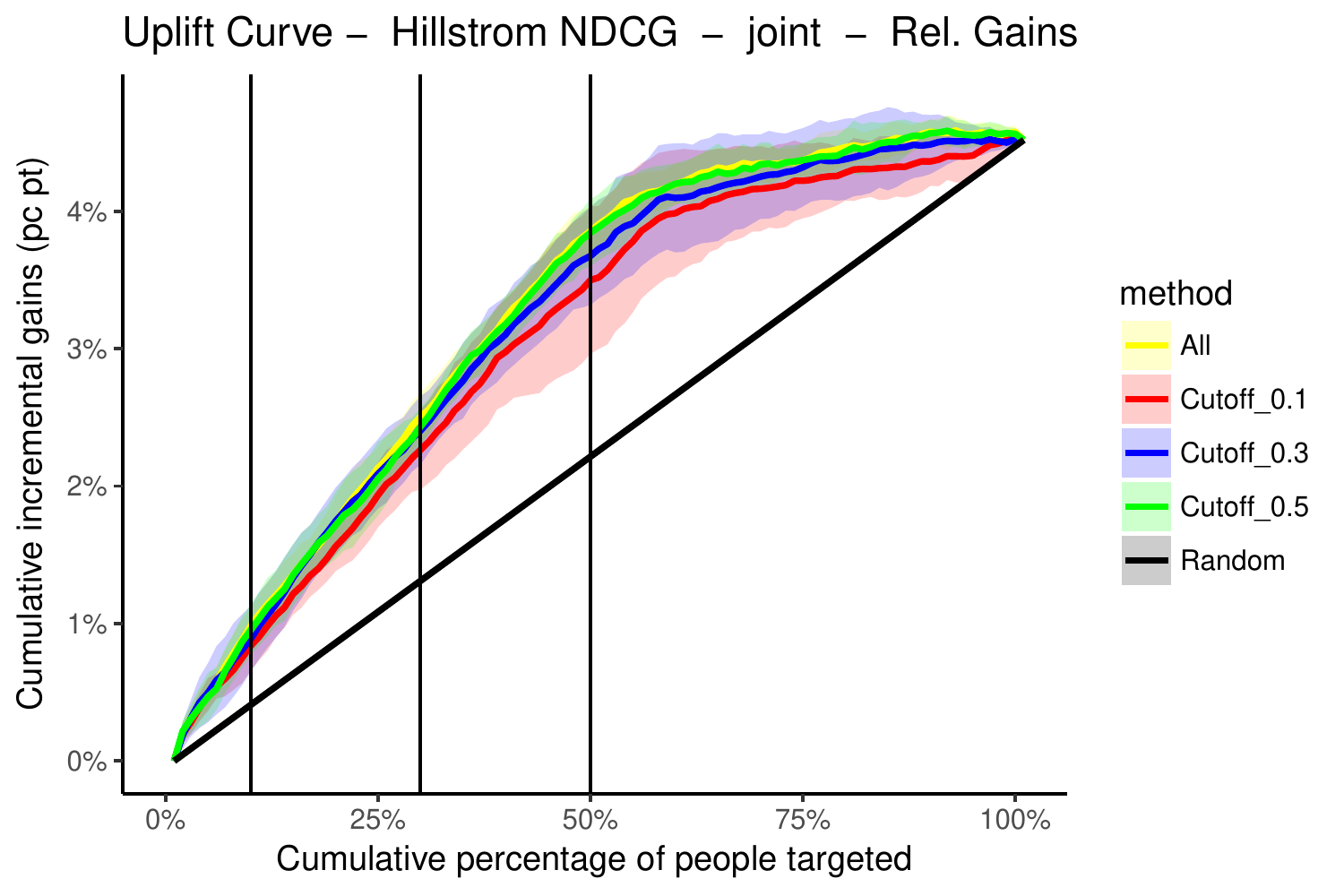}
  \caption{Uplift Curve - Hillstrom}
\end{subfigure} %
\begin{subfigure}{.32\textwidth}
  \centering
  \includegraphics[width=1\linewidth]{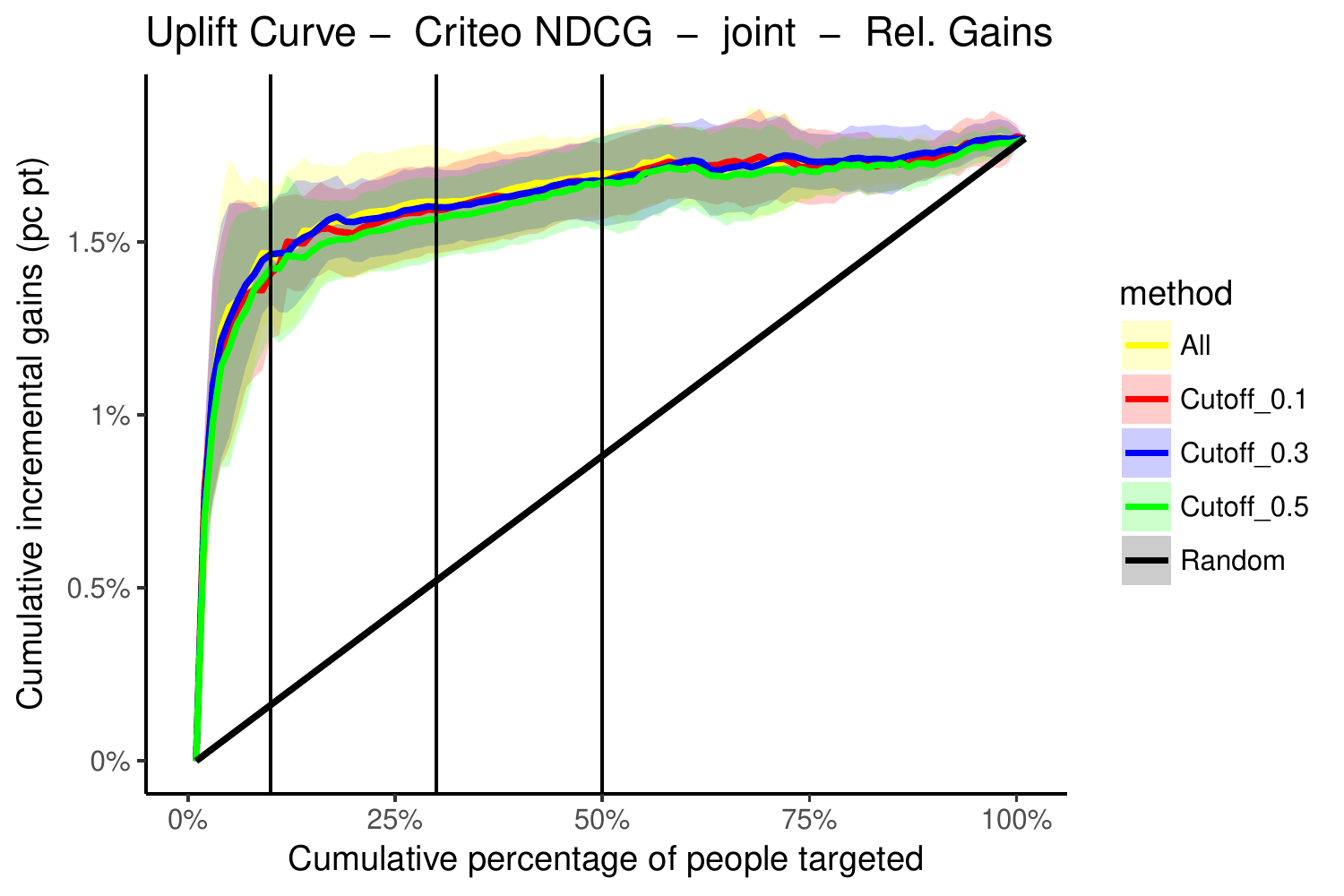}
  \caption{Uplift Curve - Criteo}
\end{subfigure}

\begin{subfigure}{.32\textwidth}
  \centering
  \includegraphics[width=1\linewidth]{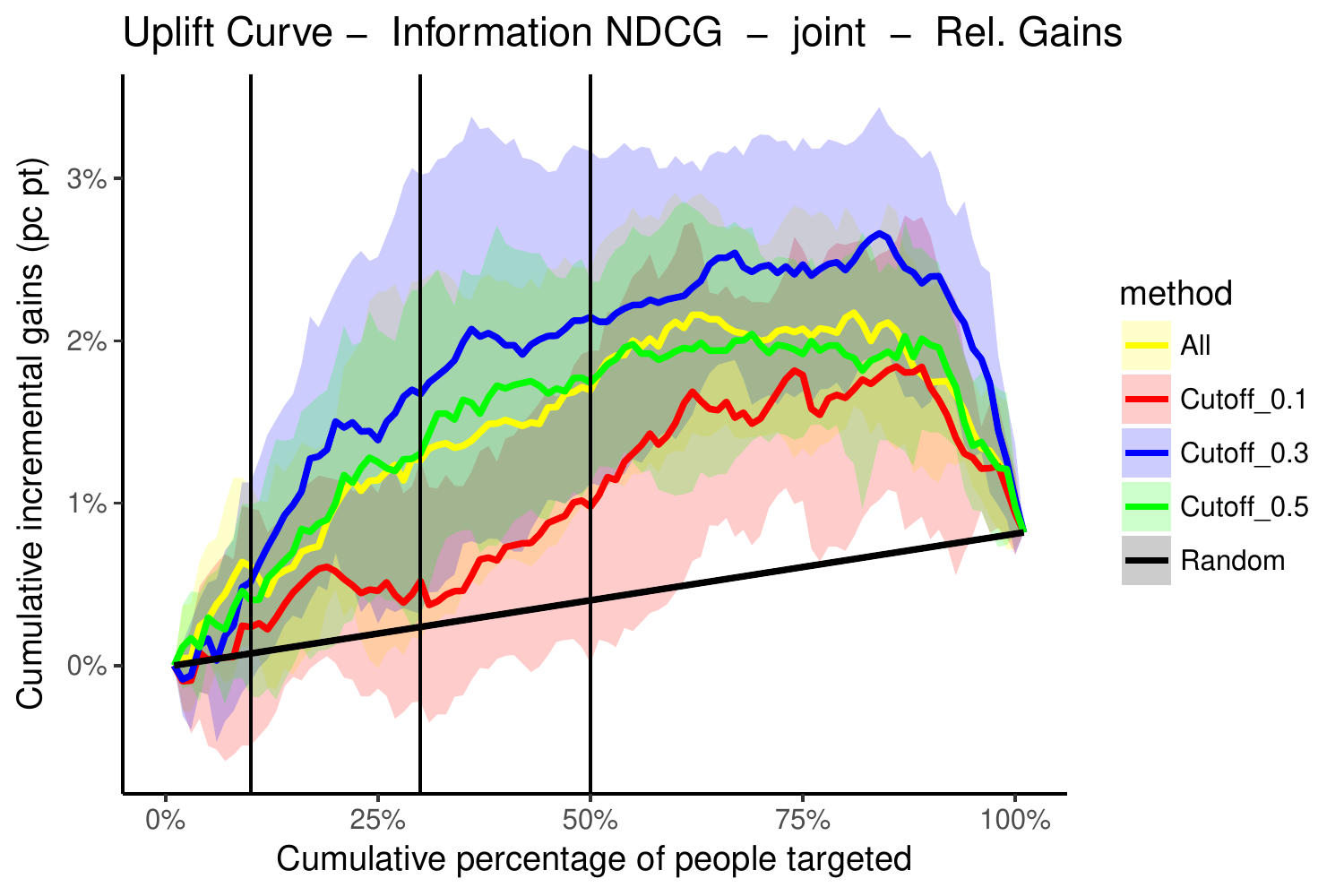}
  \caption{Uplift Curve - Information}
\end{subfigure} %
\begin{subfigure}{.32\textwidth}
  \centering
  \includegraphics[width=1\linewidth]{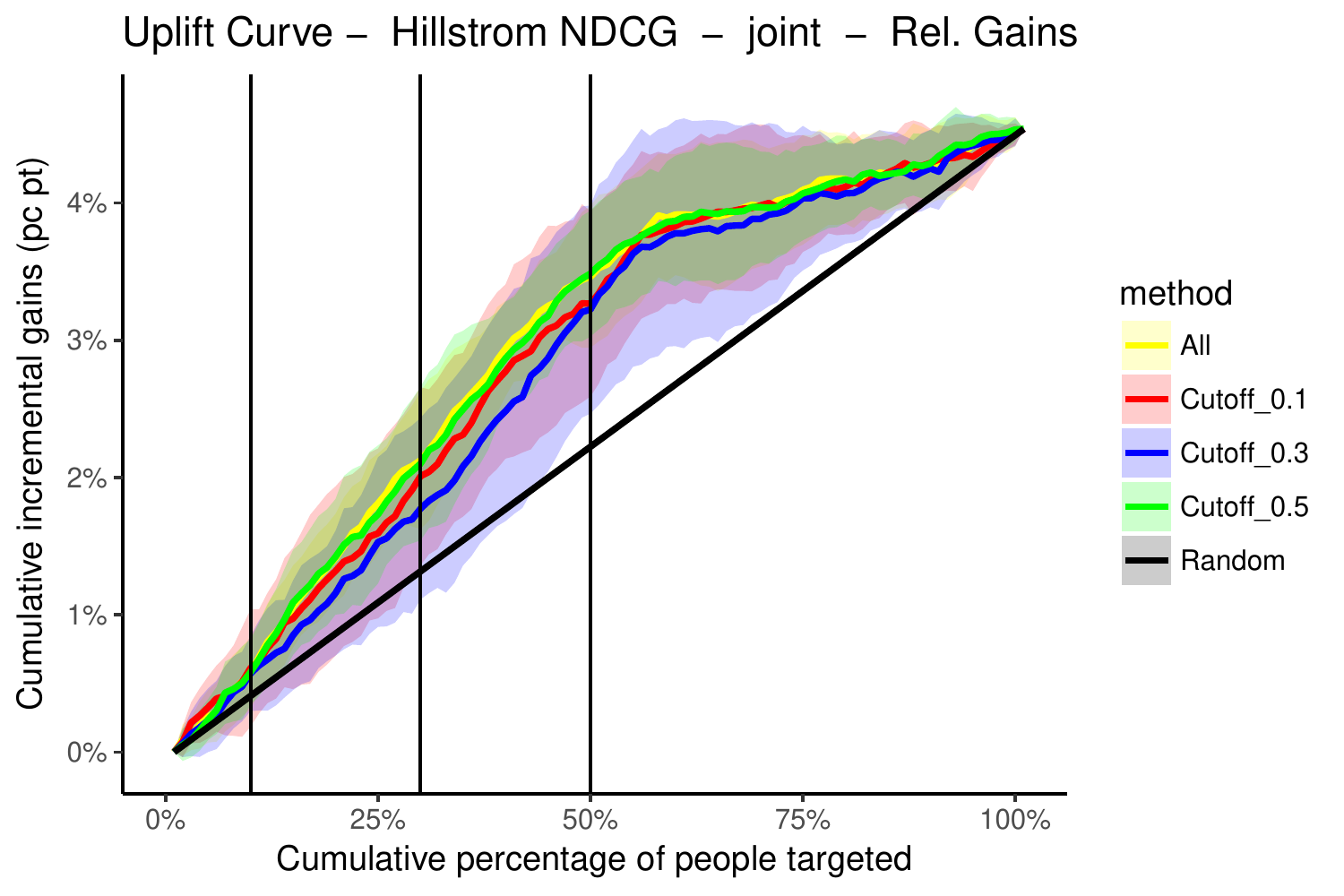}
  \caption{Uplift Curve - Hillstrom}
\end{subfigure} %
\begin{subfigure}{.32\textwidth}
  \centering
  \includegraphics[width=1\linewidth]{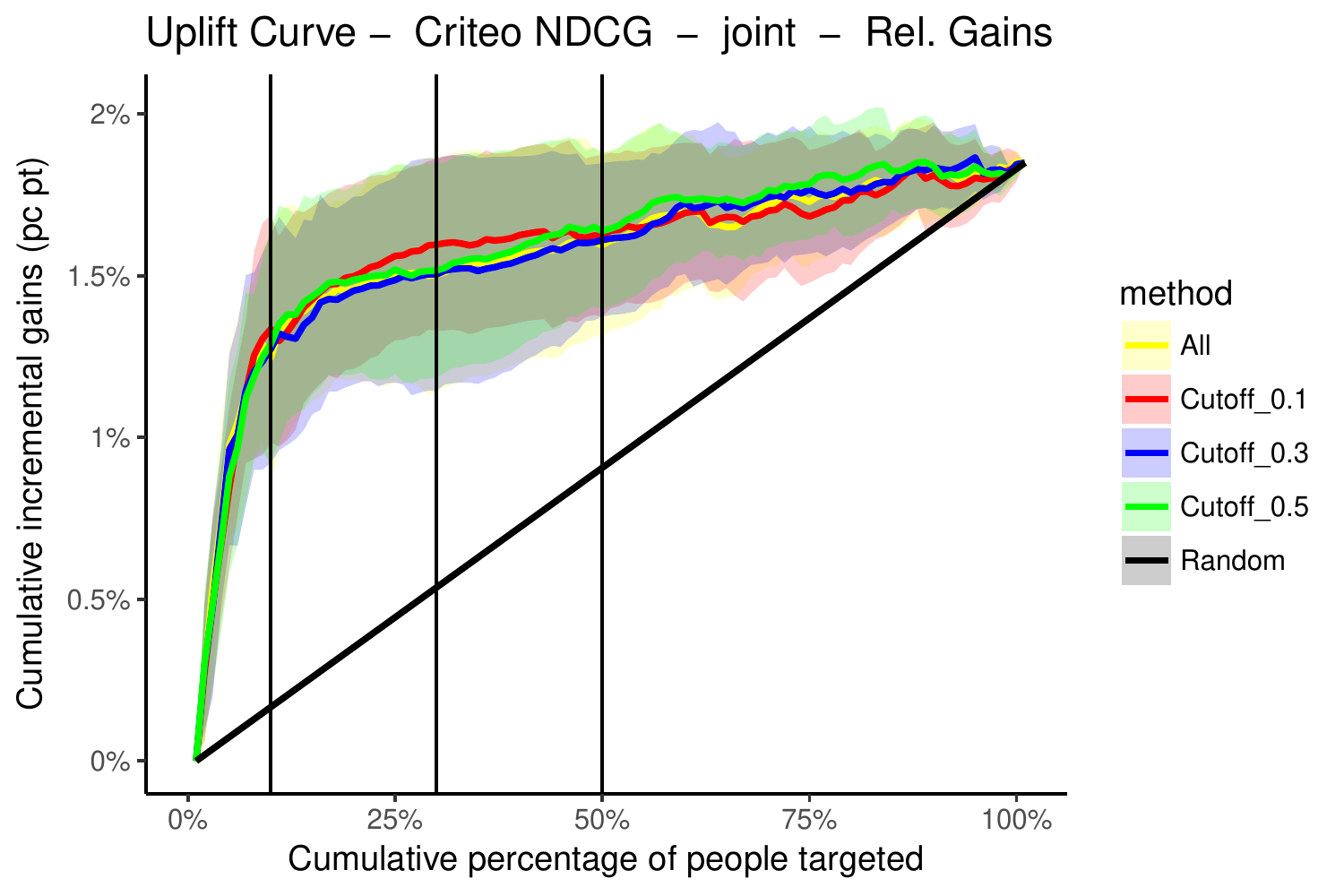}
  \caption{Uplift Curve - Criteo}
\end{subfigure}

\caption{Experiment 2 - NDCG for Joint Setting at multiple cutoffs with relative gains.}
\end{figure}

\begin{figure}[ht!]
\centering
\begin{subfigure}{.32\textwidth}
  \centering
  \includegraphics[width=1\linewidth]{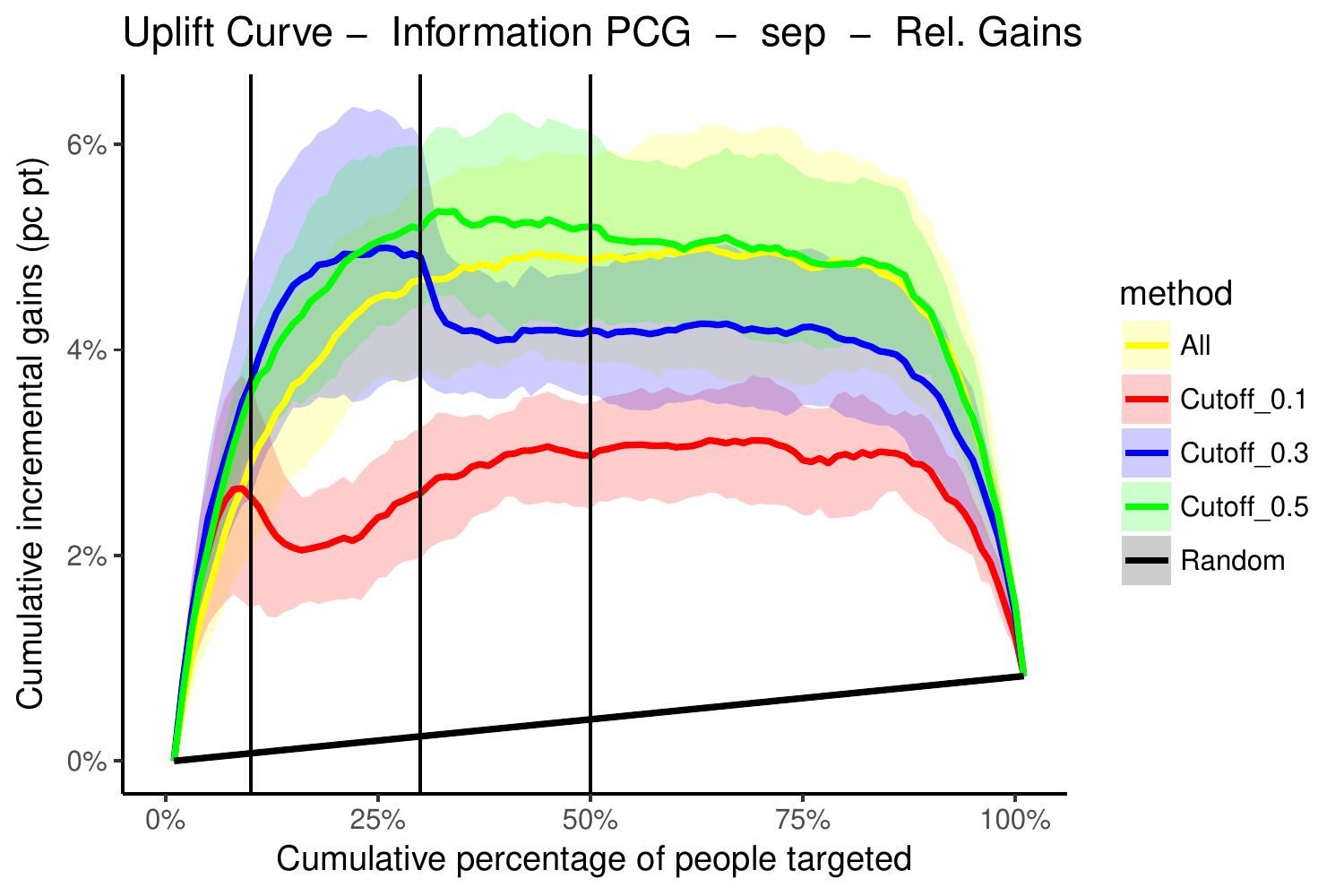}
  \caption{Uplift Curve - Information}
\end{subfigure} %
\begin{subfigure}{.32\textwidth}
  \centering
  \includegraphics[width=1\linewidth]{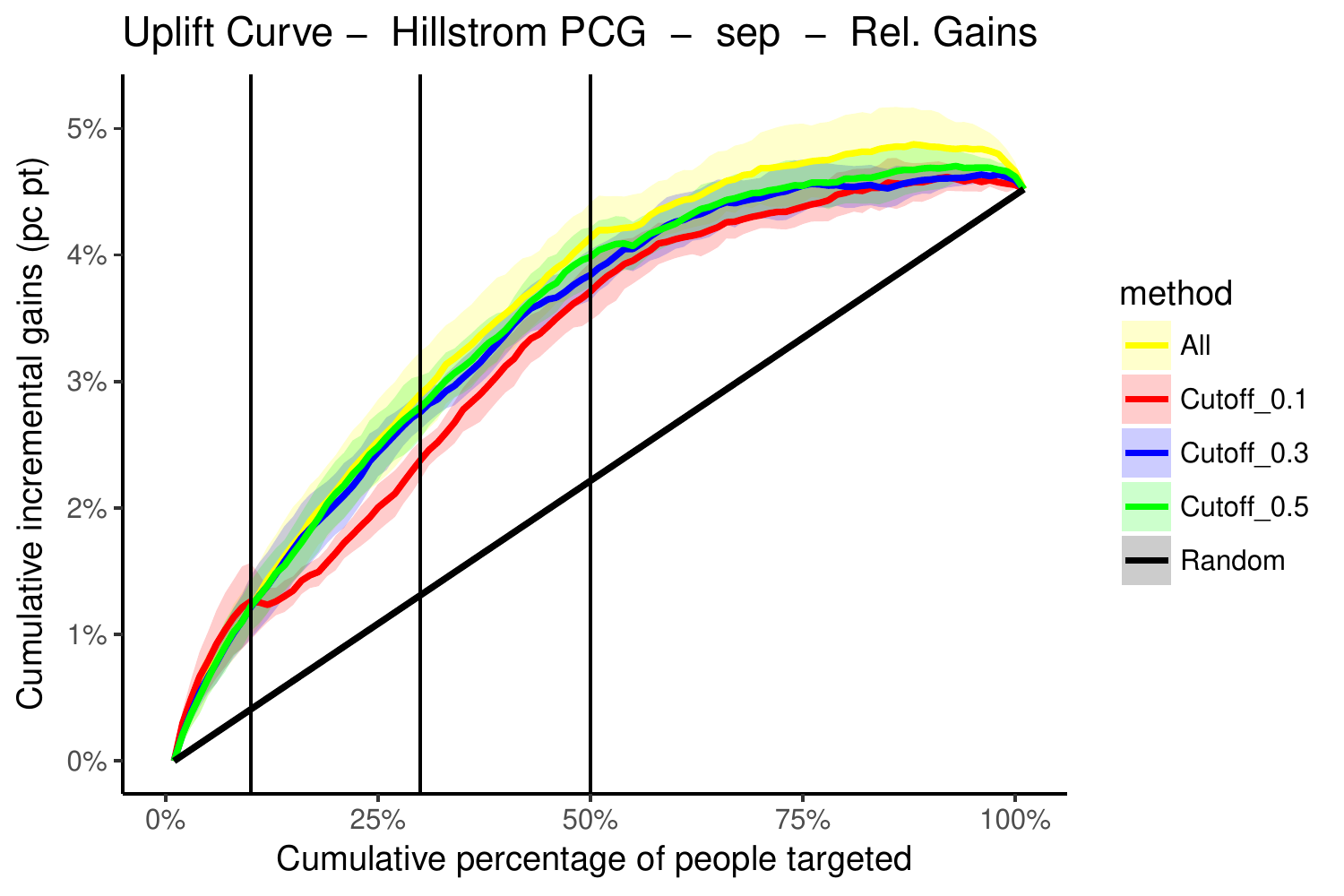}
  \caption{Uplift Curve - Hillstrom}
\end{subfigure} %
\begin{subfigure}{.32\textwidth}
  \centering
  \includegraphics[width=1\linewidth]{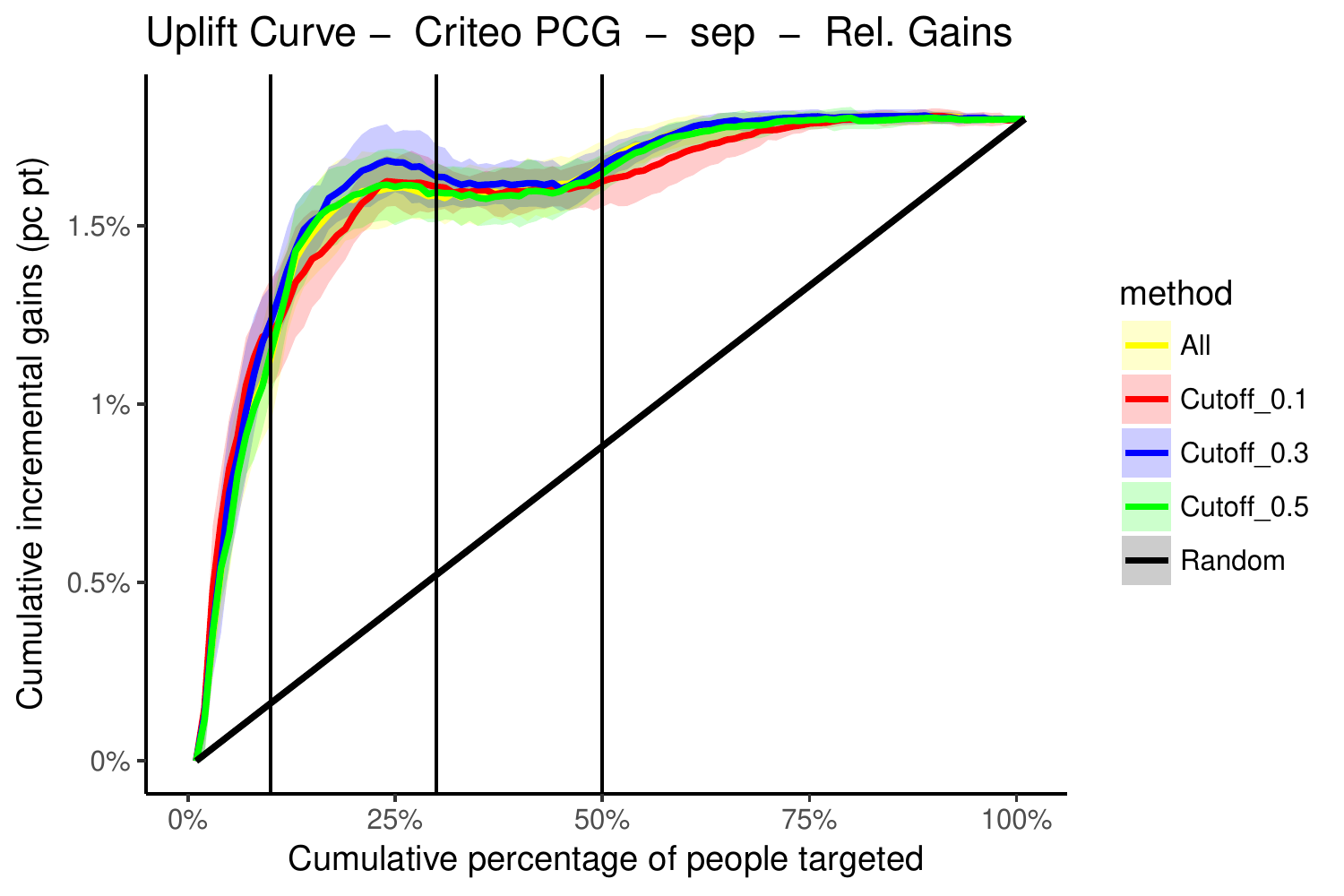}
  \caption{Uplift Curve - Criteo}
\end{subfigure}

\begin{subfigure}{.32\textwidth}
  \centering
  \includegraphics[width=1\linewidth]{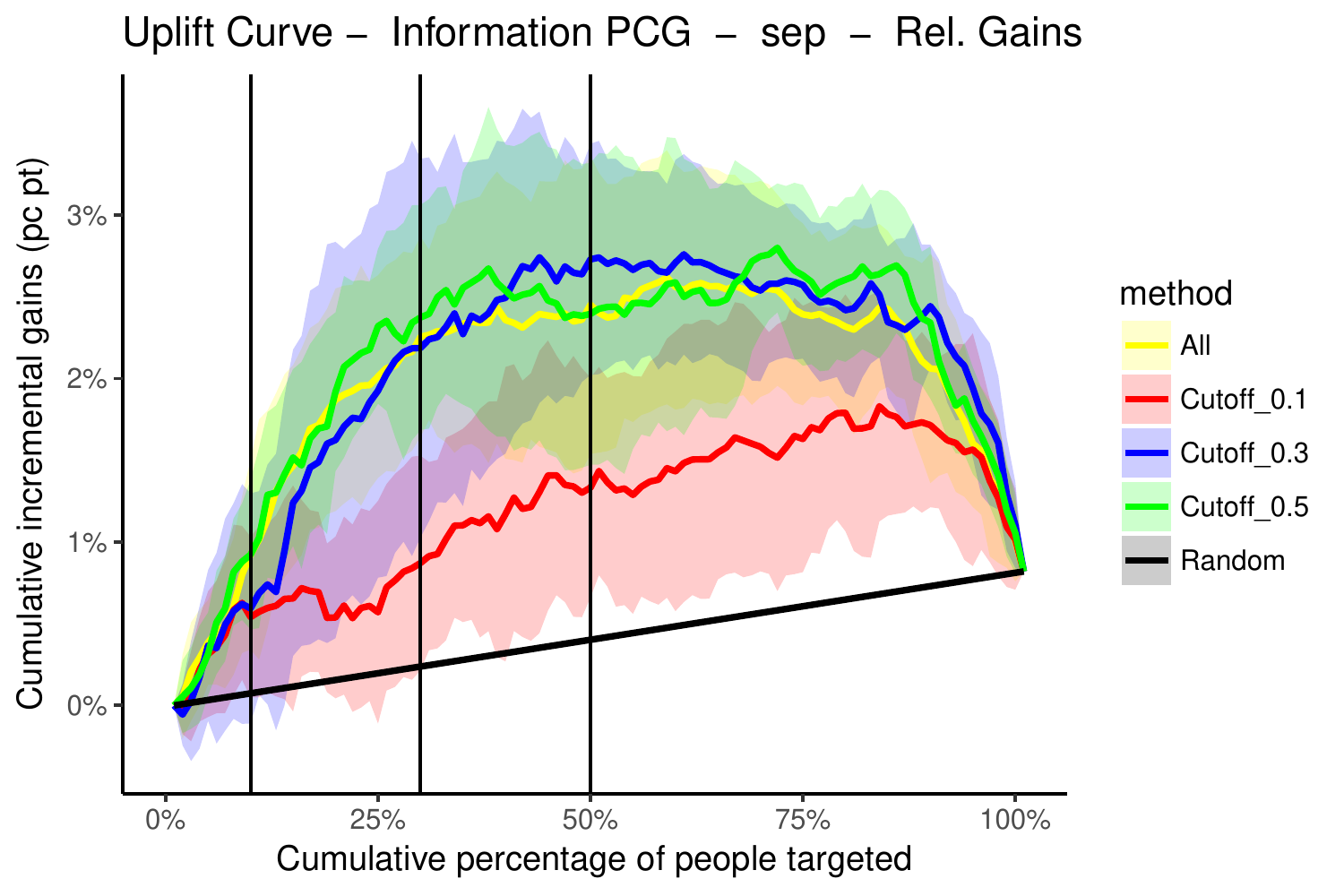}
  \caption{Uplift Curve - Information}
\end{subfigure} %
\begin{subfigure}{.32\textwidth}
  \centering
  \includegraphics[width=1\linewidth]{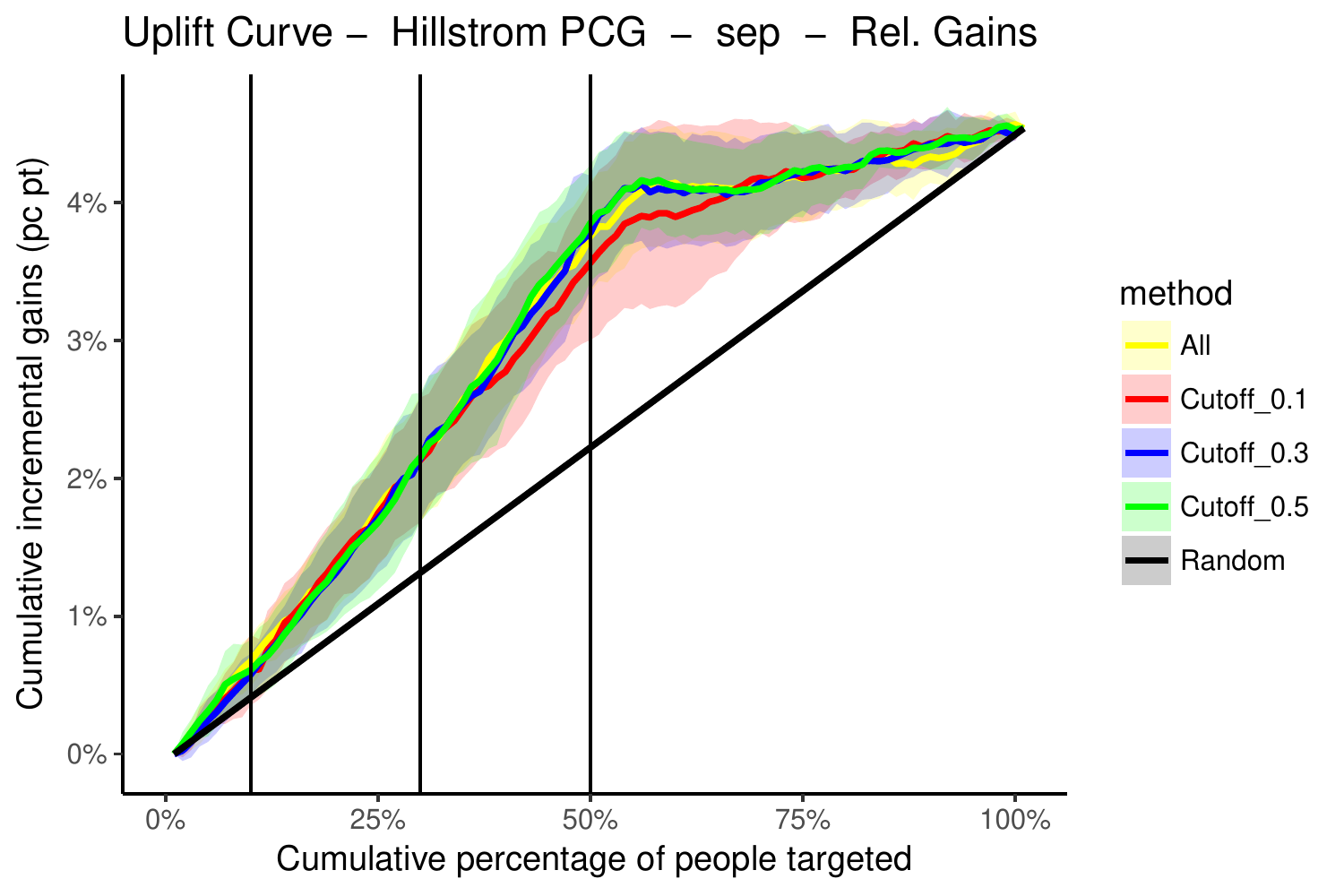}
  \caption{Uplift Curve - Hillstrom}
\end{subfigure} %
\begin{subfigure}{.32\textwidth}
  \centering
  \includegraphics[width=1\linewidth]{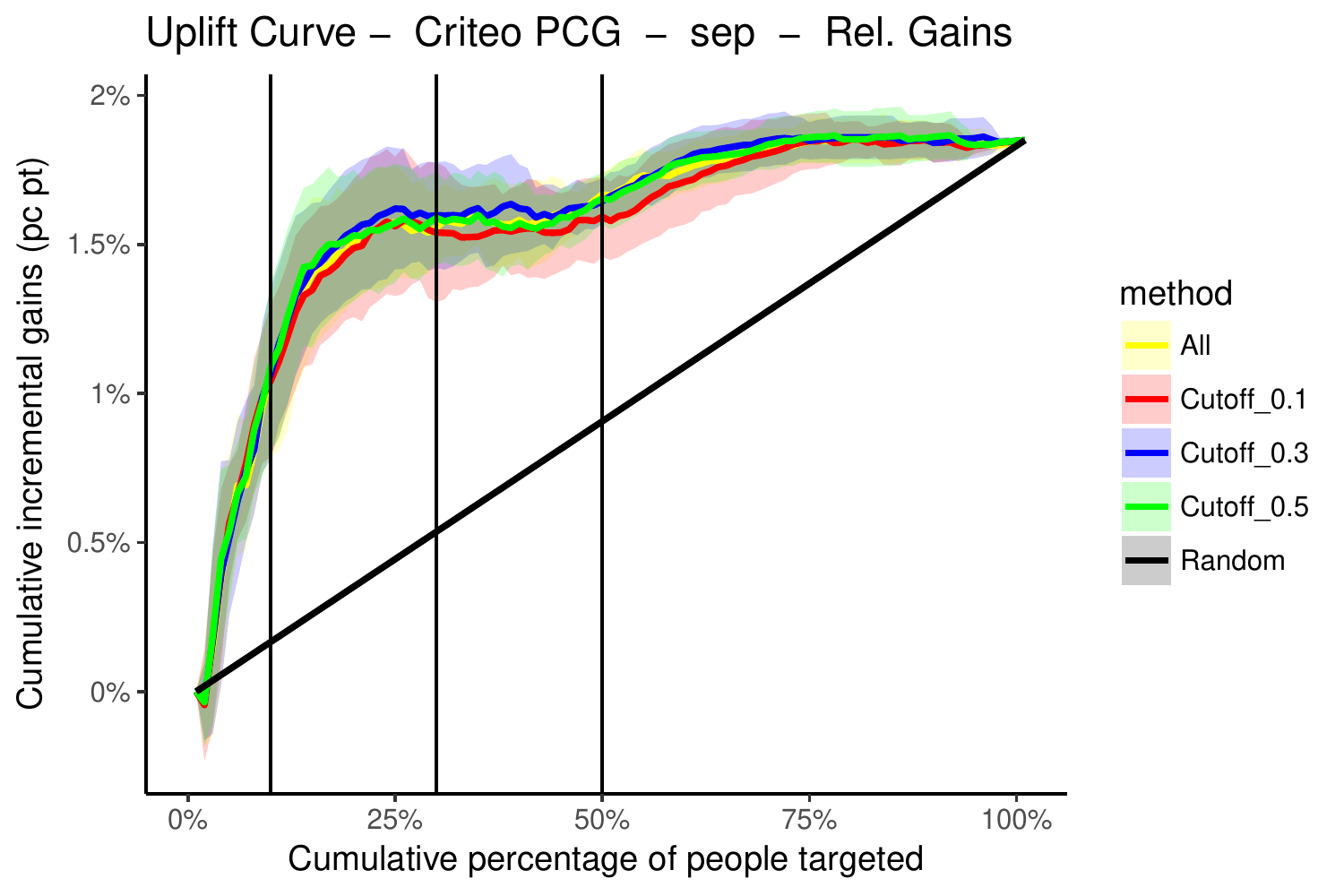}
  \caption{Uplift Curve - Criteo}
\end{subfigure}
\caption{Experiment 2 - PCG for Separate Setting at multiple cutoffs with relative gains.}
\end{figure}

\begin{figure}[ht!]
\centering
\begin{subfigure}{.32\textwidth}
  \centering
  \includegraphics[width=1\linewidth]{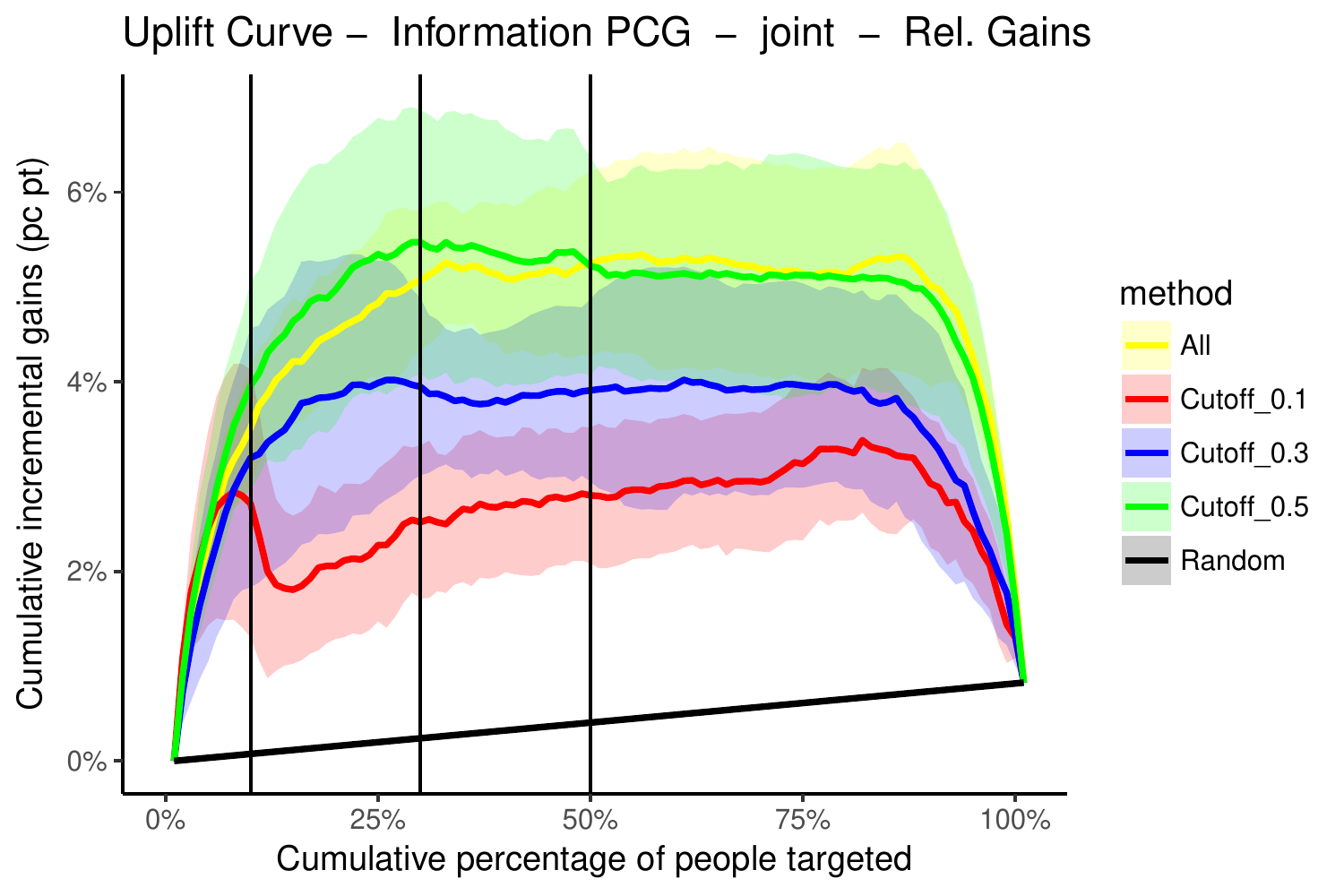}
  \caption{Uplift Curve - Information}
\end{subfigure} %
\begin{subfigure}{.32\textwidth}
  \centering
  \includegraphics[width=1\linewidth]{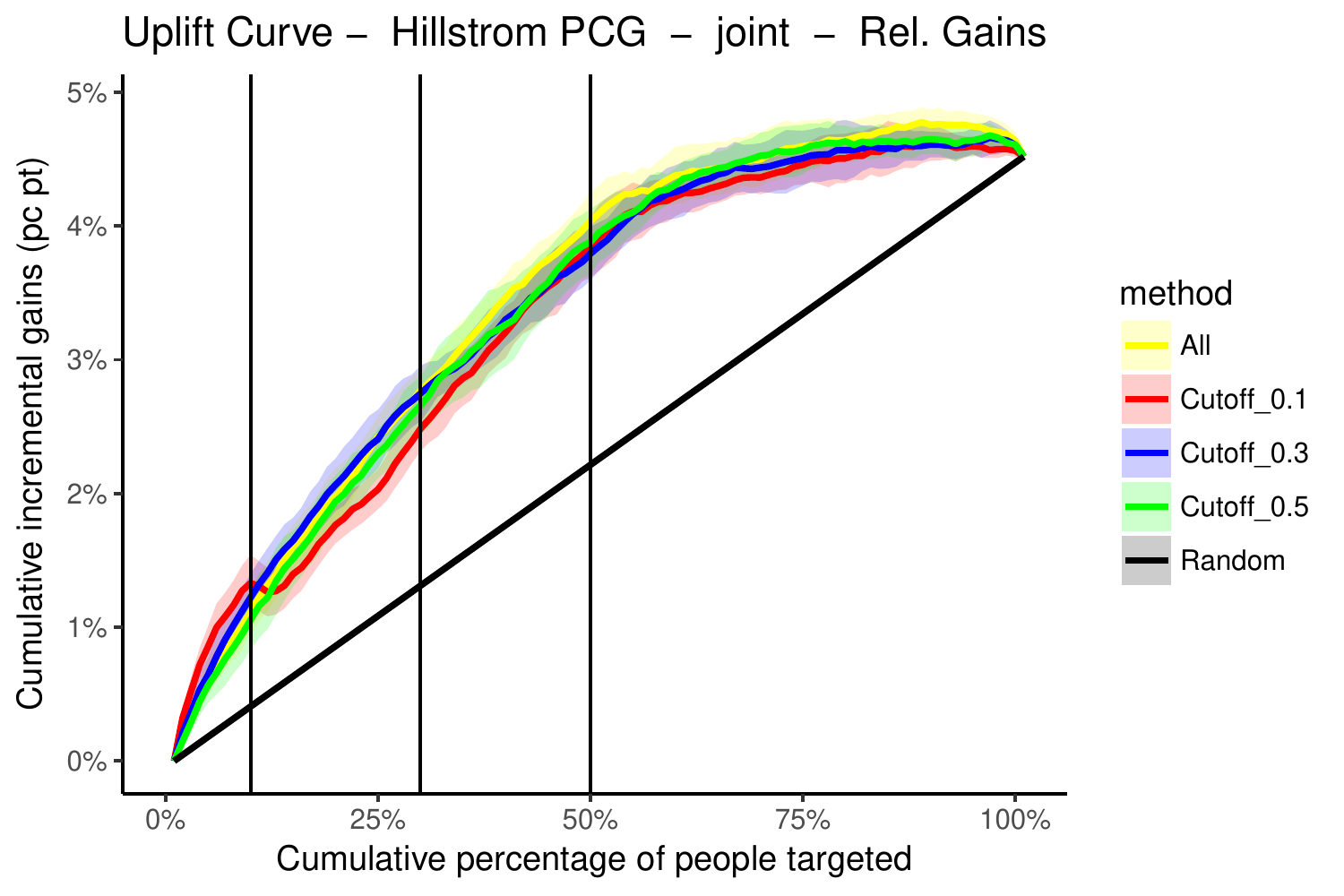}
  \caption{Uplift Curve - Hillstrom}
\end{subfigure} %
\begin{subfigure}{.32\textwidth}
  \centering
  \includegraphics[width=1\linewidth]{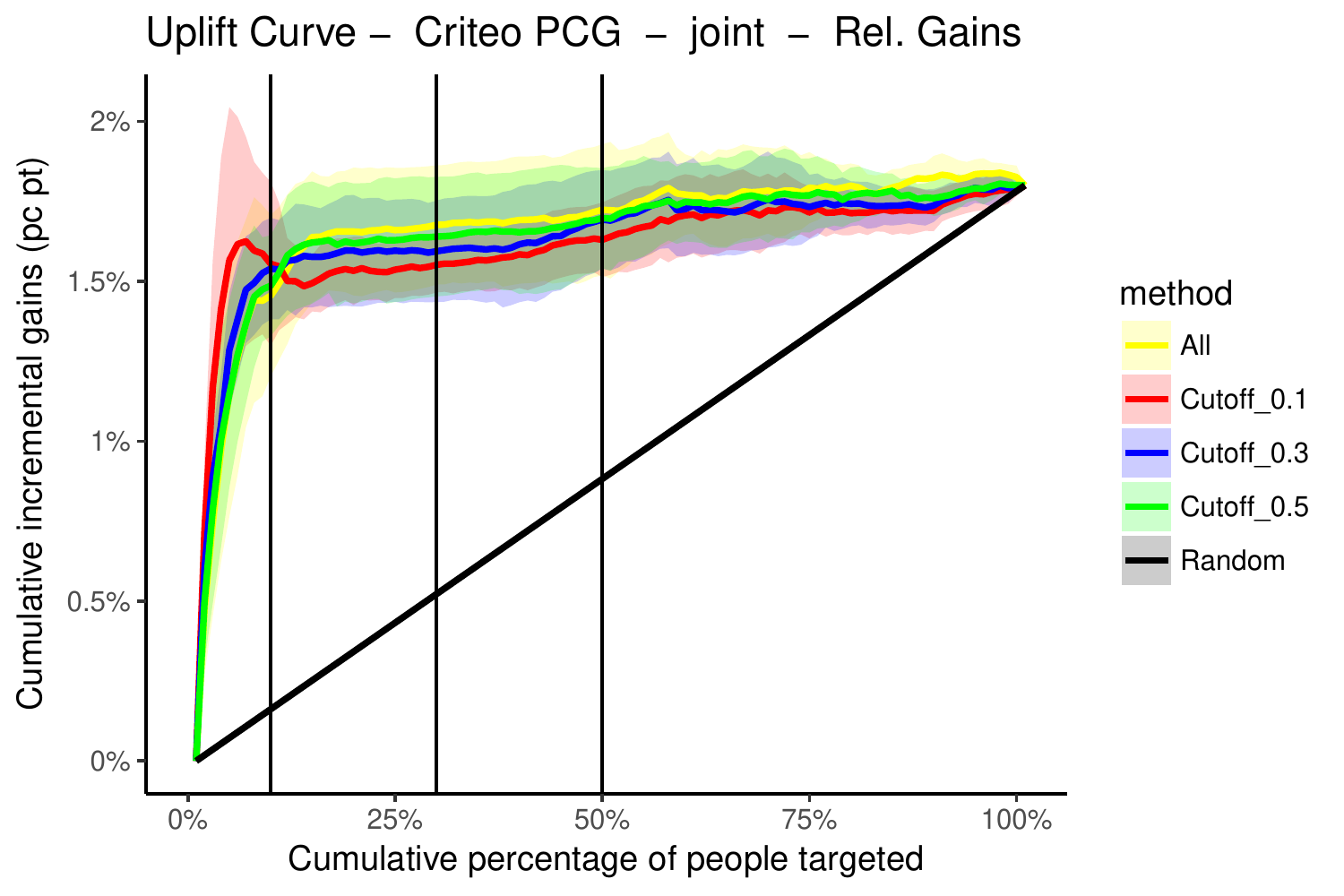}
  \caption{Uplift Curve - Criteo}
\end{subfigure}

\begin{subfigure}{.32\textwidth}
  \centering
  \includegraphics[width=1\linewidth]{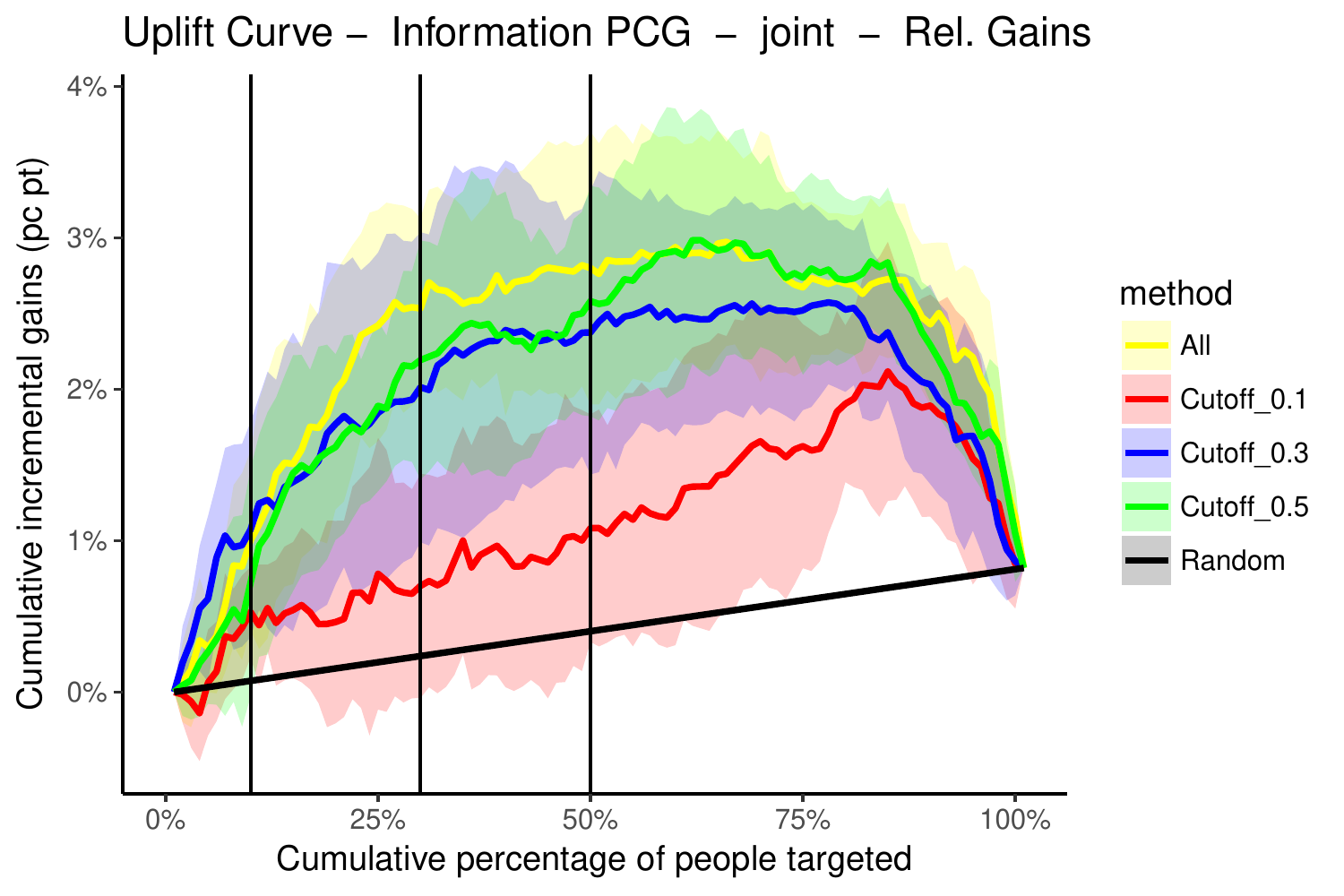}
  \caption{Uplift Curve - Information}
\end{subfigure} %
\begin{subfigure}{.32\textwidth}
  \centering
  \includegraphics[width=1\linewidth]{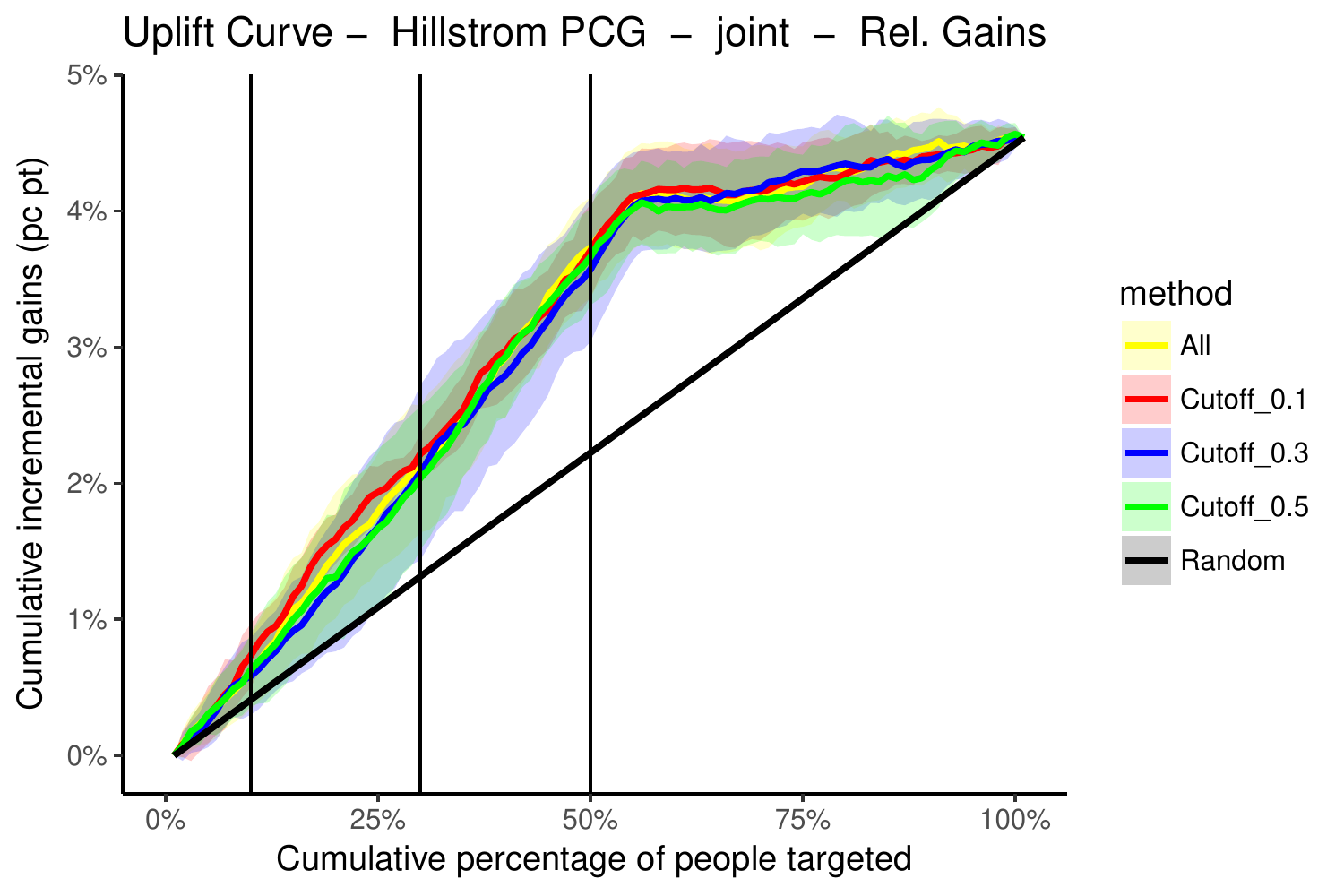}
  \caption{Uplift Curve - Hillstrom}
\end{subfigure} %
\begin{subfigure}{.32\textwidth}
  \centering
  \includegraphics[width=1\linewidth]{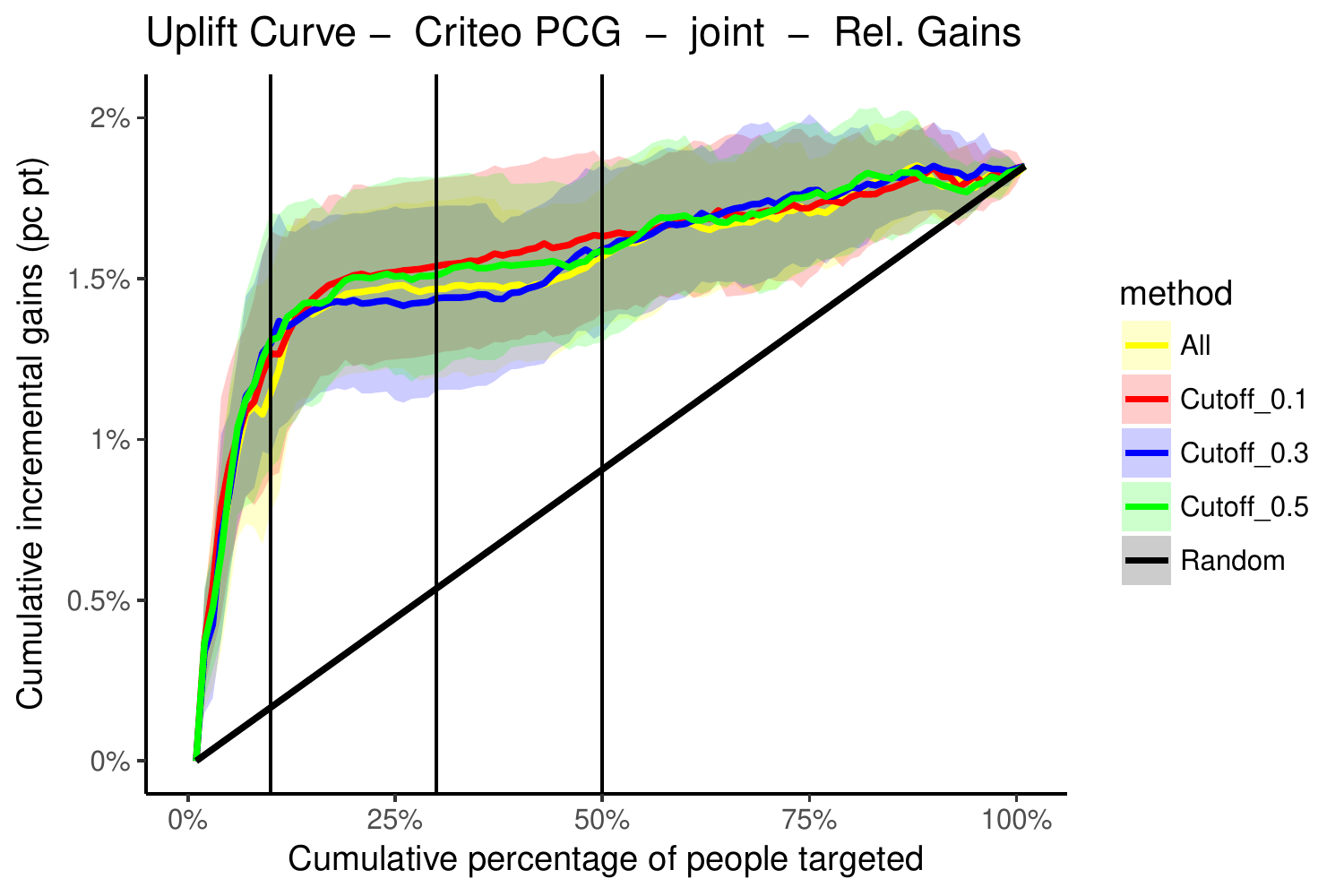}
  \caption{Uplift Curve - Criteo}
\end{subfigure}

\caption{Experiment 2 - PCG for Joint Setting at multiple cutoffs with relative gains.}
\end{figure}

\clearpage
\subsection*{Appendix E: Experiment 4 - Tables - Relative Relevance}
\begin{table*}[ht!]
\begin{center}
\caption{AUUC-values at multiple cut-offs for the Information dataset on the test set. On the left hand side the metric optimized at 10, 30, 50 and 100 percent is displayed. The AUUC values are checked at multiple cutoffs to compare the performance. Relative relevance values.}
\begin{adjustbox}{max width=\textwidth}
{\small
\begin{tabular}{|l|l|l|l|l|l|l|l|l|}
\hline
Information         & \multicolumn{4}{l|}{Relative Separate Uplift AUUC at cutoff} & \multicolumn{4}{l|}{Relative Joint Uplift AUUC at cut-off} \\ \hline
                     & 10\%    & 30\%   & 50\%   & 100\%   & 10\%   & 30\%   & 50\%  & 100\%  \\ \hline
DCG @ 10\% & 0.00011 & 0.00156 & 0.00373 & 0.01166 & 0.00004 & 0.0012 & 0.00366 & 0.01098 \\ \hline 
DCG @ 30\% & 0.00019 & 0.00203 & 0.00518 & 0.01345 & -0.00006 & 0.00086 & 0.00245 & 0.00833 \\ \hline 
DCG @ 50\% & -0.00002 & 0.00146 & 0.00418 & 0.01296 & 0.00014 & 0.00128 & 0.00338 & 0.0108 \\ \hline 
DCG @ 100\% & 0.00016 & 0.00205 & 0.0055 & 0.01507 & 0.00016 & 0.00135 & 0.00329 & 0.01052 \\ \hline 
NDCG @ 10\% & 0.00028 & 0.002 & 0.00305 & 0.00729 & 0.00006 & 0.00123 & 0.00368 & 0.01096 \\ \hline 
NDCG @ 30\% & 0.00043 & 0.00148 & 0.00204 & 0.00649 & -0.00002 & 0.00076 & 0.00263 & 0.00879 \\ \hline 
NDCG @ 50\% & 0.00037 & 0.00123 & 0.00197 & 0.00636 & 0.00014 & 0.00127 & 0.00339 & 0.01076 \\ \hline 
NDCG @ 100\% & 0.00045 & 0.00199 & 0.00355 & 0.00811 & 0.00014 & 0.00134 & 0.00314 & 0.01031 \\ \hline 
PCG @ 10\% & 0.00026 & 0.00159(*) & 0.00368(*) & 0.0115(*) & 0.0004 & 0.00232 & 0.00495 & 0.01264 \\ \hline 
PCG @ 30\% & 0.0003 & 0.00299 & 0.00731 & 0.01819 & 0.0003 & 0.00246 & 0.00611 & 0.01608 \\ \hline 
PCG @ 50\% & 0.00049 & 0.00419 & 0.00942 & 0.02143 & 0.00021 & 0.00257 & 0.00646 & 0.0169 \\ \hline 
PCG @ 100\% & 0.00037 & 0.00384 & 0.009 & 0.02178 & 0.00034 & 0.00271 & 0.00616 & 0.01573 \\ \hline 
\end{tabular}
}
\end{adjustbox}
\end{center}
\end{table*}

\begin{table*}[ht!]
\begin{center}
\caption{AUUC-values at multiple cut-offs for the Hillstrom dataset on the test set. On the left hand side the metric optimized at 10, 30, 50 and 100 percent is displayed. The AUUC values are checked at multiple cutoffs to compare the performance. Relative relevance values.}
\begin{adjustbox}{max width=\textwidth}
{\small
\begin{tabular}{|l|l|l|l|l|l|l|l|l|}
\hline
Hillstrom         & \multicolumn{4}{l|}{Relative Separate Uplift AUUC at cutoff} & \multicolumn{4}{l|}{Relative Joint Uplift AUUC at cut-off} \\ \hline
                     & 10\%    & 30\%   & 50\%   & 100\%   & 10\%   & 30\%   & 50\%  & 100\%  \\ \hline
DCG @ 10\% & 0.00033 & 0.0031 & 0.00838 & 0.02948 & 0.00029 & 0.00281 & 0.00808 & 0.02832 \\ \hline 
DCG @ 30\% & 0.00031 & 0.00308 & 0.00899 & 0.02981 & 0.00033 & 0.00288 & 0.00857 & 0.02935 \\ \hline 
DCG @ 50\% & 0.0003 & 0.00322 & 0.0091 & 0.03021 & 0.00029 & 0.00285 & 0.00846 & 0.02922 \\ \hline 
DCG @ 100\% & 0.00034 & 0.00348 & 0.00944 & 0.03032 & 0.00037 & 0.00316 & 0.00865 & 0.02954 \\ \hline 
NDCG @ 10\% & 0.00035 & 0.00284 & 0.00781 & 0.02785 & 0.00029 & 0.00281 & 0.00808 & 0.02832 \\ \hline 
NDCG @ 30\% & 0.00039 & 0.00318 & 0.00858 & 0.02827 & 0.00033 & 0.00288 & 0.00857 & 0.02935 \\ \hline 
NDCG @ 50\% & 0.00035 & 0.00323 & 0.00877 & 0.02906 & 0.00029 & 0.00285 & 0.00846 & 0.02922 \\ \hline 
NDCG @ 100\% & 0.00035 & 0.0033 & 0.00885 & 0.02874 & 0.00037 & 0.00316 & 0.00864 & 0.02954 \\ \hline 
PCG @ 10\% & 0.00032 & 0.00318 & 0.00901 & 0.0298 & 0.00029 & 0.00296 & 0.00853 & 0.0292 \\ \hline 
PCG @ 30\% & 0.00037 & 0.00323 & 0.00934 & 0.03045 & 0.00035 & 0.00313 & 0.00887 & 0.03001 \\ \hline 
PCG @ 50\% & 0.00038 & 0.00321 & 0.00935 & 0.03043 & 0.00043 & 0.00347 & 0.0095 & 0.03073 \\ \hline 
PCG @ 100\% & 0.00037 & 0.00332 & 0.00947 & 0.0306 & 0.00037 & 0.00319 & 0.00905 & 0.03027 \\ \hline  
\end{tabular}
}
\end{adjustbox}
\end{center}
\end{table*}

\begin{table*}[ht!]
\begin{center}
\caption{AUUC-values at multiple cut-offs for the Criteo dataset on the test set. On the left hand side the metric optimized at 10, 30, 50 and 100 percent is displayed. The AUUC values are checked at multiple cutoffs to compare the performance. Relative relevance values.}
\begin{adjustbox}{max width=\textwidth}
{\small
\begin{tabular}{|l|l|l|l|l|l|l|l|l|}
\hline
Criteo         & \multicolumn{4}{l|}{Relative Separate Uplift AUUC at cutoff} & \multicolumn{4}{l|}{Relative Joint Uplift AUUC at cut-off} \\ \hline
                     & 10\%    & 30\%   & 50\%   & 100\%   & 10\%   & 30\%   & 50\%  & 100\%  \\ \hline
DCG @ 10\% & 0.00061 & 0.00363 & 0.00699 & 0.01598 & 0.00095 & 0.00401 & 0.00724 & 0.01598 \\ \hline 
DCG @ 30\% & 0.00062 & 0.0035 & 0.00679 & 0.01565 & 0.0009 & 0.00385 & 0.00696 & 0.01566 \\ \hline 
DCG @ 50\% & 0.00057 & 0.00345 & 0.00668 & 0.01544 & 0.00089 & 0.00377 & 0.00684 & 0.01543 \\ \hline 
DCG @ 100\% & 0.00059 & 0.00337 & 0.00649 & 0.01514 & 0.00089 & 0.0037 & 0.00674 & 0.01541 \\ \hline 
NDCG @ 10\% & 0.00058 & 0.00341 & 0.00654 & 0.01525 & 0.00095 & 0.00401 & 0.00725 & 0.01604 \\ \hline 
NDCG @ 30\% & 0.00062 & 0.00352 & 0.0067 & 0.01549 & 0.0009 & 0.00384 & 0.00693 & 0.0156 \\ \hline 
NDCG @ 50\% & 0.0006 & 0.00341 & 0.00651 & 0.01519 & 0.00089 & 0.00378 & 0.00686 & 0.01547 \\ \hline 
NDCG @ 100\% & 0.00056 & 0.00342 & 0.00648 & 0.01527 & 0.00089 & 0.0037 & 0.00675 & 0.01543 \\ \hline 
PCG @ 10\% & 0.00063 & 0.00353 & 0.00681 & 0.01577 & 0.00087 & 0.00392 & 0.00725 & 0.01599 \\ \hline 
PCG @ 30\% & 0.00062 & 0.00368 & 0.00706 & 0.01608 & 0.00092 & 0.00383 & 0.00689 & 0.01545 \\ \hline 
PCG @ 50\% & 0.00058 & 0.00353 & 0.00676 & 0.0158 & 0.00088 & 0.00377 & 0.00684 & 0.01533 \\ \hline 
PCG @ 100\% & 0.00059 & 0.00354 & 0.00671 & 0.01575 & 0.00088 & 0.0038 & 0.00691 & 0.01554 \\ \hline 
\end{tabular}
}
\end{adjustbox}
\end{center}
\end{table*}

\clearpage
\subsection*{Appendix F: Experiment 4 - Tables - Absolute Relevance}
\begin{table*}[ht!]
\begin{center}
\caption{AUUC-values at multiple cut-offs for the Information dataset on the test set. On the left hand side the metric optimized at 10, 30, 50 and 100 percent is displayed. The AUUC values are checked at multiple cutoffs to compare the performance. Absolute relevance values.}
\begin{adjustbox}{max width=\textwidth}
{\small
\begin{tabular}{|l|l|l|l|l|l|l|l|l|}
\hline
Information         & \multicolumn{4}{l|}{Relative Separate Uplift AUUC at cutoff} & \multicolumn{4}{l|}{Relative Joint Uplift AUUC at cut-off} \\ \hline
                     & 10\%    & 30\%   & 50\%   & 100\%   & 10\%   & 30\%   & 50\%  & 100\%  \\ \hline
DCG @ 10\% & 0.00011 & 0.00136 & 0.00327 & 0.01087 & 0.00007 & 0.001 & 0.00245(*) & 0.00995(*) \\ \hline 
DCG @ 30\% & 0.00011 & 0.00203 & 0.00468 & 0.01263 & 0.00019 & 0.00285 & 0.00679 & 0.01806 \\ \hline 
DCG @ 50\% & 0.00002 & 0.00144 & 0.00446 & 0.01285 & 0.00037 & 0.00242 & 0.00572 & 0.01463 \\ \hline 
DCG @ 100\% & 0.00014 & 0.00195 & 0.0051 & 0.01396 & 0.00037 & 0.00235 & 0.00567 & 0.01563 \\ \hline 
NDCG @ 10\% & 0.00014 & 0.0013 & 0.00335(*) & 0.01168 & 0.00007 & 0.00099 & 0.00243(*) & 0.01004(*) \\ \hline 
NDCG @ 30\% & 0.00023 & 0.00235 & 0.00609 & 0.0167 & 0.00019 & 0.00283 & 0.00683 & 0.0182 \\ \hline 
NDCG @ 50\% & 0.00028 & 0.00199 & 0.00474 & 0.01338 & 0.00026 & 0.00227 & 0.00561 & 0.01476 \\ \hline 
NDCG @ 100\% & 0.00026 & 0.00264 & 0.0065 & 0.017 & 0.00035 & 0.00223 & 0.00525 & 0.01473 \\ \hline 
PCG @ 10\% & 0.00034 & 0.00168(*) & 0.00409(*) & 0.01181(*) & 0.00019(*) & 0.00136(*) & 0.00317(*) & 0.01086(*) \\ \hline 
PCG @ 30\% & 0.00035 & 0.00357 & 0.00862 & 0.02064 & 0.00072 & 0.00406 & 0.00867 & 0.01986 \\ \hline 
PCG @ 50\% & 0.00049 & 0.0043 & 0.0093 & 0.02111 & 0.00036 & 0.00372 & 0.00848 & 0.02123 \\ \hline 
PCG @ 100\% & 0.00047 & 0.00408 & 0.00878 & 0.02003 & 0.0005 & 0.00456 & 0.00998 & 0.023 \\ \hline 
\end{tabular}
}
\end{adjustbox}
\end{center}
\end{table*}

\begin{table*}[ht!]
\begin{center}
\caption{AUUC-values at multiple cut-offs for the Hillstrom dataset on the test set. On the left hand side the metric optimized at 10, 30, 50 and 100 percent is displayed. The AUUC values are checked at multiple cutoffs to compare the performance. Absolute relevance values.}
\begin{adjustbox}{max width=\textwidth}
{\small
\begin{tabular}{|l|l|l|l|l|l|l|l|l|}
\hline
Hillstrom         & \multicolumn{4}{l|}{Relative Separate Uplift AUUC at cutoff} & \multicolumn{4}{l|}{Relative Joint Uplift AUUC at cut-off} \\ \hline
                     & 10\%    & 30\%   & 50\%   & 100\%   & 10\%   & 30\%   & 50\%  & 100\%  \\ \hline
DCG @ 10\% & 0.0003 & 0.00299 & 0.00834 & 0.02949 & 0.00036 & 0.00309 & 0.00863 & 0.02894 \\ \hline 
DCG @ 30\% & 0.00031 & 0.00312 & 0.009 & 0.02996 & 0.0003 & 0.00276 & 0.00788 & 0.02795 \\ \hline 
DCG @ 50\% & 0.00032 & 0.00316 & 0.00912 & 0.03014 & 0.00032 & 0.00327 & 0.00908 & 0.02955 \\ \hline 
DCG @ 100\% & 0.00035 & 0.00345 & 0.00944 & 0.0305 & 0.00032 & 0.00332 & 0.00922 & 0.02968 \\ \hline 
NDCG @ 10\% & 0.00031 & 0.00297 & 0.00845 & 0.02824 & 0.00036 & 0.00309 & 0.00863 & 0.02894 \\ \hline 
NDCG @ 30\% & 0.00034 & 0.00308 & 0.00883 & 0.02979 & 0.0003 & 0.00276 & 0.00788 & 0.02796 \\ \hline 
NDCG @ 50\% & 0.00032 & 0.003 & 0.00881 & 0.0294 & 0.00032 & 0.00327 & 0.00908 & 0.02955 \\ \hline 
NDCG @ 100\% & 0.00032 & 0.00305 & 0.00893 & 0.02966 & 0.00032 & 0.00332 & 0.00922 & 0.02968 \\ \hline 
PCG @ 10\% & 0.00033 & 0.00324 & 0.00905 & 0.03003 & 0.00038 & 0.00362 & 0.00967 & 0.03096 \\ \hline 
PCG @ 30\% & 0.00031 & 0.00314 & 0.0092 & 0.03041 & 0.00034 & 0.00308 & 0.00889 & 0.03013 \\ \hline 
PCG @ 50\% & 0.00037 & 0.00321 & 0.00937 & 0.03065 & 0.00036 & 0.00316 & 0.00911 & 0.02999 \\ \hline 
PCG @ 100\% & 0.00038 & 0.00331 & 0.00941 & 0.03052 & 0.00035 & 0.00334 & 0.00936 & 0.03063 \\ \hline 
\end{tabular}
}
\end{adjustbox}
\end{center}
\end{table*}

\begin{table*}[ht!]
\begin{center}
\caption{AUUC-values at multiple cut-offs for the Criteo dataset on the test set. On the left hand side the metric optimized at 10, 30, 50 and 100 percent is displayed. The AUUC values are checked at multiple cutoffs to compare the performance. Absolute relevance values.}
\begin{adjustbox}{max width=\textwidth}
{\small
\begin{tabular}{|l|l|l|l|l|l|l|l|l|}
\hline
Criteo         & \multicolumn{4}{l|}{Relative Separate Uplift AUUC at cutoff} & \multicolumn{4}{l|}{Relative Joint Uplift AUUC at cut-off} \\ \hline
                     & 10\%    & 30\%   & 50\%   & 100\%   & 10\%   & 30\%   & 50\%  & 100\%  \\ \hline
DCG @ 10\% & 0.00057 & 0.00333 & 0.00641 & 0.01507 & 0.00092 & 0.00393 & 0.00717 & 0.0159 \\ \hline 
DCG @ 30\% & 0.00062 & 0.00344 & 0.00665 & 0.01534 & 0.0009 & 0.00378 & 0.0069 & 0.01568 \\ \hline 
DCG @ 50\% & 0.0006 & 0.00339 & 0.00657 & 0.01524 & 0.00088 & 0.00383 & 0.00703 & 0.01592 \\ \hline 
DCG @ 100\% & 0.00061 & 0.00339 & 0.00661 & 0.01528 & 0.00091 & 0.00383 & 0.00696 & 0.01568 \\ \hline 
NDCG @ 10\% & 0.00059 & 0.0034 & 0.00653 & 0.01524 & 0.0009 & 0.00389 & 0.00713 & 0.01578 \\ \hline 
NDCG @ 30\% & 0.00059 & 0.00344 & 0.00671 & 0.01549 & 0.0009 & 0.00378 & 0.0069 & 0.01568 \\ \hline 
NDCG @ 50\% & 0.00057 & 0.0034 & 0.00652 & 0.01523 & 0.00088 & 0.00383 & 0.00703 & 0.01592 \\ \hline 
NDCG @ 100\% & 0.00057 & 0.00347 & 0.00676 & 0.0156 & 0.00091 & 0.00383 & 0.00696 & 0.01568 \\ \hline 
PCG @ 10\% & 0.0006 & 0.00352 & 0.00662 & 0.01557 & 0.00089 & 0.00385 & 0.00703 & 0.01573 \\ \hline 
PCG @ 30\% & 0.00059 & 0.00361 & 0.00684 & 0.01596 & 0.00088 & 0.00371 & 0.0067 & 0.01548 \\ \hline 
PCG @ 50\% & 0.0006 & 0.0036 & 0.00677 & 0.01586 & 0.00089 & 0.00383 & 0.00693 & 0.01565 \\ \hline 
PCG @ 100\% & 0.00058 & 0.00352 & 0.00671 & 0.01576 & 0.00083 & 0.0037 & 0.00669 & 0.01536 \\ \hline 
\end{tabular}
}
\end{adjustbox}
\end{center}
\end{table*}
\end{document}